\documentclass[oneside,12pt]{CUEDthesisPSnPDF}

\usepackage[T1]{fontenc}
\usepackage{graphicx,times,amsmath,subfigure} % Add all your packages here
\usepackage{amsthm}
\usepackage{fullpage}

\newcommand{\ompist}{omperical}
\newcommand{\Compist}{C\ompist}
\newcommand{\compist}{c\ompist}

\newcommand{\felz}{Felzenszwalb}

\def\naive{na\"{\i}ve}

\newcommand{\etal}{\textit{et al.}}

\newcommand{\booktitle}{Notes on a New Philosophy of Empirical Science}

\newcommand{\mpagebreak}{}

\newcommand{\keycite}[1]{\cite{#1}*}

\title{\booktitle \\ (Draft Version)}
\author{Daniel Burfoot}
\collegeordept{}
\crest{\includegraphics[width=32mm]{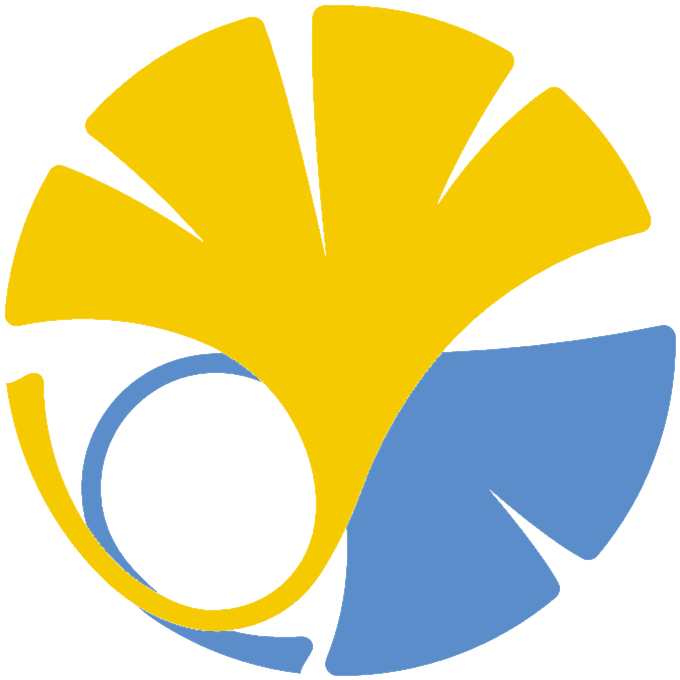}}

\university{}

\renewcommand{\submittedtext}{}
\degree{}
\degreedate{April 2011}

\usepackage{watermark}

\begin{document}

%\language{english}

 \renewcommand\baselinestretch{1.2}
\baselineskip=18pt plus1pt

% A page with the abstract on including title and author etc may be
% required to be handed in separately. If this is not so, then comment
% the below 3 lines (between '\begin{abstractseparte}' and 
% 'end{abstractseparate}'), normally like a declaration ... needs some more
% work, mind as environment abstracts creates a new page!
% \begin{abstractseparate}
%   \input{Abstract/abstract}
% \end{abstractseparate}

\newenvironment{mylisting}
{\begin{list}{}{\setlength{\leftmargin}{1em}}\item\scriptsize\bfseries}
{\end{list}}

\newenvironment{mytinylisting}
{\begin{list}{}{\setlength{\leftmargin}{1em}}\item\tiny\bfseries}
{\end{list}}

% Using the watermark package which is in StyleFiles/
% and to remove DRAFT COPY ONLY appearing on the top of all pages comment out below line
%\watermark{DRAFT COPY ONLY}

\maketitle

%set the number of sectioning levels that get number and appear in the contents
\setcounter{secnumdepth}{3}
\setcounter{tocdepth}{3}

\frontmatter

\chapter{Release Notes}
\nonumber

This document is a draft version of the book, 
	tentatively titled ``\booktitle''.
This book represents an attempt to build a philosophy of science
	on top of a large number of technical ideas
	related to information theory, machine learning, 
	computer vision, and computational linguistics.
It seemed necessary,
	in order to make the arguments convincing,
	to include brief summary descriptions of these ideas.
It is probably inevitable that the technical summaries
	will contain a number of errors or misconceptions,
	related to, for example,
	the Shi-Malik image segmentation algorithm
	or the \textsc{Bleu} metric for evaluating
	machine translation results.
While it seems unlikely that such errors could derail 
	the central arguments of the book,
	it is not impossible.
The reader is advised to exercise caution 
	and consult the relevant literature directly.

The book has been influenced by a diverse set of authors and ideas.
Especially influential references are cited with an asterisk,
	for example: \keycite{Popper:1959}.
The final version of the book will probably include
	a brief description of how each key reference
	influenced the development of the book's ideas.

This draft version contains all the major ideas
	and themes of the book.
However,
	it also contains no small number of 
	blemishes, disfluencies, and other shortcomings.
Two holes are particularly glaring.
First, the chapter on computer vision
	includes an analysis of evaluation methods for the task
	of optical flow estimation,
	as well as a comperical reformulation of the task,
	but does not describe the task itself.
Interested readers can repair this problem
	by a Google Scholar search for the term ``optical flow''.
Second,
	there should be another thought experiment
	involving Sophie and Simon near the end of Chapter 2.
In this thought experiment,
	Sophie proposes to use a birdsong synthesizer to create virtual labels,
	thereby obviating the need for Simon to label
	audio clips by hand.
This idea comes up again in the section on the evaluation
	of face detection algorithms,
	where a graphics program for face modeling is used instead.

\pagebreak
~~~
\pagebreak

\vspace{10mm}

\begin{quote}
It is a profound and necessary truth that the deep things in science are not found because they are useful, 
	they are found because it was possible to find them.

-Oppenheimer
\end{quote}

\vspace{10 mm}

\begin{quote}
It is the mark of a higher culture to value the little unpretentious truths
	which have been discovered by means of rigorous method more
	highly than the errors handed down by metaphysical ages and men,
	which blind us and make us happy.
	
-Nietzsche
\end{quote}

\vspace{10 mm}

\begin{quote}
Go as far as you can see; 
	when you get there you will be able to see farther.
	
-Carlyle
\end{quote}

\pagebreak
~~~
\pagebreak

\tableofcontents
%\listoffigures
%\printglossary  %% Print the nomenclature
\addcontentsline{toc}{chapter}{Nomenclature}

\mainmatter

%\minclude{introductn}

\chapter{Compression Rate Method}
\label{chapt:compmethod}

\section{Philosophical Foundations of Empirical Science}

In a remarkable paper published in 1964,
	a biophysicist named John Platt
	pointed out the somewhat impolitic fact
	that some scientific fields made 
	progress much more rapidly than others~\keycite{Platt:1964}.
Platt cited particle physics
	and molecular biology as exemplar fields in which progress
	was especially rapid.
To illustrate this speed he relates the following anecdote:
	
\begin{quote}
[Particle physicists asked the question]:
	Do the fundamental particles conserve mirror-symmetry or
	``parity'' in certain reactions, 
	or do they not? 
The crucial experiments were suggested: within a few months they
	were done, and conservation of parity
	was found to be excluded. 
Richard Garwin, Leon Lederman, and Marcel Weinrich 
	did one of the crucial experiments. 
It was thought of one evening at suppertime: 
	by midnight they had arranged the apparatus for it; 
	and by 4 am they had picked up the predicted
	pulses showing the non-conservation of parity. 
\end{quote}
	
\noindent
Platt attributed this rapid progress not to the superior intelligence of 
	particle physicists and molecular biologists,
	but to the fact that they used a more rigorous scientific methodology,
	which he called Strong Inference.
In Platt's view,
	the key requirement of rapid science is the ability to rapidly generate new theories,
	test them, and discard those that prove to be incompatible with evidence.
	
Many observers of fields such as artificial intelligence (AI),
	computer vision, computational linguistics, 
	and machine learning will agree that,
	in spite of the journalistic hype surrounding them,
	these fields do not make rapid progress.
Research in artificial intelligence was begun
	over 50 years ago.
In spite of the bold pronouncement made at the time,
	the field has failed to transform society.
Robots do not walk the streets;
	intelligent systems are generally brittle 
	and function only within narrow domains.
This lack of progress is illustrated by a comment by Marvin Minsky, 
	one of the founders of AI, 
	in reference to David Marr,
	one of the founders of computer vision:

\begin{quote}
After [David Marr] joined us, 
	our team became the most famous vision group in the world, 
	but the one with the fewest results. 
His idea was a disaster. 
The edge finders they have now using his theories, 
	as far as I can see, are slightly worse than the ones we had just before taking him on. 
We've lost twenty years (\cite{Crevier:1993}, pg. 189).
\end{quote}

This book argues that the lack of progress in 
	artificial intelligence and related fields
	is caused by philophical limitations, 
	not by technical ones.
Researchers in these fields 
	have no scientific methodology of power
	comparable to Platt's concept of Strong Inference.
They do not rapidly generate, test, and discard theories
	in the way that particle physicists do.
This kind of critique has been uttered before,
	and would hardly justify a book-length exposition.
Rather, the purpose of this book is to \textit{propose}
	a scientific method that can be used, at least,
	for computer vision and computational linguistics,
	and probably for several other fields as well.

To set the stage for the proposal
	it is necessary to briefly examine the unique intellectual 
	content of the scientific method.
This uniqueness can be highlighted by comparing it to a
	theory of physics such as quantum mechanics.
While quantum mechanics often seems mysterious and perplexing to beginning students,
	the scientific method appears obvious and inevitable.
Physicists 
	are constantly testing, examining,
	and searching for failures of quantum mechanics.
The scientific method itself receives no comparable interrogation.
Physicists are quite confident that quantum mechanics 
	is wrong in some subtle way:
	one of their great goals is to find a unified theory
	that reconciles the conflicting predictions made by
	quantum mechanics and general relativity.
In contrast, it is not even clear what it would mean for
	the scientific method to be wrong.

But consider the following chain of causation:
	the scientific method allowed humans to discover physics,
	physics allowed humans to develop technology,
	and technology allowed humans to reshape the world.
The fact that the scientific method succeeds 
	must reveal some abstract truth about 
	the nature of reality.
Put another way,	
	the scientific method depends implicitly on some assertions or propositions,
	and because those assertions happen to be true,
	the method works.	
But what is the content of those assertions?
Can they be examined, modified, or generalized?	

This chapter begins with an attempt to analyze and document 
	the assertions and philosophical commitments
	upon which the scientific method depends.
Then, a series of thought experiments
	illustrate how a slight change to one of the statements
	results in a modified version of the method.
This new version is based on large scale lossless data compression,
	and it uses large databases instead of experimental observation
	as the necessary empirical ingredient.
The remainder of the chapter argues
	that the new method retains all the crucial characteristics
	of the original.
The significance of the new method 
	is that allows researchers to conduct investigations
	into aspects of empirical reality that have never before
	been systematically interrogated.
For example,
	Chapter~\ref{chapt:comprvsion} that
	attempting to compress a database of natural images
	results in a field very similar to computer vision.
Similarly,
	attempting to compress large text databases
	results in a field very similar to computational linguistics.
The starting point 
	in the development is a consideration of one of the most critical
	components of science: objectivity.
	
%TODO: goal is \textit{not} to paint a fully general picture of
%	what empirical science is.
%There are many ways of doing science and
%	to describe them all would be beyond the scope of this book.
%Instead, the goal is to describe a particular set
%	of commitments and beliefs,
%	and claim that researchers who operate according 
%	to that mindset certainly \textit{are} doing science.
%Also, that mindset is highly conducive to making rapid progress,
%	and, crucially, the mindset is not at all obvious.
%The argument of second half of chapter is that the compression goal
%	exactly matches the mindset.
%So while it is nonobvious,
%	the modified scientific procedure described 
%	can also support rapid progress.
	
\mpagebreak	

\subsection{Objectivity, Irrationality, and Progress}

The history of humanity clearly indicates that
	humans are prone to dangerous flights of irrationality.
Psychologists have shown 
	that humans suffer from a wide range of cognitive blind spots,
	with names like Scope Insensitivity and Availability Bias~\cite{Kahneman:1982}.
One special aspect of human irrationality
	of particular relevance to science
	is the human propensity to enshrine theories,
	abstractions, and explanations without sufficient evidence.
Often, once a person decides that a certain theory is true,
	he begins to use that theory to interpret all new evidence.
This distortative effect prevents him from 
	seeing the flaws in the theory itself.
Thus Ptolemy believed that the Sun rotated around the Earth,
	while Aristotle believed that all matter could be decomposed
	into the elements of fire, air, water, or earth.
	
Individual human fallibility is not the only obstacle to intellectual progress;
	another powerful barrier is \textit{group} irrationality.
Humans are fundamentally social creatures;
	no individual acting alone could ever obtain substantial
	knowledge about the world.
Instead, humans must rely on a division of labor
	in which knowledge-acquisition tasks
	are delegated to groups of dedicated specialists.
This division of labor is replicated even within the scientific community:
	physicists rely extensively on the experimental and theoretical
	work of other physicists.
But groups are vulnerable to an additional set of perception-distorting
	effects involving issues
	such as status, signalling, politics, conformity pressure,
	and pluralistic ignorance.
A low-ranking individual in a large group cannot comfortably disagree
	with the statements of a high-ranking individual,
	even if the former has truth on his side.
Furthermore, 
	scientists are naturally competitive and skeptical.
A scientist proposing a new result must be prepared
	to defend it against inevitable criticism.

To overcome the problems of individual irrationality and group irrationality,
	a single principle is tremendously important:
	the principle of objectivity.
Objectivity requires that new results be validated
	by mechanistic procedure that cannot be influenced
	by individual perceptions or sociopolitical effects.
While humans may implement the validation procedure,
	it must be somehow independent of the particular
	oddities of the human mind.
In the language of computer science,
	the procedure must be like an abstract algorithm,
	that does not depend on the particular architecture
	of the machine it is running on.
The validation procedure helps to prevent individual irrationality,
	by requiring scientists to hammer their ideas
	against a hard anvil.
It also protects against group irrationality,
	by providing scientists with a strong shield
	against criticism and pressure from the group.
	
The objectivity principle is also an important requirement
	for a field to make progress.
Researchers in all fields love to publish papers.
If a field lacks an objective validation procedure,
	it is difficult to prevent people from publishing low quality papers
	that contain incorrect results or meaningless observations.
The so-called hard sciences such as mathematics, physics,
	and engineering employ highly objective evaluation procedures,
	which facilitates rapid progress.
Fields such as psychology, economics, and medical science
	rely on statistical methods to validate their results.
These methods are less rigorous, 
	and this leads to significant problems in these fields,
	as illustrated by a recent paper entitled
	``Why most published research findings are false''~\cite{Ioannidis:2006}.
Nonscientific fields such as literature and history rely on the qualitative
	judgments of practitioners for the purposes of evaluation.
These examples illustrate a striking correlation between
	the objectivity of a field's evaluation methods
	and the degree of progress it achieves.
	
\mpagebreak

\subsection{Validation Methods and Taxonomy of Scientific Activity}

The idea of objectivity,
	and the mechanism by which various fields achieve objectivity,
	can be used to define a useful taxonomy of scientific fields.
Scientific activity, broadly considered, can be categorized into three parts:
	mathematics, empirical science, and engineering.
These activities intersect at many levels,
	and often a single individual will make contributions 
	in more than one area.
But the three categories produce very distinct kinds of results,
	and utilize different mechanisms to validate the results
	and thereby achieve objectivity.

Mathematicians see the goal of their efforts as the 
	discovery of new theorems.
A theorem is fundamentally a statement of implication:
	\textit{if} a certain set of assumptions are true,
	\textit{then} some derived conclusion must hold.
The legitimate mechanism for demonstrating the validity
	of a new result is a proof.
Proofs can be examined by other mathematicians 
	and verified to be correct,
	and this process provides the field with its objective validation mechanism.
It is worthwhile to note that practical utility plays no essential role
	in the validation process.
Mathematicians may hope that their results are useful to others,
	but this is not a requirement for a theorem to be considered correct.

Engineers, in contrast,
	take as their basic goal the development of practical devices.
A device is a set of interoperating components that
	produce some useful effect.
The word ``useful'' may be broadly interpreted:
	sometimes the utility of a new device may be speculative,
	or it may be useful only as a subcomponent of a larger device.
Either way, the requirement for proclaiming success in engineering
	is a demonstration that the device \textit{works}.
It is very difficult to game this process:
	if the new airplane fails to take off
	or the new microprocessor fails to multiply numbers correctly, 
	it is obvious that these results are low-quality.
Thus, this public demonstration process provides
	engineering with its method of objective validation.
	
The third category,
	and the focus of this book, is empirical science.
Empirical scientists attempt to obtain theories of natural phenomena.
A theory is a tool that enables the scientist to make 
	predictions regarding a phenomenon.
The value and quality of a theory depends entirely on how
	well it can predict the phenomenon to which it applies.
Empirical scientists are similar to mathematicians in the purist attitude they take
	toward the product of their research:
	they may \textit{hope} that a new theory will have practical applications,
	but this is not a requirement.

Mathematics and engineering both have long histories.
Mathematics dates back at least to 500 BC,
	when Pythagoras proved the theorem that bears his name.
Engineering is even older;
	perhaps the first engineers were the men who 
	fashioned axes and spearheads out of flint
	and thus ushered in the Stone Age.
Systematic empirical science, in contrast, 
	started only relatively recently,
	building on the work of thinkers like Galileo, Descartes, and Bacon.
It is worth asking why the ancient philosophers
	in civilizations like Greece, Babylon, India, and China,
	in spite of their general intellectual advancement,
	did not begin a systematic empirical investigation of
	various natural phenomena.
	
The delay could have been caused by the fact that,
	for a long time,
	no one realized that there could be, 
	or needed to be,
	an area of intellectual inquiry that was distinct from
	mathematics and engineering.
Even today,
	it is difficult for nonscientists to appreciate
	the difference between a statement of mathematics
	and a statement of empirical science.
After all,
	physical laws are almost always expressed in mathematical terms.
What is the difference between the Newtonian statement
	$F = m a$
	and the Pythagorean theorem $a^2 + b^2 = c^2$?
These statements, though they are expressed in a similar form,
	are in fact completely different constructs:
	one is an empirical theory, the other is a mathematical law.
Several heuristics
	can be used to differentiate between the two types of statement.
One good technique is to ask if the statement 
	could be invalidated by some new observation or evidence.
One could draw a misshapen triangle that did not obey	
	the Pythagorean theorem,
	but that would hardly mean anything about
	the truth of the theorem.
In contrast, there are observations that could invalidate Newton's laws,
	and in fact such observations were made
	as a result of Einstein's theory of relativity.
There are, in turn,
	also observations that could disprove relativity.

Ancient thinkers might also have failed to see how
	it could be meaningful to make statements about
	the world that were not essentially connected
	to the development of practical devices.
An ancient might very well have believed that it was
	impossible or meaningless to find a unique optimal theory of gravity and mass.
Instead, 
	scientists should develop a toolbox of methods 
	for treating these phenomena.
Engineers should then select a tool that is well-suited
	to the task at hand.
So an engineer might very well utilize 
	one theory of gravity to design a bridge,
	and then use some other theory
	when designing a catapult.
In this mindset,
	theories can only be evaluated by incorporating them into some practical device
	and then testing the device.

Empirical science is also unique in that 
	it depends on a special \textit{process} for obtaining new results.
This process is called the scientific method;
	there is no analogous procedure in mathematics or engineering.
Without the scientific method,
	empirical scientists cannot do much more than make catalogs
	of disconnected and uninterpretable observations.
When equipped with the method,
	scientists begin to discern the structure and meaning of the observational data.
But as explained in the next section,
	the scientific method is only obvious in hindsight.
It is built upon deep philosophical commitments 
	that would have seemed bizarre to an ancient thinker.
	
\mpagebreak 

\subsection{Toward a Scientific Method}
\label{sec:toscimethd}

To understand the philosophical commitments implicit in empirical science,
	and to see why those commitments were nonobvious to the ancients,
	it is helpful to look at some other plausible scientific procedures.
To do so,
	it is convenient to introduce the following 
	simplified abstract description of the goal of scientific reasoning.
Let $x$ be an experimental configuration,
	and $y$ be the experimental outcome.
The variables $x$ and $y$ should be thought of not as numbers
	but as large packets of information
	including descriptions of various objects and quantities.
The goal of science is to find a function $f(\cdot)$ 
	that predicts the outcome of the configuration: $y = f(x)$.

A first approach to this problem,	
	which can be called the pure theoretical approach,
	is to deduce the form of $f(\cdot)$ using logic alone.
In this view,
	scientists should use the same mechanism for 
	proving their statements that mathematicians use.
Here there is no need to check the results of a prediction against 
	the experimental outcome.
Just as it is meaningless to check the Pythagorean theorem by
	drawing triangles and measuring its sides,
	it is meaningless to check the function $f(\cdot)$ 
	against the actual outcomes $y$.
Mathematicians can achieve perfect confidence in their theories
	without making any kind of appeal to experimental validation,
	so why shouldn't scientists be able to reason the same way?
If Euclid can prove,
	based on purely logical and conceptual considerations,
	that the sum of the angles of a triangle adds up to 180 degrees,
	why cannot Aristotle use analogous considerations
	to conclude that all matter is composed of the four classical elements?
A subtle critic of this approach might point out that
	mathematicians require the use of \textit{axioms},
	from which they deduce their results,
	and it is not clear what statements can play this role
	in the investigation of real-world phenomena.
But even this criticism can be answered;
	perhaps the existence of human reason is the only necessary axiom,
	or perhaps the axioms can be found in religious texts.
Even if someone had proposed to check a prediction against the actual outcome,
	it is not at all clear what this means or how to go about doing it.
What would it mean to check Aristotle's theory of the four elements?
The ancients must have viewed the crisp proof-based validation method of mathematics
	as far more rigorous and intellectually satisfying 
	than the tedious, error prone, and conceptually murky
	process of observation and prediction-checking.	
	
At the other extreme from the pure theoretical approach
	is the strategy of searching for 
	$f(\cdot)$ using a purely \textit{experimental} investigation of various phenomena.
The plan here would be to conduct a large number of experiments,
	and compile the results into an enormous almanac.
Then to make a prediction in a given situation,
	one simply looks up a similar situation in the almanac,
	and uses the recorded value.
For example, one might want to predict whether a brige will collapse
	under a certain weight.
Then one simply looks up the section marked ``bridges'' in the almanac,
	finds the bridge in the almanac that is most similar to the one in question,
	and notes how much weight it could bear.
In other words, the researchers obtain a large number of data samples $\{x_{i}, y_{i}\}$
	and define $f(\cdot)$ as an enormous lookup table.
The pure experimental approach has an obvious drawback:
	it is immensely labor-intensive.
The researchers given the task of compiling the section on bridges
	must construct several different kinds of bridges,
	and pile them up with weight until they collapse.	
Bridge building is not easy work,
	and the almanac section on bridges is only one among many.
The pure experimental approach may also be inaccurate,
	if the almanac includes only a few examples relating
	to a certain topic.
	
Obviously, neither the pure theoretical approach
	nor the pure experimental approach is very practical.
The great insight of empirical science is that one can 
	effectively \textit{combine} experimental and theoretical investigation 
	in the following way.
First, a set of experiments corresponding to configurations $\{x_{1}, x_{2} \ldots x_{N}\}$ are performed,
	leading to outcomes $\{y_{1}, y_{2} \ldots y_{N}\}$.
The difference between this process and the pure experimental approach
	is that here the number of tested configurations is much smaller.
Then, in the theoretical phase,
	one attempts to find a function $f(\cdot)$ that agrees with 
	all of the data: $y_{i} = f(x_{i})$ for all $i$.
If such a function is found,
	and it is in some sense simple, 
	then one concludes that it will \textit{generalize}
	and make correct predictions when applied to new configurations 
	that have not yet been tested.

This description of the scientific process should produce 
	a healthy dose of sympathy for the ancient thinkers
	who failed to discover it.
The idea of generalization,
	which is totally essential to the entire process, 
	is completely nonobvious and raises a number of nearly intractable
	philosophical issues.
The hybrid process assumes the existence of a finite number of observations $x_{i}$,
	but claims to produce a \textit{universal} predictive rule $f(\cdot)$.
Under what circumstances is this legitimate?
Philosophers have been grappling with this question,	
	called the Problem of Induction,
	since the time of David Hume.
Also, a moment's reflection indicates that
	the problem considered in the theoretical phase
	does not have a unique solution.
If the observed data set is finite,
	then there will be a large number of functions $f(\cdot)$ 
	that agree with it.
These functions must make the same predictions for the known data $x_{i}$,
	but may make very different predictions for other configurations.

\mpagebreak
	
\subsection{Occam's Razor}

William of Occam famously articulated the principle that bears his name
	with the Latin phrase: \textit{entia non sunt multiplicanda praeter necessitatem};
	entities must not be multiplied without necessity.
In plainer English,
	this means that if a theory is adequate to explain a body of observations,
	then one should not add gratuitous embellishments or clauses to it.
To wield Occam's Razor
	means to take a theory and cut away all of the inessential parts
	until only the core idea remains.

Scientists use Occam's Razor to deal with the problem of theory degeneracy mentioned above.
Given a finite set of experimental configurations $\{x_{1}, x_{2}, \ldots x_{N}\}$
	and corresponding observed outcomes $\{y_{1}, y_{2} \ldots y_{N}\}$,
	there will always be an infinite number of functions
	$f_{1}, f_{2}, \ldots $
	that agree with all the observations.
The number of compatible theories is infinite
	because one can always produce a new theory by adding a new 
	clause or qualification to a previous theory.
For example, one theory might be expressed in English as 
	``General relativity holds everywhere in space''.
This theory agrees with all known experimental data.
But one could then produce a new theory that says
	``General relativity holds everywhere in space except
	in the Alpha Centauri solar system, where Newton's laws hold.''
Since it is quite difficult to show the superiority of the theory of 
	relativity over Newtonian mechanics 
	even in our local solar system,
	it is probably almost impossible to show that relativity 
	holds in some other, far-off star system.
Furthermore, an impious philosopher could generate an effectively infinite
	number of variant theories of this kind,
	simply by replacing ``Alpha Centauri'' with the name of some other star.
This produces a vast number of conflicting accounts of physical reality,
	each with about the same degree of empirical evidence.

Scientists use Occam's Razor to deal with this kind of crisis
	by justifying the disqualification of the variant theories mentioned above.
Each of the variants has a gratuitous subclause,
	that specifies a special region of space where relativity does not hold.
The subclause does not improve the theory's descriptive accuracy;
	the theory would still agree with all observational data if it were removed.
Thus, the basic theory that relativity holds everywhere
	stands out as the simplest theory that agrees with all the evidence.
Occam's Razor instructs us to accept the basic theory as the current champion,
	and only revise it if some new contradictory evidence arrives.
	
This idea sounds attractive in the abstract,
	but raises a thorny philosophical problem
	when put into practice.
Formally, the razor requires one to construct a functional $\mathbf{H}[f]$
	that rates the complexity of a theory.
Then given a set of theories $\cal{F}$
	all of which agree with the empirical data,
	the champion theory is simply the least complex member  of $\cal{F}$:
	
\begin{equation*}
f^{*} = \min_{f \in \cal{F}} \mathbf{H}[f]
\end{equation*}
	
\noindent
The problem is: 
	how does one obtain the complexity functional $\mathbf{H}$?
Given two candidate definitions for the functional,
	how does one decide which is superior?
It may very well be that complexity is in the eye of the beholder,
	and that two observers can legitimately disagree
	about which of two theories is more complex.
This disagreement would, in turn, cause them to disagree
	about which member of a set of candidate theories
	should be considered the champion on the basis of the 
	currently available evidence.
This kind of disagreement appears to undermine 
	the objectivity of science.
Fortunately, in practice, the issue is not insurmountable.
Informal measures of theory complexity,
	such as the number of words required to describe a theory in English,
	seem to work well enough.
Most scientists would agree that ``relativity holds everywhere''
	is simpler than ``relativity holds everywhere except around Alpha Centauri''.
If a disagreement persists,
	then the disputants can, in most cases, settle the issue
	by running an actual experiment.

\mpagebreak

\subsection{Problem of Demarcation and Falsifiability Principle}

The great philosopher of science Karl Popper
	proposed a principle called \textit{falsifiability}
	that substantially clarified the meaning and 
	justification of scientific theorizing~\keycite{Popper:1959}.
Popper was motivated by a desire to rid the world of pseudosciences	
	such as astrology and alchemy.
The problem with this goal is that
	astrologers and alchemists may very well appear to be doing real science,
	especially to laypeople.
Astrologers may employ mathematics, 
	and alchemists may utilize much of the same equipment as chemists.
Some people who promote creationist or religiously inspired accounts 
	of the origins of life make plausible sounding arguments
	and appear to be following the rules of logical inference.
These kinds of surface similarities may make it impossible for nonspecialists
	to determine which fields are scientific and which fields are not.
Indeed, even if everyone agreed that astronomy is science 
	but astrology is not,
	it would be important from a philosophical perspective to justify
	this determination.
Popper calls this the Problem of Demarcation:
	how to separate scientific theories from nonscientific ones.

Popper answered this question by proposing the principle of falsifiability.
He required that,
	in order for a theory to be scientific,
	it must make a prediction with enough confidence that,
	if the prediction disagreed with the actual outcome of an appropriate experiment or observation,
	the theory would be discarded.
In other words,
	a scientist proposing a new theory must be willing to risk embarassment
	if it turns out the theory does not agree with reality.
This rule prevents people from constructing grandiose theories
	that have no empirical consequences.
It also prevents people from using a theory as a lens,
	that distorts all observations so as to render them compatible with its abstractions.
If Aristotle had been aware of the idea of falsifiability,
	he might have avoided developing his silly theory of the four elements,
	by realizing that it made no concrete predictions.

In terms of the notation developed above,
	the falsifiability principle requires that a theory
	can be instantiated as a function $f(\cdot)$ that applies
	to some real world configurations.
Furthermore, 
	the theory must designate a configuration $x$ and a prediction $f(x)$
	with enough confidence that if the experiment is done,
	and the resulting $y$ value does not agree with the prediction $y \neq f(x)$,
	then the theory is discarded.
This condition is fairly weak,
	since it requires a prediction for only a single configuration.
The point is that the falsifiability principle does not say
	anything about the \textit{value} of a theory,
	it only states a requirement for the theory to be considered scientific.
It is a sort of precondition,
	that guarantees that the theory can be evaluated in relation to other theories.
It is very possible for a theory to be scientific but wrong.

In addition to marking a boundary between science and pseudoscience,
	the falsifiability principle also
	permits one to delineate between statements
	of mathematics and empirical science.
Mathematical statements are \textit{not} falsifiable
	in the same way empirical statements are.
Mathematicians do not and can not use the falsifiability principle;
	their results are verified using an alternate criterion:
	the mathematical proof.
No new empirical observation or experiment could falsify
	the Pythagorean theorem.
A person who drew a right triangle and attempted to show that
	the length of its sides did not satisfy $a^2 + b^2 = c^2$
	would just be ridiculed.
Mathematical statements are fundamentally implications:
	if the axioms are satisfied,
	then the conclusions follows logically.
	
The falsifiability principle is strong medicine,
	and comes, as it were, with a set of powerful side-effects.
Most prominently, 
	the principle allows one to conclude that a theory is false,
	but provides no mechanism whatever to justify the conclusion
	that a theory is true.
This fact is rooted in one of the most basic rules of logical inference:
	it is impossible to assert universal conclusions on the basis
	of existential premises.
Consider the theory ``all swans are white''.
The sighting of a black swan, 
	and the resulting premise ``some swans are black'',
	leads one to conclude that the theory is false.
But no matter how many white swans one may happen to observe,
	one cannot conclude with perfect confidence that 
	the theory is true.	
According to Popper,
	the only way to establish a scientific theory is to 
	falsify all of its competitors.
But because the number of competitors is vast,
	they cannot all be disqualified.
This promotes a stance of radical skepticism towards
	scientific knowledge.
	
\mpagebreak
	
\subsection{Science as a Search Through Theory-Space}
\label{sec:theryspace}

Though the scientific is not monolithic or precisely defined,
	the following list describes it fairly well:

\begin{enumerate}
\item Through observation and experiment, amass an initial corpus of 
	configuration-outcome pairs $\{x_{i}, y_{i}\}$
	relating to some phenomenon of interest.
\item Let $f_{C}$ be the initial champion theory.
\item Through observation and analysis, develop a new theory,
	which may either be a refinement of the champion theory,
	or something completely new. 
	Prefer simpler candidate theories to more complex ones.
\item Instantiate the new theory in a predictive function $f_{N}$. 
	If this cannot be done, the theory is not scientific.
\item Find a configuration $x$ for which
	$f_{C}(x) \neq f_{N}(x)$, and run the indicated experiment.
\item If the outcome agrees with the rival theory,
	$y = f_{N}(x)$, then discard the old champion and set $f_{C} = f_{N}$.
	Otherwise discard $f_{N}$.
\item Return to step \#3.
\end{enumerate}

The scientific process described above makes a crucial assumption,
	which is that perfect agreement between
	theory and experiment can be observed, 
	such that $y = f(x)$.
In practice,
	scientists never observe $y = f(x)$ but rather $y \approx f(x)$.
This fact does not break the process described above,
	because even if neither theory is perfectly correct,
	it is reasonable to assess one theory as ``more correct'' than another
	and thereby discard the less correct one.
However, the fact that real experiments never agree perfectly with theoretical 
	predictions has important philosophical consequences,
	because it means that scientists are searching not for perfect truth
	but for good approximations.
Most physicists will admit that even their most refined theories
	are mere approximations,
	though they are spectacularly accurate approximations.

%The fact that scientists observe $y \approx f(x)$
%	and not $y = f(x)$ 
%	brings up another philosophical issue.
%The procedure described above suggests that the Occam's Razor
%	idea is only useful for sorting out
%	the theories that agree with all empirical observations.
%If perfect alignment between observation and prediction could be achieved,
%	then a theory that did not agree with some observations should be discarded,
%	no matter how simple it is relative to the others.
%But if theories are approximations,
%	then this principle must be refined somehow.
%If theory A is much simpler than theory B,
%	but only slightly less accurate,
%	it seems reasonable to select theory A over theory B.
%In other words,
%	there must be a way to weigh complexity 
%	against descriptive accuracy.
	
In the light of this idea about approximation,
	the following conception of science becomes possible.
Science is a search through a vast space $\cal{F}$
	that contains all possible theories.
There is some ideal theory $f^{*} \in \cal{F}$,
	which correctly predicts the outcome of all experimental configurations.
However, this ideal theory can never be obtained.
Instead,
	scientists proceed towards $f^{*}$ through a 
	process of iterative refinement.
At every moment, the current champion theory $f_{C}$
	is the best known approximation to $f^{*}$
And each time a champion theory is unseated in favor of a new candidate,
	the new $f_{C}$ is a bit closer to $f^{*}$.	
	
This view of science as a search for good approximations 
	brings up another nonobvious component of the 
	philosophical foundations of empirical science.
If perfect truth cannot be obtained,
	why is it worth expending so much effort
	to obtain mere approximations?
Wouldn't one expect that using an approximation might cause problems 
	at crucial moments?
If the theory that explains an airplane's ability to remain aloft
	is only an approximation, 
	why is anyone willing to board an airplane?
The answer is, of course,
	that the approximation is good enough.
The fact that perfection is unachievable 
	does not and should not dissuade
	scientists from reaching toward it.
A serious runner considers it deeply meaningful to attempt to run faster,
	though it is impossible for him to complete a mile in less than a minute.
In the same way,
	scientists consider it worthwhile to search for increasingly accurate approximations,
	though perfect truth is unreachable.
	
\mpagebreak
	
\subsection{Circularity Commitment and Reusability Hypothesis}

Empirical scientists follow a unique conceptual cycle in their work
	that begins and ends in the same place.
Mathematicians start from axioms and move on to theorems.
Engineers start from basic components and assemble them
	into more sophisticated devices.
An empirical scientist begins with an experiment or set of observations that produce measurements.
She then contemplates the data and attempts to understand the hidden structure of the measurements.
If she is smart and lucky, she might discover a theory of the phenomenon.
To test the theory, she uses it to make predictions regarding the \textit{original phenomenon}.
In other words, the same phenomenon acts as both the starting point
	and the ultimate justification for a theory.
This dedication to the single, isolated goal of describing
	a particular phenomenon is called the Circularity Commitment.
	
The nonobviousness of the Circularity Commitment can be
	understood by considering the alternative.
Imagine a scientific community in which theories are not justified
	by their ability to make empirical predictions,
	but by their practical utility.
For example,
	a candidate theory of thermodynamics might be evaluated
	based on whether it can be used to construct combustion engines.
If the engine works,
	the theory must be good.
This reasoning is actually quite plausible,
	but science does not work this way.
No serious scientist would suggest that because the theory of relativity is not 
	relevant to or useful for the construction of airplanes,
	it is not an important or worthwhile theory.
Modern physicists develop theories regarding a wide range of esoteric 
	topics such as quantum superfluidity and the entropy of black holes
	without concerning themselves with the practicality of those theories.
Empirical scientists are thus very similar
	to mathematicians in the purist attitude they adopt regarding their work.
	
In a prescientific age, 
	a researcher expressing this kind of dedication to pure empirical inquiry, 
	especially given the effort required to carry out such an inquiry,
	might be viewed as an eccentric crank or religious zealot.
In modern times no such stigma exists,
	because everyone can see that empirical science is eminently practical.
This leads to another deeply surprising idea, 
	here called the Reusability Hypothesis:
	in spite of the fact that scientists are explicitly unconcerned
	with the utility of their theories,
	it just so happens that those theories tend to be extraordinarily useful.
Of course, no one can know in advance
	which areas of empirical inquiry will prove 
	to be technologically relevant.
But the history of science demonstrates
	that new empirical theories often 
	catalyze the development of amazing new technologies.
Thus Maxwell's unified theory of electrodynamics
	led to a wide array of electronic devices,
	and Einstein's theory of relativity led to the atomic bomb.
The fact that large sums of public money are spent on constructing ever-larger
	particle colliders is evidence that the Reusability Hypothesis
	is well understood even by government officials and policy makers.

The Circularity Commitment and the Reusability Hypothesis 
	complement each other naturally.
Society would never be willing to fund scientific research
	if it did not produce some tangible benefits.
But if society explicitly required scientists to produce practical results,
	the scope of scientific investigation would be drastically reduced.
Einstein would not have been able to justify his research into relativity,
	since that theory had few obvious applications
	at the time it was invented.
The two philosophical ideas justify 
	a fruitful division of labor.
Scientists aim with intent concentration at a single target:
	the development of good empirical theories.
They can then hand off their theories to the engineers,
	who often find the theories to be useful in the development
	of new technologies.

\mpagebreak

\section{Sophie's Method}

This section develops a refined version of the scientific method,
	in which large databases are used instead of experimental observations
	as the necessary empirical ingredient.
The necessary modifications are fairly minor,
	so the revised version includes all of the same
	conceptual apparatus of the standard version.
At the same time,
	the modification is significant enough
	to considerably expand the scope of empirical science.
The refined version is developed through a series 
	of thought experiments relating to a fictional character named Sophie.

\subsection{The Shaman}

Sophie is a assistant professor of physics at a large American state university.
She finds this job vexing for several reasons,
	one of which is that she has been chosen by the department 
	to teach a physics class intended for students majoring in the humanities,
	for whom it serves to fill a breadth requirement. 
The students in this class, who major in subjects like literature,
	religious studies, and philosophy, tend to be intelligent
	but also querulous and somewhat disdainful of the ``merely technical'' 
	intellectual achievements of physics.

In the current semester she has become aware of the presence
	in her class of a discalced student with a large beard and often bloodshot eyes.
This student is surrounded by an entourage of similarly strange looking followers.
Sophie is on good terms with some of the more serious students in the class,
	and in conversation with them has found out that the odd student
	is attempting to start a new naturalistic religious movement and refers to himself as a ``shaman''.

One day while delivering a simple lecture on Newtonian mechanics,
	Sophie is surprised when the shaman raises his hand.
When Sophie calls on him,
	he proceeds to claim that physics is a propagandistic hoax designed by the elites
	as a way to control the population.
Sophie blinks several times, 
	and then responds that physics can't be a hoax because it makes
	real-world predictions that can be verified by independent observers.
The shaman counters by claiming that the so-called ``predictions'' made by physics 
	are in fact trivialities, and that he can obtain better forecasts by communing with the spirit world.
He then proceeds to challenge Sophie to a predictive duel,
	in which the two of them will make forecasts regarding the outcome of a simple experiment,
	the winner being decided based on the accuracy of the forecasts.
Sophie is taken aback by this but,
	hoping that by proving the shaman wrong she can break the spell he has cast on some of the other students,
	agrees to the challenge.
	
During the next class, Sophie sets up the following experiment. 
She uses a spring mechanism to launch a ball into the air at an angle $\theta$.
The launch mechanism allows her to set the initial velocity of the ball 
	to a value of $v_{i}$.
She chooses as a predictive test the problem of predicting the time $t_{f}$
	that the ball will fall back to the ground after being launched at $t_{i}=0$.
Using a trivial Newtonian calculation she concludes that $t_{f} = 2 g^{-1} v_{i} \sin(\theta)$,
	sets $v_{i}$ and $\theta$ to give a value of $t_{f} = 2$ seconds,
	and announces her prediction to the class.
She then asks the shaman for his prediction.
The shaman declares that he must consult with the wind spirits,	
	and then spends a couple of minutes chanting and muttering.
Then, dramatically flaring open his eyes as if to signify a moment of revelation,
	he grabs a piece of paper, writes his prediction on it,
	and then hands it to another student. 
Sophie suspects some kind of trick, 
	but is too exasperated to investigate and so launches the ball into the air.
The ball is equipped with an electronic timer that starts and stops when an impact is detected,
	and so the number registered in the timer is just the time of flight $t_{f}$.
A student picks up the ball and reports that the result is $t_{f} = 2.134$.
The shaman gives a gleeful laugh,
	and the student holding his written prediction hands it to Sophie.
On the paper is written $1 < t_{f} < 30$.
The shaman declares victory:
	his prediction turned out to be correct,
	while Sophie's was incorrect (it was off by $0.134$ seconds).

To counter the shaman's claim and because it was on the syllabus anyway,
	in the next class Sophie begins a discussion of probability theory.
She goes over the basic ideas,
	and then connects them to the experimental prediction made about the ball.
She points out that technically, the Newtonian prediction $t_{f}=2$ 
	is not an assertion about the exact value of the outcome.
Rather it should be interpreted as the mean of a probability distribution
	describing possible outcomes.
For example, one might use a normal distribution with mean $\mu=t_{f}=2$
	and $\sigma=.3$.
The reason the shaman superficially seemed to win the contest is that he gave  
	a probability distribution while Sophie gave a point prediction;
	these two types of forecast are not really comparable.
In the light of probability theory, 
	the reason to prefer the Newtonian prediction above the shamanic one,
	is that it assigns a higher probability to the outcome that actually occurred.
Now, plausibly, if only a single trial is used then the Newtonian theory might simply have gotten lucky, 
	so the reasonable thing to do is combine the results over many trials, 
	by multiplying the probabilities together.
Therefore, 
	the formal justification for preferring
	the Newtonian theory to the shamanic theory is that:
	
\begin{equation*}
\label{eq:newtnshamn}
\prod_{k} P_{newton}(t_{f,k}) > \prod_{k} P_{shaman}(t_{f,k})
\end{equation*}

\noindent
Where the $k$ index runs over many trials of the experiment.
Sophie then shows how the Newtonian probability predictions are both more \textit{confident}
	and more \textit{correct} than the shamanic predictions.
The Newtonian predictions assign a very large amount of probability
	to the region around the outcome $t_{f}=2$,
	and in fact it turns out that almost all of the real data outcomes fall in this range.	 
In contrast, the shamanic prediction assigns a relatively small amount 
	of probability to the $t_{f}=2$ region, 
	because he has predicted a very wide interval ($1 < t_{f} < 30$).
Thus while the shamanic prediction is correct, it is not very confident.
The Newtonian prediction is correct and highly confident,
	and so it should be prefered.
	
Sophie tries to emphasize that the Newtonian probability prediction $P_{newton}$
	only works well for the \textit{real} data.
Because of the requirement that probability distributions be normalized,
	the Newtonian theory can only achieve good results
	by reassigning probability towards the region around $t_{f} = 2$
	and away from other regions.
A theory that does not perform this kind of reassignment 	
	cannot achieve superior high performance.
	
Sophie recalls that some of the students are studying computer science
	and for their benefit points out the following.
The famous Shannon equation $L(x) = -\log_{2} P(x)$ 
	governs the relationship between the probability of an outcome
	and the length of the optimal code that should be used to represent it.
Therefore, 
	given a large data file containing the results of many trials of the ballistic motion experiment,
	the two predictions (Newtonian and shamanic) can both be used to 
	build specialized programs to compress the data file.
Using the Shannon equation, the above inequality can be rewritten as follows:

\begin{equation*}
%http://www.codecogs.com/png.latex?\sum_{k} L_{newton}(t_{f,k}) < \sum_{k} L_{shaman}(t_{f,k})
\sum_{k} L_{newton}(t_{f,k}) < \sum_{k} L_{shaman}(t_{f,k})
\end{equation*}

\noindent
This inequality indicates an alternative criterion
	that can be used to decide between two rival theories. 
Given a data file recording measurements related to a phenomenon of interest, 
	a scientific theory can be used to write a compression program
	that will shrink the file to a small size.
To decide between two rival theories of the same phenomenon,
	one invokes the corresponding compressors on a shared benchmark data set,
	and prefers the theory that achieves a smaller encoded file size.
This criterion is equivalent to the probability-based one,
	but has the advantage of being more concrete,
	since the quantities of interest are file lengths instead of probabilities.
	
\subsection{The Dead Experimentalist}

Sophie is a theoretical physicist and,
	upon taking up her position as assistant professor,
	began a collaboration with a brilliant experimental physicist 
	who had been working at the university for some time.
The experimentalist had previously completed the development 
	of an advanced apparatus that allowed the investigation of an
	exotic new kind of quantum phenomenon.
Using data obtained from the new system,	
	Sophie made rapid progress in developing a mathematical 
	theory of the phenomenon.
Tragically, 
	just before Sophie was able complete her theory,
	the experimentalist was killed in a laboratory explosion
	that also destroyed the special apparatus.
After grieving for a while,
	Sophie decided that the best way to honor her friend's memory
	would be to bring the research they had been working on to a successful conclusion.
	
Unfortunately, there is a critical problem with Sophie's plan.
The experimental apparatus had been completely destroyed,
	and Sophie's late partner was the only person in the world who could have rebuilt it.
He had run many trials of the system before his death,
	so Sophie had a quite large quantity of data.
But she had no way of generating any new data.
Thus, no matter how beautiful and perfect her theory might be,
	she had no way of testing it by making predictions.
	
One day while thinking about the problem 
	Sophie recalls the incident with the shaman.
She remembers the point she had made for the benefit of the software engineers,
	about how a scientific theory could be used to compress a real world data set
	to a very small size.
Inspired, she decides to apply the data compression principle
	as a way of testing her theory. 
She immediately returns to her office and spends 
	the next several weeks	writing Matlab code,
	converting her theory into a compression algorithm.
The resulting compressor is successful:
	it shrinks the corpus of experimental data from an initial size
	of $8.7 \cdot 10^{11}$ bits to an encoded size of $3.3 \cdot 10^{9}$ bits. 
Satisfied, Sophie writes up the theory, 
	and submits it to a well-known physics journal.

The journal editors like the theory, 
	but are a bit skeptical of the compression based method for testing the theory.
Sophie argues that if the theory becomes widely known,
	one of the other experts in the field
	will develop a similar apparatus, 
	which can then be used to test the theory in the traditional way.
She also offers to release the experimental data,
	so that other researchers can test their own theories
	using the same compression principle.
Finally she promises to release the source code of her program,
	to allow external verification of the compression result.
These arguments finally convince the journal editors to accept the paper.
	
\subsection{The Rival Theory}

After all the mathematics, software development,
	prose revisions, and persuasion necessary to complete her theory
	and have the paper accepted,
	Sophie decides to reward herself by living the good life for a while.
She is confident that her theory is essentially correct,
	and will eventually be recognized as correct by her colleagues.
So she spends her time reading novels and hanging out in coffee shops with her friends.

A couple of months later, however,
	she receives an unpleasant shock in the form of an email from a colleague
	which is phrased in consolatory language,
	but does not contain any clue as to why such language might be in order.
After some investigation she finds out that 
	a new paper has been published about the same quantum phenomenon of interest to Sophie.
The paper proposes a alternative theory of the phenomenon which 
	bears no resemblance whatever to Sophie's.
Furthermore, 
	the paper reports a better compression rate than was achieved by Sophie,
	on the database that she released.
	
Sophie reads the new paper and quickly realizes that it is worthless.
The theory depends on the introduction of a large number of additional parameters,
	the values of which must be obtained from the data itself.
In fact, a substantial portion of the paper involves
	a description of a statistical algorithm
	that estimates optimal parameter values from the data.
In spite of these aesthetic flaws,
	she finds that many of her colleagues are quite taken
	with the new paper and some consider it to be ``next big thing''.
Sophie sends a message to the journal editors describing in detail
	what she sees as the many flaws of the upstart paper.
The editors express sympathy,
	but point out that the new theory outperforms Sophie's theory
	using the performance metric she herself proposed.
The beauty of a theory is important,
	but its correctness is ultimately more important.

Somewhat discouraged, Sophie sends a polite email to the authors of the new paper,
	congratulating them on their result and asking to see their source code.
Their response, which arrives a week later,
	contains a vague excuse about how the source code is not properly documented 
	and relies on proprietary third party libraries.
Annoyed, Sophie contacts the journal editors again and asks them 
	for the program they used to verify the compression result.
They reply with a link to a binary version of the program.

When Sophie clicks on the link to download the program,
	she is annoyed to find it has a size of 800 megabytes.
But her annoyance is quickly transformed into enlightenment,
	as she realizes what happened, and that her previous philosophy contained a serious flaw.
The upstart theory is not better than hers;
	it has only succeeded in reducing the size of the encoded data 
	by dramatically increasing the size of the compressor.
Indeed, when dealing with specialized compressors,
	the distinction between ``program'' and ``encoded data'' becomes almost irrelevant.
The critical number is not the size of the compressed file,
	but the net size of the encoded data plus the compressor itself.
 
Sophie writes a response to the new paper 
	which describes the refined compression rate principle.
She begins the paper by reiterating the unfortunate circumstances
	which forced her to appeal to the principle,
	and expressing the hope that someday an experimental group
	will rebuild the apparatus developed by her late partner,
	so that the experimental predictions made by the two theories 
	can be properly tested.
Until that day arrives, standard scientific practice does not permit a decisive declaration
	of theoretical success.
But surely there is \textit{some} theoretical statement that can be made in the meantime,
	given the large quantity of data that is available.
Sophie's proposal is that the goal should be to find the theory 
	that has the highest probability of predicting a new data set,
	when it can finally be obtained.
If the theories are very simple in comparison to the data being modeled, 
	then the size of the encoded data file is a good way of 
	choosing the best theory.
But if the theories are complex,
	then there is a risk of \textit{overfitting} the data.
To guard against overfitting complex theories must be penalized;	
	a simple way to do this is to take into account
	the codelength required for the compressor itself.
The length of Sophie's compressor was negligible,	
	so the net score of her theory is just the codelength
	of the encoded data file: $3.3 \cdot 10^{9}$ bits.
The rival theory achieved a smaller size of $2.1 \cdot 10^{9}$ for the encoded data file,
	but required a compressor of $6.7 \cdot 10^{9}$ bits to do so,
	giving a total score of $8.8 \cdot 10^{9}$ bits.
Since Sophie's net score is lower, her theory should be prefered.
	
\mpagebreak

\section{Compression Rate Method}

In the course of the thought experiments discussed above,
	the protagonist Sophie articulated a refined
	version of the scientific method.
This procedure will be called the Compression Rate Method (CRM).
The web of concepts related to the CRM will be called the 
	\textit{\compist}~philosophy of science,
	for reasons that will become evident in the next section.
The CRM consists of the following steps:

\begin{enumerate}
\item Obtain a vast database $T$ relating to a phenomenon of interest.
\item Let $f_{C}$ be the initial champion theory.
\item Through observation and analysis,
	develop a new theory $f_{N}$,
	which may be either a simple refinement of $f_{C}$ 
	or something radically new.
\item Instantiate $f_{N}$ as a compression program.
	If this cannot be done, then the theory is not scientific.
\item Score the theory by calculating $L(T|f_{N}) + \mathbf{H}[f_{N}]$,
	the sum of the encoded version of $T$ and the length of the compressor.
\item If $L(T|f_{N}) + \mathbf{H}[f_{N}] < L(T|f_{C}) + \mathbf{H}[f_{C}]$,
	then discard the old champion and set $f_{C} = f_{N}$.
	Otherwise discard $f_{N}$.
\item Return to step \#3.
\end{enumerate}

It is worthwhile to compare the CRM
	to the version of the scientific method given in Section~\ref{sec:theryspace}.
One improvement is that in this version
	the Occam's Razor principle plays an explicit role,
	through the influence of the $\mathbf{H}$ term.
A solution to the Problem of Demarcation is also built into 
	the process in Step \#4.
The main difference is that the
	empirical ingredient in the CRM is a large database,
	while the traditional method employs experimental observation.
	
The significance of the CRM can be seen by understanding the relationship
	between the target database $T$ and the resulting theories.
If $T$ contains data related to the outcomes of physical experiments,
	then physical theories will be necessary to compress it.
If $T$ contains information related to interest rates, 
	house prices, global trade flows, and so on,
	then economic theories will be necessary to compress it.
One obvious choice for $T$ is simply an enormous image database,
	such as the one hosted by the Facebook social networking site. 
In order to compress such a database one must develop 
	theories of \textit{visual reality}.
The idea that there can be an empirical science of visual reality
	has never before been articulated,
	and is one of the central ideas of this book.
A key argument, contained in Chapter~\ref{chapt:comprvsion},
	is that the research resulting from the application of the CRM
	to a large database of natural images will
	produce a field very similar to modern computer vision.
Similarly, Chapter~\ref{chapt:linguistic} argues that
	the application of the CRM to a large text corpus
	will result in a field very similar to computational linguistics.
Furthermore,
	the reformulated versions of these fields will have
	far stronger philosophical foundations,
	due to the explicit connection between the CRM
	and the traditional scientific method.
	
It is crucial to emphasize the deep connection 
	between compression and prediction.	
The real goal of the CRM is to evaluate
	the predictive power of a theory,
	and the compression rate is just a way of quantifying that power.
There are three advantages to using the compression rate
	instead of some measure of predictive accuracy.
First, 
	the compression rate naturally accomodates a model complexity
	penalty term.
Second,
	the compression rate of a large database is an objective quantity,
	due to the ideas of Kolmogorov complexity 
	and universal computation, discussed below.
Third, the compression principle provides an 
	important verificational benefit.
To verify a claim made by an advocate of a new theory,
	a referee only needs to check the encoded file size,
	and ensure that the resulting decoded data
	matches exactly the original database $T$.	

Most people express skepticism as their first reaction to the plan 
	of research embodied by the CRM.
They generally admit that it may be possible to use the method 
	to obtain increasingly short codes for the 
	target databases.
But they balk at accepting the idea that the method
	will produce anything else of value.
The following sections argue that the philosophical commitments
	implied by the CRM are exactly analogous
	to those long accepted by scientists working
	in mainstream fields of empirical science.
\Compist~science is nonobvious in the year 2011
	for exactly the same kinds of reasons that
	empirical science was nonobvious in the year 1511.

\mpagebreak
	
\begin{figure*}
\centering
\subfigure{\includegraphics[width=.4\textwidth]{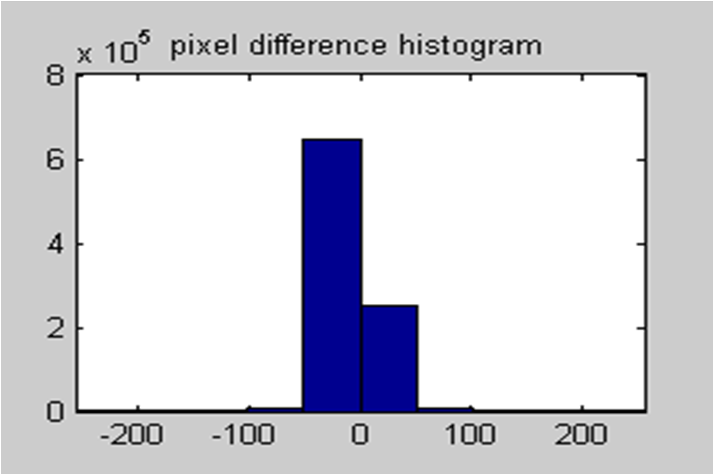}}
\subfigure{\includegraphics[width=.4\textwidth]{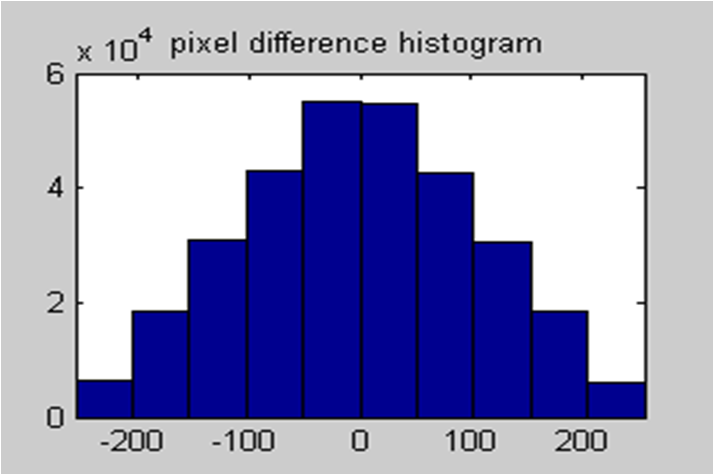}}
\caption[Pixel difference histograms for a natural and a random image.]{
Histograms of differences between values of neighboring pixels
	in a natural image (left) and a random image (right).
The clustering of the pixel difference values around 0 in the natural image
	is what allows compression formats like PNG to achieve compression.
Note the larger scale of the image on the left; 
	both histograms represent the same number of pixels.
}
\label{fig:pixeldiffn}
\end{figure*}

\subsection{Data Compression is Empirical Science}
	
The following theorem is well known in data compression.
Let $C$ be a program that losslessly compresses bit strings $s$,
	assigning each string to a new code with length $L_{C}(s)$.
Let $U_{N}(s)$ be the uniform distribution over $N$-bit strings.
Then the following bound holds for all compression programs $C$:

\begin{equation}
E_{(s \sim U_{N})}[L_{C}(s)] \geq N
\end{equation}

\noindent
In words the theorem states that
	no lossless compression program can achieve average codelengths smaller than $N$ bits,
	when averaged over all possible $N$ bit input strings.
Below,
	this statement is referred to as the ``No Free Lunch'' (NFL) theorem
	of data compression
	as it implies that one can achieve compression for some strings $s$
	only at the price of inflating other strings.
At first glance, 
	this theorem appears to turn the CRM proposal into nonsense.
In fact, the theorem is the keystone of the \compist~philosophy
	because it shows how lossless, large-scale compression
	research must be essentially empirical in character.
	
To see this point, consider the following apparent paradox.
In spite of the NFL theorem, 
	lossless image compression programs exist and have been in widespread use for years.
As an example, the well-known Portable Network Graphics (PNG) compression algorithm
	seems to reliably produce encoded files that are 
	40-50\% shorter than would be achieved by a uniform encoding.
This apparent success seems to violate the No Free Lunch theorem. 	

The paradox is resolved by noticing that the images used to evaluate
	image compression algorithms are not drawn from a uniform
	distribution $U_{N}(s)$ over images.
If lossless image formats were evaluated based on 
	their ability to compress random images, 
	no such format could ever be judged successful.
Instead, the images used in the evaluation process
	belong to a very special subset of all possible images:
	those that arise as a result of everyday human photography.
This ``real world'' image subset, though vast in absolute terms,
	is miniscule compared to the space of all possible images.
So PNG is able to compress a certain image subset,
	while inflating all other images.
And the subset that PNG is able to compress
	happens to overlap substantially with the real world image subset.
	
The specific empirical regularity used
	by the PNG format is that in real world images,
	adjacent pixel values tend to have very similar values.
A compressor can exploit this property by encoding the differences between neighboring
	pixel values instead of the values themselves.
The distribution of differences is very narrowly clustered around zero,
	so they can be encoded using shorter average codes (see Figure~\ref{fig:pixeldiffn}).
Of course, this trick does not work for random images,
	in which there is no correlation between adjacent pixels.

The NFL theorem indicates that in order to succeed,
	a \compist researcher must follow 
	a strategy analogous to the procedure of physics.
First, she must attempt to discover some 
	structures or patterns present in real world images.
Then she must develop a mathematical theory
	characterizing that structure,
	and build the theory into a compressor.
Finally, she must demonstrate that the theory
	corresponds to reality,
	by showing that it achieves an improved compression rate.

To make statements about the world,
	physicists need to combine mathematical and empirical reasoning;
	neither alone is sufficient.
Consider the following statement of physics:
	when a ball is tossed into the air,
	its vertical position will be described by the equation:
	$y(t) = g t^2 + v_{0} t + y_{0}$.
That statement can be decomposed into a mathematical and an empirical component.
The mathematical statement is:
	\textit{if} a quantity's evolution in time is governed by the differential 
	equation $\frac{d^2 y}{d t^2} = k$, where $k$ is some constant,
	\textit{then} its value is given by the function $y(t) = k t^2 + v_{0} t + y_{0}$,
	where $v_{0}$ and $y_{0}$ are determined by the initial conditions.
The empirical statement is:
	if a ball is thrown into the air, its vertical position 
	will be governed by the differential equation $\frac{d^2 y}{d t^2} = g$,
	where $g$ is the acceleration due to gravity.
By combining these statements together,
	the physicist is able to make a variety of predictions.
	
Just like physicists,
	\compist~researchers must combine
	mathematical statements with empirical statements
	in order to make predictions.
Because of the NFL theorem,
	pure mathematics is never sufficient to reach conclusions of the form:
	``Algorithm Y achieves good compression.''
Mathematical reasoning can only be used to make implications:
	``If the images exhibit property X, 
	then algorithm Y will achieve good compression''.
In order to actually achieve compression,
	it is necessary to demonstrate the empirical fact
	that the images actually have property X.
This shows why the \compist~proposal is not fundamentally about saving disk space or bandwidth;
	it is fundamentally about characterizing the properties
	of images or other types of data.

\mpagebreak

\subsection{Comparison to Popperian Philosophy}

The \compist~philosophy of science bears a strong family resemblance to the Popperian one,
	and inherits many of its conceptual advantages.
First, the compression principle provides a clear answer to the Problem of Demarcation:
	a theory is scientific if and only if it can be 
	used to build a compressor for an appropriate kind of database.
Because of the intrinsic difficulty of lossless data compression,
	the only way to save bits is to explicitly reassign probability
	away from some outcomes and toward other outcomes.
If the theory assigns very low probability to an outcome
	which then occurs,
	this suggests that the theory has low quality and should be discarded.
Thus, the probability reassignment requirement is 
	just a graduated or continuous version of the falsification requirement.	
The falsifiability principle means that a researcher hoping to prove the value of his new theory 
	must risk embarassment if his predictions turn out to be incorrect.
The compression principle requires a researcher 
	to face the potential for embarassment 
	if his new theory ends up \textit{inflating} the database.

One difference between Popperian view and \compist~view
	is that the former appears to justify stark binary assessments
	regarding the truth or falsehood of a theory,
	while the latter provides only a number which can be compared to other numbers.
If theories are either true or false, 
	then the compression principle is no more useful than the falsifiability principle.
But if theories can exist on some middle ground between absolute truth and its opposite,
	then it makes sense to claim that one theory is relatively more true than another,
	even if both are imperfect.
The compression principle can be used to justify such claims.
Falsifiability consigns all imperfect theories to the same garbage bin;
	compression can be used to rescue the valuable theories from the bin,
	dust them off, and establish them as legitimate science.

The falsifiability idea seems to imply that theories can be evaluated in isolation:
	a theory is either true or false,
	and this assessment does not depend on the content of rival theories.
In contrast, while the compression idea
	assigns a score to an individual theory,
	this score is useful only for the purpose of comparison.
This distinction may be conceptually significant to some people,
	but in practice it is unimportant.
Science is a search for good approximations;
	science proceeds by incrementally improving the quality of the approximations.
The power of the falsifiability requirement is that it 
	enables a rapid search through the theory-space
	by ensuring that theories can be decisively compared.
The compression requirement provides exactly the same benefit.
When a researcher proposes a new theory and shows that it
	can achieve a smaller compressed file size for the target database,
	this provides decisive evidence that the new theory is superior.
Furthermore, both principles allow a research community to identify
	a champion theory.
In the Popperian view,
	the champion theory is the one that has withstood all attempts at falsification.
In the \compist~view,
	the champion theory is the one that achieves the smallest codelength
	on the relevant benchmark database.
	
One of the core elements of Popper's philosophy
	is the dedication to the continual testing,
	examination, and skepticism of scientific theories.
A Popperian scientist is never content 
	with the state of his knowledge.
He never claims that a theory is true;
	he only accepts that there is currently no evidence
	that would falsify it.
The \compist~philosopher takes an entirely analogous stance.
To her, a theory is never true or even optimal,
	it is only the best theory that has thus far been discovered.
She will never claim,
	``the probability of event X is 35\%''.
Instead, 
	she would state that 
	``according to the current champion theory, the probability of event X is 35\%''.
She might even make decisions based on this probability assignment.
But if a new theory arrives that provides a better codelength,
	she immediately replaces her probability estimates 
	and updates her decision policy 
	based on the new theory.
	
The Popperian commitment to continual examination and criticism of theoretical knowledge 
	is good discipline,
	but the radical skepticism it promotes is probably a bit too extreme. 
A strict Popperian would be unwilling to use Newtonian physics
	once it was falsified,
	in spite of the fact that it obviously still works 
	for most problems of practical interest.
The compression principle promotes a more nuanced view.
If a claim is made that a theory provides a good description of a certain phenomenon, 
	and the claim is justified by demonstrating a strong compression result, 
	then the claim is valid for all time. 
It is possible to develop a new theory that achieves a better compression rate,
	or to show that the previous theory does not do as well on another related database.
These circumstances might suggest that the old theory should no longer be used.
But if the old theory provided a good description
	of a particular database,
	no future developments will change that fact.
This captures the intuition that Newtonian physics
	still provides a perfectly adequate description
	of a wide range of phenomena;
	Eddington's solar eclipse photographs 
	simply showed that there are some phenomena to which
	it does not apply.

\mpagebreak

\subsection{Circularity and Reusability in Context of Data Compression}

Just like empirical scientists do,
	\compist~researchers adopt a the Circularity Commitment
	to guide and focus their efforts.
A \compist~researcher
	evaluates a new theory based on one and only one criterion:
	its ability to compress the database for which it was developed.
A community using a large collection of face images
	will be highly interested in various tools
	such as hair models, eyeglass detectors, and theories of lip color,
	and only secondarily interested in potential
	applications of face modeling technology.
If the researchers chose to introduce some additional considerations 
	to the theory comparison process,
	such as the relevance of a theory to a certain type of practical task,
	they would compromise their own ability to discard
	low quality theories and identify high quality ones.

Some truly purist thinkers
	may consider large scale data compression as an intrinsically interesting goal.
\Compist~researchers will face many challenging problems,
	involving mathematics, algorithm design, 
	statistical inference, and knowledge representation.
Furthermore, researchers will receive a clear signal indicating
	when they have made progress, and how much.
For a certain type of intellectual,
	these considerations are very significant,
	even if there is no reason to believe that the investigation
	will yield any practical results.
	
In this light,
	it is worth comparing the proposed field of large scale lossless data compression 
	with the established field of computer chess.
Chess is an abstract symbolic game with very little connection to the real world.
A computer chess advocate would find it quite difficult to convince
	a skeptical audience that constructing powerful chess
	programs would yield any tangible benefit.
However, like the compression goal,
	the computer chess goal is attractive
	because it produces a variety of subproblems,
	and also provides a method for making decisive comparisons
	between rival solutions.
For these reasons, 
	computer scientists devoted a significant amount of effort to the field,
	leading some to claim that chess was ``the drosophila of AI research''.
Furthermore, 
	these efforts were incredibly successful,
	and led to the historic defeat of the top ranked human grandmaster,
	Gary Kasparov, by IBM's Deep Blue in 1997.
Most scientists would agree that this event was an important advance
	for human knowledge,
	even if it did not lead to any practical applications.
Because of its similar methodological advantages,
	\compist~research has a similar potential to advance
	human knowledge.
	
For the reader who is unmoved by the argument about the intrinsic interest
	of compression science,
	it is essential to defend the validity of the Reusability Hypothesis 
	in the context of data compression.
The hypothesis really contains two separate pieces.
First,
	theories employ abstractions,
	and good theories use abstractions that correspond to reality.
So the abstraction called ``mass'' is not just a clever computational trick,
	but represents a fundamental aspect of reality.
These real abstractions are useful \textit{both} for 
	compression and for practical applications.
The second piece of the Reusability Hypothesis is that,
	while theories based on \naive~or simplistic characterizations of reality
	can achieve compression,
	the \textit{best} codelengths will be achieved by
	theories that use real abstractions.
So by vigorously pursuing the compression goal,
	researchers can identify the real abstractions governing
	a particular phenomenon,
	and those abstractions can be reused for practical applications.
	
The following examples illustrate
	the idea of the Reusability Hypothesis.
Consider constructing a target database by setting up	
	a video camera next to a highway and recording the resulting image stream.
One way to predict image frames
	(and thus compress the data)
	would be to identify batches of pixels corresponding to a car, 
	and use an estimate of the car's velocity to interpolate the pixels forward.
A compressor that uses this trick thus implicitly contains
	abstractions related to the concepts of ``car'' and ``velocity''.
Since these are real abstractions, 
	the Reusability Hypothesis states that the specialized compressor
	should achieve better compression rates than a more generic one.
Another good example of this idea relates to text compression.
Here, the Reusability Hypothesis states that 
	a specialized compressor making use of abstractions such
	as verb conjugation patterns, parts of speech,
	and rules of grammar will perform better than a generic compressor.
If the hypothesis is true,
	then the same division of labor
	between scientists and engineers 
	that works for mainstream fields will work here as well.
The \compist~scientists obtain various abstractions by following the compression principle,
	and hand them off to the engineers,
	who will find them very useful for developing applications
	like automatic license plate readers and machine translation systems.

\mpagebreak

\subsection{The Invisible Summit}

An important concept related to the Compression Rate Method is 
	called the Kolmogorov complexity.
The Kolmogorov complexity $K_{A}(s)$ of a string $s$ is the length
	of the shortest program that will output $s$ when
	run on a Turing machine $A$.
The key property of the Kolmogorov complexity
	comes about as a consequence 
	of the idea of universal computation.
If a Turing machine (roughly equivalent to a programming language)
	is of sufficient complexity,
	it becomes \textit{universal}: 
	it can simulate any other Turing machine,
	if given the right simulator program.
So given a string $s$ and a short program $P_{A}$
	that outputs it when run on Turing machine A,
	one can easily obtain a program $P_{B}$
	that outputs $s$ when run on (universal) Turing machine B,
	just by prepending a simulator program $S_{AB}$ to $P_{A}$,
	and $|P_{B}| = |S_{AB}| + |P_{A}|$.
Now, the simulator program is fixed by the definition of the two Turing machines.
Thus for very long and complex strings,
	the contribution of the simulator to the total program length becomes insignificant,
	so that $|P_{B}| \approx |P_{A}|$,
	and thus the Kolmogorov complexity is effectively \textit{independent}
	of the choice of Turing machine.

Unfortunately or not,
	a brief proof shows that the Kolmogorov complexity is incomputable:
	a program attempting to compute $K(s)$ cannot be guaranteed
	to terminate in finite time.
This is not surprising,
	since if a method for computing the Kolmogorov complexity were found,
	it would be immensely powerful.
Such a program would render theoretical physicists unnecessary.
Experimental physicists could simply compile a large database of observations,
	and feed the database to the program.
Since the optimal theory of physics corresponds provides the best explanation,
	and thus the shortest encoding, of the data,
	the program would automatically find the optimal theory of physics
	on its way to finding the Kolmogorov complexity.
	
Another way of seeing the impossibility of finding $K(s)$
	is by imagining what it would mean to find the Kolmogorov complexity
	of the Facebook image database.
To compress this database to the smallest possible size,
	one would have to know $P^{*}(I)$ : 
	the probability distribution generating the Facebook images.
While $P^{*}(I)$ may look innocuous,
	in fact it is a mathematical object of vast complexity,
	containing an innumerable quantity of details.
To begin with,
	it must contain a highly sophisticated model of the human face.
It must contain knowledge of hair styles and facial expressions.
It must capture the fact that lips are usually reddish in color,
	and that women are more likely to enhance this color using lipstick.
Moving on from there, 
	it would require knowledge about other things people like to photograph,
	such as pets, natural scenery, weddings, and boisterous parties.
It would need to contain details about the appearance of babies,
	such as the fact that a baby usually has a pink face,
	and its head is large in proportion to the rest of its body.
All this knowledge is necessary because,
	for example, $P^{*}(I)$ must assign higher probability, and shorter codelength,
	to an image featuring a woman with red lips,
	than to an image that is identical in every
	way except that the woman has green lips.
	
While calculating $K(s)$ is impossible in general,
	one \textit{can} find upper bounds to it.
Indeed, the Compression Rate Method is just
	the process of finding a sequence of increasingly tight
	upper bounds on the Kolmogorov complexity of the target database.
Each new champion theory corresponds to a tighter upper bound.
In the case of images,
	a new champion theory corresponds to 
	to a new model $P_{C}(I)$ of the probability of an image.
Every iteration of theory refinement 
	packages more realistic information into the model $P_{C}(I)$,
	thereby bringing it closer to the unknowable $P^{*}(I)$.
This process is exactly analogous
	to the search through the theory space carried 
	out by empirical scientists.
Both empirical scientists and \compist~scientists
	recognize that their theories are mere approximations.
The fact that perfect truth cannot be obtained simply does not matter:
	it is still worthwhile to climb towards the invisible summit.

\mpagebreak
	
\subsection{Objective Statistics}

Due to the direct relationship between 
	statistical modeling and data compression (see Appendix~\ref{chapt:infoappend}),
	\compist~research can be regarded as a subfield of statistics.
A traditional problem in statistics starts
	with a set of $N$ observations $\{x_{1}, x_{2} \ldots x_{N}\}$ of some quantity,
	such as the physical height of a population.
By analyzing the data set,
	the statistician attempts to obtain a good
	estimate $P(x)$ of the probability of a given height.
This model could be,
	for example, a Gaussian distribution with a given mean and variance.
\Compist~research involves an entirely analogous process.
The difference is that instead of simple single-dimensional numbers,
	\compist~statisticians analyze complex data objects
	such as images or sentences,
	and attempt to find good models of the probability of such objects.

All statistical inference must face a deep conceptual issue
	that has been the subject of acrimonious debate
	and philosophical speculation since the time of David Hume,
	who first identified it.
This is the Problem of Induction:
	when is it justified to jump from a limited set of specific observations
	(the data samples) to a universal rule describing
	the observations (the model)?
This problem has divided statisticians into two camps,
	the Bayesians and the frequentists,
	who disagree fundamentally about the meaning and justification
	of statistical inference.
A full analysis of the nature of this disagreement would require its own book,
	but a very rough summary is that, 
	while the Bayesian approach has a number of conceptual benefits,
	it is hobbled by its dependence on the use of \textit{prior} distributions.
A Bayesian performs inference
	by using Bayes rule to update a prior distribution in response to evidence,
	thus producing a posterior distribution,
	which can be used for decision-making and other purposes.
The critical problem is that is no objective way to choose a prior.
Furthermore, 
	two Bayesians who start with different priors
	will reach different conclusions,
	in spite of observing the same evidence.
The use of Bayesian techniques to justify scientific conclusions
	therefore deprives science of objectivity.
		
Any data compressor must implement a mapping
	from data sets $T$ to bit strings of length $L(T)$.
This mapping defines
	an implicit probability distribution $P(T) = 2^{-L(T)}$.
It appears,
	therefore, that \compist~statisticians make the same
	commitment to the use of prior distributions
	as the Bayesians do.
However, there is a crucial subtlety here.	
Because the length of the compressor itself is taken
	into account in the CRM,
	the prior distribution is actually defined by 
	the choice of programming language
	used to write the compressor.
Furthermore,
	\compist~researchers use their models to describe
	vast datasets.
Combined, 
	these two facts imply that \compist~statistical inference is objective.
This idea is illustrated by the following thought experiment.
	
Imagine a research subfield which has established a database $T$
	as its target for CRM-style investigation.
The subfield makes slow but steady progress for several years.
Then, out of the blue,
	an unemployed autodidact from a rural village in India
	appears with a bold new theory.
He claims that his theory, instantiated in a program $P_{A}$,
	achieves a compression rate which is dramatically superior
	to the current best published results.
However, among his other eccentricities, 
	this gentleman uses a programming language he himself developed,
	which corresponds to a Turing machine $A$.
Now, the other researchers of the field are well-meaning but skeptical,
	since all the previously published results used a standard language 
	corresponding to a Turing machine $B$.
But it is easy for the Indian maverick to produce a compressor
	that will run on $B$: 
	he simply appends $P_{A}$ to a simulator program $S_{AB}$,
	that simulates $A$ when run on $B$.
The length of the new compressor is $|P_{B}| = |P_{A}| + |S_{AB}|$,
	and all of the other researchers can confirm this.
Now, assuming the data set $T$ is large and complex enough
	so that $|P_{A}| \gg |S_{AB}|$,
	then the codelength of the modified version is 
	effectively the same as the original: $|P_{B}| \approx |P_{A}|$.
This shows that there can be no fundamental disagreement
	among \compist~researchers regarding the quality of a new result.
	
\mpagebreak
	
\section{Example Inquiries}

This section makes the makes the abstract discussion above 
	tangible by describing several concrete proposals.
These proposals begin with a method of constructing a target database,
	which defines a line of inquiry.
In principle, researchers can use any large database 
	that is not completely random 
	as a starting point for a \compist~investigation.
In practice, 
	unless some care is exercised in the construction of the target dataset,
	it will be difficult to make progress.
In the beginning stages of research,
	it will be more productive to look at data sources
	which display relatively limited amounts of variation.
Here are some examples inquiries that might provide 
	good starting points:
	
\begin{itemize}
\item Attempt to compress the immense image database hosted	
	by the popular Facebook social networking web site.
One obvious property of these images is that they contain 
	many faces.
To compress them well,
	it will be necessary to develop a computational understanding 
	of the appearance of faces.
\item Construct a target database by packaging together
	digital recordings of songs, concerts, symphonies, opera,
	and other pieces of music. 
This kind of inquiry will lead to theories of the structure of music,
	which must describe harmony, melody, pitch, rhythm 
	and the relationship between these variables
	in different musical cultures.
It must also contain models of the sounds produced by different instruments,
	as well as the human singing voice.
\item Build a target database by recording from microphones positioned
	in treetops. 
A major source of variation in the resulting data will be bird vocalizations.
To compress the data well, 
	it will be necessary 
	to differentiate between bird songs and bird calls,
	to develop tools that can identify species-characteristic vocalizations,
	and to build maps showing the typical ranges of various species.
In other words, this type of inquiry will be a computational version
	of the traditional study of bird vocalization carried out by ornithologists.
\item Generate a huge database of economic data showing changes in home prices,
	interest and exchange rate fluctuations, business inventories,
	welfare and unemployment applications, and so on.
To compress this database well,
	it will be necessary to develop economic theories
	that are capable of predicting, for example,
	the effect that changes in interest rates have on home purchases.
\end{itemize} 

Since the above examples involve empirical inquiry
	into various aspects of reality,
	any reader who believes in the intrinsic value of science
	should regard them as at least potentially interesting.
Skeptical readers, on the other hand,
	may doubt the applicability of the Reusability Hypothesis here,
	and so view an attempt to compress these databases
	as an eccentric philosophical quest.
The following examples are more detailed,
	and give explicit analysis of what kinds of theories
	(or computational tools) will be needed,
	and how those theories will be more widely useful.
An important point,
	common to all of the investigations,
	is that a single target database can be used to develop
	and evaluate a large number of methods.

It should be clear that,
	if successful, these example inquiries 
	should lead to practical applications.
The study of music may help composers to write better music,
	allow listeners to find new music that suits their taste,
	and assist music publishing companies in determining
	the quality of a new piece.
The investigation of bird vocalization, if successful, 
	should be useful to environmentalists and bird-watchers
	who might want to monitor the migration and population fluctuation
	of various avian species.
The study of economic data is more speculative,
	but if successful should be of obvious interest to 
	policy makers and investors.
In the case of the roadside video data described below,
	the result will be sophisticated visual systems
	that can be used in robotic cars.
Also mentioned below is an inquiry into the structure of English text,
	which should prove useful for speech recognition
	as well as for machine translation.

\mpagebreak
	
\subsection{Roadside Video Camera}
\label{sec:cncrtprpsl}

Consider constructing a target database by setting up a video camera
	next to a highway,
	and recording video streams of the passing cars.
Since the camera does not move, 
	and there is usually not much activity on the sides of highways, 
	the main source of variation in the resulting video will be the automobiles.
Therefore, in order to compress the video stream well, 
	it will be necessary to obtain a good computational understanding 
	of the appearance of automobiles.
	
A simple first step would be to take advantage of the fact that cars
	are rigid bodies subject to Newtonian laws of physics.
The position and velocity of a car must be continuous functions of time.
Given a series of images at timesteps $\{t_{0}, t_{1}, t_{2} \ldots t_{n}\}$
	it is possible to predict the image at timestep $t_{n+1}$
	simply by isolating the moving pixels in the series 
	(these correspond to the car), 
	and interpolating those pixels forward into the new image,
	using basic rules of camera geometry and calculus.
Since neither the background nor the moving pixel blob changes much between frames, 
	it should be possible to achieve a good compression
	rate using this simple trick.

Further improvements can be achieved by detecting and exploiting patterns
	in the blob of moving pixels.
One observation is that the wheels of a moving car have
	a simple characteristic appearance:
	a dark outer ring corresponding to the tire,
	along with the off-white circle of the hubcap at the center.
Because of this characteristic pattern,
	it should be straightforward to build a wheel detector
	using standard techniques of supervised learning.
One could then save bits by representing the wheel pixels
	using a specialized model,
	akin to a graphics program,
	which draws a wheel of a given size and position.
Since it takes fewer bits to encode the size and position parameters
	than to encode the raw pixels of the wheel,
	this trick should save codelength.
Further progress could be achieved by conducting 
	a study of the characteristic appearance of the surfaces of cars.
Since most cars are painted in a single color,
	it should be possible to develop a specialized algorithm
	to identify the frame of the car.
Another graphics program could be used to draw the frame of the car,
	using a variety of parameters related to its shape.
Extra attention would be required to handle the complex reflective appearance
	of the windshield,
	but the same general idea would apply.
Note that the encoder always has the option of ``backing off'';
	if attempts to apply more aggressive encoding methods fail
	(e.g., if the car is painted in multiple colors),
	then the simpler pixel-blob encoding method can be used instead.
	
% TODO: need to figure out synonyms for car "model", and another word
% to use instead of "module".
Additional progress could be achieved by recognizing that
	most automobiles can be categorized 
	into a discrete set of categories (e.g., a 2009 Toyota Corolla).
Since these categories have standardized dimensions,
	bits could be saved by encoding the category of a car
	instead of information related to its shape.
Initially, the process of building category-specific modules 
	for the appearance of a car might be difficult and time-consuming.
But once one has developed modules for the Hyundai Sonata, Chevrolet Equinox,
	Honda Civic, and Nissan Altima,
	it should not require much additional work to construct
	a module for the Toyota Sienna.
Indeed, it may be possible to develop a learning algorithm that,
	through some sort of clustering process,
	would automatically extract,
	from large quantities of roadside video data,
	appearance modules for the various car categories.
	
\mpagebreak
	
\subsection{English Text Corpus}

Books and other written materials constitute another interesting source
	of target data for~\compist~inquiry.
Here one simply obtains a large quantity of text,
	and attempts to compress it.
One tool that will be very useful for the compression of English text
	is an English dictionary.
To see this, consider the following sentence:

\begin{quote}
John went to the liquor store and bought a bottle of \_\_\_\_.
\end{quote}

Assume that the word in the blank space has $N$ letters,
	and the compressor encodes this information separately.
A \naive~compressor would require $\log (26^N) = N \log 26$ bits to encode the word,
	since there are $26^N$ ways to form an $N$-letter word.
A compressor equipped with a dictionary can do much better.
First it looks up all the words of length $N$,
	and then it encodes the index of the actual word
	in this list.
This costs $\log (W_{N})$, where $W_{N}$ is the number of words of
	length $N$ in the dictionary.
Since most combinations of letters such as ``yttu'' and ``qwhg''
	are not real words, $W_{N} < 26^N$ and bits are saved.

By making the compressor smart,
	it's possible to do even better.
A smart compressor should know that the word ``of'' is usually followed by a noun.
So instead of looking up all the $N$-letter words,
	the compressor could restrict the search to only nouns.
This cuts down the number of possibilities even further,
	saving more bits.
An even smarter compressor
	would know that in the phrase ``bottle of X'',
	the word X is usually a liquid.
If it had an enhanced dictionary which contained information about
	various properties of nouns,
	it could restrict the search to $N$-letter nouns that represent liquids.
Even better results could be obtained by noticing that the bottle is purchased
	at a liquor store, and so probably represents some kind of alcohol.
This trick would require that the enhanced dictionary contains
	annotations indicating that words such as ``wine'',
	``beer'', ``vodka'', are types of alcoholic beverages.
It may be possible to do even better by analyzing the surrounding text.
The word list may be narrowed even further
	if the text indicates that John is fond of brandy,
	or that his wife is using a recipe that calls for vodka.
Of course, these more advanced schemes are far beyond the current 
	state of the art in natural language processing,
	but they indicate the wide array of techniques that can in theory
	be brought to bear on the problem.

\mpagebreak
	
\subsection{Visual Manhattan Project}
\label{sec:vismanproj}

Consider constructing a database target
	by mounting video cameras on the dashboards 
	of a number of New York City taxi cabs,
	and recording the resulting video streams.
Owing to the vivid visual environment of New York City,
	such a database would exhibit an immense amount of complexity and variation.
Several aspects of that complexity could be then analyzed and studied in depth.

One interesting source of variation in the video would come from the pedestrians.
To achieve good compression rates for the pixels representing pedestrians,
	it would be necessary to develop theories
	describing the appearance of New Yorkers.
These theories would need to include details about
	clothing, ethnicity, facial appearance, hair style, walking style, 
	and the relationship between these variables.
A truly sophisticated theory of pedestrians would need to take into
	account time and place:
	it is quite likely to observe a suited investment banker in the financial district on a weekday afternoon,
	but quite unlikely to observe such a person in the Bronx in the middle of the night.

Another source of variation would come from the building and storefronts of the city.
A first steps towards achieving a good compression rate for these pixels
	would be to construct a three-dimensional model of the city.
Such a model could be used not only to determine the location 
	from which an image frame was taken,
	but also to predict the next frame in the sequence.
For example, the model could be used to predict that,
	if a picture is taken at the corner of 34th Street and Fifth Avenue,
	the Empire State Building will feature very prominently.
Notice that a \naive~representation of the 3D model 
	will require a large number of bits to specify,
	and so even more savings can be achieved by compressing
	the model itself.
This can be done by analyzing the appearance of typical building
	surfaces such as brick, concrete, and glass.
This type of research might find common ground 
	with the field of architecture, 
	and lead to productive interdisciplinary investigations.
	
A third source of variation would come from the other cars.
Analyzing this source of variation would lead to an investigation
	very similar to the roadside video camera inquiry mentioned above.
Indeed, if the roadside video researchers are successful,
	it should be possible for the taxi cab video researchers
	to reuse many of their results.
In this way, researchers can proceed in a virtuous circle,
	where each new advance facilitates the next line of study.
	
\mpagebreak
	
\section{Sampling and Simulation}

Sampling is a technique whereby one uses a statistical model
	to generate a data set that is ``typical'' of it.
For example, imagine one knows that the distribution
	of heights in a certain population is a Gaussian 
	with a mean of 175 cm and a standard deviation of 10 cm.
Then by sampling from a Gaussian distribution with these parameters,
	one obtains a set of numbers that are similar
	to what might be observed if some actual measurements were done.
Most of the data would cluster in the 165-185 cm range,
	and it would be extremely rare to observe a sample larger than 205 cm.

The idea of sampling suggests a useful technique
	for determining the quality of a statistical model:
	one samples from the model,
	and compares the sample data to the real data.
If the sample data looks nothing like the real data,
	then there is a flaw in the model.
In the case of one-dimensional numerical data 
	this trick is not very useful.
But if the data is complex and high-dimensional,
	and humans have a good understanding of its real structure,
	the technique can be quite powerful.
As an example of this, 
	consider the following two batches of pseudo-words:
	
\begin{quote}
a
abangivesery
ad
allars
ambed
amyorsagichou
an
and
anendouathin
anth
ar
as
at
ate
atompasey
averean
cath
ce
d
dea
dr
e
ed
eeaind
eld
enerd
ens
er
evedof
fod
fre
g
gand
gho
gisponeshe
greastoreta
har
has
haspy
he
heico
ho
ig
iginse
ill
ilyo
in
ind
io
is
ite
iter
itwat
ju
k
le
lene
lilollind
lliche
llkee
ly
mang
me
mee
mpichmm
n
nd
nder
ng
ngobou
nif
nl
noved
o
ond
onghe
oounin
oreengst
otaserethe
oua
ptrathe
r
rd
re
reed
reroved
sern
sinttlof
suikngmm
t
tato
tcho
te
th
the
toungsshes
ver
wit
y
ythe

\end{quote}

\begin{quote}
a
ally
anctyough
and
andsaid
anot
as
aslatay
astect
be
beeany
been
bott
bout
but
camed
chave
comuperain
deas
dook
ed
eveny
fel
filear
firgut
for
fromed
gat
gin
give
givesed
got
ha
hard
he
hef
her
heree
hilpte
hoce
hof
ierty
imber
in
it
jor
like
lo
lome
lost
mader
mare
mise
moread
od
of
om
ome
onertelf
our
out
over
owd
pass
put
qu
rown
says
seectusier
seeked
she
shim
so
soomereand
sse
such
tail
the
thingse
tite
to
tor
tre
tro
uf
ughe
umily
upeeperlyses
upoid
was
wat
we
were
wers
whith
wird
wirt
with
wor

\end{quote}

These words were created by sampling from two different 
	models $P(\alpha_{i}|\alpha_{i-1}, \ldots \alpha_{1})$
	of the conditional probability of a letter given a history of preceding letters.
The variable $\alpha_{i}$ stands for the $i$th letter of the word.
To produce a word,
	one obtains the first letter by sampling from 
	the unconditional distribution $P(\alpha_{1})$.
Then one samples from $P(\alpha_{2}|\alpha_{1})$
	to produce the second letter, 
	and so on.
A special word-ending character is added to the alphabet,
	and when this character is drawn, the word is complete.

The two models were both constructed using a large corpus of English text.
The first model is a simplistic bigram model,
	where the probability of a letter depends only 
	on the immediately preceding letter.
The second model is an enhanced version of the bigram model,
	which uses a refined statistical characterization of English words,
	that incorporates, for example,
	the fact that it is very unlikely for a word to have no vowel.
Most people will agree that the words from the second set
	are more similar to real English words
	(indeed, several of them are real words).
This perceptual assessment justifies the conclusion that the second model is in some
	sense superior to the first model.
Happily, it turns out that the second model also 
	achieves a better compression rate than the first model,
	so the qualitative similarity principle agrees
	with the quantitative compression principle.
While the second model is better than the first,
	it still contains imperfections.
One such imperfection relates to the word ``sse''.
The double-s pattern is common in English words,
	but it is never used to begin a word.
It should be possible to achieve improved
	compression rates by correcting this deficiency in the model.
	
All compressors implicitly contain a statistical model,
	and it is easy to sample from this model.
To do so one simply generates 
	a random bit string
	and feeds it into the decoder.
Unless the decoder is trivially suboptimal,
	it will map any string of bits to a legitimate outcome
	in the original data space.
This perspective provides a nice interpretation of what compression means.
An ideal encoder maps real data
	to perfectly random bit strings,
	and the corresponding decoder maps random bit strings to real data.

\mpagebreak
	
\subsection{Veridical Simulation Principle of Science}

Modern video games often attempt to illustrate scenes involving complex physical processes,
	such as explosions,
	light reflections, 
	or collisions between nonrigid bodies (e.g. football players).
In order to make these scenes look realistic,
	video game developers need to include ``physics engines''
	in their games.
A physics engine is a program that simulates 
	various processes using the laws of physics.
If the physics used in the simulators did not correspond to real physics,
	the scenes would look unrealistic:
	the colliding players would fall too slowly, 
	or the surface of a lake would not produce an appropriate reflection.

This implies that there is a connection between scientific theories
	and veridical simulation.
Can this principle be generalized?
Suspend disbelief for a moment and imagine that,
	perhaps as a result of patronage from an advanced alien race,
	humans had obtained computers before the development of physics.
Then scientists could conduct a search for a good theory of mechanics
	using the following method.
First, they would write down a new candidate theory.
Then they would build a simulator based on the theory,
	and use the simulator to generate various scenes,
	such athletes jumping,
	rocks colliding in mid-air,
	and water spurting from fountains.
The new theory would be accepted and the old champion discarded
	if the former produced more realistic 
	simulations than the latter.
	
As a more plausible example,
	consider using the simulation principle to guide
	an inquiry into the rules of grammar and linguistics.
Here the researchers write down candidate theories
	of linguistics,
	and use the corresponding simulator to generate sentences.
A new theory is accepted if the sentences it generates
	are more realistic and natural than those 
	produced by the previous champion theory.
This is actually very similar to Chomsky's 
	formulation of the goal of generative grammar;
	see Chapter~\ref{chapt:linguistic} for further discussion.
	
This notion of science appears to meet many of 
	the requirements of empirical science
	discussed previously in the chapter.
It provides a solution to the Problem of Demarcation:
	a theory is scientific if it can be used
	to build a simulation program for a particular phenomenon.
It gives scientists a way to make decisive theory comparisons,
	allowing them to search efficiently through the space of theories.
It involves a kind of Circularity Commitment:
	one develops theories of a certain phenomenon
	in order to be able to construct convincing
	simulations of the same phenomenon.
Sophie could plausibly have answered the shaman's critique
	of physics by demonstrating that a simulator
	based on Newtonian mechanics produces more realistic
	image sequences than one based on shamanic revelation.

In comparison to the compression principle,
	the veridical simulation principle has one obvious disadvantage:
	theory comparisons depend on qualitative human perception.
If the human observers have no special ability to judge the
	authenticity of a particular simulation,
	the theory comparisons will become noisy and muddled.
The method may work for things like basic physics,
	text, speech, and natural images,
	because humans have intimate knowledge of these things.
But it probably will not work for phenomena 
	which humans do not encounter in their everyday lives.

The advantage of the simulation principle
	compared to the compression principle
	is that it provides an indication of where
	and in what way a model fails to capture reality.
The word sampling example above showed
	how the model failed to capture the fact
	that real English words do not start with a double-s.
If a model of visual reality were used to generate images,
	the unrealistic aspects of the resulting images
	will indicate the shortcomings of the model.
For example, if a certain model does not handle shadows correctly,
	this will become obvious when it produces
	an image of a tree that casts no shade.
The compression principle does \textit{not} provide
	this kind of indication.
For this reason, the simulation principle
	can be thought of as a natural complement
	to the compression principle,
	that researchers can use to find out where to look for further progress.

Another interesting aspect of the veridical simulation principle
	is that it can be use to define a challenge
	similar to the Turing Test.
In this challenge, 
	researchers attempt to build simulators that	
	can produce samples that are veridical enough
	to fool humans into thinking they are real.
The outcome of the context is determined
	by showing a human judge two data objects,
	one real and one simulated.
The designers of the system
	win if the human is unable to tell which object is real.

To see the difficulty and interest of this challenge,
	consider using videos obtained in the course
	of the Visual Manhattan Project inquiry of Section~\ref{sec:vismanproj}
	as the real world component.
The statistical model of the video data
	would then need to produce samples 
	that are indistinguishable 
	from real footage of the streets of New York City.
The model would thus need to contain
	all kinds of information and detail relating to the visual environment of the city,
	such as the layout and architecture of the buildings,
	and the fashion sense and walking style of the pedestrians.
This is, of course, exactly the kind of information needed to compress the video data.
This observation provides further support
	for the intuitive notion that while the simulation principle and the compression principle
	are not identical, they are at least strongly aligned.

It will require an enormous level of sophistication 
	to win the simulation game,
	especially if the judges are long term inhabitants of New York.
A true New Yorker 
	would be able to spot very minor deviations from veridicality,
	related to things like the color of the sidewalk carts
	used by the pretzel and hot dog vendors,
	or to subtle changes in the style of clothing worn
	by denizens of different parts of the city.
A true New Yorker might be able to spot a fake video
	if it failed to include an appropriate degree of strangeness.
New York is no normal place and a real video strream will reflect
	that by showing celebrities, business executives, 
	beggars, transvestites, fashion models, inebriated artists, and so on.
In spite of this difficulty,
	the alignment between the compression and simulation principles
	suggest that there is a simple way to make systematic progress:
	get more and more video data, and improve the compression rate.

\mpagebreak

\section{Comparison to Physics}

Physics is the exemplar of empirical science,
	and many other fields attempt to imitate it.
Some researchers have deplored the influence of so-called ``physics envy''
	on fields like computer vision and artificial intelligence~\cite{Brooks:1999a}.
This book argues that there is nothing wrong with imitating physics.
Instead, the problem is that previous researchers failed to understand the essential character of physics,
	and instead copied its superficial appearance.
The superficial appearance of physics is its use of sophisticated mathematics;
	the essential character of physics is its obsession with reality.
A physicist uses mathematics for one and only one reason:
	it is useful in describing empirical reality.
Just as physicists do,
	\compist~researchers adopt as their fundamental goal 
	the search for simple and accurate descriptions of reality.
They will use mathematics,
	but only to the extent that it is useful in achieving the goal.
		
Another key similarity between physics
	and \compist~science involves the justification of research questions.
Some skeptics may accept that CRM research is legitimate science,
	but believe that it will be confined to a narrow set of technical topics. 
After all,
	the CRM defines only one problem:
	large scale lossless data compression.
But notice that physics also defines only one basic problem:
	given a particular physical configuration,
	predict its future evolution.
Because there is a vast number of possible configurations of matter and energy,
	this single question is enormously productive,
	justifying research into 
	such diverse topics as black holes,
	superconductivity, quantum dots, Bose-Einstein condensates,
	the Casimir effect, and so on.
Analogously,
	the single question of \compist~science
	justifies a wide range of research,
	due to the enormous diversity of empirical regularities
	that can be found in databases of natural images,
	text, speech, music, \textit{etc}.
The fact
	that a single question provides a \textit{parsimonious justification}
	for a wide range of research
	is actually a key advantage of the philosophy.

Both physics and \compist~science
	require candidate theories to be tested against empirical observation
	using hard, quantitative evaluation methods.
However, there is an important difference 
	in the way the theory-comparisons work.
Physical theories are very specific.
In physics, any new theory must agree with the current champion
	in a large number of cases,
	since the current champion has presumably been validated
	on many configurations.
To adjudicate a theory contest,
	researchers must find a particular configuration
	in which the two theories make opposing predictions,
	and then run the appropriate experiment.
In \compist~science,
	the predictions made by the champion theory are neither correct or incorrect,
	they are merely good.	
To unseat the champion theory,
	it is sufficient for a rival theory to make better predictions on average.

\chapter{Compression and Learning}
\label{chapt:comprlearn}

\section{Machine Learning}

Humans have the ability to develop amazing skills 
	relating to a very broad array of activities.
However, almost without exception,
	this competence not innate,
	and is achieved only as a result of extended learning.
The field of machine learning takes this observation
	as its starting point.
The goal of the field is to develop algorithms
	that improve their performance over time
	by adapting their behavior based on the data they observe.

The field of machine learning appears to have achieved 
	significant progress in recent years.
Researchers produce a steady stream of new learning systems
	that can recognize objects,
	analyze facial expressions,
	translate documents from one language to another,
	or understand speech.
In spite of this stream of new results,
	learning systems still have frustrating limitations.
Automatic translations systems often produce gibberish,
	and speech recognition systems often cause more
	annoyance than satisfaction.
One particularly glaring illustration of the limits of machine learning
	came from a ``racist'' camera system that was supposed to detect faces,
	but worked only for white faces,
	failing to detect black ones~\cite{Simon:2009}.
The gap between the enormous ambitions of the field
	and its present limitations
	indicates that there is some mountainous conceptual barrier impeding progress.
Two views can be articulated regarding the nature of this barrier.
	
According to the first view,
	the barrier is primarily \textit{technical} in nature.
Machine learning is on a promising trajectory that will ultimately 
	allow it to achieve its long sought goal. 
The field is asking the right questions;
	success will be achieved by improving the answers to those questions. 
The limited capabilities of current learning systems reflect
	limitations or inadequacies of modern theory and algorithms.
While the modern mathematical theory of learning is advanced,
	it is not yet advanced enough.
In time, new algorithms will be found that are far more powerful
	than current algorithms such as 
	AdaBoost and the Support Vector Machine~\cite{Freund:1997,Vapnik:1996}.
The steady stream of new theoretical results and improved algorithms will
	eventually yield a sort of grand unified theory of learning
	which will in turn guide the development of truly intelligent machines. 
	
In the second view, the barrier is primarily \textit{philosophical} in nature.
In this view,
	progress in machine learning is tending toward a sort of asymptotic limit.
The modern theory of learning provides a comprehensive answer
	to the problem of learning as it is currently formulated.
Algorithms
	solve the problems for which they are designed 
	nearly as well as is theoretically possible. 
The demonstration of an algorithm that provides an improved convergence rate
	or a tighter generalization bound may be interesting from an intellectual perspective, 
	and may provide slightly better perfomance on the standard problems.
But such incremental advances
	will produce true intelligence.
To achieve intelligence,
	machine learning systems must make a \textit{discontinuous} 
	leap to an entirely new level of performance.
The current mindset is analogous to the researchers in the 1700s 
	who attempted to expedite ground transportation by breeding faster horses,
	when they should actually have been searching for 
	qualitatively different mode of transportation.
The problem, then,
	is in the philosophical foundations of the field,
	in the types of questions considered by its practitioners
	and their philosophical mindset.	
If this view is true, 
	then to make further progress in machine learning, 
	it is necessary to formulate the problem of learning in a new way. 
This chapter presents arguments in favor of the second view.
	
\mpagebreak

\subsection{Standard Formulation of Supervised Learning}

There are two primary modes of statistical learning:
	the supervised mode and the unsupervised mode.
The present discussion will focus primarily on the former;
	the latter is discussed in Appendix~\ref{chapt:relappendx}.
The supervised version can be understood by considering a typical
	example of what it can do.
Imagine one wanted to build a face detection system capable
	of determining if a digital photo contains an image of a face.
To use a supervised learning method,
	the researcher must first construct a \textit{labeled} dataset,
	which is made up of two parts.
The first part is a set of $N$ images $X = \{x_{1}, x_{2} \ldots x_{N}\}$.
The second part is a set of binary labels
	$Y = \{y_{1}, y_{2} \ldots y_{N}\}$,
	which indicate whether or not a face is present in each image.
Once this database has been built,
	the researcher invokes the learning algorithm,
	which attempts to obtain a predictive rule $h(\cdot)$
	such that $h(x) = y$.
Below, this procedure is refered to as the ``canonical'' form
	of the supervised learning problem.
Many applications can be formulated in this way, 
	as shown in the following list:
	
\begin{itemize}
\item Document classification: the $x_{i}$ data are the documents,
	and the $y_{i}$ data are category labels such as ``sports'',
	``finance'', ``political'', etc.
\item Object recognition: the $x_{i}$ data are images,
	and the $y_{i}$ data are object categories such as 
	``chair'', ``tree'', ``car'', etc.
\item Electoral prediction:
	each $x_{i}$ data is a package of information relating to current
	political and economic conditions, and the $y_{i}$ is a binary
	label which is true if the incumbent wins.
\item Marital satisfaction: 
	each $x_{i}$ data is a package of vital statistics relating to 
	particular marriage 
	(frequency of sex, frequency of argument, religious involvement, education levels, etc)
	and the corresponding $y_{i}$ is a binary label which is true
	if the marriage ends in divorce.
\item Stock Market prediction:
	each $x_{i}$ is a set of economic indicators such as interest rates,
	exchange rates, and stock prices for a given day;
	the $y_{i}$ is the change in value of a particular stock on the \textit{next} day.
\end{itemize}

\mpagebreak

\subsection{Simplified Description of Learning Algorithms}
\label{sec:simpdescla}

For readers with no background in machine learning,
	the following highly simplified description should
	provide a basic understanding of the basic ideas.
One starts with a system $S$ that performs some task,
	and a method for evaluating the performance
	$S$ provides on the task.
Let this evaluation function be denoted as $\mathbf{E}$,
	and $\mathbf{E}[S]$ be the performance of the system $S$.
In terms of the canonical task mentioned above,
	the system is the predictive rule $h(\cdot)$,
	and the evaluation function is just the squared difference
	between the predictions and the real data:
	
\begin{equation*}
\mathbf{E}[h] = \sum_{i} (h(x_{i}) - y_{i})^{2}
\end{equation*}

A key property of the system is that it be mutable.
If a system $S$ is mutable,
	then a small perturbation will produce a new system $S'$
	that behaves in nearly the same way as $S$.
This mutability requirement prevents one from defining $S$ to be, for example,
	the code of a computer program,
	since a slight random change to a program will usually break it completely.
To construct systems that can withstand these minor mutations without
	suffering catastrophic failures,
	researchers often construct the system by introducing
	a set of numerical parameters $\theta$.
If the behavior of $S(\theta)$ changes smoothly with changes in $\theta$,
	then small changes to the system can be made by making small changes to $\theta$.
There are, of course, other ways to construct mutable systems.
Given a mutable system and an evaluation function, 
	then the following procedure can be used to search for a high-performance system:
	
\begin{enumerate}
\item Begin by setting $S = S_{0}$, 
	where $S_{0}$ is some default setup (which can be \naive).
\item Introduce a small change to $S$, producing $S'$.
\item If $\mathbf{E}[S'] > \mathbf{E}[S]$, 
	keep the change by setting $S = S'$. 
	Otherwise, discard the modified version.
\item Return to step \#2.
\end{enumerate}

Many machine learning algorithms can be understood as refined
	versions of the above process.
For example, the backpropagation algorithm
	for the multilayer perceptron 
	uses the chain rule of calculus to find the 
	derivative of the $\mathbf{E}(S(\theta))$ with 
	respect to $\theta$~\cite{Rumelhart:1986}.
Many reinforcement learning algorithms
	work by making smart changes to a policy that
	depends on the parameters $\theta$~\cite{Barto:1997}.
Genetic algorithms,
	which are inspired by the idea of natural selection,
	also roughly follow the process outlined above.

\mpagebreak

\subsection{Generalization View of Learning}

Machine learning researchers have developed 
	two conceptual perspectives by which to approach the canonical task.
The first and more popular perspective is called the Generalization View.
Here the goal is to obtain,
	on the basis of the limited $N$-sample data set,
	a model or predictive rule that works well for new,
	previously unseen data samples.
The Generalization View is attractive for obvious practical purposes: 
	in the case of the face detection task, for example, 
	the model resulting from a successful learning process
	can be used in a system which requires the ability
	to detect faces in previously unobserved images (e.g. a surveillance application).
The key challenge of the Generalization View is that 
	the \textit{real} distribution generating the data is unknown.
Instead, one has access to the empirical distribution
	defined by the observed data samples.	
	
\begin{figure*}[t]
\centering
\subfigure{\includegraphics[width=.4\textwidth]{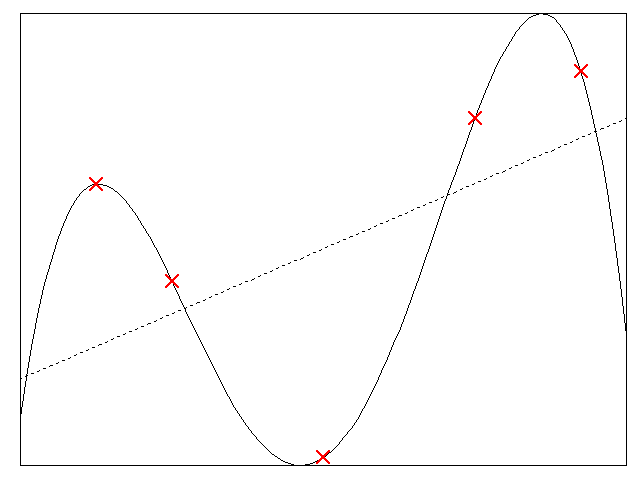}}
\subfigure{\includegraphics[width=.4\textwidth]{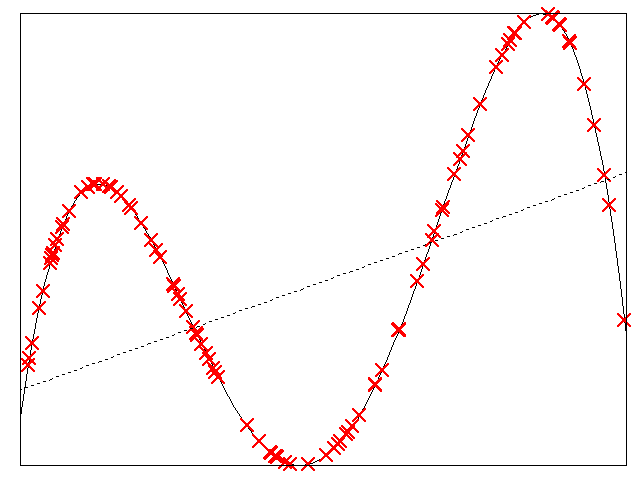}}
\caption[Pixel difference histograms for a natural and a random image.]{
Illustration of the idea of model complexity and overfitting.
In the limited data regime situation depicted on the left,
	the line model should be preferred to the curve model,
	because it is simpler.
In the large data regime,
	however, the polynomial model can be justified.
}
\label{fig:quartpolyn}
\end{figure*}

In the early days of machine learning research,
	many practitioners thought that the sufficient condition 
	for a model to achieve good performance on the real distribution
	was that it achieved good empirical performance:
	it performed well on the observed data set.
However,
	they often found that their models would 
	perform very well on the observed data, 
	but fail completely when applied to new samples.
There were a variety of reasons for this failure,
	but the main cause was the phenomenon of \textit{overfitting}.
Overfitting occurs when
	a researcher applies a complex model to solve a
	problem with a small number of data samples.
Figure~\ref{fig:quartpolyn} illustrates the problem of overfitting.
Intuitively, 
	it is easy to see that when there are only five data points,
	the complex curve model should not be used,
	since it will probably fail to generalize to any new points.
The linear model will probably not describe new points exactly,
	but it is less likely to be wildly wrong.
While intuition favors the line model,
	it is not immediately obvious how to formalize that intuition:
	after all, the curve model achieves better empirical performance
	(it goes through all the points).
	
The great conceptual achievement of statistical learning
	is the development of methods by which to overcome overfitting.
These methods have been formulated in many different ways,
	but all articulations share a common theme:
	to avoid overfitting, one must \textit{penalize} complex models.
Instead of choosing a model solely on the basis of its empirical performance,
	one must optimize a tradeoff between the empirical performance and model complexity.
In terms of Figure~\ref{fig:quartpolyn},
	the curve model achieves excellent empirical performance,
	but only because it is highly complex.
In contrast the line model achieves a good balance of performance and simplicity.
For that reason,
	the line model should be preferred in the limited-data regime.
In order to apply the complexity penalty strategy,
	the key technical requirement is a method 
	for quantifying the complexity of a model.

Once a suitable expression for a model's complexity is obtained,
	some further derivations yield
	a type of expression called a \textit{generalization bound}.
A generalization bound is a statement of the following form:
	if the empirical performance of the model is good,
	and the model is not too complex,
	then with high probability its real performance 
	will be only slightly worse.
The caveat ``with high probability'' can never be done away with,
	because there is always some chance that the empirical data
	is simply a bizarre or unlucky sample of the real distribution.
One might conclude with very high confidence that a coin is biased
	after observing 1000 heads in a row,
	but one could never be completely sure.
	
While most treatments of model complexity and generalization bounds
	require sophisticated mathematics,
	the following simple theorem can illustrate the basic ideas.
The theorem can be stated in terms of the notation used for the canonical task 
	of supervised learning mentioned above. 
Let $C$ be a set of hypotheses or rules that take 
	a raw data object $x$ as an argument and output a prediction $h(x)$ of its label $y$.
In terms of the face detection problem,
	$x$ would be an image, and $y$ would be a binary flag
	indicating whether the image contains a face.
Assume it is possible to find a hypothesis $h^{*} \in C$ 
	that agrees with all the observed data: 
	
\begin{displaymath}
h^{*}(x_{i}) = y_{i} \qquad i = 1 \ldots N 
\end{displaymath}

\noindent
%TODO: redo this to put $\epsilon^{*}$ on the left hand side.
Now select some $\epsilon$ and $\delta$ such that the following inequality holds: 

\begin{equation}
\label{eq:inqvaliant}
N \geq \frac{1}{\epsilon} \log \big( \frac{|C|}{\delta} \big) 
\end{equation}

\noindent
Then with probability $1 - \delta$,
	the error rate of the hypothesis will be at most $\epsilon$
	when measured against the real distribution. 
Abstractly, the theorem says that if 
	the hypothesis class is not too large compared to the number of data samples, 
	and some element achieves good empirical performance, 
	then with high probability its performance on the real (full) distribution
	will be not too much worse.
The following informal proof of the theorem may illuminate	
	the core concept.

To understand the theorem,
	imagine you are searching through a barrel of apples (the hypotheses),
	looking for a good one.
Most of the apples are ``wormy'':
	they have a high error rate on the real distribution.
The goal is to find a ripe, tasty apple;
	one that has a low error rate on the real distribution.
Fortunately, 
	most of the wormy apples can be discarded because
	they are \textit{visibly} old and rotten,
	meaning they make errors on the observed data.
The problem is that there might be a ``hidden worm'' apple that looks tasty -
	it performs perfectly on the observed data - 
	but is in fact wormy.
Define a wormy apple as one that has real error rate larger than $\epsilon$.
Now ask the question:
	if an apple is wormy,
	what is the probability it looks tasty?
It's easy to find an upper bound for this probability:
	
\begin{equation*}
\label{eq:wormyproby}
P(\mathrm{hiddenworm}) \leq (1 - \epsilon)^{N}
\end{equation*}

\noindent
This is because,
	if the apple is wormy,
	the probability of not making a mistake on one sample is $\leq (1 - \epsilon)$,
	so the probability of not making a single mistake on $N$ samples is $\leq (1 - \epsilon)^N$.
Now the question is: 
	what is the probability that there are no hidden worms in the entire hypothesis class?
Let $HW_{k}(\epsilon)$ be the event that the $k$th apple is a hidden worm.
Then the  probability that there are no 
	hidden worms in the hypothesis class is:
	
\begin{eqnarray*}
P(\mathrm{no hidden worms}) &=& P( \neg [ HW_{1}(\epsilon) \vee HW_{2}(\epsilon) \vee  HW_{3}(\epsilon) \ldots ]) \\
&=& 1 - P( [ HW_{1}(\epsilon) \vee HW_{2}(\epsilon) \vee  HW_{3}(\epsilon) \ldots ]) \\
&\geq& 1 - \sum_{k} P( HW_{k}(\epsilon)) \\
&=& 1 - |C| P( HW(\epsilon)) \\
&\geq& 1 - |C| (1 - \epsilon)^N 
\end{eqnarray*}

\noindent
The first step is true because $P(\neg A) = 1 - P(A)$,
	the second step is true because $P(A \vee B) \leq P(A)+P(B)$,
	the third step is true because there are $|C|$ hypotheses,
	and the final step is just a substitution of Inequality~\ref{eq:wormyproby}.
Then the result follows by letting $\delta = |C| (1 - \epsilon)^N$,
	noting that $\log(1-\epsilon) \approx -\epsilon$,
	and rearranging terms.

A crucial point about the proof is that it makes no guarantee
	whatever that a good hypothesis (tasty worm-free apple)
	will actually appear.
The proof merely says that, 
	if the model class is small and the other values are reasonably chosen,
	then it is unlikely for a hidden worm hypothesis to appear.
If the probability of a hidden worm is low,
	\textit{and} by chance a shiny apple is found,
	then it is probable that the shiny apple is actually worm-free.
	
A far more sophisticated development of the ideas
	of model complexity and generalization 
	is due to the Russian mathematician Vladimir Vapnik~\keycite{Vapnik:1998}.
In Vapnik's formulation the goal is to minimize the real (generalization) risk $R$,
	which can be the error rate or some other function.
Vapnik derived a sophisticated model complexity term
	called the VC dimension,
	and used it to prove several generalization bounds.
A typical bound is:

\begin{equation*}
R(h_{i}) \leq R_{emp}(h_{i}) + \frac{\log(|C'|) - \log(\delta)}{N} 
	\bigg ( 1 + \sqrt{1 + \frac{2 N R_{emp}(h_{i})}{\log(|C'|) - \log(\delta)}} \bigg) 
\end{equation*}

\noindent
Where $R(h_{i})$ is the real risk of hypothesis $h_{i}$
	and $R_{emp}(h_{i})$ is the empirical risk,
	calculated from the observed data.
The bound, 
	which holds for all hypotheses simultaneously,
	indicates the conditions under which the real risk will 
	not exceed the empirical risk by too much.
As above, the bound holds with probability $1-\delta$,
	and $N$ is the number of data samples.
The term $\log(|C'|)$ is the log size of the VC-dimension,
	which plays a conceptually similar role
	to the simple $\log(|C|)$ term in the previous theorem.
Vapnik's complex inequality shows the same basic idea as the simple theorem above:
	the real performance will be good if the empirical performance is good
	and the log size of the hypothesis class is small 
	in comparison with the number of data samples.
Proofs of theorems in the VC theory also use a similar strategy:
	show that if the model class is small,
	it is unlikely that it includes a ``hidden worm'' hypothesis
	which has low empirical risk but high real risk.
Also, none of the VC theory bounds guarantee that a good 
	hypothesis (low $R_{emp}(h_{i})$) will \textit{actually} be found.

% TODO: need to massage these two theorems, especially the simple one,
% into a more coherent form.
% Basically want the 
The problem of overfitting is easily understood 	
	in the light of these generalization theorems.
A \naive~approach to learning attempts to minimize the empirical risk
	without reference to the complexity of the model.
The theorems show that a low empirical risk, by itself,
	does not guarantee low real risk.
If the model complexity terms $\log(|C|)$ and $\log(|C'|)$ 
	are large compared to the number of samples $N$,
	then the bounds will become too loose to be meaningful.
%TODO: want to say "right hand side of inequality will become large, 
% so that the generalization bound tells us nothing.
In other words,
	even if the empirical risk is reduced to a very small quantity,
	the real risk may still be large.
The intuition here is that because such a large number of hypotheses was tested, 
	the fact that one of them performs well on the empirical data 
	is meaningless. 
If the hypothesis class is very large, 
	then some hypotheses can be expected to perform well merely by chance. 

The above discussion seems to
	indicate that complexity penalties 
	actually apply to model classes, not to individual models.
There is an important subtlety here.
In both of the generalization theorems mentioned above,
	all elements of the model class were treated equally,
	and the penalty depended only on the size of the class.
However, it is also reasonable to 
	apply different penalties to different elements of a class.
Say the class $C$ contains two subclasses $C_{a}$ and $C_{b}$.
Then if $|C_{b}| > |C_{a}|$,
	hypotheses drawn from $C_{b}$ must receive a larger penalty,
	and therefore require relatively better 
	empirical performance in order to be selected.
For example, 
	in terms of Figure~\ref{fig:quartpolyn},
	one could easily construct an aggregate class
	that includes both lines and polynomials.
Then the polynomials would receive a larger penalty,
	because there are more of them.
	
While more complex models must receive larger penalties,
	they are never prohibited outright.
In some cases it very well may be worthwhile
	to use a complex model,
	if the model is \textit{justified} by a large amount 
	of data and achieves good empirical performance.
This concept is illustrated in Figure~\ref{fig:quartpolyn}:
	when there are hundreds of points that all fall on the complex curve,
	then it is entirely reasonable to prefer it to the line model.
The generalization bounds also express this idea,
	by allowing $\log(|C|)$ or $\log(|C'|)$
	to be large if $N$ is also large.

\mpagebreak

\subsection{Compression View} 

The second perspective on the learning problem
	can be called the Compression View.
The goal here is to compress a data set
	to the smallest possible size.
This view is founded upon the insight, 
	drawn from information theory, 
	that compressing a data set to the smallest possible size 
	requires the best possible model of it. 
The difficulty of learning comes from the fact that the bit cost 
	of the model used to encode the data must itself be accounted for. 
In the statistics and machine learning literature,
	this idea is known as the Minimum Description 
	Length (MDL) principle~\cite{Wallace:1968,Rissanen:1978}.
	
The motivation for the MDL idea can best be seen by contrasting
	it to the Maximum Likelihood Principle,
	one of the foundational ideas of statistical inference.
Both principles apply to the problem of how to choose
	the best model $M^{*}$ out of a class $\cal{M}$
	to use to describe a given data set $D$.
For example, the model class $\cal{M}$ could be the set of all Gaussian distributions,
	so that an element $M$ would be a single Gaussian,
	defined by a mean and variance.
The Maximum Likelihood Principle suggests 
	choosing $M^{*}$ so as to maximize the likelihood of the data given the model:
	
\begin{equation*}
M^{*} = \max_{M \in \cal{M}} P(D|M)
\end{equation*}

\noindent
This principle is simple and effective in many cases,
	but it can lead to overfitting.
To see how, 
	imagine a data set made up of 100 numbers $\{x_{1}, x_{2}, \ldots x_{100}\}$.
Let the class $\cal{M}$ be the set of Gaussian mixture models.
A Guassian mixture model is just a sum of normal distributions
	with different means and variances.
Now, one simple model for the data could be built by finding the mean
	and variance of the $x_{i}$ data and using a single Gaussian
	with the given parameters.
A much more complex model
	can be built by taking a sum of 100 Gaussians,
	each with mean equal to some $x_{i}$ and near-zero variance.
Obviously, this ``comb'' model is worthless:
	it has simply overfit the data and will fail badly 
	when a new data sample is introduced.
But it produces a higher likelihood than the single Gaussian model,
	and so the Maximum Likelihood principle
	suggests it should be selected.
This indicates that the principle contains a flaw.

The Minimum Description Length principle 
	approaches the problem by imagining the following scenario.
A sender wishes to transmit a data set $\{x_{1}, x_{2}, \ldots x_{100}\}$
	to a receiver.
The two parties have agreed in advance on the model class $\cal{M}$.
To do the transmission,
	the sender chooses some model $M^{*} \in \cal{M}$
	and sends enough information to specify $M^{*}$ to the receiver.
The sender then encodes the $x_{i}$ data using a code based on $M^{*}$.
The best choice for $M^{*}$ minimizes the net codelength required:

\begin{eqnarray*}
M^{*} &=& \min_{M \in \cal{M}} L(M) + L(D|M) \\
	&=& \min_{M \in \cal{M}} L(M) - \log P(D|M)
\end{eqnarray*}

\noindent 
Where $L(M)$ is the bit cost of specifying $M$ to the receiver,
	and $L(D|M) = -\log P(D|M)$ is the cost of encoding the data given the model.
If if it were not for the $L(M)$ term,
	the MDL principle would be exactly the same as the Maximum Likelihood principle,
	since maximizing $P(D|M)$ is the same as minimizing $-\log P(D|M)$.
The use of the $L(M)$ term penalizes complex models,
	which allows users of the MDL principle to avoid overfitting the data.
In the example mentioned above,
	the Gaussian mixture model with 100 components
	would be strongly penalized,
	since the sender would need to transmit a mean/variance parameter
	pair for each component.
	
The MDL principle can be applied to the canonical task
	by imagining the following scenario.
A sender has the image database $X$ and the label database $Y$, 
	and wishes to transmit the latter to a receiver. 
A crucial and somewhat counterintuitive point is that 
	the receiver already has the image database $X$.
Because both parties have the image database, 
	if the sender can discover a simple relationship between the images and the labels, 
	he can exploit that relationship to save bits. 
If a rule can be found that accurately predicts $y_{i}$ given $x_{i}$, 
	that is to say if a good model $P(Y|X)$ can be obtained, 
	then the label data can be encoded using a short code.
However, in order for the receiver to be able to perform the decoding,
	the sender must encode and transmit information
	about how to build the model.
More complex models will increase the total number of bits that must be sent.
The best solution, therefore, 
	comes from optimizing a tradeoff between 
	empirical performance and model complexity.
	
\mpagebreak
	
\subsection{Equivalence of Views}	

The Compression View and the Generalization View adopt
	very different approaches to the learning problem.
Profoundly, however, 
	when the two different goals are formulated quantitatively,
	the resulting optimization problems are quite similar.
In both cases,
	the essence of the problem is to balance a tradeoff
	between model complexity and empirical performance.
Similarly, 
	both views justify the intuition relating 
	to Figure~\ref{fig:quartpolyn}
	that the linear model should be preferred
	in the low-data regime,
	while the polynomial model should be preferred in the high-data regime.

The relationship between the two views can be further
	understood in the context of the simple hidden worm theorem described above.
As stated,
	this theorem belongs to the Generalization View.
However,
	it is easy to convert it into a statement of the Compression View.
A sender wishes to transmit to a receiver
	a database $Y$ of labels which are related to a set $X$
	of raw data objects.
The receiver already has the raw data $X$.
The sender and receiver agree in advance on the hypothesis class $C$ and
	an encoding format based on it that works as follows. 
The first bit is a flag that indicates whether a good hypothesis $h^{*}$ was found.
If so, the sender then sends the index of the hypothesis in $C$, 
	using $\log_{2} |C|$ bits. 
The receiver can then look up $h^{*}$ and apply it to the images $x_{i}$
	to obtain the labels $y_{i}$. 
Otherwise, the sender encodes the labels $y_{i}$ normally at a cost of $N$ bits. 
This scheme achieves compression under two conditions:
	a good hypothesis $h^{*}$ is found 
	and $\log |C|$ is small compared to the number of samples $N$.
These are exactly the same conditions required for generalization
	to hold in the Generalization View approach to the problem.
	
This equivalence in the case of the hidden worm theorem 
	could be just a coincidence.
But in fact there are a variety of theoretical statements
	in the statistical learning literature
	that suggest that equivalence is actually quite deep.
For example,
	Vapnik showed that if that if a model class $C$ could
	be used to compress the label data,
	then the following inequality relates the achieved compression rate $K(C)$
	to the generalization risk $R(C)$:

\begin{equation}
R(C) < 2 (K(C) \log 2 - N^{-1} \log \delta)
\end{equation}

\noindent
The second term on the right, $N^{-1} \log \delta$, is for all practical cases
	small compared to the $K(C)$ term,
	so this inequality shows a very direct relationship
	between compression and generalization.
This expression is strikingly simpler than 
	any of the other VC generalization bounds.

There are many other theorems in the machine learning 
	literature that suggest the equivalence of the Compression View 
	and the Generalization View.
For example,
	a simple result due to Blumer~\etal~
	relates the \textit{learnability} of a hypothesis class 
	to the existence of an Occam algorithm for it.
In this paper,
	the key question of learning is whether 
	a good approximation $h^{*}$ can be found
	of the true hypothesis $h^{T}$,
	when both functions are contained in a hypothesis class $H$.
If a good approximation (low $\epsilon$) can be found
	with high probability (low $\delta$) using a 
	limited number of data samples (small $N$),
	the class is called learnable.
Functions in the hypothesis class $H$ 
	can be specified using some finite bit string;
	the length of this string is the complexity of the function.
To define an Occam algorithm,
	let the unknown true function $h^{T}$ have complexity $W$,
	and let there be $N$ samples $(x_{i}, h^{T}(x_{i}))$.
The algorithm then produces 
	a hypothesis $h^{*}$ of complexity $W^{c} N^{\alpha}$,
	that agrees with all the sample data,
	where $c \geq 1$ and $0 \leq \alpha < 1$
	are constants.
Because the complexity of $h^{*}$ grows sublinearly with $N$,
	then a simple encoding scheme such as the one mentioned above
	based on $H$ and the Occam algorithm
	is guaranteed to produce compression for large enough $N$.
Blumer~\etal~show that if an Occam algorithm exists,
	then the class $H$ is learnable.
More complex results by the same authors are given in~\ref{LearnabilityVapnik}
	(see section 3.2 in particular).
	
While the Generalization View and the Compression View may be equivalent,
	the latter approach has a variety of conceptual advantages.
First of all,
	the No Free Lunch theorem of data compression
	indicates that no completely general compressor can ever succeed.
This shows that
	all approaches to learning must discover and exploit
	special empirical structure in the problem of interest.
This fact does not 
	seem to be widely appreciated in the machine learning literature:
	many papers advertise methods
	without explicitly describing the conditions required
	for the methods to work.
Also, because the model complexity penalty $L(M)$
	can be interpreted as a prior over hypotheses,
	the Compression View clarifies the relationship
	between learning and Bayesian inference.
This relationship is obscure in the Generalization View,
	leading some researchers to claim that 
	learning differs from Bayesian inference
	in some kind of deep philosophical way.

Another significant advantage of the Compression View is that
	it is simply easier to think up compression schemes
	than it is to prove generalization theorems.
For example,
	the Generalization View version of the hidden worm theorem
	requires a derivation and some modest level of mathematical sophistication
	to determine the conditions for success,
	to wit, that $log |C|$ is small compared to $N$ and a 
	good hypothesis $h^{*}$ is found.
In contrast, 
	in the Compression View version of the theorem,	
	the requirements for success become obvious immediately
	after defining the encoding scheme.
The equivalence between the two views suggest that a fruitful procedure
	for finding new generalization results
	is to develop new compression schemes,
	which will then automatically imply associated generalization bounds.
	
\mpagebreak

\subsection{Limits of Model Complexity in Canonical Task}

In the Compression View,
	an important implication regarding model complexity
	limits in the canonical task is immediately clear.
The canonical task is approached by finding a short program
	that uses the image data set $X$ to compress the label data $Y$.
The goal is to minimize the net codelength of the compressor 
	itself plus the encoded version of $Y$. 
This can be formalized mathematically as follows:

\begin{equation*}
M^{*} = \arg \min_{M \in \cal{M}} [ L(M) - \log_{2} P_{M}(Y|X) ]
\end{equation*}

\noindent
Where $M^{*}$ is the optimal model, 
	$L(M)$ is the codelength required to specify model $M$,
	and $\cal{M}$ is the model class.
Now assume that $\cal{M}$ contains some trivial model $M_{0}$,
	and assume that $L(M_{0}) = 0$.
The intuition of $M_{0}$ is that
	it corresponds to just sending the data in a flat format,
	without compressing it at all.
Then, in order to justify the choice of $M^{*}$ over $M_{0}$,
	it must be the case that:

\begin{equation*}
L(M^{*}) - \log_{2} P_{M^{*}}(Y|X) < \log_{2} P_{M_{0}}(Y|X)
\end{equation*}

The right hand side of this inequality is easy to estimate.
Consider a typical supervised learning task
	where the goal is to predict a binary outcome,
	and there are $N=10^{3}$ data samples
	(many such problems are studied in the machine learning literature,
	see the review~\cite{Dietterich:2000}).
Then a dumb format for the labeled data
	simply uses a single bit for each outcome,
	for a total of $N=10^{3}$ bits.
The inequality then immediately implies that:

\begin{equation*}
L(M^{*}) < 10^{3}
\end{equation*}
	
\noindent
This puts an absolute upper bound on the complexity
	of any model that can ever be used for this problem.
In practice, the model complexity must really be quite a 
	bit lower to get good results.
Perhaps the model requires 200 bits to specify
	and the encoded data requires 500 bits,
	resulting in a savings of 300 bits.
200 bits corresponds to 25 bytes.
It should be obvious to anyone who has ever written
	a computer program 
	that no model of any complexity can be specified
	using only 25 bytes.

\mpagebreak

\subsection{Intrinsically Complex Phenomena}
\label{sec:cmplxphnmn}

Consider the following thought experiment.
Let $T = \{t_{1}, t_{2} \ldots t_{N}\}$ be some database of interest,
	made up of a set of raw data objects.
Let $\cal{M}$ be the set of programs
	that can be used to losslessly encode $T$.
A program $m$ is an element of $\cal{M}$, 
	its length is $|m|$. 
Furthermore let $L_{m}(T)$ be the codelength 
	of the encoded version of $T$ produced by $m$.
For technical reasons, 
	assume also that the compressor is \textit{stateless},
	so that the $t_{i}$ can be encoded in any order,
	and the codelength for each object will be the same
	regardless of the ordering.
This simply means that
	any knowledge of the structure of $T$ must be included in $m$ at the outset,
	and not learned as a result of analyzing the the $t_{i}$.
Now define:

\begin{eqnarray*}
L^{*}_{T}(x) &=& \min_{|m| < x} \big( |m| + L_{m}(T) \big )
\end{eqnarray*}

\noindent
So the quantity $L^{*}_{T}(x)$ is the shortest codelength for $T$ that 
	can be achieved using a model of length less than $x$.
This quantity cannot actually be found,
	for basically the same reason that the Kolmogorov complexity
	of a string cannot be computed.
But set this fact aside for a 
	moment and consider what the graph of $L^{*}_{T}(x)$
	would look like if somehow it could be calculated.
	
Clearly, the shape of the function $L^{*}_{T}(x)$ depends on the characteristics
	of the data set $T$.
If $T$ is made up of completely random noise,
	then $L^{*}_{T}(x)$ will be a flat line, 
	because random data cannot be compressed.
On the other hand,
	consider the case of constructing $T$
	by running many trials of a simple physics experiment,
	involving some basic process such as electric currents
	or ballistic motion.
In that case a very short program encoding the relevant
	physical laws such as $V = IR$
	would achieve a good compression rate,
	so $L^{*}_{T}(x)$ would drop substantially for small $x$.

A third type of dataset
	would fall in between these two extremes.
In this case, 
	$L^{*}_{T}(x)$ would decline only gradually:
	good compression rates could be achieved,
	but only for large $x$.
Such a dataset would represent an \textit{intrinsically} complex phenomenon.
This kind of phenomenon could be understood and predicted well,
	but only with an highly complex model (large $|m|$).
Because $L^{*}_{T}(x)$ cannot actually be computed,
	it is impossible to prove that any given naturally occuring dataset is complex.
But common sense suggests complex data sets exist,
	and may be common.
The following quote from an interview with Vladimir Vapnik provides
	further support for the proposition:
	
\begin{quote}
I believe that something drastic has happened in computer science and machine learning. 
Until recently, philosophy was based on the very simple idea that the world is simple.
In machine learning, for the first time, we have examples where the world is not simple. 
For example, when we solve the ``forest'' problem (which is a low-dimensional problem) and use 
	data of size 15,000 we get 85\%-87\% accuracy. 
However, when we use 500,000 training examples we achieve 98\% of correct answers. 
This means that a good decision rule is not a simple one, it cannot be described by a very few parameters. 
This is actually a crucial point in approach to empirical inference.

This point was very well described by Einstein who said ``when the solution is simple, God is answering''. 
That is, if a law is simple we can find it. 
He also said ``when the number of factors coming into play is too large, scientific methods in most cases fail''. 
In machine learning we are dealing with a large number of factors. 
So the question is: what is the real world? 
Is it simple or complex? 
Machine learning shows that there are examples of complex worlds~\keycite{Vapnik:2008}. 
\end{quote}
	
The forest dataset mentioned by Vapnik is part of the UCI machine learning repository~\cite{Asuncion:2007},
	a set of benchmark problems widely used in the field.
It is one of the largest datasets in the repository.
Most of the other datasets are much smaller:
	there are many with less than 1000 samples,
	and only a few with more than 10000.
This scarcity of data is caused by the simple fact that
	generating labelled datasets tends to require a substantial amount of labor.
For example, 
	one of the problems in the repository requires the algorithm to guess a person's income
	based on factors such as education level, occupation, age, and marital status.
In order to produce a single valid $\{x, y\}$ data point for this problem,
	a person must fill out a lengthy questionnaire.

There are many standard learning problems,
	in the UCI and elsewhere,
	for which good performance cannot be achieved.
Researchers often conclude that this is due to the intrinsic 
	difficulty of the problem.
But the results reported by Vapnik on the Forest problem
	suggest a new diagnosis:
	perhaps the problems \textit{can} be solved well,
	but only by using a complex model.
Many of the problems in the UCI repository
	include a relatively small number of samples;
	most have less than 15,000,
	and some have less than 1,000.
In this low-data regime, 
	it is impossible to use a complex model without overfitting,
	so these problems cannot be solved well.

This new diagnosis is part of a larger analysis 
	of the limitations of supervised learning.
Supervised learning appears to be based on the assumption
	that the world, and the problems it contains, are \textit{simple}.
If the world is simple,
	then simple models can describe it well.
Since simple models can be justified based on small labeled data sets,
	the empirical content of such data sets 
	is sufficient to understand the world.	
But the idea of intrinsically complex phenomena
	casts doubt on the assumption of the simplicity of the world.
If such phenomena exist,
	simple models will be inadequate to describe them well.
In that case it is essential to obtain large data sets,
	on the basis of which complex models can be justified.

This analysis may not be very satisfying,
	because building large labeled data sets 
	is an expensive and time-consuming chore.
But that is only because of the necessity of 
	using human intelligence to provide 
	labels for the data.
In contrast to labeled data,
	unlabeled data is easy to acquire in large quantities.
Furthermore,
	raw data objects (e.g. images) 
	have far larger information content than the labels.
This suggests that,
	in order to construct complex models and thereby understand the complex world,
	it is necessary to find a way to exploit
	the vast information content of raw unlabeled data sets.
The compression principle provides exactly such a mechanism.

\mpagebreak
	
\subsection{\Compist~Reformulation of Canonical Task}
\label{sec:compistref}

The Compression Rate Method
	suggests a qualitatively different approach 
	to the canonical task of supervised learning. 
The new approach is to separate the learning process 
	into two phases. 
In the first phase,
	the goal is to learn a model of the raw data objects themselves.
In the face detection example,
	the first phase involves a study of face images.
The results of the first phase
	are reused in the second phase
	to find the relationship between the data objects and the labels.
	
This line of attack totally changes the nature of the problem
	by vastly expanding the amount of data being modeled.
A typical digital images might have about $640 \cdot 480 \cdot 3 \approx 9.2 \cdot 10^{5}$ pixels,
	each of 8 bits apiece,
	for a total of about $8 \cdot 10^6$.
If there are $10^{5}$ images,
	then the total information content of the database is about $8 \cdot 10^{11}$.
This enormous expansion in the amount of data being modeled
	justifies a correspondingly huge increase in the complexity of the models used.
For example, one can easily justify a model requiring $10^8$ bits to encode 
	(larger by far than the models used in traditional statistical learning),
	by showing that it achieves a savings of $10^{10}$ bits on the large database.

Another vast increase in the quantity of data available
	comes from the realization that it is no longer necessary 
	to use labeled data.
A major bottleneck in supervised learning research is the difficulty of 
	constructing labeled databases,
	which require significant expenditures of human time and effort.
The MNIST handwritten digit recognition database, 
	which is one of the largest in common use, 
	contains $6 \cdot 10^{5}$ labeled samples~\cite{Lecun:2006}. 
In contrast, 
	the popular web image sharing site Flickr.com recently announced
	that the size of its image database
	had reached $5\cdot 10^9$~\cite{Wikipedia:Flickr}.

The beginning of this chapter suggested
	that algorithms like AdaBoost and the Support Vector Machine
	solve the problems for which they were designed
	nearly as well as is theoretically possible.
A skeptic could argue that this claim is absurd,
	because there are many problems which the Support Vector Machine cannot solve,
	but the human brain can.
A \compist~researcher would respond that in fact
	the human brain is solving a \textit{different} problem.
The human brain learns from a \textit{vast} data set
	containing visual, audial, linguistic, 
	olfactory, and tactile components.
This allows the brain to circumvent the model complexity limitations
	inherent in the supervised learning formulation,
	and obtain a highly sophisticated model.
The brain then reuses components from this complex model 
	to handle every specific learning problem,
	such as object recognition and face detection.
If the brain had to solve a limited data classification problem
	for which none of its prelearned components were relevant,
	it would fare just as badly as a typical machine learning algorithm.	
	
It may not be obvious why having a good model of $P(X)$ will 
	assist in learning $P(Y|X)$. 
The key is to think in terms of description languages, 
	or compression formats, rather than probability distributions. 
A good compression format for real world images will involve 
	elements corresponding to abstractions such as ``face'',
	``person'', ``tree'', ``car'', and so on. 
Transforming the image from the raw, pixel-based representation	
	into the abstraction-based representation
	should considerably simplify the problem
	of finding the relationship between image and label.
The face detection problem
	becomes trivial if the abstract representation 
	contains a ``face'' element.

Similar articulations of the idea that more complex models can 
	be justified when modeling raw data have appeared in the machine learning literature. 
For example, Hinton \textit{et al.} note that:

\begin{quote}
Generative models can learn low-level features without requiring feedback from the label, 
	and they can learn many more parameters than discriminative models without overfitting. 
In discriminative learning, 
	each training case constrains the parameters only by as many bits of information 
	as are required to specify the label. 
For a generative model, each training case constrains the parameters by the 
	number of bits required to specify the input~\keycite{Hinton:2006b}.
\end{quote}

\noindent
In the same paper, the authors provide clear evidence 
	that an indirect approach (learn $P(X)$, then $P(Y|X)$)
	can produce better results than a purely direct approach 
	(learn $P(Y|X)$ directly).
Note that a generative model is exactly one half of a compression program:
	the decoder component.
Hinton develops the generative model philosophy at greater length in~\keycite{Hinton:2007}.
Hinton's philosophy is one of the key influences on this book;
	see the Related Work section for further discussion.

The \compist~approach also bears some similarity 
	to the family of methods called unsupervised learning.
The goal of unsupervised learning is to discover useful 
	structure in collections of raw data objects with no labels.
Unlike supervised learning which can be fairly clearly defined,
	unsupervised learning is an extremely broad area,
	making comparisons difficult.
Roughly, there are three qualities that make the \compist~approach different.
First, there is no standard quantity
	that all unsupervised learning researchers
	attempt to optimize;
	this makes it difficult to compare competing solutions.
\Compist~researchers focus exclusively on the compression rate,
	allowing strong comparisons.
Second, 
	unsupervised learning methods are advertised as general purpose
	and widely applicable,
	whereas \compist~methods are targeted at specific datasets
	and emphasize the \textit{empirical} character of the research.
Similarly, \compist~research calls for the construction 
	of large specialized data sets,
	such as the Visual Manhattan Project.
Further discussion is contained in the Related Work appendix.

\mpagebreak
	
\section{Manual Overfitting}

The above discussion involves the problem of overfitting,
	and the ways in which machine learning methods can be used to prevent overfitting.
There is, however, another conceptual issue that machine learning researchers must face,
	which is in some sense even more subtle than overfitting.
This can be called ``manual overfitting'',
	and is illustrated by the following thought experiment.

\subsection{The Stock Trading Robot}

Sophie's brother Thomas also studied to be a physicist,
	but after receiving his doctorate he decides
	that he would rather be rich than tenured,
	and gets a job as a quantitative analyst at an investment bank.
He joins a team that is working on building an automated trading system.
Thomas receives a bit of training in financial engineering
	and machine learning.
When his training is complete, 
	he is given a database of various economic information
	including interest rates, unemployment figures, stock price movements,
	and so on.
This database is indexed by date,
	so it shows, for example, the change in interest rates on January 1st,
	the change on January 2nd, and so on.
Thomas also receives a smaller, second database
	showing the changes in the yen-dollar exchange rates,
	also indexed by day.
Thomas' task is to write a program that will 
	use the economic indicator information to 
	predict the changes in the yen-dollar exchange rate.	
	
Thomas, as a result of growing up in his sister's intellectual shadow,
	and in spite of having a PhD. in physics from a prestigious university,
	has never been confident in his own abilities.
He has read about the generalization theorems of machine learning,
	but is still puzzled by some of the more esoteric mathematics.
So, he decides to attempt to solve the prediction problem using 
	the simplest possible generalization theorem,
	the Hidden Worm theorem mentioned above.

Thomas finds it easy to formulate economic prediction problem
	in terms of the canonical task of supervised learning.
The $x_{i}$ data is the batch of economic indicators for a given day.
The corresponding $y_{i}$ data is a binary label that is true or false if
	the yen-dollar exchange goes up on the following day.
Thomas builds a hypothesis class $C$ by writing down a family of simple rules,
	and forming hypotheses by combining the rules together in various ways.
Each rule is just a binary function, 
	which returns true if a given economic indicator exceeds a certain threshold.
So, for example, a rule might be true if the change in Treasury bill yields
	was greater than 5\% on a certain day.
The specific value of the threshold is a parameter;
	at first, Thomas uses $K=200$ different choices for possible threshold values.
There are also $E=1000$ economic indicators,
	so the number of rules is $KE=2 \cdot 10^{5}$.
Somewhat arbitrarily, Thomas decides to form hypotheses 
	by combining $D=4$ basic rules using the logical pattern
	$(R_{1} \vee R_{2}) \wedge (R_{3} \vee R_{4})$.
This implies that the total size of the hypothesis class is $(KE)^{D}$, 
	and the log size is given by:
	
\begin{eqnarray*}
\log |C| &=& D (\log K + \log E) \\
&\approx& 48.8
\end{eqnarray*}

\noindent
Lacking any strong guidance regarding how to choose these values,
	Thomas decides that the rule should be in error at most 1\% of the time,
	and he wants to be at least 95\% confident that the result holds.
In the terms of the Valiant theorem,
	these imply values of $\epsilon = .01$ and $\delta = .05$.
The two databases also contain $N=500$ days worth of data.
Rearranging Inequality~\ref{eq:inqvaliant},
	he finds that an upper bound on the size of hypothesis class is:

\begin{eqnarray*}
\log |C|_{max} &=& N \epsilon - \log \delta \\
&\approx& 45
\end{eqnarray*}

\noindent
This indicates that there is a problem:
	his original choice for the model class is too large,
	and so using it may cause overfitting.
To deal with the problem, 
	Thomas decides to reduce the number of thresholds from
	$K=200$ to $K=10$.
This new class has $\log |C| \approx 36.8$,
	so generalization should hold if a good hypothesis $h^{*} \in C$ is found.
	
Thomas writes a program that will find $h^{*}$ if it exists,
	by testing each hypothesis against the data.
He notices that a \naive~implementation will require quite a bit of time 
	to check every hypothesis, since $|C|$ is large in absolute terms.
However, he is able to find a variety of computational tricks
	to expedite the computation.
Finally, he completes the learning algorithm,
	invokes it on the data set,
	and leaves it to run for the night.
After a night of troubled dreams,
	Thomas is delighted to discover,
	when he arrives at work the next day,
	that the program was successful in finding
	a good hypothesis $h^{*}$.
This hypothesis correctly predicts the movement of the yen-dollar exchange rate
	for all of the 500 days recorded in the database.
Excited, he begins working on a presentation describing his result.
	
When Thomas presents his method and result at the next group meeting,
	his boss listens to the talk with a blank expression.
On the final slide, Thomas states his conclusion that, with 95\% probability, 
	the hypothesis he discovered should be correct 99\% of the time.
The boss does not really understand what Thomas is saying,
	but his basic management policy is to always push his employees
	to work harder.
So he declares that the 95\% number is not good enough:
	``We need a better guarantee! Do it again!''
Then he storms out of the room.

Thomas, at this point, is somewhat vexed. 
He is not sure how to respond to the boss' demand to improve the generalization guarantee.
A coworker mentions that she had been working with the database before it was given to Thomas,
	and she was surprised to find that it contained any predictive power at all:
	she thought the economic indicator information was too noisy,
	and unrelated to the yen-dollar exchange rate.
Thomas tries the same approach a couple more times,
	using different kinds of hypothesis classes.
Most of them fail to find a good hypothesis,
	but eventually he finds another class that is small enough to produce
	a tighter generalization probability (smaller $\delta$), 
	and also contains a good $h^{*}$.
Relieved, he sends an email to his boss to inform him of the discovery.
The boss congratulates him and promises him a big bonus if the hypothesis works out
	in the actual trading system.

Later on, as Thomas is reviewing his notes on the project,
	he finds something vaguely disturbing:
	the first $h^{*}$, from the original presentation,
	is suspiciously similar to the second $h^{*}$,
	that came from a smaller class and so produced a smaller $\delta$.
Indeed, on further examination, he notices that the two are almost exactly the same,
	except for slighty different threshold values.
What can this mean?
He goes over the math again, and thinks to himself:
	what if I had just cut the size of the original $C$ by 90\%,
	while making sure that it still contained the good hypothesis?
If $|C|$ goes down by 90\%, that means $\delta$ can also be reduced by 90\%
	while leaving the rest of the numbers unchanged.
Then the second analysis will produce a much better generalization probability
	than the first one, even though they both use the same hypothesis!
	
\mpagebreak 

\subsection{Analysis of Manual Overfitting}

The above thought experiment illustrates the problem of \textit{manual overfitting}.
Thomas attempted to use a model complexity penalty and the associated generalization theorem
	to guard against overfitting.
But the theorem has a crucial implicit requirement:
	the model class must not be changed after looking at the data.
By testing many different model classes and cherry-picking one that works,
	Thomas violated this requirement.
What he has really done, in effect,
	is implement a learning algorithm \textit{by hand}.
Instead of using only the algorithm to choose the best hypothesis,
	Thomas performed a lot of the hypothesis-selection work himself.
And this manual selection process implicitly uses a much larger hypothesis class,
	thus causing overfitting.

To see this more clearly,
	imagine Thomas decides to use 1000 model classes
	$C_{1}, C_{2} \ldots C_{1000}$.
He writes a program that checks each model class to determine if it contains 
	a good hypothesis $h^{*}$.
The program finds a good $h^{*}$ in $C_{45}$.
Thomas then concludes that generalization will hold,
	since $\log |C_{45}|$ is small compared to $N$.
But this is nonsense,	
	because he has effectively tested a much larger class.
	
The problem of manual overfittig causes deep conceptual problems
	for the traditional mindset of engineering and computer programming.
The typical approach taken by a programmer when faced with a new task
	is to proceed toward a solution through a series of iterative refinements.
She writes an initial prototype without too much painstaking care,
	runs it, and observes the errors that are produced.
Then she diagnoses the errors,
	makes an appropriate modification to the code,
	and tries again.
She repeats this process until there are no more errors.
This process of repeated code/test/debug loops works fine for normal programming tasks.
Abstractly,
	this kind of trial and error process is utilized by almost
	all successful human endeavors.
But this approach just does not work for machine learning -
	at least, it doesn't work unless the dataset is replenished
	with new samples every time the hypothesis class is changed.
This is almost never done, because of the difficulty of creating labeled datasets.	

%TODO - this thought experiment sucks.
The fallacy of manual overfitting becomes
	self-evident when the learning problem
	is properly phrased in terms of the Compression View.
To see this,
	imagine that Thomas has a colleague who is departing tomorrow for Madagascar.
The colleague has the large database of economic indicator information,
	and will take it with him.
But he also needs the yen-dollar database,
	which has not yet been prepared.
Before the departure,
	Thomas and his colleague agree on a compression format based on a class $C$,
	which includes many rules for predicting the yen-dollar information
	from the rest of the economic indicators.	
Then the colleague leaves,
	and a while later, 
	the yen-dollar database becomes available.
Thomas finds a good rule $h^{*} \in C$,
	and so the format achieves a good compression rate.
However, at this point,
	Thomas' boss intervenes,
	complaining that the compression rate isn't good enough
	(it is expensive to send data to Madagascar).
Having looked at the data,
	Thomas can easily build a new compressor,
	based on some new model class $C_{2}$,
	that achieves much better compression rates.
But to use this new compressor,
	he would have to send it to his colleague in Madagascar,
	thereby defeating the whole point of the exercise.

\mpagebreak

\subsection{Train and Test Data}

Some people might argue that the problem of manual overfitting is not important,	
	because in practice people do not report results obtained from generalization theorems.
In practice researchers report actual results on so-called ``test'' data sets.
The idea is that any benchmark database is partitioned into
	two subsets: the ``training'' data and the ``test'' data.
Researchers are licensed to use the training data for whatever purposes they want.
The usual case is to use the training data to experiment with different approaches, 
	and to find good parameters for their statistical models.
The test data, however, is supposed to be used only for reporting
	the actual results achieved by the algorithm.
The idea is that if the learning algorithm does not have access to the test data, 
	then the performance on the test data is a good approximation
	of its actual generalization ability.
The training/test strategy is fraught with methodological difficulty, however,
	as is illustrated by the following ominous warning on the web page 
	that hosts the Berkeley Segmentation Dataset, 
	a well known benchmark in computer vision:

\begin{quote}
Please Note: Although this should go without saying, we will say it anyway.  
To ensure the integrity of results on the test data set, 
	you may use the images and human segmentations in the training set for tuning your algorithms 
	but your algorithms should not have access to any of the data (images or segmentations) 
	in the test set until your are finished designing and 
	tuning your algorithm~\cite{Martin:2001a}.
\end{quote}

\noindent
The concern underlying this warning
	is that a problem analogous to manual overfitting
	can occur even when using the train/test strategy.
Here, manual overfitting occurs
	when a researcher tries many different model classes,
	invokes a learning algorithm to select optimal 
	hypotheses $h^{*}$ based on the training data,
	measures the performance of $h^{*}$ on the test data,
	and then selects the model class
	that provides the best performance.
Again,
	this procedure gives the illusion that it prevents overfitting,
	because the (computer) algorithm does not 
	observe the test data when selecting the optimal hypothesis.
But what has actually happened is that the developer
	has implemented a learning process by hand,
	and that learning process ``cheats'' by looking at the test data.
	
There are several reasons why the approach advocated by the warning 
	probably does not actually work.
Most basically,
	the prohibition against iterative tuning and design is deeply
	at odds with the natural tendencies of computer programmers.
If the purpose of a system is to produce good segmentations,
	and the quality of the produced segmentations 
	is defined by the score on the test data,
	then if a system does not achieve a good test set score,
	it must have a bug and so require further development and refinement.
Moreover, researchers are often under intense pressure to produce results.
A PhD student whose thesis involves a new segmentation method
	may not be allowed to graduate if his algorithm does not produce
	state of the art results.
Also, there is no way to prevent publication bias (or positive results bias).
A researcher may try dozens of different algorithms;
	if only one of them succeeds in achieving good performance
	on the test data, then that is the one that gets turned into a paper and submitted.
Finally, the underlying point about manual overfitting is that it does not
	matter whether an algorithm or a human does the hypothesis selection.
If the hypothesis is selected from too large a class, 
	overfitting will result.
The idea of benchmarking is that is provides the community with a way 
	of finding the ``best'' approach.
But this just means that now it is the research community that does the selection,
	instead of an algorithm or a human.
	
\mpagebreak	

\subsection{\Compist~Solution to Manual Overfitting}

Overfitting is a subtle and insidious problem,
	and it took statisticians many years to fully understand it
	and to propose safeguards against it.
Arguably,
	this process is still going on today.
Manual overfitting is a similarly subtle problem,
	which the field has not yet begun to truly recognize.
Perhaps future researchers will propose
	elegant solutions to the problem.
There is, however,
	a direct method to prevent manual overfitting that is available today.

The solution is: pick a model class \textit{now},
	and use it for all future learning problems.
This solution is, of course, quite drastic.
If the model class selected today is of only moderate complexity,
	then some learning problems will be rendered insolvable
	simply because of prior limitations on the expressivity of the class.
If, on the other hand,
	the model class is highly complex,
	it potentially includes a hypothesis that will work
	well for all problems of possible interest.
But such a large and complex class would be extremely prone to overfitting.
In order to avoid overfitting when using such a class,
	it would be necessary to use a huge amount of data.
	
The \compist~answer to manual overfitting is exactly this strategy.
\Compist~research employs the same model class, 
	based on a programming language, 
	to deal with all problems.
Since this model class is extremely (in fact, maximally) expressive,
	it is necessary to use vast quantities of data
	to prevent overfitting.
As described above,
	this requirement is accommodated by modeling
	the raw data objects, such as images or sentences,
	instead of the labels.
Because the raw data objects have much greater information content,
	overfitting can be prevented.

Notice how the \compist~approach \textit{rehabilitates}
	the natural code/test/debug loop used by programmers.
\Compist~researchers can propose as many candidate
	models as time and imagination allow,
	with no negative consequences.
In the \compist approach to learning,
	there is only one basic problem:
	find a good hypothesis (compressor)
	to explain the empirical data.
This search is extremely difficult,
	but it is perfectly legitimate to 
	perform the search using a combination of human insight
	and brute computational force,
	since the model class never changes and complex models
	are appropriately penalized.	
In contrast,
	the traditional approach to learning
	involves two simultaneous goals.
The researcher must find a small class $C$
	that is expressive enough to solve the problem of interest,
	and develop an algorithm that can find a good 
	hypothesis $h^{*} \in C$ based on the data.
This suggests the legitimacy of experimenting with different
 	model classes and associated algorithms.
But the idea of manual overfitting implies that this procedure
	contains a logical error.

\section{Indirect Learning}
	
\subsection{Dilbert's Challenge}

Consider the following scene that takes place in the 
	dystopian corporate world portrayed in the Dilbert cartoon.
Dilbert is sitting at his cubicle,
	working sullenly on some irrelevant task,
	when the pointy-haired boss comes up to him and delivers the following
	harangue:
	
\begin{quote}
I've been told by the new strategy consultants that exactly one year from today, 
	our company will need to deploy a highly robust and scalable web services application.
Deploying this application on time is absolutely critical to the survival of the company.
Unfortunately, the consultants won't know what the application
	is supposed to do for another 11 months.
Drop everything you're doing now and get started right away.
\end{quote}

At first, this seems like a unreasonable demand.
Surely it is impossible for anyone - 
	let alone the beleaguered Dilbert - 
	to develop any kind of significant software system in a single month.
In a month, 
	a team of highly skilled and motivated developers 
	might be able to develop a \textit{prototype}.
And software designs are a bit like pancakes:
	one should always throw away the first batch.
Even if, by some miracle, Dilbert got the design right on the first attempt,
	he would still have to do several iterations of polishing,
	debugging, and quality assurance before the software was ready for a public release.
	
Still, after thinking about the problem for a while,
	Dilbert begins to believe that it is actually an interesting challenge.
He begins to see the problem in a new light:
	the specifications are not completely unknown, merely incomplete.
For example, the boss mentioned that the application will need to be ``scalable''
	and will involve ``web services''.
Dilbert has read stories about how highly skilled programmers,
	by working nonstop and using powerful tools, 
	were able to build sophisticated software systems very rapidly.
Dilbert decides to use the 11 month preparation period 
	to familiarize himself with various technologies
	such as web application frameworks, debugging tools, 
	databases, and cloud computing services.
He also resolves to do a series of high-intensity training simulations,
	where he will attempt to design, implement, and test a 
	software system in the space of a single month.
Dilbert has no illusions that this preparation will allow him
	to perform as well as he could with a full year of development time.
But he is confident that the preparation
	will make his ultimate performance much better 
	than it would otherwise be.
	
Situations that are conceptually similar to Dilbert's Challenge
	arise very often in the real world.
The military, for example, 
	has no way to predict when the next conflict will take place,
	who the opponents will be, 
	or under what conditions it will be fought.
To deal with this uncertainty,	
	the military carries out very general preparations.
Sometimes these preparations are ineffective.
During the occupation of Iraq,
	the United States Army deployed a large number of heavily armored Abrams tanks,
	which contributed very little to the urban counterinsurgency efforts
	that constituted the core of the military mission.
Overall, though, the military's preparation activities
	are generally successful.

Another example of a Dilbert Challenge appears in the context	
	of planning for the long-term survival of humanity.
Many people who think about this topic are concerned
	about catastrophic events such as asteroid impacts 
	that could destroy human civilization or completely wipe out humanity.
While each individual category of event is unlikely,
	the overall danger could be significant,
	due to the large number of different catastrophes:
	plague, nuclear war, resource depletion, environmental disaster, and so on.
By the time the dire event is near enough to be predicted with certainty,
	it might be too late to do anything about it.
The futurists must therefore attempt to prepare for an as-yet-unspecified danger.

A full consideration of Dilbert Challenges that appear in practice,
	and methods for solving them,
	is the subject for another book.
The relevance of the idea to the present discussion
	is that the precise nature of real-world cognitive tasks
	are often unknown in advance.
For this reason, learning should be thought of as a general
	preparation strategy, 
	rather than a direct attempt to solve a particular problem.
This idea is illustrated in the following thought experiment.
	
\mpagebreak
	
\subsection{The Japanese Quiz}

\begin{figure}
\begin{centering}
\includegraphics[width=\textwidth]{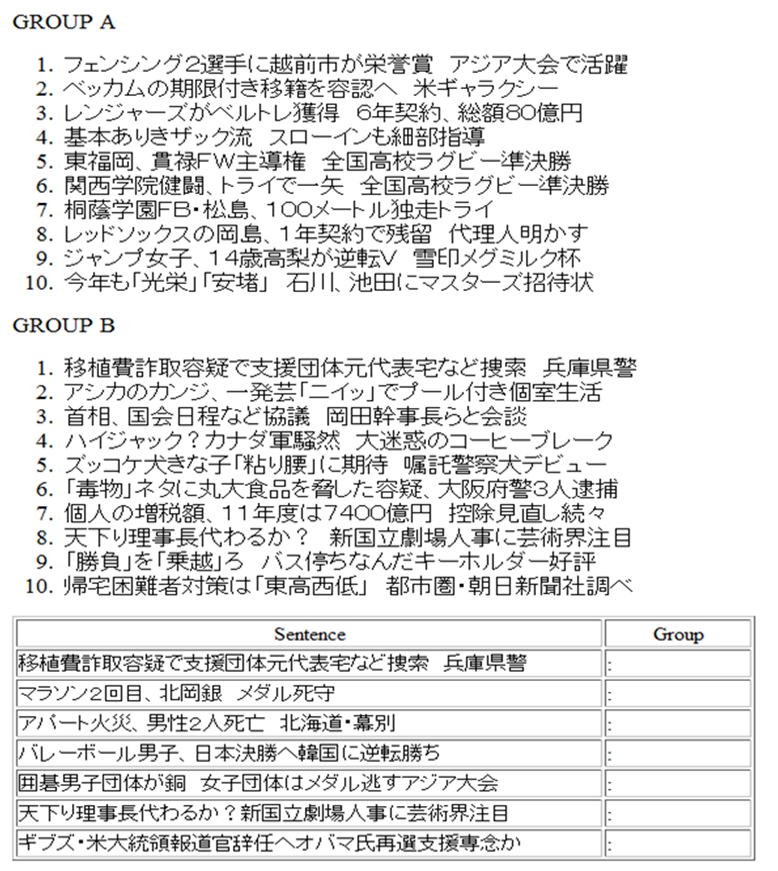} 
\caption{
Japanese language quiz.
}
\label{fig:jsentences}
\end{centering}
\end{figure}

Sophie is in the city of Vancouver for a conference,
	and has decided to do a bit of sightseeing.
She is wandering around the city, 
	pondering for the thousandth time the implications of the double-slit experiment,
	when a group of well-dressed elderly Asian gentlemen
	approach and request her attention.	
She is vaguely worried that the are members of some strange religious sect,
	but they claim to be members of the 
	Society for the Promotion of Japanese Language and Literature,
	and they are organizing a special contest,
	which has a prize of ten thousand dollars.
They show Sophie a sheet of paper, 
	on which 26 Japanese sentences are written (Figure~\ref{fig:jsentences}).
The first 20 sentences are grouped into two categories marked A and B,
	while the remaining 6 are unmarked.
The challenge, they inform Sophie,
	is to infer the rule that was used to categorize the first twenty sentences, 
	and to use that rule to guess the category of the second set of sentences.

Sophie studies the sentences intently for several minutes.
Her spirits sink rapidly as she realizes the 
	quiz is nearly impossible.
The complex writing is completely incomprehensible to her.
She makes an educated guess by attempting to match  
	unlabeled sentences to similar labeled ones,
	but she has very little confidence in her guess.
She marks her answers on the paper and submits it to one of the gentlemen.
He looks it over for a moment, 
	and then regretfully informs Sophie that she did not succeed.
The sentences, he explains are newspaper headlines.
The sentences of category A relate to political affairs,
	while category B is about sports.
The correct response, therefore, is AABBAB.

The Japanese apologize for the difficulty of the test,
	but then inform Sophie that they will be giving the quiz
	again next year.
She looks at them for a moment, puzzled.
``Do you really mean,'' she asks, 
	``that I can come back here next year and take the same test,
	and if I pass I will win ten thousand dollars?''
Somewhat abashed, the gentlemen respond that it will of course not be \textit{exactly} the same test.
There will be a new set of sentences and a new rule used to categorize them.
But the idea will be basically the same;
	it will still be a Japanese test.
``So,'' Sophie asks, suspiciously, 
	``what is to prevent me from spending the whole year studying Japanese,
	so that I will be ready,
	and then coming back next year to collect the ten thousand dollars?''
The gentlemen reply that there is nothing to prevent her from doing this,
	and in fact giving people an incentive to study Japanese
	is exactly the point of the Society!
	
\hrulefill

The Japanese Quiz thought experiment raises several interesting issues.
First, notice that the challenge facing Sophie is conceptually analogous to the 	
	challenge facing Dilbert mentioned above.
In both cases, the protagonist must find a way to exploit
	a long preparation period in order to 
	perform well on a test that is only incompletely specified in advance.
The difference is that in Sophie's case,
	it is immediately obvious that a preparation strategy exists and,
	if pursued diligently, 
	will produce much improved results on the test.

The second issue relates to the decisive role played 
	by background knowledge in the quiz.
Any skilled reader of the Japanese language will be able to solve
	the quiz easily.
At the same time, 
	a person without knowledge of Japanese
	will be unable to do much better than random guessing.
An untrained person sees the sentences as merely a collection of odd scribbles,
	but the proficient Japanese reader sees meaning,
	structure, and order.	
It is as if,
	in the mind of the Japanese reader,
	the sentences are transformed into a new, abstract representation,
	in which the difference between the two groups of sentences
	is immediately obvious.
In other words,
	two people can observe exactly the same pool of data,
	and yet come to two utterly different conclusions,
	one of which is far superior to the other.

The Japanese Quiz thought experiment poses something of a challenge 
	for the field of machine learning.
It is very easy to phrase the basic quiz in terms of the 
	standard formulation:
	the sentences correspond to the raw data objects $x_{i}$
	and the categories are the labels $y_{i}$.
Using this formulation, any one of the many algorithms for supervised learning
	can be applied to the problem.
However, while the algorithms can be applied,
	they will almost certainly fail,
	because the number of samples is too small.
Solving the quiz requires a sophisticated understanding of Japanese,
	and such understanding cannot be obtained from only twenty data samples.
But the fact that the learning algorithms will fail is not the problematic issue,
	since the task cannot be solved by an intelligent unprepared human, either.
The problematic issue relates to the preparation strategy for next year's test.
For a human learner, 
	it is obvious that an intense study of Japanese
	will produce a decisive advantage in the next round.
This suggests that the same strategy should work
	for a machine learning system.
The problem is that machine learning theory provides no standard way to operationalize
	the strategy ``study Japanese''.
Since the preparation strategy will provide a decisive performance advantage,
	the fact that the philosophy of machine learning provides no standard way to formalize it
	indicates that the philosophy has a gaping hole.	
	
In contrast,
	the \compist~philosophy provides a clear 
	way to formalize the preparation process in computational terms.
To prepare for the quiz means to obtain 
	a large corpus of raw Japanese text
	and perform a CRM-style investigation using it as a target database.
A system that achieves a strong compression rate 
	on the text corpus can be said to ``understand'' Japanese 
	in a computational sense.
The abstractions and other knowledge useful for compressing the text
	will also be useful for solving the quiz itself;
	this is just a restatement of the Reusability Hypothesis of Chapter~\ref{chapt:compmethod}.

\mpagebreak	
	
\subsection{Sophie's Self-Confidence}

Sophie has always been a bit of a radical, and,
	upon returning from the conference,
	decides to neglect her research and teaching duties
	and dedicate all of her free time to learning Japanese.
She is also an autodidact and a loner,
	so she does not bother to sign up for classes.
Instead, she decides to hole herself up in the library 
	with a large quantity of educational material:
	textbooks, dictionaries, audio guides, and so on.
She also brings a lot of Japanese pop culture products:
	TV shows, movies, romance novels, comic books, and radio programs.

At the beginning of her studies, 
	Sophie is very aware of her own lack of understanding.
The complicated \textit{kanji} 
	characters are indistiguishable and meaningless to her.
When listening to the spoken language,
	the phonemes blend together into an unintelligible stream
	from which she cannot pick out individual words.
Once she begins to study, of course, the situation changes.
She learns to read the simplified \textit{hiragana} and \textit{katakana} syllabaries, 
	as well as a couple of the basic \textit{kanji}.
She memorizes some of the most common words,
	and learns the basic structure of a Japanese sentence,
	so that she can usually pick out the verb and its conjugation pattern.
	
After a couple of months,
	Sophie has absorbed most of the material in the textbooks.
At this point she begins to rely almost exclusively on cultural material.
She watches television programs,
	and is often able to glean new words and their meaning by 
	matching the spoken Japanese to the English subtitles.
With the help of an online dictionary,
	she translates articles from Japanese newspapers into English.
She creates groups of similar \textit{kanji},
	and invents odd stories for each group,
	so that she can remember the meaning of a character by recalling the story.
She cultivates a taste for Japanese pop music,
	which she listens to on her MP3 player at the gym.

Finally, after a year of intense study,
	Sophie's comprehension of the language is nearly complete.
She knows the meaning and pronounciation of the two thousand most important \textit{kanji},
	and she is familiar with all of the standard grammatical rules.
She can read books without consulting a dictionary;
	if she comes across a new word, she can usually infer its meaning from the context
	and the meaning of the individual \textit{kanji} from which it is constituted.
She can watch television shows without subtitles and understand all of the dialogue;
	she is even able to understand Japanese comedy.
At the conclusion of her year's worth of study,
	she is completely confident in her own command of the language.
	
\hrulefill

Sophie's self-confidence in her new language ability raises an 
	important philosophical question:
	how does she \textit{know} she understands Japanese?
Note that she is not taking classes,
	consulting a tutor, or taking online proficiency tests.
These activities would constitute a source of \textit{external} guidance.
It would be easy for Sophie to justify her self-confidence if,
	for example, 
	she took a weekly online quiz, 
	and noticed that her scores progressively increased.
But she is not doing this,
	and is nevertheless very confident in her own improvement.
Sophie may be exceptionally bright and driven,
	but her ability to guage her own progress is not unique.	
Anyone who has ever learned a foreign language
	has experienced the sensation of gradually increasing competence.

The premise of machine learning is that the same basic principles of learning
	apply to both humans and artificial systems.
In machine learning terms,
	Sophie's brain is a learning system
	and her corpus of Japanese language material is an unlabeled dataset.
The thought experiment indicates, therefore,	
	that it is possible to gauge improvements in the competence
	of a learning system based solely on its response to the raw data,
	without appealing to an external measurement
	such as its performance on a labeled learning task
	or its suitability for use in a practical application.
Sophie can justifiably claim proficiency at Japanese 
	without taking a quiz that requires her to 
	categorize sentences by topic or 
	identify the part of speech of a particular word.
Analogously, 
	it should be possible to demonstrate the competence 
	of a learning system without requiring it  
	to solve a sentence categorization task
	or a part-of-speech tagging problem.

Sophie's self-confidence is easily accounted for by the \compist~philosophy.
The philosophy equates the competence of a system 
	with the compression rate it achieves on a large corpus of raw data.
The system can gauge its own performance improvement
	because it can measure the encoded file size.
A human language learner adopts a mindset analogous to the 
	Circularity Commitment of empirical science:
	the value of a new idea or insight depends solely
	on the extent to which the idea is useful
	in producing a better description of the language.
Furthermore, 
	the belief that substantial open-ended learning will result
	in good performance on quizzes or other applications
	is a simple restatement of the Reusability Hypothesis.
A determined search for a good theory of English must, for example,
	discover significant knowledge relating to part-of-speech tagging,
	since such information is very important in describing the structure of text.
The point of these arguments is not to claim that the brain 
	is actually performing data compression,
	although there is a theory of neuroscience that claims 
	just that~\cite{Barlow:1961,Atick:1992}.
However, 
	in order to explain the self-confidence effect,
	whatever quantity the brain is in fact optimizing as it learns
	must be similar to the compression rate in the sense that
	it can be defined based on the raw data alone.
	
\mpagebreak

\subsection{Direct and Indirect Approaches to Learning}

The traditional perspective on the learning problem
	can be thought of as the ``direct'' approach.
The philosophy of the direct approach can be summed up 
	by the following maxim, due to Vapnik:
	
\begin{quote}
When you have a limited data set, 
	attempt to estimate the quantities of interest directly, 
	without building a general model as an intermediate step~\keycite{Vapnik:1998}.
\end{quote}

\noindent
The reason one ought not to construct a general model
	is that such a model is more complex,
	and thus will run the risk of overfitting
	when the data is limited.
Because of the relationship between model complexity and generalization,
	practitioners of the direct approach must be fanatical about
	finding the simplest possible model.
Indeed, the most successful modern learning algorithm,
	the Support Vector Machine,
	is successful exactly because of its ability to 
	find an especially simple model
	(called a separating hyperplane)~\cite{Vapnik:1996}.
In the direct approach, 
	aspects of the data other than the quantity of immediate interest
	are considered unimportant or inaccessible.
When developing a face detection system,
	the researcher of the direct school does not spend much time considering
	the particular shape of nostrils or the colors of the eyes. 
Insofar as it goes,
	the direct approach cannot be critiqued:
	Vapnik's maxim does indicate the surest way to
	find a \textit{specific} relationship of interest.
	
The \compist~philosophy of learning
	suggests an alternative, ``indirect'' approach. 
This approach is based on the following maxim,
	due to Thomas Carlyle:
	
\begin{quote}
Go as far as you can see; when you get there you will be able to see farther.
\end{quote}

\noindent
Carlyle's maxim suggests that instead of focusing on one specific quantity of interest,
	learning researchers should adopt a mindset of general curiousity.
The researcher should go as far as she can see,
	by learning anything and everything that can be learned
	on the basis of the available data.
In the study of faces, 
	the \compist~is happy to learn about noses, eyes, eyebrows,
	lips, facial hair, eyewear, and so on.
In the study of the Japanese language,
	the \compist~is happy to learn 
	how varying levels of politeness can be expressed by verb conjugation,
	how a single \textit{kanji} can have multiple pronounciations,
	how Japanese family names often evoke natural settings,
	and so on.	
This mindset is analogous to the Circularity Commitment of empirical science,
	where a theory is evaluated based on its ability to describe the data (e.g. face images) well,
	not on its practical utility.
At the same time,
	the \compist~endorses the Reusability Hypothesis,
	which in this context says that a systematic study of 
	the raw data will lead to practical applications.
Knowledge of the shapes of eyebrows and nostrils 
	will help to solve the problem of face detection.
Knowing how Japanese verbs conjugate
	will help to solve the problem of part-of-speech tagging.

\section{Natural Setting of the Learning Problem}

\subsection{Robotics and Machine Learning}

The most obvious application of machine learning methods
	is for the development of autonomous robotic systems.
The field of robotics has blossomed in recent years,
	and there are now many groups across the world
	working to construct various types of robots.
However,
	the interaction between the field of machine learning
	and the field of robotics is not as productive as one might imagine.
In an ideal world, 
	the relationship between roboticists and learning researchers
	would mirror the relationship between physicists and mathematicians.
The latter group works to develop a toolbox of theory and 
	mathematical machinery.
The former group then selects the appropriate tool 
	from the large array of available choices.
Just as Einstein adapted Riemannian geometry for use in the development
	of general relativity,
	roboticists should be able to select, 
	from an array of techniques in machine learning,
	one or two methods that can then be applied to 
	facilitate the development of autonomous systems.
	
Unfortunately, in practice, the theory offered by machine learning
	does not align with the needs of roboticists.
Roboticists often attempt to use machine learning techniques in spite of this mismatch,
	usually with disappointing results.
In other cases roboticists simply ignore ideas in machine learning
	and develop their own methods.
The Roomba, one of the few successful applications of robotics,	
	employs simple, low-level algorithms that produce spiraling,
	wall-following, or even random walk motion patterns.
Other commercial applications of robotics are not autonomous,
	but instead require a human user to control the system remotely.

Part of the reason for the disconnect between the two fields is that
	robots must take actions,
	and the problem of action selection
	cannot be easily transformed into the canonical form of supervised learning.
Another subfield of machine learning,
	called reinforcement learning, 
	is more applicable to the problem of action selection.
This subfield is also hindered by deep conceptual problems,
	as discussed below.
	
\mpagebreak

\subsection{Reinforcement Learning}

All humans are intimately familiar with the opposing qualia called 
	pleasure and pain.
These sensations play a crucial role in guiding our behavior.
A painful stimulus produces a change in behavior 
	that will make it less likely for the experience to be repeated.
After being stung by a bee,
	one learns to avoid beehives.
Conversely a pleasant stimulus produces a behavior shift
	that makes it more likely to experience the sensation again in the future.
The psychological study of how behavior changes in response to pleasure and pain
	is called behaviorism.
% TODO: make sure this is true...?
The most radical theories developed in this area,
	associated with the psychologist B.F. Skinner,
	assert that all brain function can be understood in terms
	of reinforcement~\cite{Skinner:1957}.
	
Critics of this perspective promote a more nuanced view.
While there can be little doubt that reinforcement learning plays a role,
	many aspects of human behavior cannot be accounted for
	by behavioral principles alone.
One such aspect, 
	pointed out by Noam Chomsky in a famous critique of the behaviorist theory,
	is language learning~\cite{Chomsky:1967}.
Parents neither punish their children for making grammatical mistakes,
	nor reward them for speaking properly.
In spite of this, children are able to learn very sophisticated grammatical rules,
	suggesting that humans possess a learning mechanism
	that does not depend on the reward signal.

While few psychologists accept behaviorism as a full explanation of human intellgence,
	the idea has recently taken on a new significance 
	to researchers in a subfield of machine learning called Reinforcement Learning (RL).
These researchers
	attempt to use the reward principle 
	as a tool to guide the behavior of intelligent systems.
The basic idea is simple.
One first defines a reward function for a system,
	that produces positive values when the system performs as desired,
	and negative values when the system fails.
One then applies a learning algorithm that iteratively 
	modifies a controller for the system,
	until it finds a version of the controller that succeeds
	in obtaining a large reward.
There are many such algorithms,
	but they are mostly similar to the simplified algorithm
	mentioned in Section~\ref{sec:simpdescla},
	where $\mathbf{E}$ is the reward function,
	$S(\theta)$ is the controller,
	and the parameters $\theta$ are iteratively refined to maximize reward.
If the algorithm works,
	then the researcher obtains a good controller for the system of interest,
	without needing to do very much work.
	
Because of the connection it makes to human psychology,
	and because of the prospect of automatically obtaining good controllers,
	the idea of reinforcement learning seems quite attractive.
Roboticists often attempt to apply RL algorithms
	to learn controllers for their robots.
Unfortunately, 
	researchers have been frustrated in their ambitions
	by a variety of deep obstacles.
The most obvious issue is that RL algorithms 
	typically represent the world as a number of discrete states.
The algorithms work well if the number of states is not too large.
However, in the real world applications such as robot control,
	the system of interest is multidimensional,
	and the number of states grows exponentially with the number of dimensions.
Take for example a humanoid robot with 53 joints.
Even if each joint can only take 5 different values,
	then the total number of states is $5^{53} \approx 1.1 \cdot 10^{37}$.
And this huge number only suffices to describe the configuration of the robot;
	in practice it is necessary to describe the state of the world as well.
This issue is called the Curse of Dimensionality in the RL literature~\cite{Barto:1997}.

While the Curse of Dimensionality is bad enough,
	the RL approach to learning has another critical conceptual limitation.
This is the fact that while the reward signal can theoretically be used to guide
	behavioral learning (i.e. decision-making),
	it cannot plausibly be used to drive \textit{perceptual} learning.
Intelligent adaptive behavior cannot possibly be achieved without
	a good perceptual system.
The perceptual system must be able to transform an incoming image
	into an abstract description containing the concept ``charging rhinoceros''
	in order for the decision-making system
	to determine that ``run away'' is the optimal action.
The necessity of using a mechanism other than the reinforcement signal
	to guide perceptual learning is illustrated
	by the following thought experiment.
	
\mpagebreak
	
\subsection{Pedro Rabbit and the Poison Leaf}

In the woods behind Thomas' house
	there lives a rabbit named Pedro.
Pedro is a simple creature,
	who lives entirely in the moment,
	avoiding pain and seeking pleasure.
He spends his days eating grass and leaves, 
	avoiding foxes, and chasing after pretty female rabbits.
Though his world is quite narrow,
	he is very familiar with it.
As he wanders, 
	he intently studies the things he observes.
He becomes very knowledgeable,
	about the landscape and the other wildlife.
He also becomes an expert about foliage,
	which is very important to him because it constitutes his diet.
He notices that plant leaves 
	have certain kinds of characteristics
	that tend to repeat over and over again.
For example,
	some leaves are needle-shaped,
	while others are circular or hook-shaped.
Some leaf edges are jagged,
	while others are smooth,
	and others have tiny hairs.

One day Pedro goes a bit further afield than usual.
As he is grazing,
	he comes upon a new type of plant 
	that he has never before encountered.
He takes a moment to study the shape of its leaves,
	which are hook-shaped,
	with wavy indentations on the edges.
The veins run up and down the length of the leaf.
He has never seen this particular combination before.
However, this does not surprise him much,
	as the forest is large and he often encounters new types of plants.
He eats a few of the leaves,
	and then continues on his way.
	
A couple of hours later, 
	Pedro becomes intensely sick.
He grows feverish and starts to see double.
He can barely manage to drag himself
	back to his burrow before he collapses.

When Pedro wakes up the next day,
	his entire body is throbbing with pain.
Recalling his violent sickness of the day before,
	he is surprised that he is still alive.
He makes a solemn vow to his personal rabbit deity
	to do whatever it takes to avoid 
	experiencing that kind of awful sickness again.
Since the pain seems to radiate outward from his stomach,
	he realizes that the sickness must have
	been caused by something he ate.
Then he remembers the unusual new plant,
	and instantly concludes that it was the cause of his present grief.
After another day of resting,
	he starts to get hungry and slowly drags himself up
	to go in search of food.
	
Pedro slowly wanders around the forest for a while,
	until he spies a plant that looks
	like it will make a good lunch.
But just as he is about to bite off a leaf,
	he stops, petrified with terror.
The leaf is green, thin,
	and grows on a bush -
	just like the one that poisoned him.
He slowly turns around,
	and finds another leaf.
It is also green, thin, and growing from a bush.
Pedro realizes he has quite a dilemma.
He has no choice but to eat leaves,
	or else he will starve.
But he cannot bear to risk eating another poison leaf.

Thinking over the issue slowly in his rabbit fashion,
	he thinks that perhaps
	there is a connection between the appearance
	of a leaf and whether or not it is poisoned.
This might allow him to determine
	whether a leaf was poisoned or not without eating it.
But even if this were true,
	how can he possibly recognize the poisonous leaf if he sees it again?
He has only seen it once!

Then Pedro recalls the unique appearance of the leaf,
	and its strange combination of hook shape,
	wavy edge indentation,
	and longitudinal venation pattern.
By fixing that particular combination of characteristics in his mind,
	he realizes that he \textit{can} recognize the poison leaf.
With a flood of relief,
	he goes back to his old lifestyle of leaf-grazing,
	confident that if he ever encounters the poison leaf again,
	he will be able to avoid it.

\hrulefill
	
This thought experiment is interesting because it shows a shortcoming
	of the Reinforcement Learning philosophy.
In terms of reinforcement,
	Pedro has received a long series of minor positive rewards
	from eating regular leaves,
	followed by a massive negative reward from eating the poison leaf.
Furthermore, his life depends
	on \textit{rapidly} learning to avoid the poison leaf in the future:
	if it takes him multiple tries to learn an avoidance
	behavior, he will probably not survive.
It also presents a dilemma for supervised learning,
	because Pedro needs to be able to obtain the correct
	classification rule on the basis of a \textit{single} positive example.

In order to achieve the necessary avoidance behavior,
	Pedro must have a sophisticated perceptual understanding
	of the shapes of leaves.
If Pedro saw a leaf merely as a green-colored blob,
	it would be impossible for him to decide
	whether or not it was poisonous.
It is only because Pedro has a sophisticated perceptual system,
	that can identify various abstractions in the raw sensory data 
	such as the shape, edge type, and venation pattern of the leaf,
	that Pedro is able to determine the correct classification rule.
The problem is that this sophisticated perceptual system
	can never be developed on the basis of reward adaptation alone.
That is because,
	in all his previous experience,
	it \textit{was} sufficient for the perceptual system
	to report only the presence of a thin, green-colored blob
	growing on a bush.
The precise appearance of the leaf
	never mattered in terms of reward,
	so reward adaptation could never be used to optimize the perceptual system.
This implies that in order to learn
	the perceptual system,
	Pedro's brain must have relied on some optimization
	principle other than reinforcement adaptation.

\mpagebreak

\subsection{Simon's New Hobby}

Thomas has a 14 year old son named Simon, 
	who is intelligent and a bit eccentric.
Simon has a tendency to become obsessed with various hobbies,
	focusing on them to the exclusion of all else.
Thomas worries about these obsessions,
	but can't bring himself to forbid them,
	and even feels compelled to fund them liberally using
	his investment banker's salary.
Sophie and Simon are fond of one another,
	and Thomas often tries to get Sophie to influence 
	Simon in various ways.

A couple of years ago,
	Simon became obsessed with bird-watching.
He bought expensive binoculars, read all the field guides,
	and went on bird-watching expeditions to nearby state parks.
On these expeditions he would find a good observation place
	and lie in wait for hours,
	hoping to catch sight of a Grey Heron or a Piratic Flycatcher.
At the moment of sighting such a creature,
	he felt a dizzying burst of exhilaration.
Eventually, though,
	impatience proved to be a stronger motivating force.
Simon enjoyed seeing the birds,
	but the torment of the long wait was simply too much.

Recently,
	Simon has acquired a new hobby: digital surveillance.
Sophie hears bits and pieces about this from Thomas,
	in the form of terse emails.
``Simon has got cameras set up all over the house
	and in the back yard.''
``Now he's got microphones, too, he's recording everything we say.''
``Sophie, now he's attached the camera to a remote control car,
	he's driving it around and spying on people and looking for birds.''
``He's got some kind of pan-tilt-zoom camera set up
	and he's controlling it with his computer.''
``He's got a whole room full of hard drives. He says he has a terabyte of data.''
``Sophie, you've got to help me here. 
The kid has a remote control helicopter flying around looking for birds,
	I'm afraid it's going to crash into someone's house.''

Sophie wants to help her brothers,
	but also likes to encourage Simon's hobbies.
On her next visit,
	she suggests to him that he write a computer program
	that can automatically recognize a bird from its call.
By connecting this program to the treetop microphones he has set up,
	he could be alerted if a interesting bird was nearby,
	and go see it without suffering through a long wait.
Simon is very excited by this idea and so 
	Sophie buys him a copy of Vapnik's book
	``The Nature of Statistical Learning Theory''~\keycite{Vapnik:1998}.
She tells him that it is a book about machine learning,
	and some of the ideas might be useful 
	in developing the bird call recognition system.
Thomas, observing this, expresses some skepticism:
	Simon can't possibly understand the generalization theorems,
	since nobody understands the generalization theorems.
Sophie suggests that Thomas might be surprised by the intellectual powers
	of highly motivated youngsters,
	and that anyway it is better for Simon to be studying mathematics
	than crashing his mini-helicopter into houses.

A few days later, though, Sophie receives a telephone call from Simon.
``Aunt Sophie,'' he says, ``this book is \textit{useless} to me!
It says right on the first page that it's about \textit{limited} data.
I don't care about limited data problems.
I've got 15 microphones recording at 16 kilobits per second, 24 hours a day.''
Sophie explains to Simon that to use the statistical learning ideas
	to build a classifier to recognize the bird calls,
	he will need to label some of the audio clips by hand.
So while he might have huge amounts of raw data,
	the amount of data that is really useful will be limited.
``So, let's say I label 100 ten second audio clips.
That's about 16 minutes, 
	but I have thousands of hours worth of data,
	what am I supposed to do with it?''
Sophie responds that this portion cannot really be used.
Simon listens to this explanation sullenly.
Then he asks how many audio clips he will need to label.
Sophie doesn't know, but suggests labelling a hundred,
	and seeing if that works.
Then he asks how he is supposed to find the bird calls in the audio data:
	that is exactly the problem that the detector is supposed to solve for him.
Sophie doesn't have a good answer for this either.
Simon does not seem happy about the situation.
``I don't know about this, Aunt Sophie... 
	it just seems really inefficient, and \textit{boring}.''

\mpagebreak

\subsection{Foundational Assumptions of Supervised Learning}

Supervised learning theorists make a number of important assumptions
	in their approach to the problem.
By examining these assumptions,
	one can achieve a better understanding of the mismatch
	between the tools of machine learning and the needs
	of robotics researchers.
To a great extent,
	the philosophical commitments of supervised learning are expressed
	by the first sentence of the great work by Vapnik
	(emphasis in original):

\begin{quote}
In this book we consider the learning problem as a problem of finding a desired dependence
	using a \textit{limited} number of observations~\cite{Vapnik:1998}.
\end{quote}

\noindent
According to Vapnik,
	learning is about finding a specific ``desired dependence'',
	and all attention is devoted to discovering it.
In practice,
	the dependence is almost always the relationship between
	a raw data object such as an image,
	and a label attached to it.
In this view, 
	there is very little scope for using raw data objects that have
	no attached label.
Because all data objects need a label,
	and it is typically time-consuming and expensive to produce labeled datasets,
	the number of data samples is fundamentally limited.
This restriction on the amount of the available data
	constitutes the core challenge of learning.	
Because of the limited amount of data,
	it is necessary to use extremely simple models
	to avoid overfitting.
A third assumption is embedded in the word ``observations'',
	which in practice implies the data samples are independent 
	and identically distributed (IID).
The next section argues that in the natural setting of the learning problem,
	none of these assumptions are valid.

Other machine learning researchers have
	proposed different formulations of the learning problem
	that depend on different assumptions.
Focusing specifically on Vapnik's formulation is a simplification,
	but for several reasons it is not entirely unreasonable.
First, Vapnik's formulation \textit{is} very nearly standard;
	many other approaches are essentially similar.
Many machine learning researchers
	work to extend the VC theory or apply it to new areas~\cite{Joachims:1998}.
Second, Vapnik's theory is the basis for the Support Vector Machine (SVM) algorithm,
	one of the most powerful and popular learning algorithms currently in use.
Much research in fields such as computer vision and speech recognition
	use the SVM as an ``off-the-shelf'' component.

\mpagebreak
	
\subsection{Natural Form of Input Data}

There are several perspectives from which one could approach 
	the study of learning.
A psychologist might study learning to shed light on the function of the human brain.	
A roboticist might wish to use learning to construct autonomous systems.
And a pure observer,
	like Simon, might be interested in learning as a way 
	to make sense of a corpus of data.
These problems are not identical,
	but they do share several important characteristics.
Neither the supervised learning approach
	nor the reinforcement learning approach is 
	exactly appropriate for these problems.
The reason is that those formalisms make very specific
	assumptions about the form of the input 
	data that is available to the learning system.
The basic characteristics of the natural form of the 
	input data are described below.
	
\subsubsection{Data is a Stream}
\label{sec:datastream}

For robots and organisms operating in the real world,
	and also for pure observers,
	data arrives in the form of a stream.
Each moment brings a new batch of sensory input.
Each new batch of data is highly dependent on the previous batch,
	so any attempt to analyze the data as a set of independent samples
	is misleading from the outset.
	
Consider, for example, a camera placed in the jungle.
The camera records a 12 hour video sequence in
	which a tiger is present for five seconds.
How much data about the tiger does this represent?
If one considers that the appearance of the tiger constitutes
	a single data sample,
	then it will be very difficult to learn anything without overfitting,
	because of the small size of the data set.
Perhaps, on the other hand,
	one notes that the five seconds corresponds to 150 image frames.
Using this substantially increased figure for the size of the data
	justifies the use of a much more complex model.
But it also causes problems with the traditional formulation,
	since the 150 frames are certainly not independent:
	each frame depends strongly on the previous frames.
One could take this even further, 
	and say that within each $640 \cdot 480$ image there are 
	768 subwindows of size $20 \cdot 20$, 
	so there are really $768*150 = 115,200$ positive data samples.
The success of any given approach to learning from this data set
	will depend strongly on the data partitioning scheme employed,
	but the standard formulation of machine learning
	gives no instruction about which scheme is correct.

\subsubsection{The Stream is Vast}
\label{sec:vastdata}

Not only does data arrive in the form of a stream,
	it is a \textit{enormous} stream.
A cheap digital web camera,	
	available from any electronics or computer store,
	can easily grab frames at 30 Hz, 
	corresponding to data input rates on the order of megabytes per second.
And cameras are not the only source of data:
	robotic systems can also be equipped with microphones, 
	laser range finders, odometers, gyroscopes, and many other devices. 
Humans experience a similarly huge and multi-modal deluge of sensory data.

Standard formulations of machine learning do not take into account
	the vast size of the input data.
Indeed, Vapnik's formulation explicitly assumes that the input data is limited,
	and that the limitations constitute the core of the challenge of learning.
The reinforcement learning formulation is also maladapted 
	to the natural form of the input data.
RL theory assumes that observations are drawn from a set of symbols
	and delivered to the learning system.
In principle, the theory works regardless of the form the observations take.
But in practice, learning will be disastrously inefficient
	unless the observations transformed from raw data into
	abstract perceptions that can used for decision making.
If the learning system receives an observation labelled ``tiger'',
	it may reasonably conclude that the correct action is ``escape''.
But an observation delivered merely as a batch of ten million pixels
	is nearly useless.
		
\subsubsection{Supervision is Scarce}
\label{sec:scarcesupervision}

Biological organisms,
	real-world robots,
	and observational learning systems all receive data in a vast, mult-modal stream.
These entities may use different types of supervision mechanisms
	to produce adaptive behavior.
An organism is guided by a natural reward signal that is determined
	by its biological makeup.
A real robot can be guided by an artificial reward signal that
	is decided by its developer.
In the context of observational learning,
	the analogue of the reward signal is the set
	of labels attached to the data.
These labels can be used to guide the learning system
	to reproduce some desired classification behavior.
The key point is that in all three cases,
	the information content of the supervisory data is small
	compared to the vast sensory input data.
	
Simple introspection should be sufficient to prove this claim
	in the context of real organsisms.
Humans certainly experience moments of intense pleasure and pain,
	but most of our lives is spent on a hedonic plateau.
On this plateau the reward signal is basically constant:
	one is content and comfortable, neither ecstatic nor miserable.
Since the reward signal is roughly constant, very little can be learned from it.
At the same time, however, our sensory organs are flooded with complex information,
	and this data supports extensive learning.

The same thing holds for information received from parents, teachers or supervisors.
A child may receive some explicit instruction in language from his mother,
	but most instruction is purely implicit.
The mother does not provide a comparative chart illustrating
	the differences between grammatical and ungrammatical speech;
	she only provides many examples of the former.
The feedback received by students from their teachers also 
	has low information content.
An aspiring young writer could not plausibly plan on mastering the 
	art of writing from the comments made by her professors
	on her term papers.
Children learn difficult motor skills such as running, jumping 
	and hand-eye coordination without any explicit instruction.

In the Reinforcement Learning approach,
	a natural tendency is to attempt to have robots
	learn sophisticated behavior
	by using complex reward functions.
Imagine that a researcher who has built a humanoid robot,	
	and now wishes to train it to fetch coffee in an office environment.
The simplest method would be to give the robot a big reward when it 
	arrives with the coffee. 
This scheme implies, however, 
	that the robot will have no guidance at all until it successfully
	completes a coffee-fetching mission.
Until then, it will explore its state space more or less at random
	(it may, for example, spend a long time trying to do hand stands).
In this case the inclination of the RL researcher is
	to develop a more complex reward signal.
The robot may be given a small reward just for obtaining the coffee,
	since this is an important first step toward delivering it.
But this is still not adequate,
	since it will require a long time,
	and many broken cups and much spilled coffee, 
	before the robot ends up with a full cup in its hand as a
	result of random exploration.
One could then go further,
	defining all kinds of small rewards for each small component of the full task,
	such as standing, walking, navigating the hallways, grasping the cup, and so on.
One would also need to define negative rewards for spilling the coffee
	or bumping into people.
	
This complex reward strategy may work, but it is unattractive for several reasons.
First, it requires enormous efforts on the part of the researcher.
It almost discards the entire point of machine learning,	
	which is for computers to discover solutions on their own.
Also, the resulting system will be extremely brittle.
If it researcher decides to change course in mid-development
	and teach the robot to mow lawns instead of fetching coffee,
	a huge amount of the previous work will have to be discarded.
Third, it still does not address the problem of learning perception.

\mpagebreak

\subsection{Natural Output of Learning Process}

\subsubsection{Dimensional Analysis}

In physics there is a fascinating technique called dimensional analysis,
	that allows one to find the rough outline of a solution
	without doing any work at all.
Dimensional analysis relies on the fact that
	most quantities used in physics have units,
	such as kilograms, seconds, or meters.
Dimensional analysis provides a nice way of checking for obvious
	mistakes in a calculation.
Say the quantity $A$ is measured in meters,
	and the quantity $B$ is measured in seconds.
Now, if the result of a calculation involves a term of the form $(A+B)$,
	then there is obviously a mistake in the calculation,
	since it is nonsense to add meters to seconds.
	
Dimensional analysis actually permits even more powerful inferences.
Consider the standard problem where a ball oscillates	
	due to the influence of a spring.
There are no other forces,
	and the system obeys Hooke's law so that $F = -kx$,
	where $F$ is the force from the spring, 
	$x$ is the displacement of the ball.
The problem contains two parameters:
	$m$, the mass of the ball, measured in $kg$,
	and $k$, the spring constant,
	measured in $\frac{kg}{sec^{2}}$.
The goal is to find $T$, the period of the oscillation,
	which is measured in $sec$.
There is only one way to combine $k$ and $m$ to get 
	a quantity that has the appropiate units:
	$\sqrt{\frac{m}{k}}$.
That is in fact the right answer, up to a constant factor
	(the actual answer is $T = 2 \pi \sqrt{\frac{m}{k}}$).
So dimensional analysis tells us the form of the answer,
	without requiring any actual calculation at all.

\subsubsection{Predicting the Stream}

Dimensional analysis suggests that the correct form of an answer
	can be deduced merely from the inputs and outputs of a problem.
In the natural setting of the learning problem,
	the input is a vast, multimodal data stream with minimal supervision.
The output is some computational tool
	that can be used to produce adaptive behavior.
There seems to be one \textit{natural} answer:
	the learning algorithm should attempt to predict the stream.

The problem of prediction is extremely rich,
	in the sense that a wide array of potential methods
	can be used to contribute to a solution.
The most basic prediction scheme is trivial:
	just predict each pixel in an image sequence based
	on its value at the previous timestep.
Conversely, perfect prediction is impossible - 
	it would require, for example, 
	the ability to predict the outcomes of historical events,
	since such events interact with day to day experience.
Between the two extremes exist a wide variety of possible
	methods, techniques, and theories that will yield
	varying levels of predictive power.

The relationship between compression and prediction has already been established.
Because of this connection,
	the quality of a prediction system can be evaluated by 
	measuring the compression rate it achieves on a data stream 
	produced by the phenomenon of interest.
The \compist~approach is appropriate
	for all three types of learning described above.
Simon can conduct a \compist~investigation
	of the audio data produced by his treetop microphones.
Such an investigation could lead
	to a method for recognizing a bird species 
	from its song.	
A biological agent could conduct a CRM-like investigation
	based on the rich sensory data it observes.
In the case of Pedro Rabbit,
	such an investigation could produce a sophisticated 
	understanding of the appearance of plant leaves,
	\textit{before} such an understanding becomes critical for survival.
An autonomous robot could,
	in principle, be programmed to conduct a CRM-style investigation
	of the data it observes.
In practice, however,
	it is probably more reasonable for a group of \compist~researchers to 
	develop visual systems ahead of time,
	and then port the resulting systems onto the robot when it is built.

% TODO: put this back in
%\subsection{Virtual Labels and Compression}
%
%After thinking about Simon's dillemma for a while,
%	Sophie recalls her own experience with the data compression problem.
%She calls Simon up and explains to him
%	how data compression can be viewed as a way of doing empirical 
%	science on the phenomenon of interest.
%Since most of the variation in the audio data comes from the bird songs,
%	attempting to compress the streams will require theories
%	of bird songs.
%``But,'' she says, ``I'm not sure how exactly you would go about compressing the audio data.''

\mpagebreak

\subsection{Synthesis: Dual System View of Brain}

The previous discussion expressed some skepticism related
	to the ability of the reinforcement learning principle
	to provide a full account of intelligence.
However, it would be foolish to completely dismiss the significance
	of the reward signal;
	humans and other organisms certainly do change their behavior
	in response to pleasure and pain.
It is easy to reconcile the reinforcement learning approach
	with the \compist~approach by 
	noting that the ability to predict is extremely
	useful for maximizing reward.	
Given a powerful prediction system,
	it is easy to connect it to a simple decision-making algorithm
	to produce a reward-maximizing agent.
To see this, consider an agent that has experienced a large amount of data
	with visual, audial, and reward components.
The sensory data is also interleaved with a record
	of the actions taken by the agent.
Assume that the agent's learning process has obtained a very good predictive model of the data.
Then the following simple algorithm can be applied:

\begin{itemize}
\item Create a set of proposal plans.
\item For each plan
	use the predictive model to estimate the reward resulting
	from executing it in the current state.
\item Execute the plan corresponding to the highest predicted reward.
\end{itemize}

This decision-making algorithm is not very elegant.
It also incomplete,
	since it does not specify a technique for 
	creating the list of proposal plans.
Moreover,
	the strategy may fail badly
	if the reward prediction function is inaccurate.
However,
	these limitations are not entirely discouraging,
	because, introspectively, 
	humans seem to be very poor planners.
Humans are often paralyzed with indecision when presented with difficult choices,
	and frequently make decisions simply by imitating others,
	or by repeating their own past behavior.
This suggests that \naive~methods for constructing
	the list of proposal plans may provide adequate performance.
The fact that humans can act in a reasonably adaptive way	
	without using a very sophisticated planning system
	suggests that the inelegant algorithm given above can achieve
	comparably acceptable performance.

The above argument implies that prediction is \textit{sufficient} 
	for obtaining reasonably good reward maximizing behavior.
A moment's reflection shows
	that in many cases prediction is also \textit{necessary}.
A system
	that does not make predictions could, perhaps,
	be used for very controlled problems such 
	as maximizing the production rate of a paper factory.
But real agents
	need to make predictions to deal with
	situations they have never before encountered.
It is essentialy to know in advance that jumping off a cliff
	will be harmful,
	without actually performing the experiment.
This is obvious, 
	but standard reinforcement learning algorithms cannot learn 
	to avoid bad states without visiting them at least once.
	
The more general principle here is the 
	idea that learning can take place without input from a
	teacher, critic, supervisor, or reward signal.
This is true even if the ultimate goal is to please the teacher
	or maximize the reward signal.
The poison leaf thought experiment 
	indicated that in order to survive,
	Pedro needed extensive knowledge 
	of the shapes of leaves \textit{before} he consumed the poisoned one.
Such extensive knowledge could not be produced by a purely reward-based 
	learning process, 
	since before eating the poison leaf the reward signal 
	was not correlated with the appearance of leaves.
The poison leaf thought experiment
	reiterates the lesson of the Japanese quiz:
	for some cognitive tasks,
	good performance is only possible when extensive background learning
	has already been conducted.

These ideas suggest a new view of how the brain operates.
In this view,
	the brain contains two learning systems.
The first is an \textit{oracular} system,
	whose goal is to make predictions
	(particularly action-conditional predictions).
To do this, the oracular system learns 
	about a large number of abstractions such as space, time,
	objects, relationships, and so on.
Because it learns from the vast sensory data stream,
	the oracular system can grow to be quite sophisticated,
	and is able to obtain highly confident predictions
	about a wide variety of events.
The oracular system learns in an unsupervised manner
	and attempts to maximize a quantity analogous 
	to the compression rate.

To complement the oracular system,
	the brain also contains an executive system.
The purpose of this system is to exploit the oracular system
	to produce adaptive behavior.
This system uses the predictions and abstractions produced
	by the oracular system to help in optimizing the reinforcement signal.
The executive system learns from the reward signal,
	which it attempts to maximize.
But because the reward signal has relatively low information content,
	the executive system does not achieve a high level of sophistication.
It produces clear signals about how to act only in special situations,
	such as when confronted with a charging rhinoceros.
In other situations it produces only vague inclinations or urges.

This two-module explanation for intelligence is
	attractive from both an explanatory perspective and an engineering perspective.
In terms of explanation,
	it shows how children can learn language at a high level of proficiency,
	in spite of the fact that they are not punished for making
	grammatical mistakes or much rewarded for speaking correctly. 
It provides a way for Pedro Rabbit to avoid the poison leaf after his initial encounter with it,
	since the general purpose system has provided him
	with an advanced perceptual understanding of the shape of leaves.
It also accounts for the contrast between
	the strong confidence with which the brain can make perceptual predictions
	and the ambivalence with which it makes decisions.
	
In terms of engineering,
	the two-component system introduces a useful modular decomposition
	by splitting the notion of intelligence into two parts.
The oracular system performs general purpose, open-ended learning,
	which produces a variety of high-level abstractions.
To a first approximation,
	the learning process takes into account only the predictive power
	of the abstractions it considers,
	not their adaptive significance.
In turn,
	the design of the executive system is considerably
	simplified by the fact that it can rely 
	on the oracular system to deliver high quality
	abstractions and predictions.
The conceptual simplification introduced by the two-system view
	can also be thought of as a modular decomposition
	between the low-level perceptual inputs and the decision making function.
In the unsimplified case,
	the executive system computes an action based on
	the current state and history of all the perceptual input channels.
But this computation is probably intractable,
	due to the vast amount of input data.
The simplification is to introduce a set of intermediate variables
	corresponding to high-level abstractions.
Then the influence of the low-level perceptions
	on decision making is mediated by the abstraction variables.
	
\mpagebreak

%\minclude{comprstats}

\chapter{Compression and Vision}
\label{chapt:comprvsion}

The issues raised in the preceding chapters now propel our attention
	to the field of computer vision.
Vision is uniquely relevant to the \compist~philosophy
	because visual data exhibits two crucial properties:
	it is easy to obtain in large quantities,
	and it has rich and extensive structure. 
The potential practical applications of computer vision 
	provide additional motivation,
	as does the fact that a large fraction of the human brain 
	is dedicated to visual processing.
	
The goal of computer vision research is to build machines 
	that can see as well as humans.
For many who are not familiar with the field, 
	this seems like a rather simple goal. 
But it is in fact profoundly difficult, 
	and in spite of several decades of research,
	current vision systems cannot perform
	even at the level of human children.
One might diagnose a number of factors
	contributing to this lack of progress.
It could be that the mathematical theory employed by the field is not 
	sophisticated enough, or that modern computers are not fast enough. 
If this is true then vision researchers are basically on the right track,
	are making gradual progress,
	and will eventually achieve their goals.
	
This book presents an alternative view of the situation,
	which is that vision research is hobbled by
	the weakness of its philosophical foundations.
In particular, 
	the field lacks convincing answers to two very concrete methodological questions.
The first is the question of evaluation:
	given a pool of candidate solutions to a certain task,	
	how does one select the best one?
There is a substantial amount of work in the area of empirical evaluation,
	but current methods are not completely satisfying,
	either from a conceptual or a practical standpoint.
The philosophical difficulties involved in the evaluation of computer vision solutions
	stands in stark contrast to the conceptual ease with which 
	two theories of physics can be compared,
	by finding an experimental configuration for which the two theories 
	make conflicting predictions and running the experiment.

Behind the question of evaluation lies a deeper issue 
	which may be called the ``meta-question'':
	what are the important scientific questions facing the field?
While individual researchers may have their own philosophies,
	the community is not united by a convincing answer.
The situation is well illustrated 
	by the following passage from Wikipedia
	describing the state of the art:

\begin{quote}
Computer vision covers a wide range of topics which are often related to other disciplines,
	and consequently there is no standard formulation of ``the computer vision problem''. 
Moreover, there is no standard formulation of how computer vision problems should be solved.
Instead, there exists an abundance of methods for solving various well-defined computer vision tasks, 
	where the methods often are very task specific and seldom can be generalized over a 
	wide range of applications~\cite{Wikipedia:Comp2}.
\end{quote}

The argument of this chapter is that the compression principle
	can repair these issues by providing a standard formulation
	of the computer vision problem.
In this view, computer vision becomes the \compist~science of visual reality.
Research proceeds by obtaining progressively better compression rates
	on benchmark databases of natural images,
	and by expanding the size and scope of those databases.
The first step in this argument,
	contained in Section~\ref{sec:repcompvis},
	involves an overview of some traditional computer vision tasks,
	and a summary of influential papers addressing those tasks.
Vision researchers have developed a variety of schemes
	to evaluate the performance of candidate solutions on standard tasks;
	Section~\ref{sec:mthoddiffc} contains a discussion of these schemes.
This discussion makes it clear that
	the evaluation schemes contain serious conceptual and practical deficiencies.
Section~\ref{sec:critanalcv} presents a more general critique of the 
	scientific philosophy of computer vision,
	motivating a proposal to reform that philosophy.
Finally, Section~\ref{sec:equivalence} shows how most standard computer vision
	tasks can be reformulated as specialized image compression techniques.

\section{Representative Research in Computer Vision}
\label{sec:repcompvis}

The following sections contain a discussion and analysis 
	of several representative research papers in computer vision.
Most of the papers made strong impacts 
	on subsequent technical work in the field,
	and some of them represent the current state of the art.
But the purpose of presenting the papers is not to provide
	the reader with a comprehensive technical overview of computer vision methods.
Such an overview would require its own book or perhaps bookshelf.
Instead, the purpose is to illustrate the principles 
	and philosophical commitments that guide research in the field.
Since most vision researchers share the same general mindset,
	this can be done by describing a few key papers.

\mpagebreak

\subsection{Edge Detection}
	
Humans naturally perceive the visual world in terms
	of edges, lines, and boundaries.
The goal of an edge detection algorithm is to replicate this aspect of perception,
	by marking positions in an image that 
	correspond to edges.
Different authors have formalized this goal in various ways,
	though everyone agrees that edge detection is more than
	simply finding positions where the image intensity changes rapidly.
The consensus seems to be that the detector
	should identify perceptually salient edges,
	and should produce detection results
	that are useful for higher-level applications.
The typical approach to the edge detection problem
	involves a new computational definition of the task,
	and an attempt to show that this new definition 
	leads to results that agree with human perception or are practically useful.

A simplistic approach to the edge detection problem
	would be to find the positions in the image
	where the intensity gradient is high.
The limitations of this simple approach 
	can be understood by considering the issue of scale.
Many images exhibit sharp intensity variations on a fine scale
	that are not really important.
An example would be an image showing 
	a patch of grass next to a sidewalk.
The significant edge in this image is the one separating
	the grass from the sidewalk.
However, the region containing the grass will also
	contain sharp intensity variations,
	corresponding to individual blades of grass.
The ideal edge detector would ignore these fine scale variations, 
	and report only the significant edge between
	the grass and sidewalk.
The problem is to formulate this idea of ignoring fine-scale
	variations in computational terms
	that can be programmed into an algorithm.
	
One standard way of dealing with the issue of scale is to 
	blur the image as a preprocessing step.
Blurring transforms the grass region into a homogeneous
	patch of greenish-brown pixels.
This wipes out the fine scale edges,
	leaving behind the coarse scale grass-sidewalk edge.	
Blurring also has the advantage of making the detector 
	robust to noise.
Unfortunately, the use of blurring introduces a new set of problems.
One such problem is that blurring decreases the accuracy with
	which the position of the coarse scale edges are detected.
Also, in some cases, it might be necessary to find the fine scale edges,
	and so throwing them away is the incorrect behavior.

Marr and Hildreth,
	in a paper boldly entitled ``Theory of Edge Detection'',
	presented an early and influential approach to the problem~\cite{Marr:1980}.
The paper contains two parts;
	the first part proposes an algorithm,
	and the second part shows how the algorithm might be implemented
	by neurons in the brain.
The algorithm begins by blurring the image at various scales,
	for the reasons noted above.
The second step searches for places where image intensity changes rapidly,
	since these correspond to edge locations.
A useful trick for finding these positions is given by the rules of calculus.
Rapid intensity changes correspond to extrema 
	in the first derivative of the intensity function.
An extremum in the first derivative corresponds to a
	zero in the second derivative.
So the algorithm blurs the image and then searches for zeroes of the second derivative.
	
This approach would be sufficient if an image were one-dimensional.
But images are two-dimensional, 
	so when calculating intensity derivatives,
	one must specify the direction in which the derivative is taken.
A key idea of the paper is the realization that,
	under reasonable conditions,
	it is acceptable to use the Laplace operator $\nabla^{2}$,
	which is just the sum of the second derivatives in two directions.
The Laplace operator is invariant under rotations of the coordinate system,
	so using it will mark edges in the same position for an image
	and a rotated copy of the image.

The paper also presents the idea of a ``primal sketch'',
	which is formed by calculating the edge positions at multiple scales 
	(blurring levels),
	and combining the responses together into a network
	of line segments.
The authors argue that this representation is complete:
	the network of segments, 
	along with the amplitude values indicating the intensity of an edge,
	is sufficient to reconstruct the original image.
Since the line segments can be thought of as discrete units,
	the transformation from the raw image to the primal sketch 
	amounts to a transformation from the analog world 
	of continuous values to the discrete world of symbols.
Early artificial intelligence researchers
	thought intelligence was mostly about symbol processing,
	so they considered this transformation important
	because it served as a bridge between raw perception
	and logical, symbolic reasoning.

The second part of the paper is entitled ``Implications for Biology'',	
	which attempted to show that the proposed algorithm
	could actually be computed by neurons in the visual cortex.
The paper analyzes known neurophysiological findings,
	and argues that various patterns of neural activity can be interpreted 
	as performing the computations called for by the algorithm.
For example, a specific proposal is that 
	a certain type of neurons called lateral geniculate X-cells
	encode the result of applying the Laplace operator to the blurred image.
The concluding discussion uses the theory to make some neuroscientific predictions.

Another important paper in this area is entitled
	``A Computational Approach to Edge Detection'',
	by John Canny~\cite{Canny:1987}.
Canny showed how a standard scheme of edge detection
	could be refined, 
	making it robust to noise and more likely to report
	perceptually salient edges.
The standard way to detect edges or other patterns
	is to apply a filter to an image,
	and then look for maxima in the filter's response.
For Canny, the main shortcoming of this scheme 
	is that it is highly sensitive to noise.
To Canny and other like-minded researchers,
	noise was not caused just by a flaw in the measuring apparatus (camera),
	but also by real distractions or uninteresting aspects of an image.	
The problem of scale can be thought of as a special case of the noise problem;
	the fine-scale edges of individual blades of grass are thought
	of as ``noise'' corrupting the homogeneous green of the grass patch.
Noise is a problem because it can cause simplistic filters to report false edges,
	or to produce incorrect estimates for the locations of real edges.
Canny wrote down a mathematical expression describing these two sources of error,
	and then derived a set of optimal filters
	that would minimize the expression.
Different types of edges required a different filters.
For example, one filter is optimal for roof edges (light-dark-light)
	and another is optimal for step edges (light-dark).
	
The main idea of the Canny paper is the derivation of the set of optimal filters,
	but it also describes several additional topics.
Canny shows that, contra Marr-Hildreth,
	there are actually strong advantages in using 
	directionally oriented edge detectors.
The paper also describes the problem
	of searching for edges at multiple scales.
The problem now becomes how to integrate the responses
	from the filters at multiple scales and orientations
	into one single result;
	a scheme called feature synthesis is proposed to do this.
A third topic mentioned in the paper is a technique
	for estimating the noise in an image using a method called Wiener filtering.
It is very useful to know the noise power,
	because if there is a lot of noise,
	then the filter response theshold for identifying a particular position
	as an edge should be raised.
Positions that show filter responses below the threshold
	are marked as spurious edges and ignored.

\mpagebreak
	
\subsection{Image Segmentation}

Humans tend to perceive images in terms of regions.
Most people will perceive an image of a car, a tree, and a pedestrian
	by packaging the pixels into regions corresponding to 
	those objects.

The goal in the image segmentation task is to imitate
	this perceptual grouping tendency.
An image segmentation algorithm 
	partitions the pixels of an image into non-overlapping regions.
Ideally, this partitioning should reflect real object boundaries,
	so that all the pixels corresponding to a car are 
	packaged together, including tires and windows.
Unfortunately, this level of performance has not yet been reached.
In practice, most algorithms perform segmentation
	based on similarities in pixel color,
	so that the black pixels of the car tire would be packaged into one region,
	while the windshield pixels would go into another.
Arguably, the output of an image segmentation algorithm
	might also be useful as an input to a higher-level vision system.
	
One of the most important papers on the topic of 
	image segmentation is ``Normalized Cuts and Image Segmentation'' 
	by Shi and Malik~\cite{Shi:2000}.
This paper proposes to use ideas from the mathematical field of graph theory
	to approach the segmentation problem.
A graph consists of a set of nodes,
	some of which are connected by edges.
A graph is called connected if
	for any pair of nodes in the graph,
	there is a path (i.e. sequence of edges) connecting them.
Given a connected graph,
	a cut is a set of edges which, when removed,
	render the graph disconnected.
The problem of finding the minimum cut - 
	the cut involving the smallest set of edges - 
	is a well-studied problem in graph theory
	and efficient algorithms exist to find it.

To use the minimum cut idea to do segmentation,
	Shi and Malik construct a graph by creating a node for each pixel,
	and placing an edge between two nodes representing neighboring pixels.
Each edge has a weight corresponding to the
	degree of similarity between two pixels.
Then the minimum cut of the resulting graph separates the image
	into two maximally-dissimilar regions. 	
However, one refinement is necessary.
By itself, the minimum cut rule tends to create cuts
	that isolate single pixels,
	since typically a pixel can be cut off by removing 
	only four edges.
Shi and Malik repair this undesirable behavior by proposing
	to use a normalized cut:
	
\begin{equation*}
Ncut(A, B) = \frac{cut(A, B)}{assoc(A, V)} + \frac{cut(A, B)}{assoc(B, V)}
\end{equation*}

\noindent 
Where $cut(A, B)$ is the weight of the edges separating the two subregions,
	and $assoc(A, V)$ is the weight of all the edges between $A$
	and the full graph $V$.
If $A$ is a single pixel, then the $cut / assoc$ ratio will be large,
	since all the edges included in $cut$ will also be included in $assoc$.
Thus the minimum normalized cut for the graph
	should correspond to a good boundary separating 
	two distinct regions in the image.
Unfortunately the inclusion of the $assoc$ terms
	make the problem harder than the basic minimum cut problem.
The authors define a relaxed version of the basic minimum cut problem,
	where an edge can be fractionally included in a cut,
	instead of completely in or out.
This relaxed problem can be cast as an eigenvector problem,
	and the authors give a specialized algorithm for solving it.

Another, more recent contribution to the image segmentation literature
	is a paper entitled ``Efficient Graph-Based Image Segmentation''
	by \felz~and Huttenlocher~\cite{Felzenszwalb:2004}.
These authors also use graph theory concepts to perform segmentation,
	but formalize the problem in a different way.
Again, each pixel in the image corresponds to a graph node,
	adjacent pixels in the image are linked by edges in the graph,
	and edges have weights measuring the similarity between pixels.
The paper defines two key quantities
	related to a pair of regions $C_{1}$ and $C_{2}$.
The first is a region-difference score $Dif(C_{1}, C_{2})$,
	which is just the minimum weight edge 
	connecting the two regions.
The second key quantity is the following:

\begin{equation*}
\min (Int(C_{1}) + \tau(C_{1}), Int(C_{2}) + \tau(C_{2}))
\end{equation*}

\noindent
Where $Int(C)$ is an internal coherence score of the region $C$,
	and $\tau(C)$ is an arbitrary auxiliary function.
The function $Int(C)$ attempts to quantify the internal similarity
	of the pixels in a region.
This is defined as the maximum weight edge
	of the minimum spanning tree of the region.
The minimum spanning tree of a group of nodes is the minimum set of edges 
	required to make the group connected.
If the minimum spanning tree contains a high weight node,
	it implies that the region has low internal coherence.

The basic segmentation principle of the paper is that
	a boundary should exist between two regions
	if the region-difference score is smaller
	than the paired internal coherence score.
Roughly,
	this means that two regions should be merged
	if the internal coherence of the resulting region is no
	smaller than the internal coherence of the original two regions.
The authors point out that an arbitrary non-negative function 
	$\tau(C)$ of the pixels in a region can be added
	to the internal coherence score without
	breaking the algorithm.
This provides a nice element of flexibility,
	since $\tau(C)$ can be used to prefer regions with a certain shape, color, or texture.
This could allow the algorithm to be adapted
	for the specialized purpose of, for example, detecting faces.

\felz and Huttenlocher present an algorithm
	that can perform segmentation according to the principle defined above,
	and an analysis of its properties.
The algorithm starts by creating a list of all the edges in the graph,
	ordered so that low weight edges,
	corresponding to very similar pixel-pairs, come first.
The algorithm goes through the list of edges,
	and for each pixel-pair,
	determines if the regions containing the two pixels should be merged.
This is done by comparing the 
	edge weight to the internal coherence of the two regions
	containing the two pixels.
The authors also prove several theorems,
	which basically say that the segmentations produced by the algorithm
	are neither too coarse nor too fine,
	according to a particular definition of these concepts.
Finally, the authors show that the algorithm runs in time $O(N \log N)$,
	where $N$ is the number of pixels in the image.
This is quite fast;
	potentially fast enough to segment video sequences in real time,
	something that most other segmentation algorithms are too slow to do.

Both of the papers described above
	present only qualitative evidence to demonstrate
	the value of their proposals.
The experimental validation sections 
	show the results of applying the algorithm to a couple of images.
The reader is expected to
	conclude that the algorithm is good 
	based on a visual inspection of the output.
There is no real attempt to empirically evaluate the results of the algorithm,
	or to compare it to other proposed solutions to the segmentation task.
	
\mpagebreak	

\subsection{Stereo Matching}

Humans have two eyes located very close to one another.
In normal situations,
	the two eyes observe images that are similar, but measurably different.
This difference depends critically on depth,
	due to the way an image is formed by collapsing a 3D scene onto a 2D plane.
If a pen is held at a large distance from a viewer,
	the pen will appear at nearly the same position in both eye-images,
	but a pen held an inch from the viewer's nose will 
	appear at markedly different positions.
Thus, the depth of the various objects in a scene
	can be inferred by analyzing the differences in the stereo image pair.	
Vision researchers can emulate the conditions of human stereo observation
	by mounting two cameras on a bar, 
	separated by a known distance.
	
The crucial difficulty in stereo vision to find matching points,
	which are positions in the two images
	that correspond to the same location in the 3D world.
For example, if the 2D positions of the tip of a pen
	can be found in both images, 
	this is a matching point, 
	since the pen-tip occupies a specific location in the 3D world.
From a single matched pair,
	a disparity function can be calculated,
	which can then be used to find the 3D depth of the point.
Typically, though, the goal is to find a depth map,
	which represents the depth of all the points in the image.
Finding large numbers of matching points is difficult for several reasons.
Some points in 3D space might be visible in one eye-image,
	but occluded in the other image.
If the image shows an object with very few distinguishing features,
	such as a white wall, 
	then it becomes hard to find matching points.
Some surfaces are specular,
	meaning that they reflect light in a mirror-like way,
	as opposed to a diffuse, incoherent way.
Specularity can therefore	
	cause the two cameras to observe very different
	amounts of light coming from the same point in 3D space.

%TODO: expand this desciption to more than one paragraph
An early effort in the area of stereo matching
	is a famous paper by Lucas and Kanade,
	``An Iterative Image Registration Technique with an Application to Stereo Vision''
	~\cite{Lucas:1981}.
This paper begins with a technique for solving a simplified version of 
	the stereo matching problem, called image registration.
Here it is assumed that the one of the images is a translated
	and linearly scaled copy of the other,
	and the task is to find the translation and scaling factor.
In principle this could be done by an exhaustive search,
	but it would require a very long time,
	so the paper gives a more efficient algorithm.
The algorithm starts with a guess of the correct translation.
Then, it approximates the effect of making a small change to the translation
	by finding the derivative of one image at every point.
That assumption allows an optimal update to the translation-guess
	to be computed, and then the algorithm repeats with the new guess.
The same basic idea works to find the scaling factor.
Once the scaling factor and the translation vector are known,
	a straightforward calculation can be used to find
	the depth of the object.
	
A more recent paper on the subject is ``Stereo Matching Using Belief Propagation''
	by Sun, Shum, and Zheng~\cite{Sun:2003}.
This paper attempts to solve a more challenging version of the matching problem:
	it attempts to match each pixel with a partner in the other image.
The paper formulates the matching task
	as a probabilistic inference problem,
	where three probabilistic variables are attached to each pixel.
One variable corresponds to the pixel depth,	
	another is true if there is a depth discontinuity at the pixel,
	and the third is true if the pixel is occluded.
The intuition behind the model is that it is very unlikely
	for one pixel to have a depth of one meter,
	if the neighboring pixel has a depth of five meters,
	unless the discontinuity variable is true for the pixel.
Also, if a pixel is the same color as its neighbors,
	it is very unlikely that a discontinuity exists at the given point.
The authors formalize
	these relationships using a model called a Markov Random Field (MRF).	
The MRF expresses the probability of a variable configuration
	using purely local functions:
	the probability of a certain pixel depth value
	depends only on the depth values of neighboring pixels.

The authors propose to describe the relationship
	between the depth, occlusion, and discontinuity variables
	using a probabilisic model called a Markov Random Field (MRF).
The MRF assigns a probability to a configuration of variables
	based on purely local functions,
	so that, for example,
	the probability of a certain pixel depth value
	depends only on the depth values of neighboring pixels.
So the MRF
	model defines a relationship between
	the observed variables (the actual pixel values)
	and the hidden variables (depth, etc).
The problem now is in a general form:
	find the highest probability configuration of some hidden variables
	given evidence from the observed variables.
There are many techniques for approaching this general problem.
An algorithm called Belief Propagation can solve the problem efficiently,
	but it is only guaranteed to work for a restricted class
	of simplified models.
The authors propose to use Belief Propagation anyway,
	and accept that the resulting solution will be only an approximation.
	
\pagebreak

\subsection{Object Recognition and Face Detection}

Object recognition is the problem of identifying objects
	in an image.
In contrast to the general purpose tasks like segmentation and edge detection,
	the object recognition task is directly practical.
Systems implementing face detection - 
	a special case of object recognition - 
	have been implemented in digital cameras.
A vision system with the ability to recognize
	pedestrians, cars, traffic lights, and other objects
	would be an important component of robotic automobiles.

Most object recognition systems in computer vision
	follow the same basic recipe.
The first ingredient is a database 
	containing many images marked with a label
	designating the type of object the image shows.
Researchers either construct these database themselves,
	or use a shared benchmark.
The second ingredient is a standard learning algorithm
	such as AdaBoost or the Support Vector Machine~\cite{Freund:1997,Boser:1992}.
Unfortunately, these classifiers cannot be applied
	directly to the images,
	because images are far too high-dimensional.
The images must be preprocessed somehow
	in order to be used as inputs to the learning algorithms.
The search for smart preprocessing techniques
	constitutes one of the main areas of research in the subfield. 
One common strategy is to define \textit{features},
	which are simple functions of an image or an image region.
The feature vector calculated from an image
	can be thought of as a statistical signature of the image.
Hopefully, if a good set of features can be found,
	this signature will contain enough information to recognize the object,
	while using only a small number of values (say, 50).

Compared to general purpose tasks like image segmentation,
	performance on the object recognition task 
	is easy to evaluate: 
	look at the error rate an algorithm achieves on the given database.
Often, a shared database is used,
	so it is possible to directly compare a new algorithm's performance
	to previously published results.
The question of whether these quantitative results are in fact meaningful
	is taken up at greater length in Section~\ref{sec:mthoddiffc}.
	
An important recent paper in the face detection literature
	is entitled ``Robust Real-Time Face Detection''
	by Viola and Jones~\cite{Viola:2004}.
The authors use the basic formula 
	for object recognition mentioned above,
	with the well-known AdaBoost algorithm as a classifier~\cite{Freund:1997}.
AdaBoost is way of combining many weak classifiers
	that individually perform only slightly better than random
	into one strong classifier that yields high performance.
A key innovation of the paper is 
	the use of a special set of rectangular filter functions as features,
	along with a computational trick called the Integral Image.
The Integral Image allows for extremely rapid computation 
	of the rectangular features.
This is important because the goal of the paper is not
	just to determine if an image contains a face,
	but also where in the image the face is.
This means that the algorithm needs to scan every subwindow 
	to determine if it contains a face or not.
Because there are many subwindows,
	it is essential that each subwindow be processed quickly.
Another performance-enhancing trick is called the attentional cascade,
	which is a modification of the AdaBoost classifier
	where, if a subwindow looks unpromising after the initial
	weak classifiers are applied,
	further analysis is abandoned.
	
A more recent paper on object recognition is 
	``SVM-KNN: Discriminative Nearest Neighbor Classification 
	for Visual Category Recognition'' by
	Zhang, Berg, Maire, and Malik~\cite{Zhang:2006}.
This paper combines two methods:
	$K$-Nearest Neighbor (KNN) and the Support Vector Machine (SVM).
KNN works as follows:
	given a new point to be classified,
	it finds the $K$ samples in the training data that are ``closest''
	to the new point, according to some distance function.
To find a guess for the category of a new point,
	each neighbor ``votes'' for its own label.
In spite of (or because of) this simplicity,
	KNN works well in practice,
	and offers several attractive properties.
KNN works without any modification for multiclass problems
	(e.g. where there are many possible labels),
	and its performance becomes optimal as the number
	of samples in the training data increases.

The Support Vector Machine
	is based on the VC-theory discussed in Chapter~\ref{chapt:comprlearn},
	and is one of the most powerful modern learning algorithms~\cite{Vapnik:1996}.
The SVM works by projecting the training data points into a very high-dimensional feature space,
	and then finding a hyperplane that separates the points 
	corresponding to different categories with the largest possible margin.
Importantly, 
	the hyperplane computed in this way will be determined by a relatively small number
	of critical training samples; these are called the support vectors.
Because the model is determined by the small set of support vectors,
	it is simple and so,
	for reasons discussed in Chapter~\ref{chapt:comprlearn},
	has good generalization power.
However, there are a couple of drawbacks involved in using the SVM for object recognition.
First, the SVM is designed to solve the binary classification problem,
	and so applying it to the multiclass problem requires 
	some awkward modifications.
Second, the time required for training an SVM 
	(the search for the optimal hyperplane)
	scales badly with the number of training samples.

So the idea of the paper is to combine the two classifier schemes.
To classify a new point,
	the first step is to find the $K$ nearest neighbors in the training data,
	just as with standard KNN.
But instead of using those points to vote for a label,
	they are used to train a SVM classifier.
This classifier is used only for the given query point;
	to classify the next query point,
	a new set of neighbors is found and a new SVM is trained.
This hybrid scheme combines the good scaling and multiclass properties of KNN
	with the strong generalization power of the SVM.

The main missing ingredient to complete the algorithm 
	is to define a distance function for the KNN lookup.
A distance function simply takes two arguments (typically images),
	and gives a value indicating how close together they are.
The paper discusses several types of distance functions
	that can be used for various classification problems.

\mpagebreak

\section{Evaluation Methodologies in Computer Vision}
\label{sec:mthoddiffc}

A defining feature of the hard sciences
	such as mathematics, physics, and engineering
	is that they contain rigorous and objective
	procedures by which to validate new results.
A new mathematical theorem is justified by its proof,
	a new physical theory is justified by the empirical predictions it makes,
	and a new engineering device is justified by the demonstration
	that it works as advertised.
Because these fields possess rigorous evaluation methods,
	they are able to make systematic progress.
Given the importance of rigorous evaluation,
	it is worth inquiring into the methods used in computer vision.
Under a superficial examination,
	these methods may appear to be quite rigorous.
Certainly they are quantitative,
	and in some cases seem to provide a clear basis for 
	preferring one method over another.
However, a more critical analysis reveals that evaluation is 
	one of the central conceptual difficulties in computer vision.	
	
Since each computer vision task requires its own evaluation procedure (or ``evaluator'')
	to judge candidate solutions, and some have more than one,
	there are at least as many evaluators as there are tasks.
A comprehensive survey is therefore beyond the scope of this book,
	which instead provides a discussion that should 
	illustrate the basic issues.
Before proceeding to the discussion,
	it is worth noting that, historically at least,
	the field has exhibited very bad habits 
	regarding the issue of empirical evaluation.
Shin \etal~note that, 
	of 23 edge detection papers that were published in four journals between 1992 and 1998, 
	not a single one gave results using ground truth data from real images~\cite{Shin:1998}.
Even when empirical evaluations are carried out,
	it is often by a research group that has developed
	a new technique and is interested in highlighting its performance
	relative to existing methods.
	
\mpagebreak
	
\subsection{Evaluation of Image Segmentation Algorithms}	

% TODO: bad pacing here, break up into two paragraphs
The subfield of image segmentation dates back to at least 1978~\cite{Ohlander:1978}.
In spite of this long history,
	vision researchers have never agreed 
	on the precise computational definition of the task.
Neither have they agreed whether the goal 
	is to emulate a human perceptual process,
	or to serve as a low-level component for a high-level application.
There is also a glaring conceptual difficulty
	in the task definition,
	which is that there clearly exist some images 
	that are impossible or meaningless to segment.
Given an image of the New York skyline,
	should one segment the buildings all into one region,
	or assign them all to their own region?
If there is a cloud in the sky,	
	should it be segmented into its own region,
	or is it conceptually contained within one generic ``sky'' region?
While these kinds of questions linger,
	there are surely some segmentation questions that can be definitively answered.
Shi and Malik show an image of a baseball player diving through
	the air to make a catch.
It seems obvious that the pixels corresponding to the player's body
	should be segmented into a separate region
	from the green grass and the backfield wall.
Authors of vision papers
	present these kinds of test images,
	along with the algorithm-generated segmentation results,
	for the reader's inspection.
It is assumed that the reader will confirm 
	the quality of the generated segmentation.

This method is not very rigorous,
	and everyone is aware of this.
A slightly more rigorous methodology
	involves using automatic evaluation metrics.
These metrics, 
	recently surveyed by Zhang~\etal,
	perform evaluation by computing various quantities
	directly from an image and the machine-generated segmentation~\cite{Zhang:2008}.
For example,
	some metrics compute a measure of intra-region uniformity,
	since in good segmentations the regions should be mostly uniform.
Another type of metric computes a measure of the inter-region disparity;
	others score the smoothness of the region boundaries.
These methods are attractive because
	they do not require either human evaluation or human-labeled reference segmentations.
However, 
	use of such automatic metrics is conceptually problematic,
	because there is no special reason to believe a given metric
	correlates well with human judgment.
The justification of these metrics is also somewhat circular,
	since the algorithms themselves often optimize similar quantities.
Furthermore, most of the metrics do not work very well.
Zhang~\etal~report an experiment in which
	7 out of 9 metrics gave higher scores to machine-generated segmentations
	than to human-generated ones more than 80\% of the time~\cite{Zhang:2008}.
	
In 2001, a real empirical benchmark appeared in the form 
	of the Berkeley Segmentation Dataset,
	which contains ``ground truth'' obtained by asking human test subjects
	to segment images by hand~\cite{Martin:2001}.
While the approach to evaluation associated with this dataset
	is an important improvement over previous techniques, several issues remain.
One obvious problem with this approach is that it is 
	highly labor-intensive and task-specific,
	so the ratio of effort expended to understanding achieved seems low.
A larger issue is that the segmentation problem has no precisely defined correct answer:
	different humans will produce substantially different responses 
	to the same image.
Even this might not be so bad;
	one can define an aggregate or average score and 
	plausibly hope that using enough data will damp out chance fluctuations
	that might cause a low quality algorithm to achieve a high score or vice versa.
But still another conceptual hurdle must be cleared:
	given two segmentations, one algorithm- and one human-generated,
	there is no standard way to score the former by comparing it to the latter.
Some scoring functions assign high values to degenerate responses, 
	such as assigning each pixel to its own region,
	or assigning the entire image to a single region~\cite{Martin:2002}.
The question of how to score a segmentation by comparing it to 	
	a human-produced result has become a research problem 
	in its own right, 
	resulting in a proliferation of scoring methods~\cite{Martin:2002,Estrada:2005,Unnikrishnan:2007}.
A more technical but still important issue is the problem of parameter choice.
Essentially all segmentation algorithms require 
	the choice of at least one, and usually several, parameters,
	which strongly influence the algorithm's output.
This complicates the evaluation process for obvious reasons.

\mpagebreak

\subsection{Evaluation of Edge Detectors}
\label{sec:edgedteval}

The task of edge detection is conceptually similar to segmentation,
	and faces many of the same issues when it comes to empirical evaluation.
One interesting paper in this area
	is promisingly titled
	``A Formal Framework for the Objective Evalution of Edge Detectors'',
	by Dougherty and Bowyer~\cite{Dougherty:1998} (1998).
This paper begins with the following remarks:

\begin{quote}
Despite the fact that edge detection is one of the most-studied
	problems in the field of computer vision,
	the field does not have a standard method of evaluating
	the performance of an edge detector.
The current prevailing method of showing a few images 
	and visually comparing subsections of edges images 
	lacks the rigor necessary to find the fine-level performance differences
	between edge detectors...
This lack of objective performance evaluation has 
	resulted in an absence of clear advancement in the ``state of the art''
	of edge detection,
	or a clear understanding of the relative behavior
	of different detectors.
\end{quote}

\noindent
To remedy this unfortunate situation,
	the authors propose a framework for edge detection based on
	manually labeled ``ground truth''.
To obtain the ground truth,
	a human studies an image and labels the edge positions by hand.
This strategy is labor intensive:
	the authors note that labeling a $512 \cdot 512$ image
	requires about 3-4 hours.
Because of the time requirements,
	the evaluation is done using only 40 images,
	which are grouped into four categories:
	Aerial, Medical, Face, and generic Object.
The manual labeling strategy is also somewhat subjective,
	since the goal is not to find \textit{all} the edges
	but only the perceptually salient ones.

Dougherty and Bowyer grapple with the issue,
	common to almost all vision evaluation research,
	of parameter choice.
Most vision algorithms employ several parameters,
	and their output can depend on the parameter choices
	in complicated ways.
So if an algorithm performs well on a certain test with one parameter setting,
	but poorly on the same test when a different setting is used,
	should the algorithm receive a high score or a low score?
Dougherty and Bowyer propose to score an algorithm based on its ``best-case'' performance.
They use an iterative sampling procedure
	where many parameter settings are tested,
	and the best setting is used to score the method.
Measuring performance is also somewhat problematic,
	since most algorithms can reduce their false positive rate 
	by increasing their false negative rate and vice versa.
The authors solve this by using a Receiver Operating Characteristic (ROC) curve,
	which shows the relationship between false positives and false negatives,
	and defining the best performance as the one with the 
	smallest area under the curve.

Based on this evalution scheme,
	Dougherty and Bowyer rank the performance of six detector algorithms,
	and conclude that the Canny detector and the Heitger detector
	are the best.
They claim that the ranking is fairly consistent across the image categories.
However,
	it is not clear if this kind of conclusion can be justified
	on the basis of such a limited dataset.
Do 40 images constitute a meaningful sample from the space of images?
Another conceptual issue is the connection between
	human perception and practical applications.
Even if an edge detector's output agrees with human perception,
	it may not also be useful as a low-level component
	of a higher-level system.

In addition to these conceptual issues,
	the evaluation reported by Dougherty and Bowyer appears to have some technical problems,
	as described in a 2002 paper by Forbes and Draper~\cite{Forbes:2000}.
These authors point out that the ROC-style evaluator is extremely noisy,
	because very small changes to the target image
	can cause the evaluator to assign wildly different scores.
This noisiness is caused by the hypersensitivity of
	the edge detectors to small changes in the target image.
Forbes and Draper use a graphics program to generate images,
	which allows them to probe the detectors' responses
	to minor changes in the target image.
They mention that one edge detector (the \textsc{Susan} detector)
	goes from receiving the best score to the worst score
	when the resolution of the target image is changed from 50-90 pixels to 130-200 pixels.
They also highlight one example where
	a one-pixel change in the resolution of the target image
	leads to a huge shift in the response of the Canny detector.
In general, 
	the Canny detector appears to be highly sensitive to 
	the choice of parameter setting.
This implies that the results of Dougherty and Bowyer
	showing the superiority of the Canny detector
	were probably just a result of the parameter search scheme they employed.
If a detector's response changes dramatically as a function of the parameters,
	then it is more likely that there exists some parameter setting
	for which it receives a high score.	

\mpagebreak
	
\subsection{Evaluation of Object Recognition}
	
In theory,
	a vision researcher can evaluate an object recognition system
	using a very straightforward procedure:
	construct a labeled benchmark database,
	and count the number of errors made by each candidate solution.
In practice,
	as discussed below,
	there are a number of practical and conceptual pitfalls
	that can make evaluation difficult.
This section describes two well-known benchmark databases,
	that appeared in 2002-3.
There are more recent databases,
	but at the time of this writing,
	it is too early to tell whether
	they have had a strong effect on the field.
	
In the context of object recognition,
	the labeled benchmark database actually plays \textit{two} critical roles.
This is because the majority of object recognition systems
	are based on learning algorithms.
The training portion of the database
	enables the learning algorithms to fit statistical models
	to the data.
The testing portion 
	is used to actually evaluate the system's performance.

A prototypical example of a benchmark used in computer vision
	is the UIUC Image Database for Car Detection~\cite{Agarwal:2004}.
The purpose of this database is to evaluate
	car detector systems,
	which are obviously an important component of autonomous cars.
The UIUC contains only two categories:
	car and non-car.
In this sense it is similar to the one used by 
	Viola-Jones to train their face detection system.
The developers constructed the database in the standard way:
	they went outside and took a lot of photos,
	about half of which included cars.
This is somewhat labor intensive:
	the UIUC contains only about 1500 images;
	a database that was twice as big
	would require about twice as much time to construct.
The images are greyscale and of size $100 \cdot 40$,
	which is quite small.

The Caltech 101 is another influential database in the subfield of object recognition.
This database is differs from the UIUC one
	in a number of ways.
Caltech 101 was developed to evaluate \textit{generic}
	object recognition ability,
	not a specific detector (such as car- or face-detection).
The Caltech 101 is so named because it contains 101 categories;
	this is far more than was considered by previous research,
	which used at most six~\cite{Feifei:2007}.
Furthermore,
	the developers of the Caltech 101 used a unique process to construct it:

\begin{quote}
The names of 101 categories were generated by flipping 
	through the pages of the [dictionary],
	picking a subset of categories 
	that were associated with a drawing.
[Then] we used Google Image Search
	engine to collect as many images as possible for each category.
Two graduate students ...
	then sorted through each category,
	mostly getting rid of irrelevant images
	(e.g. a zebra-patterned shirt for the ``zebra'' category)~\cite{Feifei:2007}.
\end{quote}
	
\noindent
This method of rapidly generating an image database 
	is quite clever,
	because it allows a large number of images to be labeled rapidly.
However, it does have an important drawback,
	related to the way images are used in web pages.
The Google Image Search engine does not select images
	based on an analysis of their contents.
Instead,	
	it searches the surrounding text for the query term.
So if the user searches for ``monkey'',
	then the engine first finds pages that include that term,
	and then extract and return any images
	that those pages include.
The issue is that a web page describing monkeys
	will probably include images 
	in which the monkey is prominently displayed
	in the central foreground.
The monkey will not be peripheral,
	enshadowed, or half-concealed by the leaves of a tree.
This implies that the 
	images are actually relatively easy to recognize.
Another issue related to this procedure is that it selects	
	a somewhat strange set of categories;
	Caltech 101 includes categories for 
	``brontosaurus'', ``euphonium'', and ``Garfield''.

\mpagebreak

While the Caltech 101 has been very influential,
	it also has some important shortcomings.
Several of these were noted in a paper by Ponce \etal~\cite{Ponce:2006}
	(the \etal includes 12 other well known computer vision researchers).
One such shortcoming is that the objects are 
	shown in very typical poses, 
	without occlusion or complex background.
Ponce \etal~note:

\begin{quote}
Even though Caltech 101 is one of the most diverse datasets available 
	today in terms of the amount of \textit{inter-class} variability it encompasses,
	it is unfortunately lacking in several important sources of 
	\textit{intra-class} variability.
Namely, most Caltech 101 objects are of uniform size and orientation
	within their class, and lack rich backgrounds~\cite{Ponce:2006}.
\end{quote}

\noindent
Also, 
	if an average image is constructed by averaging together
	all the images in a class,
	this average image will have a very characteristic appearance.
For example, 
	the average image of the ``menorah'' class is very
	easily recognizable as a menorah.

The Caltech 101 is very diverse,
	but that diversity comes at a price:
	many image categories occur only a small number of times.
Many systems are trained using on the order of 20 or 30
	examples per category.
Because of this small amount of per-category data,
	the field is vulnerable to manual overfitting
	as described in Chapter~\ref{chapt:comprlearn}.
This is described by Ponce~\etal:

\begin{quote}
The (relatively) long-time public availability 
	of image databases makes it possible for researchers
	to fine-tune the parameters of their recognition algorithms
	to improve their performance.
Caltech 101 may,
	like any other dataset,
	be reaching the end of its useful shelf life~\cite{Ponce:2006}.
\end{quote}

One odd phenomenon related to the Caltech 101 is that several techniques can
	exhibit excellent classification performance but very poor localization performance:
	they can guess that the image contains a chair, but don't know where the chair is.
This is partly because, in the Caltech 101 and several other benchmark databases,
	the identity of the target object correlates highly with the background.
The correlation improves the performance of ``global'' methods 
	that use information extracted from the entire image.
To some extent this makes sense, 
	because for example cows are often found in grassy pastures,
	but a recognizer that exploits this information may fail to 
	recognize a cow in a photograph of an Indian city.
Ponce \etal~analyze this issue and conclude that:

\begin{quote}
This illustrates the limitations as evaluation platforms (sic?)
	of datasets with simple backgrounds,
	such as CogVis, COIL-100, and to some extent, Caltech 101:
Based on the evaluation presented in this section,	
	high performance on these datasets do not necessarily
	mean high performance on real images with varying backgrounds~\cite{Ponce:2006}.
\end{quote}

\noindent
CogVis and COIL-100 are two other object recognition datasets used
	in computer vision~\cite{Leibe:2001, Nene:1996}.
The fact that good benchmark performance does not equate to good 
	real-world performance illustrates a challenging 
	methodological issue for computer vision.
It shows that database development,
	which one might think of as a rote mechanical procedure,
	is actually quite difficult and risky.
Often a benchmark database will be developed and studied,
	but the results of that study will not lead
	to improvements in the state of the art.
In other words,
	each benchmark produces a modest chance of allowing
	the field to take a small step forward.
Given the effort required to construct databases,
	this implies that the aggregate amount of benchmark development work
	that will be required to achieve real success in computer vision
	is actually immense.

There are, of course,
	newer and more sophisticated databases
	that attempt to repair some of the problems mentioned above.
For example,
	the successor to Caltech 101,
	the Caltech 256,
	shows objects in a diverse range of poses and backgrounds,
	so that the average image is a homogeneous blob.
Another set of methods involve web-based
	games in which human players are tasked
	to label images, as part of the game.
The games are lightly entertaining,
	and so can potentially exploit ``bored human intelligence''
	to label a large number of images.
At the time of this writing,
	it is not yet clear whether these methods 
	will provide the necessary impetus for further progress in object recognition.
	
\mpagebreak

\subsection{Evaluation of Stereo Matching and Optical Flow Estimation}
	
This section packages together a discussion of evaluation metrics
	for stereo matching and optical flow algorithms,
	because the evaluation schemes are actually conceptually similar.
The key property that distinguishes these tasks from others previously
	considered is that ground truth can be obtained
	using automatic methods.
This ground truth is objective and does not require human input.
A clean evaluation can be performed by comparing the ground truth
	to the algorithm output.
Interestingly, 
	there is also a second type of evaluation method
	that can be used for both the stereo matching and optical flow tasks.
This method is based on interpolation,
	as discussed below.
	
The methods for obtaining ground truth for these tasks is based on the use
	of a sophisticated experimental apparatus
	\cite{Baker:2007,Scharstein:2002,Scharstein:2003,Szeliski:1999}.
For the stereo matching problem,
	ground truth can be obtained using an apparatus that employs structured light~\cite{Scharstein:2003}.
Here, information about stereo correspondence is inferred
	from the special patterns of light cast on the scene by a projector.
The projector casts several different patterns,
	such that each pixel can be uniquely identified
	in both cameras by its particular illumination sequence.
For example,
	if ten patterns are cast,
	then one pixel might get the  code 0010101110
	while a nearby pixel gets the code 0010100010.
Matching points 
	are found by comparing the pixel codes found in each image.
A conceptually similar scheme can be used to obtain ground truth for
	the optical flow problem.
Here, an experimental setup is used in which an object sprinkled with flourescent
	paint is moved on a computer-controlled motion stage,
	while being photographed in both ambient and ultraviolet lighting~\cite{Baker:2007}.
Since the motion of the stage is known,
	the actual motion of the objects in the scene
	can be computed from the reflection pattern of the ultraviolet light.
	
Once the ground truth has been obtained,
	it is a conceptually simple matter to evaluate a solution
	by comparing its output to the correct answer.
While the evaluation schemes used for these tasks
	are methodologically superior to others used in computer vision,
	they still suffer from an important drawback,
	which is the difficulty of using the experimental apparatus.
The setting up the apparatus requires a nontrivial amount of human labor,
	which implies that only a small number of image sequences are used.
A well-known benchmark, hosted on the Middlebury Stereo Vision Page,
	contains a total of 38 sequences~\cite{Scharstein:2010}.
It is not clear if the general performance of an algorithm
	on arbitrary images 
	can be well estimated using such a small number of sequences.
Additionally,
	since most vision algorithms include several parameters,
	it is hard to rule out manual overfitting as the source
	of any good performance that might be obtained on the benchmark.

For both the optical flow problem and the stereo matching problem,
	there exists an evaluation scheme that
	is strikingly simpler than the ground-truth based method
	described above~\cite{Baker:2007,Scharstein:2002,Szeliski:1999}.
The basic idea here is to use \textit{interpolation}.
For the stereo matching problem,
	the experimentalists use a trinocular camera system:
	a set of three cameras mounted on a bar
	at known inter-camera spacings.
This system generates an image set $A, B, C$.
The peripheral images $A$ and $C$ 
	are fed into the stereo matching algorithm.
If the algorithm is successful in computing
	the depth map and occlusion points,
	then it should also be able to infer the image $B$.
Thus the evaluation proceeds simply by comparing
	the algorithm-inferred image $\hat{B}$ to the real image $B$.
A variety of scores can be used for the comparison;
	Baker \etal~use:
	
\begin{equation*}
\bigg [ \sum_{x, y} \frac{(\hat{B}(x,y) - B(x,y))^{2}}{||\nabla B(x,y)||^{2} + \epsilon} \bigg ]^{1/2}
\end{equation*}

\noindent
Where $||\nabla B(x,y)||$ is the magnitude of the image gradient 
	at a certain point.

The interpolation-based scheme for evaluating optical flow
	algorithms is conceptually identical~\cite{Baker:2007}.
The only difference is that now the image sequence $A, B, C$
	is separated in time rather than space.
A video camera observes a moving scene,
	and produces an image sequence.
Again, the extremal images $A$ and $C$
	are fed to the optical flow estimation algorithm,
	which then attempts to infer the central image $B$.
Algorithms are evaluated by comparing the 
	guess for the central image $\hat{B}$ to the real image $B$.

These interpolation-based evaluation schemes
	are far simpler than the scheme based on ground truth,
	because no special apparatus is necessary to collect the ground truth.
Section~\ref{sec:compformofe}
	shows that the interpolation metric
	is actually equivalent to a certain kind of compression score.

\mpagebreak

\section{Critical Analysis of Field} 
\label{sec:critanalcv}
	
This section contains a brief critical analysis of the field of computer vision.
Before beginning,
	it is worthwhile to identify the conditions
	that allow such a critique to be reasonable even in principle.
If computer vision were a mature field like traditional computer science,
	a critique of its philosophical foundations
	would be an utterly worthless exercise.
However,
	the limitations of modern vision systems suggest that there 
	is a some deep conceptual obstacle hindering progress.
Therefore,
	the critique should be understood not as a disparagement of previous research,
	but as an attempt to discern the nature of the obstacle.
Furthermore, 
	it would be feckless and immature to complain about the limitations
	of computer vision without proposing some plan 
	to repair those limitations.
This book contains exactly such a plan.	
	
\subsection{Weakness of Empirical Evaluation}

The most obvious failing of computer vision is
	the weakness of its methods of empirical evaluation.
This shortcoming is widely recognized
	and has been lamented by several authors~\cite{Haralick:1986,Jain:1991}.
Many papers attempt to do empirical evaluation
	by showing the results of a applying an algorithm
	to a couple of test images.
By looking at the system-generated results,
	the reader is supposed to verify that 
	the algorithm has done a good job.
Vision researchers are implicitly arguing
	that because an algorithm performs well on some small
	number of test images,
	it performs well in general.
But this argument is clearly flawed.
It may be that the researchers hand-picked
	images on which their algorithm performed well,
	or got lucky in the selection of those images.
Or, more likely,
	it may mean that the design of the algorithm was tweaked
	until it produced good results on the test images.

As discussed above,
	there is a recent trend in computer vision towards
	the use of benchmark databases,
	such as the Berkeley Segmentation Dataset and the Caltech 101,
	for empirical evaluation.
These benchmarks suffer from a wide variety
	of conceptual and practical shortcomings,
	several of which have already been described.
One general issue
	is the problem of meta-evaluation.
There is no guarantee that an evaluator scheme will
	produce good assessments of the quality of a candidate solution.
This implies that it is 
	necessary to conduct a meta-evaluation process,
	to determine which evaluator produces the best information
	about the quality of a solution.
But then how is the word ``best'' to be defined?
Is it the method that correlates with human judgment?
Or is the method that produces the optimum response
	performance when used as a subcomponent for a higher-level system?
Is there any reason to believe these measures will agree?
The fact that evaluators themselves could be low-quality
	exacerbates the fact that they often require a huge 
	investment of time and effort to develop.
Say a researcher has developed a new evaluation strategy,
	that requires a large database of human labeled ground truth.
If there is a strong possibility that the evaluator will fail,
	then the risk-reward ratio for the project
	may become unacceptably high.
	
Another shortcoming of evaluation in computer vision
	is related to the idea of Goodhart's Law~\cite{Goodhart:1957}.
Goodhart's original formulation of this law is:
	``Any observed statistical regularity will tend to collapse 
	once pressure is placed upon it for control purposes.''
To see the relevance of this idea for computer vision,	
	assume that under normal conditions,
	a certain evaluator produces a noisy but informative
	estimate of the quality of a method.
So there is an observed statistical regularity (or correlation)
	between the evaluation score and the real quality of the solution.
Then if methods were developed in perfect ignorance
	of the evaluator,
	the evaluator would produce a reasonable ranking
	of the various methods.
In reality what happens is that researchers know about
	the evaluation schemes,
	and this knowledge guides the development process.
Anyone who wants to assert the quality of a new technique
	will need to show that it performs well
	according to the evaluator.
Since the regularity is now being used for control purposes,
	Goodhart's Law suggests that it will collapse.
	
The weakness of evaluation in computer vision
	is strongly related to the fact
	that the field does not conceive of itself as an empirical science.
In empirical sciences,
	researchers eventually obtain the correct description 
	of a particular phenomenon, 
	and then move on to new problems.
Instead of conceiving of their work as empirical science,
	vision researchers see themselves as producing a suite of tools.
Each tool has a particular set of conditions or applications
	for which it works well.
The only real way to evaluate a low-level method
	such as an edge detection algorithm
	is to connect it to a real-world application
	and measure the performance of the latter.
To say that one algorithm is better than another would
	be like saying a screwdriver is better than a hammer.
In this mindset,
	the fact that the field produces a river of solutions to various tasks,
	without any strong way of validating the quality of those solutions,
	is not a problem.

\mpagebreak

\subsection{Ambiguity of Problem Definition and Replication of Effort}

A critical reader of the computer vision literature
	is often struck by the fact that different 
	authors formulate the same problem in very different ways.
The problem of edge detection means
	something very different to Marr and Hildreth than it does to Canny.
Marr and Hildreth seem very concerned
	with transforming from the continuous world to the symbolic world.
Canny, on the other hand,
	seems primarily concerned with making the edge detection
	process robust to noise.
For Viola and Jones,
	in the object detection problem it is important to also locate
	the object in the image by scanning every subwindow;
	this means the detection for a single subwindow must be extremely fast.
In contrast Zhang~\etal~(SVM-KNN paper),
	are only concerned with determining an object's identity,
	not locating it.
This proliferation of formulations and different versions of the problem
	hinders progress for obvious reasons.

The cause of this ambiguity in problem definition 
	is that computer vision has no standard formulation
	or parsimonious justification.
Compare this to the situation in physics.
Physicists employ the same basic justification for all of their research:
	predict the behavior of a certain system or experimental configuration.
While this justification is parsimonious,
	it nevertheless leads to a wide array of research,
	because there are a huge number of physical systems and experimental configurations.
Research in computer vision has no comparable justification.
Vision papers are often justified by a large number of incompatible ideas.
Introductory sections of vision papers 
	will often include discussions of psychological phenomena such
	as mirror neurons, Gestalt psychology, neuroscience of the visual cortex,
	and so on.
They will also often include completely orthogonal practical justifications,
	arguing that certain low-level systems will be useful
	for later, high-level applications.

The lack of precise problem definitions
	leads to an enormous replication of effort.
A Google Scholar search for papers with the phrase ``image segmentation'' in the title
	returns more than 15,000 hits.
Hundreds of different techniques have been applied to the problem, 
	including fuzzy methods, expectation-maximization,
	robust analysis, level sets, watershed transforms,
	random walks, neural networks, Gibbs random fields,
	genetic algorithms, cellular automata, and more.
This is all in spite of the fact that the image segmentation 
	problem is not well defined and still
	has no good evalution scheme.
A search for the phrase ``edge detection'' returns more than 7,000 hits.
In principle,
	this immense proliferation of research is not \textit{a priori} bad.
In physics, 
	any new unexplained phenomenon might elicit a large number
	of competing explanations.
The difference is that physicists 
	can eventually determine which explanation is the best.

One crucial aspect of the success of the field of physics
	is that physicists are able to build
	on top of their predecessors' work.
Thus Newton used ideas originally developed
	by Galileo and Kepler to construct his theory of mechanics.
Newtonian mechanics was then used by Ampere and Faraday 
	in their electrical research.
Their laws were combined together and embellished by Maxwell.
Maxwell's theory of electromagnetics then served 
	as an impetus for the theory of special relativity.
Each discovery depends on a set of conceptual predecessors.
An implication of this is that physics research becomes increasingly esoteric
	and difficult for nonspecialists to understand.
Computer vision does not work like this;
	researchers rarely build on top of previous work.
One might imagine, for example, 
	that recent work in image segmentation might reuse
	previously discovered edge detection methods.
But this is not true, in general:
	the \felz~and Huttenlocher paper,
	published in 2004,
	uses no concepts developed
	by other vision researchers.
Their paper can be understood by anyone with a bit of background
	in computer science.
The same general theme is true for the other papers described above.
They sometimes depend on sophisticated previous results,
	but those results come from fields like statistics and machine learning,
	not from computer vision.
So the Sun~\etal~stereo matching paper (2005) depends on the Markov Random Field model,
	and the Viola-Jones face detection paper (2004) depends on the AdaBoost algorithm,
	but neither exploits any previous result in computer vision.

\mpagebreak
	
\subsection{Failure of Decomposition Strategy}

Several of systems discussed in the previous section
	perform a task that can be described as basic or low-level.
Few people would claim that these systems have practical value as isolated programs.
Image segmentation, for example,
	is rarely useful as a standalone application.
Rather, the idea is that these low-level systems will eventually function
	as components of larger, more sophisticated vision applications.
With a few exceptions, these higher-level applications have not yet appeared,
	but it is plausible to believe they will appear
	once the low-level components have achived a high level of reliability.
So during the modern phase, researchers will develop good 
	algorithms for tasks like image segmentation, edge detection,
	optical flow estimation, and feature point extraction.
Then, in a future era,
	these algorithms will be packaged together somehow
	into powerful and practical vision systems that deliver useful functionality.
	
This strategy, though it may seem plausible in the abstract,
	is in fact fraught with philosophical difficulty.
The issue is that,
	in advance of any foreknowledge of how the futuristic systems will work,
	there is no way of knowing what subsystems will be required.
Future applications may, plausibly, operate by first applying
	low-level algorithms to find the important edges and regions of an image,
	and then performing some advanced analysis on those components.
Or they may function in some entirely different way.
It is almost as if, by viewing birds,
	researchers of an earlier age anticipated the arrival of artificial flight,
	and proposed to pave the way to that application
	by developing artificial feathers.
		
\mpagebreak
	
\subsection{Computer Vision is not Empirical Science}

Chapter~\ref{chapt:compmethod} proposed a simple taxonomy
	of scientific activity involving three categories:
	mathematics, empirical science, and engineering.
Each of these categories produces a different kind of contribution,
	and each demands a different kind of justification 
	for new research results.
The field of computer vision can be classified 
	in the above scheme by analyzing
	the types of statements it makes
	and the analytical methods used to validate those statements. 
Consider the following abbreviated versions of the statements
	presented in three of the papers mentioned above:
	
\begin{quote}
VJ\#1: By using rectangular features along with the Integral Image trick,
	the feature computation process can be sped up dramatically. \\
VJ\#2: The rectangular features, when used with the AdaBoost classifier,
	produce good face detection results. \\
SM\#1: The $Ncut$ formalism leads to a relaxed eigenvector problem,
	which can be solved using a specialized algorithm. \\ 
SM\#2: By using the $Ncut$ formalism, good segmentation results can be obtained. \\
SSZ\#1: By introducing special hidden variables representing occlusion
	and discontinuity, and using a Markov Random Field model,
	the Belief Propagation algorithm can be used to provide
	approximate solutions to the stereo matching problem. \\ 
SSZ\#2: The model defined in this way will produce good 
	stereo matching results.
\end{quote}
	
\noindent
In the above list,
	the first statement (\#1) in each pair is a statement of mathematics.
These statements are deductively true
	and cannot be objected to or falsified.
Thus, an important subset of the results produced in the field
	are mathematical in character.
However, most people would agree
	that computer vision is not simply a branch of mathematics.

The second set of statements in the above list
	relates somehow to the real world.
But these results are best understood
	as statements of engineering:
	a particular device (algorithm) performs
	the task for which it is designed.
While these statements do, perhaps, 
	contain assertions about empirical reality,
	these assertions are never made explicit.
Perhaps the Viola-Jones result contains some implication 
	about the visual structure of faces,
	but the implication is entirely indirect.
Furthermore,
	the Popperian philosophy requires that a theory
	expose itself to falsification.
The vision techniques described above do not do so;
	no new evidence can appear that will falsify the Shi-Malik approach 
	to image segmentation.
If these methods are discarded by future researchers,
	it will be because some other technique 
	achieved a higher level of performance on the relevant task.
	
Another key aspect of empirical science
	mentioned in Chapter~\ref{chapt:compmethod}
	is that practitioners adopt a Circularity Commitment 
	to guide their research.
Empirical scientists are interested in constructing
	theories of various phenomena,
	and using those theories to make predictions about the same phenomena.
They feel no special need to ensure that their theories
	are useful for other purposes.
Clearly, vision scientists do not adopt this commitment.
Vision scientists never study images out of intrinsic curiousity.
Rather, they study images to find ways to develop
	practical applications such as face detection systems.
If an inquiring youngster asks a physicist about the world,
	the physicist might respond with a long speech 
	involving topics such as atoms, gravitation, conservation laws,
	quantum superfluidity, and the fact that only
	one fermion can occupy a given quantum state.
But if the youngster puts the same question to a vision scientist,
	the latter will have very little to say.

The failure of vision scientists to make explicit empirical claims
	is related to their background and training.
Most computer vision researchers have a 
	background in computer science (CS).
For this reason,
	they formulate vision research as the application
	of the CS mindset to images.
To understand this influence,
	consider the QuickSort algorithm,
	which is an exemplary piece of CS research. 
The key innovation behind the research is the design of the algorithm itself.
The algorithm is provably correct and works
	for all input lists.
Its quality resides in the fact that,
	in most cases, 
	it runs more quickly than other sorting methods.
Using this kind of research as an exemplar,
	it is not surprising that computer vision researchers
	attempt to obtain algorithms for image segmentation or edge detection
	that work for all images.
The influence of CS also explains why vision researchers do not 
	attempt to study the empirical properties of images:
	the development of QuickSort required no knowledge 
	of the empirical properties of lists.

Two other factors,
	in addition to the influence of the CS mindset,
	prevent vision researchers from conducting a
	systematic study of natural images.
First,
	there is little communal awareness in the field
	that there exists mechanisms (e.g. the compression principle)
	that could be used to guide such a study.
Second, there is not much reason to believe
	such a study would actually produce anything of value.
In other words,
	it is not widely apparent 
	that a version of the Reusability Hypothesis would hold
	for the resulting inquiry.
Of course, 
	the nonobviousness of the Reusability Hypothesis was
	one of the key barriers holding back the
	advent of empirical science.
The argument of this book, then, 
	is that the conceptual obstacle hindering progress in computer vision
	is simply a reincarnation of one that so long delayed the development
	of physics and chemistry.

\mpagebreak

\subsection{The Elemental Recognizer}

Imagine that, perhaps as a result of patronage from an advanced alien race,
	humanity had come into the possession of computers and digital cameras
	before the advent of physics and chemistry.
Aristotle,
	in his book ``On Generation and Corruption'',
	wrote that all materials are composed of varying quantities 
	of the four classical elements:
	air, fire, earth, and water.
An ancient Greek vision scientist might
	quite reasonably propose to build a vision system for 
	the purpose of classifying an object according to 
	its primary and secondary elemental composition.
	
The system would work as follows.
First, the researcher would take many pictures of various everyday objects, 
	such as trees, foods, togas, urns, farm animals, and so on.
Then he would enlist the aid of his fellow philosophers,
	asking them to classify the various objects according to their
	elemental composition.
The other philosophers, having read Aristotle, 
	would presumably be able to do this.
They may not agree in all cases,
	but that should not matter too much,
	as long as there is a general statistical consistency in the labeling 
	(most people will agree that a tree is made up of earth and water).

Now that this benchmark database is available,
	the vision philosopher uses it to test prototype implementations
	of the elemental recognizer system.
The philosopher takes the standard approach to building the system.
His algorithm consists of the following three steps:
	preprocessing, feature extraction, and classification.
The preprocessing step may consist of thresholding,
	centering, or filtering the image.
The feature extraction step somehow transforms the image 
	from a large chunk of data into simple vector
	of perhaps 20 dimensions.
Then he applies some standard learning algorithm,
	to find a rule that predicts the elemental composition labels
	from the feature vectors.
	
The research on elemental recognition could 
	be justified on various grounds.
Some philosophers may claim that it is important to develop automatic elemental composition recognition systems
	to facilitate progress in other fields.
Some thinkers
	may view it as a good inspiration for new mathematical problems.
Purist philosophers may view it as intrinsically interesting to find out the 
	elemental compositions of objects, 
	while others might regard it as an important low level
	processing task that will be helpful for higher level tasks.
	
The point of this thought experiment is that
	this process will work, perhaps rather well,
	in spite of the fact that the idea
	of the four classical elements  
	is not even remotely scientific.
The standard approach to visual recognition contains no
	mechanism that will indicate that the elemental categories
	are not real.
Instead of learning something about reality,
	the system learns to imitate the murky perceptual process
	which assigns objects to elemental categories.	
	
This idea sounds ridiculous to a modern observer,
	but only because he knows that Aristotelian conception
	of element composition is completely false.
How could the ancient Greek vision scientists discover this fact?
Is it possible that a vision researcher working alone, 
	with no knowledge of modern chemistry or physics,
	could articulate a principle by which to determine if the elemental composition
	idea is scientific or unscientific?
The scientific philosophy of computer vision 
	can evaluate the ability of various methods to identify elemental composition,
	but it cannot judge the elemental composition idea itself.	
	
\mpagebreak

\section{\Compist~Formulation of Computer Vision}
\label{sec:equivalence}

This book proposes a new way to carry out computer vision research:
	apply the Compression Rate Method 
	to a large database of natural images.
Computer vision, in this view,
	becomes the systematic empirical study of visual reality.
A \compist vision researcher proceeds by
	studying the image database, developing a theory describing the structure of the images,
	building this theory into a compressor,
	and demonstrating that the compressor achieves a short codelength.
This formulation provides a parsimonious justification for research:
	a theorem, technique, or observation is a legitimate topic
	of computer vision researcher if it is useful for compressing the image database.
The field advances by obtaining increasingly short codes for benchmark databases,
	and by expanding the size and scope of those databases.

This approach to vision has a number of conceptual and methodological advantages.
It allows for clean and objective comparisons between
	competing solutions.
These decisive comparisons will allow the vision community
	to conduct an efficient search through the theory-space.
It allows researchers to use large unlabeled databases,
	which are relatively easy to construct,
	instead of the small labeled datasets
	that are used in traditional evaluation procedures.
Because of the large quantity of data being modeled,
	the theories obtained through the \textsc{crm} can be enormously complex
	without overfitting the data.
Perhaps most importantly,
	it allows computer vision to become a hard empirical science like physics.

It should be noted that some ingenuity must be exercised 
	in constructing the benchmark databases.
Visual reality is immensely complex,
	and it will be impossible to handle all of this complexity
	in the early stages of research.
Instead, researchers should construct image databases
	that exhibit a relatively limited amount of variation.
Chapter~\ref{chapt:compmethod} proposed using 
	a roadside video camera to generate a database,
	in which the main source of variation would be the passing automobiles.
Many other setups,
	that include enough variation to be interesting
	but not so much that it becomes impossible to handle,
	can be imagined.
	
One immediate objection to the proposal is that,
	while large scale lossless data compression of natural images may be interesting, 
	it has nothing to do with computer vision.
The following arguments counter this objection.	
The key insight is that most computer vision tasks
	can be reformulated as specialized compression techniques.

\mpagebreak
	
\subsection{Abstract Formulation of Computer Vision}

Computer vision is often described as the inverse problem of computer graphics.
The typical problem of graphics is to produce,
	given a scene description $D_{L}$ wrtten in some description language $L$,
	the image $I$ that would be created if a photo were taken of that scene.
The goal of computer vision is to perform the reverse process:
	to obtain a scene description $D_{L}$ 
	from the raw information contained in the pixels of the image $I$.
This goal can be formalized mathematically 
	by writing $I = G(D_{L}) + I_{C}$
	where $G(D_{L})$ is the image constructed by the graphics program
	and $I_{C}$ is a correction image that makes up for any discrepancies.
Then the goal is to make the correction image as small as possible:

\begin{eqnarray*}
D_{L}^{*} &=& \arg \min_{D_{L}} C_{disc}(I_{c}) \\
 &=& \arg \min_{D_{L}} C_{disc}(I - G(D_{L}))
\end{eqnarray*}

Here $C_{disc}$ is some cost function which is minimized for the zero image,
	such as the sum of the squared values of each correction pixel.
The problem with this formulation 
	is that it ignores one the major difficulties of computer vision,
	which is that the inverse problem is underconstrained:
	there are many possible scene descriptions that can produce the same image.
So it is usually possible to trivially generate any target image 
	by constructing an arbitrarily complex description $D_{L}$.
As an example, if one of the primitives of the description language is a sphere, 
	and the sphere primitive has properties that give its color and position relative to the camera, 
	then it is possible to generate an arbitrary image by 
	positioning a tiny sphere of the necessary color at each pixel location. 	
The standard remedy for the underconstrainment issue 
	is regularization~\cite{Poggio:1989}.
The idea here is to introduce a function $h(D_{L})$ 
	that penalizes complex descriptions.
Then one optimizes a tradeoff between descriptive 
	accuracy and complexity:
	
\begin{equation}
D_{L}^{*} = \arg \min_{D_{L}} C_{disc}(I_{C}) + \lambda h(D_{L})
\end{equation}

\noindent
Where the regularization parameter $\lambda$ controls how 
	strongly complex descriptions are penalized.
While this formulation works well enough in some cases,
	it also raises several thorny questions related
	to how the two cost functions should be chosen.
If the goal of the process is to obtain descriptions that appear 
	visually ``correct'' to humans, 
	then presumably it is necessary to take considerations
	of human perception into account when choosing these functions.
At this point the typical approach is for the practitioner to choose
	the functions based on taste or intuition,
	and then show that they lead to qualitatively good results.

It turns out that the regularization procedure
	can be interpreted as a form of Bayesian inference.
The idea here is to view the image as evidence
	and the description as a hypothesis explaining the evidence.
Then the goal is to find the most probable hypothesis given the evidence:

\begin{eqnarray}
D_{L}^{*} &=& \arg \max_{D_{L}} p(D_{L}|I) \nonumber \\
&=& \arg \max_{D_{L}} p(I|D_{L}) p(D_{L}) \nonumber \\
&=& \arg \min_{D_{L}} -\log p(I|D_{L}) -\log p(D_{L}) \nonumber \\
&=& \arg \min_{D_{L}} C_{disc}(I_{C}) + h(D_{L}) \label{eq:regularized}
\end{eqnarray}

In words, by identifying the conditional probability of an image 
	given a description with the discrepancy cost function
	($-\log p(I|D_{L}) = C_{disc}(I_{C})$),
	and the prior probability of a description with the regularization function
	($-\log p(D_{L}) = h(D_{L})$),
	the regularized optimization procedure is transformed into a
	Bayesian inference problem.	
This arrangement has the benefit of eliminating the $\lambda$ parameter,
	but sheds no light on the problem of selecting the two crucial functions.

But more insight can be gained by analyzing the problem in terms of data compression.
Consider a sender and a receiver who have agreed to transmit images
	using an encoding scheme based on the graphics program and
	the associated description language $L$.
The sender first transmits a scene description $D_{L}$,
	which the receiver feeds to the graphics program to 
	construct the uncorrected image $G(D_{L})$.
The sender then transmits the correction image $I_{C}$,
	allowing the receiver to losslessly recover the original image.
The parties have agreed on a prior distribution $p(D_{L})$ for the scene descriptions, 
	and a method of encoding the correction image that requires 
	a codelength of $C_{enc}(I_{C})$.
The goal is to find a good $D_{L}^{*}$ that minimizes the total codelength:

\begin{equation}
\label{eq:mdl}
D_{L}^{*} = \arg \min_{D_{L}} \big( C_{enc}(I_{c}) -\log_{2} p(D_{L}) \big)
\end{equation}	

This formulation of the problem is thus equivalent to Equation~\ref{eq:regularized},
	showing that the general problem of computer vision can 
	be formulated in terms of compression.
If the procedure is only going to be applied to a single image,
	then this perspective is not much better than the previous one.
But if many images are going to be sent,
	then this formulation provides a clean principle for
	selecting the prior and the cost function:
	they should be chosen in such a way as to minimize the total cost
	for the entire database.

%\begin{figure}
%\begin{centering}
%\includegraphics[width=.5\textwidth]{backnforth.png} 
%\caption{
%Inverse relationship of graphics and vision.
%}
%\label{fig:backnforth}
%\end{centering}
%\end{figure}

This abstract analysis shows that there is 
	another, deeper problem in computer vision that is rarely addressed
	because the standard problem is hard enough.
This is the problem of choosing a description language $L$.
It is not obvious how the traditional conceptual framework
	of computer vision can be used to solve the problem of choosing $L$.
In contrast the CRM provides a direct answer:
	given two description languages $L_{a}$ and $L_{b}$,
	prefer the one that can be used to obtain better compression rates.
Note how this criterion simultaneously evaluates two computational tools: 
	the description language and the algorithmic methods used to 
	obtain actual descriptions.
Perhaps counterintuitively, 
	the \compist may get better results from using a simplistic
	description language instead of a more realistic, full-bodied one,
	if the former supports a better interpretation algorithm.
	
The two formulations of the vision problem discussed above exhibit
	very different answers to the question of why it is 
	important to obtain good scene descriptions.
In traditional computer vision, 
	a good scene description is of interest for qualitative, humanistic reasons. 
This motivation makes it very difficult to evaluate methods, 
	since human input is required to determine the quality of a result. 
In contrast, in \compist vision research, 
	a good scene description is of interest for quantitative, mechanistic reasons. 
This takes the human out of the evaluation loop, 
	making it much easier to compare techniques.

\mpagebreak
	
\subsection{Stereo Correspondence}

The previous section showed that it was possible to reformulate
	a very abstract version of the vision problem in terms of compression.
This and the following two sections show how this reformulation can work
	for \textit{specific} vision problems.
But for the case of the stereo correspondence problem,
	the argument has already been made,
	by David Mumford (emphasis in original):
	
\begin{quote}
I'd like to give a more elaborate example to show how MDL can lead you to
	the correct variables with which to describe the world using an old and familiar
	vision problem: the stereo correspondence problem. 
The usual approach to stereo vision is to apply our knowledge of the three-dimensional structure of the
	world to show how matching the images $I_{L}$ and $I_{R}$ from the left and right eyes
	leads us to a reconstruction of depth through the ``disparity function'' $d(x,y)$
	such that $I_{L}(x+d(x,y),y)$ is approximately equal to $I_{R}(x,y)$. 
In doing so, most algorithms take into account the ``constraint'' that most surfaces in the world are
	smooth, so that depth and disparity vary slowly as we scan across an image.
The MDL approach is quite different. 
Firstly, the raw perceptual signal comes as two sets of $N$ pixel values $I_{L}(x,y)$ and $I_{R}(x,y)$ 
	each encoded up to some fixed accuracy by $d$ bits, totaling $2 d N$ bits. 
But the attentive encoder notices how often pieces of the left image code nearly 
	duplicate pieces of the right code: this is a common pattern that 
	cries out for use in shrinking the code length. 
So we are led to code the signal in three pieces:~first the raw left image $I_{L}(x,y)$; 
	then the disparity $d(x,y)$; and finally the residual $I_{R}(x, y)$.
The disparity and the residual are both quite small, 
	so instead of $d$ bits, these may need only a small number $e$ and $f$
	bits respectively.
Provided $d > e + f$, we have saved bits.
In fact, if we use the constraint that surfaces are mostly smooth,
	so that $d(x,y)$ varies slowly, we can further encode $d(x,y)$ 
	by its average value $d_{0}(y)$ on each horizontal line
	and its $x$-derivative $d_{x}(x,y)$ which is mostly much smaller.
The important point is that MDL coding leads you to introduce the third coordinate of space, 
	i.e. to discover three-dimensional space!
A further study of the discontinuities in $d$, and the ``non-matching'' pixels
	visible to one eye only goes further and leads you to \textit{invent a description}
	of the image containing labels for distinct objects,
	i.e. to \textit{discover that the world is usually made up of discrete objects}
	~\cite{Mumford:1994}.
\end{quote}

Note how a single principle (compression)
	leads to the rediscovery of structure in visual reality
	that is otherwise taken for granted
	(authors of object recognition papers do not typically feel obligated
	to justify the assumption that the world is made up of discrete objects).
Mumford's thought experiment also emphasizes the intrinsic scalability 
	of the compression problem:
	first one discovers the third dimension, 
	and then that the world is made up of discrete objects.

\mpagebreak
	
\subsection{Optical Flow Estimation}
\label{sec:compformofe}

Another traditional task in computer vision is optical flow estimation.
This is the problem of finding the apparent motion 
	of the brightness patterns in an image sequence.
The optical flow problem can be reformulated as a specialized compression technique
	that works as follows.
Consider a high frame rate image sequence (say, 100 Hz).
Because of the high frame rate,
	the scene does not change much between frames.
Thus, a good way to save bits would be to encode and transmit full frames
	at a lower rate (say, 25 Hz), 
	and use an interpolation scheme to predict the intermediate frames.
The predicted pixel value would then be used as the mean for the 
	distribution used to encode the real value,
	and assuming the predictions were good, substantial bit savings
	would be achieved while maintaining losslessness.
Now, a ``dumb'' interpolation scheme could just linearly interpolate
	a pixel value by using the start and end frames.
But a smarter technique would be to infer the motion of the pixels
	(i.e. the optical flow) and use this information to do the interpolation.
	
The simplest encoding distribution to use would be a Gaussian with unit variance
	and mean equal to the predicted value.
In that case, the codelength required to encode a pixel
	would be simply the squared difference between the prediction
	and the real outcome, 
	plus a constant corresponding to the normalization factor.
A smarter scheme 
	might take into account the local intensity variation -
	if the intensity gradient is large, 
	it is likely that the prediction will be inaccurate,
	and so a larger variance should be used for the encoding distribution.
The resulting codelength for a single interpolated frame would be:

\begin{equation}
\sum_{x, y} \frac{(I(x,y) - I_{GT}(x,y))^{2}}{||\nabla I_{GT}(x,y)||^{2} + \epsilon}  + k(I_{GT}(x,y))
\end{equation}
	
Where $I(x,y)$ is the real frame and $I_{GT}(x,y)$
	is the image predicted from the optical flow,
	and the $k(\cdot)$ is a term corresponding to the normalization factor.
Since shorter codelengths are achieved by 
	improving the accuracy with which $I_{GT}$ predicts $I$,
	this shows that improvements in the optical flow estimation
	algorithm will lead to improvements in the compression rate.
Indeed, with the exception of the $k(\cdot)$ terms,
	the above expression is equivalent to an evaluation metric
	for optical flow algorithms proposed by~\cite{Baker:2007}
	(compare their Equation 1 to the above expression).
The compression-equivalent scheme is much simpler than the other metric 
	proposed by~\cite{Baker:2007},
	which involves the use of a complicated experimental apparatus 
	to obtain ground truth.
The compression metric permits
	\textit{any} image sequence to be used as empirical data.
		
\mpagebreak
	
\subsection{Image Segmentation}

As mentioned above,
	idea of image segmentation is to 
	partition an image into a 
	\textit{small} number of \textit{homogeneous} regions with 
	\textit{simple} boundaries.
Each of the italicized words is crucial for the problem to make sense at all.
The regions must be homogeneous, 
	otherwise one can simply draw arbitrary boundaries.
The boundaries must be simple,
	because otherwise it would be easy to segment
	similar pixels together by drawing complex jagged boundaries.
Thirdly the number of regions should be small,
	or one can solve the problem simply by creating thousands
	of mini-regions of four or five pixels each.

The segmentation problem can be formulated as a compression problem
	by using a compressor that encodes an image in terms 
	of a set of regions.
A specialized model is fit to each region,	
	and then used to encode the pixels of the region.
Because the model is region-specific,
	it describes the distribution of the pixels in the region
	better than a generic model would,
	and therefore can achieve a substantial codelength savings.
However, several conditions must be met for this to work. 
First, the pixels assigned to a region must be very similar to one another. 
Second, the format requires that the contours of the regions must also be encoded, 
	so that the decoder knows what model to use for each pixel.
To reduce this cost,
	it is necessary to use simple region boundaries,
	so that they can be encoded with a short code.
Thirdly,
	the sender is also obligated to send the parameters required
	to define each region-specific model,
	so the total number of regions should be kept low.
These three considerations supply cleanly justified definitions
	for the problematic words (homogeneous, simple, small) mentioned above. 

Several authors have adopted the compression-based 
	approach to segmentation~\cite{Kanungo:1994,Leclerc:1989}. 
The following is a brief discussion of a method proposed by Zhu and Yuille~\cite{Zhu:1996}. 
Zhu and Yuille formulate the segmentation problem as a minimization of 
	a codelength functional:

\begin{eqnarray}
\sum_{i}^{M} \left\{ \lambda  - \log P \left( {I_{x,y}:(x,y) \in R_{i} }|\alpha_{i} \right) + 
	\frac{\mu}{2} \int_{\partial R_{i}} ds \right\}
\label{eq:zhumdl}
\end{eqnarray}

\noindent
Here $R_{i}$ denotes the $i$th region with boundary $\partial R_{i}$,
	$\alpha_{i}$ is a set of parameters specifying a region-specific model,
	$\lambda$ is a per-region overhead cost,
	and there are $M$ total regions.
The goal is to find choices for $M$ and $R_{i}$ 
	that minimize the functional,
	which represents the cost that would be required to encode
	the image using a segmentation-based compression format.
The overhead cost of specifying a region-specific model is $\lambda$.
The cost of specifying the boundary $\partial R_{i}$ 
	is given by the contour integral term.
The cost of encoding the pixels in region $R_{i}$
	using the specialized model defined by $\alpha_{i}$
	is given by the $\log P(\ldots)$ term.
The minimization is achieved 
	by balancing the need to package similar pixels together,
	while at the same time using a small total number of regions
	with simple boundaries.
The minimization problem is a competition between the need 
	to package similar pixels together so that narrow, region-specific
	model distributions will describe them well,
	and the need to use a small number of regions with simple boundaries.
	
Zhu and Yuille focus most of their effort 
	on the development of an algorithm for 
	finding a good minimum of Equation~\ref{eq:zhumdl}.
The result is an algorithm where the regions
	``compete'' over boundary pixels;
	the winner of the competition is the region
	that can encode the pixel with the shortest code.
Little effort is expended
	on examining Equation~\ref{eq:zhumdl} itself.
For example,
	the cost of a region boundary in Equation~\ref{eq:zhumdl}
	is proportional to its length.
But this seems inefficient;
	it should be possible to define a polygonal boundary
	by specifying a small number of vertices.
Also, the region-specific models $P(\cdot|\alpha_{i})$ 
	are chosen to be simple multivariate Gaussians.
This choice is not supported by any kind of empirical evidence.
Finally, the paper does not report any actual compression results,
	only segmentation results, and only on a few images.

The reason for these oversights is obvious:
	Zhu and Yuille view compression as merely a trick
	that can be used to obtain good segmentation.
If the authors had adopted the compression principle
	as their primary goal,
	they would have had to ask an awkward question:
	is segmentation really a good tool for describing images?
That is to say, is segmentation really a scientific idea?
Is it empirically useful to describe images in terms of a set of regions?
This question cannot be answered without empirical investigation.

At this point it is worth	
	noting that the edge detection task discussed above
	has no direct reformulation as a compression problem.
However, 
	it seems very likely that having a good edge detector
	will be useful for the image segmentation problem.
Edge detector algorithms can therefore be justified
	in \compist research by showing how they contribute
	to the performance of segmentation-based compressors.

\mpagebreak

\subsection{Face Detection and Modeling}
\label{sec:facedetmod}

Consider a \compist inquiry that uses
	the image database hosted by the popular internet 
	social networking site Facebook.
This enormous database contains many images of faces.
Faces have a very consistent structure,
	and a computational understanding of that structure
	will be useful to compress the database well.
% TODO cite SIGGRAPH paper of Blanz Vetter
There is a significant literature on modeling faces~\cite{Blanz:2003,Decarlo:1998},
	and several techniques exist that can produce convincing
	reproductions of face images from models with a
	small number of parameters.
Given a starting language $L$,
	by adding this kind of model based face rendering technique
	a new language $L_{f}$ can be defined that contains the ability 
	to describe scenes using face elements.
Since the number of model parameters required is generally small
	and the reconstructions are quite accurate,
	it should be possible to significantly compress the Facebook
	database by encoding face elements instead of raw pixels when appropriate.

However it is not enough just to add face components 
	to the description language.
In order to take advantage of the new face components of the language 
	to achieve compression,
	it is also necessary to be able to obtain good descriptions
	$D_{L_{f}}$ of images that contain faces.
If unlimited computational power were available,
	then it would be possible to test each image subwindow to determine
	if it could be more efficiently encoded by using the face model.
But the procedure of extracting good parameters for the face model is relatively expensive,
	so this brute force procedure is inefficient.
A better scheme would be to use a fast classifier for face detection
	such as the Viola-Jones detector described above~\cite{Viola:2004}.
The detector scans each subwindow,
	and if it reports that a face is present,
	the subwindow is encoded using the face model component.
Bits are saved only when the detector correctly predicts that
	the face-based encoder can be used to save bits for the subwindow.
A false negative is a missed opportunity,
	while a false positive incurs a cost related to the inappropriate
	use of the face model to encode a subwindow.
In other words, the face model implicitly defines a virtual label $V(W)$ 
	for each subwindow $W$:

\begin{equation}
V(W) = D(W) - F(W)
\end{equation}

\noindent
Where $D(W)$ is the default cost of encoding the window,
	and $F(W)$ is the cost using the face model.
These virtual labels depend only on the face model, the original encoder, and the image data,
	so they can be generated with no human effort.
The face detector can be trained using the virtual labels,
	and the performance of the combined detector/modeler system
	can be evaluated using the overall compression rate.
Since one important bottleneck in machine learning
	is the limitation on the amount of labeled data that is available,
	this technique should be very useful.
	
The same basic strategy can be used to evaluate object recognition systems,
	though the state of the art of generic object modeling
	is less advanced. 
Indeed, the virtual label strategy can be used 
	whenever a synthesizer for some kind of data can be found.
For example, one could attempt to train 
	a speech detection system
	using a large audio database and a voice synthesizer.
	
\mpagebreak

\chapter{Compression and Language}
\label{chapt:linguistic}

\section{Computational Linguistics}

\Compist~research begins by
	finding a large pool of structure-rich data
	that can be obtained cheaply.
In addition to images,
	another obvious source of such data is text.
The study of text 
	is also interesting from a humanistic perspective,
	due to the crucial role language plays in human society.
The line of inquiry resulting
	from applying the Compression Rate Method
	to large text corpora
	has much in common with the modern field known as computational linguistics (\textsc{CL}).

Computational linguistics is the field dedicated to
	computer-based analysis and processing of text.
In many ways, 
	the field is similar to computer vision in mindset and philosophical approach.
\textsc{CL} researchers employ many of the same mathematical tools 	
	as vision researchers,
	such as Hidden Markov Models, Markov Random Fields,
	and the Support Vector Machine algorithm.
The two fields also share a mindset:
	they both produce hybrid math/engineering results,
	and are strongly influenced by the field of computer science.
Both fields suffer from a ``toolbox'' mentality:
	researchers produce a large number of candidate solutions for each task,
	but lack convincing methods by which to evaluate those solutions.
	
Computational linguistics differs from computer vision
	in that it has a sister field - traditional linguistics - 
	from which it borrows many ideas.
Linguistic theories sometimes play a role in the 
	development such as parsers or machine translation systems.
The parser developed by Collins, described below,
	includes ideas such as \textit{Wh}-movement and 
	the complement/adjunct distinction~\cite{Collins:2003}.
Another area in which traditional linguistic theory
	plays a role is in the development
	of parsed corpora such as the Penn Treebank.
A treebank is a set of sentences with attached parse trees;
	such a resource allows researchers to use learning systems
	to construct parsers.
To construct a treebank,
	developers must select a theory of grammar
	that defines the set of part-of-speech (POS) tags
	and syntactic tags (noun phrase, adverb phrase, \textit{Wh}-adjective phrase, etc).

However,
	the influence of standard linguistics theory 
	on computational linguistics is not as great
	as one might expect.
Often, CL researchers find it more effective to ignore linguistic knowledge
	and simply employ brute force computational or statistical techniques.
Thus, 
	the machine translation system developed by Brown~\etal,
	discussed below,
	uses minimal linguistic content~\cite{Brown:1994}.
This is actually considered an advantage,
	because it makes the system easier to port
	to a new language-pair.
Similarly,
	researchers in the field of language modeling
	often attempt to deploy model based on sophisticated linguistic concepts,
	only to find that such models are outperformed
	by simple $n$-grams.

One of the major seismic shifts in natural language research
	was the transition from rule-based systems
	to statistical systems.
Rule-based systems were constructed by assembling a group of linguists
	and transferring their knowledge,
	expressed in the form of rules, into a computer system.
These packages  were often fairly complex,
	because most linguistic rules have exceptions.
Even a basic rule such as ``all English sentences must have a verb''
	is often broken in practice.
The complexity of the rule-based systems led to brittleness,
	fragility, and difficulty of maintenance.
Furthermore, 
	the rule-based approach was unattractive because
	the methods developed for one language did not 
	often transfer to another language.
This made it expensive to add new languages to a system.

The field of statistical natural language processing
	emerged as a response to the limitations of the rule-based system.
This area was substantially pioneered by
	a group of researchers at IBM in the 90s~\cite{Brown:1994,Berger:1996,DellaPietra:1997}.
A statistical system attempts to avoid making explicit commitments
	to a certain type of structure.
Instead, the system attempts to \textit{learn} the structure of text
	by analyzing a corpus.
Thus, a major ingredient for any learning-based natual language system
	is a corpus of text.
In the case of statistical machine translation,
	it is essential to have a \textit{bilingual} or parallel corpus:
	one which contains sentences from one language
	side by side with sentences from the other language.
The system learns how to translate
	by analyzing the relationship between
	the two side-by-side sentences.

From the perspective of \compist research,
	a particularly interesting \textsc{CL} problem
	is language modeling.
The goal here is to find good statistical models of lanuage,
	which achieve a low \textit{cross-entropy} score
	on a text database.
Cross-entropy is just the negative log-likelihood 
	of the text data using the model.
In other words,
	it is just the compression rate without the model complexity penalty.
Section~\ref{sec:statlangmd}
	provides an analysis of this subfield and its relationship
	to \compist~research.
Two other standard CL topics,
	statistical parsting and machine translation,
	are discussed in Sections~\ref{sec:secparsing} and~\ref{sec:statmachtrn}.
Some other tasks,
	such as document classification and word sense disambiguation,
	are discussed briefly in Section~\ref{sec:addtremark}.
Again,
	the goal is not to provide a comprehensive survey of the field,
	or even to describe individual papers in any depth.
Instead, 
	the idea is to give nonspecialist readers
	an impression of the basic issues in a given area.
Each section contains a description of the task,
	followed by an analysis of the mechanisms 
	used to evaluate candidate solutions.
A brief critique of the research is given,
	focusing primarily on the evaluation schemes.
Then a \compist~reformulation of the problem is given.

\mpagebreak

\section{Parsing}
\label{sec:secparsing}

An important part of the structure of natural language
	is grammar.
Grammatical rules govern how different
	elements of a sentence can fit together.
Consider the sentence ``John loves Mary''.
That sentence can be parsed as follows:

\begin{verbatim}
(S (NP (NNP John))
   (VP (VPZ loves)
       (NP (NNP Mary))) 
\end{verbatim}

\noindent
Here NNP and VPZ are part of speech (POS) tags for nouns and verbs,
	while S, NP, and VP are syntactic tags
	representing sentence, noun phrase, and verb phrase structures.
The goal of parsing is to recover
	both the POS tags and the syntactic tags,
	given the words of the sentence.
Thus, 
	the parsing problem includes the POS tagging problem.

In principle,
	an important part of designing a parser
	is to determine the set of \textsc{pos} tags
	and syntactic tags one wishes to obtain.
While there is agreement about the basics,
	different linguists endorse different theories of grammar,
	which in turn utilize different abstractions.
However, 
	this problem has been solved in practice
	by the appearance of the Penn Treebank~\cite{Marcus:1993}.
This is a corpus of text for which parse
	trees have been provided by human annotators
	(see further discussion below).
Thus,
	most parsers simply utilize the same set
	of tags as the Penn Treebank.
	
A very common approach to the parsing problem
	is based on an idea called the Probabilistic Context Free Grammar (PCFG).
A PCFG is a list of symbols,
	and a set of rewrite rules that can be used to transform those symbols.
For example,
	a PCFG could specify that an NP can transform into an adjective and a noun.
Another rule might allow the NP to transform into a determiner-adjective-noun sequence.
Each rewrite rule has a probability attached to it,
	allowing the system to express the fact
	that some grammatical structures are more common than others.
The rewrite rules of a PCFG can include recursion.
For example,
	an NP might be rewritten as a PP and a NP.
Recursion allows the PCFG to produce an infinite number of sentences.

% TODO - exponential or superexponential?
One key difficulty in parsing is that the number of possible 
	parse trees is exponential in the length of the sentence.
This means that even if a parser could determine
	the validity of a particular parse tree with perfect accuracy,
	it would be unable to test all possible parses trees.
Instead, the parser must employ some search strategy
	that narrows down the number of parses that actually get examined.
This is a bit like reconstructing the details
	of a crime from examining evidence left at the scene.
Some explanations make more sense than others,
	but if one does not glimpse the possibility of a certain explanation,
	one might end up concluding that a less-likely explanation is true.

\mpagebreak
	
One influential paper in this area is 
	``Learning to Parse Natural Language with Maximum Entropy Models''
	by Adwait Ratnaparkhi~\cite{Ratnaparkhi:1999}.
Ratnaparkhi uses the idea of \textit{shift-reduce} parsing,
	a standard technique used in computer science
	to compile software from human readable form to machine code.
A shift-reduce parser,
	as the name indicates, applies two basic operations.
The shift operation
	pushes an input element onto the stack,
	where it awaits further processing.
The reduce operation
	joins an input element together 
	with the element on top of the stack,
	producing an element of a new type.
In grammatical terms,
	this means taking two elements such as
	\textit{the} (determiner) and \textit{dog} (noun)
	and joining them to produce a noun phrase.

The main issue with applying this technique to natural language parsing,
	as opposed to software parsing,
	is that in the former there is ambiguity about
	the correct way of combining elements together.
To deal with this ambiguity,
	Ratnaparkhi's parser does not use a deterministic rule 
	to decide whether to combine two elements (reduce).
Instead, it considers multiple options.
In other words, 
	when it comes to a fork in the road,
	it pursues both paths,
	at least for a little while.
Each operation, or fork in the road, is assigned a probability.
And a full derivation,	
	or a path through the woods,
	has a net probability determined by the product
	of each individual operation.
The algorithm attempts to find the path with
	the highest net probability.

The probabilities assigned to single operations 
	are obtained using a Maximum Entropy (MaxEnt) model.
MaxEnt models are of the form:

\begin{equation}
p(a|b) = \frac{1}{Z(b)} \exp \big ({\sum_{i} \lambda_{i} f_{i}(a, b)} \big)
\end{equation}

\noindent
Where $p(a|b)$ is the probability of an operation $a$
	in a context $b$, $f_{i}$ are a set of \textit{context functions},
	the $\lambda_{i}$ are coefficients related to the context functions,
	and $Z(b)$ ensures normalization.
The operations $a$ represent either shift or reduce.
The context $b$ contains information related to
	the current group of words being examined
	and the nearby subtrees.
The key property of MaxEnt is that it allows the user
	an enormous amount of flexibility in defining
	the context functions $f_{i}$.
The researcher can test out many different kinds of context functions
	and the training algorithm will automatically
	select the optimal parameters $\lambda_{i}$.

Ratnaparkhi also presents an algorithm
	for searching for good parse trees.
The algorithm is a type of breadth first search (BFS)
	that prunes a potential parse candidate
	if its probability,
	assigned by the MaxEnt model, is too low.
This pruning is necessary because
	the total number of parse trees is vast.

\mpagebreak

A more recent paper on parsing
	is ``Head-Driven Statistical Models for Natural Language Parsing''
	by Michael Collins~\cite{Collins:2003}.
Abstractly,
	Collins strategy is to define a joint probability model $P(T,S)$,
	where $T$ is a parse tree and $S$ is a sentence.
Then, given a particular sentence, his algorithm maximizes 
	the conditional probability of the tree given the sentence:
	
\begin{equation*}
T^{*} = \arg \max_{T} P(T|S) = \arg \max_{T} \frac{P(T,S)}{P(S)} = \arg \max_{T} P(T,S)
\end{equation*}
	
Given this formulation,
	there are two basic problems:
	how to define the model $P(T,S)$,
	and how to perform the actual maximization.
Collins spends most of his attention on the former problem,
	noting that there are standard algorithms
	that can be applied to handle the latter.
Collins employs the PCFG framework to define $P(T,S)$,
A PCFG model defines the joint probability 
	as simply the product of each of the 
	expansions used in the tree.
Each expansion is a transformation of the form $\alpha \rightarrow \beta$.
For example, a verb phrase could transform
	into a verb and a noun phrase.
If the parse tree includes $n$ expansions,
	then the full joint probability is given by the expression:

\begin{equation*}
P(T,S) = \prod_{i=1}^{n} P(\beta_{i} | \alpha_{i})
\end{equation*}

\noindent
Where $\alpha_{i}$ and $\beta_{i}$ are the pre- and post-expansion
	structures involved in the $i$th step of the derivation
	embodied by the parse tree.
It is worth noting
	the close relationship between this approach to parsing
	and the problem of modeling the standalone
	probability of a sentence, $P(S)$,
	which can be obtained from $P(T,S)$ by marginalizing over $T$.

To complete the model,
	it is necessary to find the probabilities
	of a particular expansion $P(\beta|\alpha)$.
If a parsed corpus is available,
	this can be done by simple counting:
	
\begin{equation*}
P(\beta|\alpha) = \frac{Count(\alpha \rightarrow \beta)}{Count(\alpha)}
\end{equation*}

While the procedure described above is almost complete,
	the basic PCFG model is too simplistic 
	to provide good performance.
Collins' main focus in the paper is on various strategies
	for making the model more realistic.
One technique is to use a \textit{lexicalized} PCFG,
	where each nonterminal (node in the parse tree)
	includes not only a syntactic tag
	(noun phrase, prepositional phrase, etc)
	but also a head word and an associated POS tag.
So for the sentence ``Last week, IBM bought Lotus'',
	the root node is $\textsc{S(bought, VBD)}$.
The root node then expands
	into $\textsc{NP(week, nn)}$, $\textsc{NP(ibm, nnp)}$, 
	and $\textsc{VP(bought, vbd)}$.
The PCFG formalism can easily accomodate this;
	it simply means there are many more symbols and expansions.
The point of this modification is that specific words
	often contain very useful information that can affect
	the probabilities of various derivations.

The major drawback of using the lexicalized PCFG approach
	is that it vastly increases the number of derivational rules,
	leading to severe sparse data issues.
The specific derivation
	of $\textsc{S(bought, vbd)} \rightarrow 
		\textsc{NP(week, nn)} \cdot \textsc{NP(ibm, nnp)} \cdot \textsc{VP(bought, vbd)}$
	probably occurs only once in the corpus.
Collins makes a number of independence assumptions
	to alleviate the sparse data problems
	and reduce the number of parameters in the model.
To do this, 
	he notes that all production rules are of the form:
	$\alpha \rightarrow \{L_{1} \ldots L_{n}, H, R_{1} \ldots R_{m} \}$
	where $\{L_{i}\}$, $H$, and $\{R_{j}\}$ 
	designate the left side components, the head word component,
	and the right side components respectively.
So in the example given above,
	the $H$ is the node associated with \textit{bought},
	$L_{1}$, $L_{2}$ are the components associated with \textit{week}
	and \textit{IBM}, and there are no components on the right side.
Then the simplifying independence assumption
	is that the probability for each node in $L,H,R$
	depends only on the parent $P$.
This technique makes it much easier to estimate
	the relevant probabilities.
	
However, it turns out that this assumption is actually too strong,
	and makes it impossible to capture certain 
	linguistic structures.
Collins describes three increasingly sophisticated models,
	that capture increasingly complex information
	about the structure of sentences.
The first model introduces 
	the notion of linguistic distance into the model,
	allowing it to take into account
	the history of previously applied expansions.
Roughly,
	this means that when expanding a node,
	information about the node's parent node
	or sibling nodes can be used to modify
	the expansion probabilities.

The second model introduced by Collins 
	attempts to handle the distinction between
	adjuncts and complements.
A \naive~parse of the sentence
	``Last week IBM bought Lotus''
	would identify both \textit{Last week}
	and \textit{IBM} as noun phrases.
But these two elements actually have distinct roles:
	\textit{IBM} is the subject of the verb \textit{bought},
	while \textit{Last week} is an adjunct modifying the verb.
Collins introduces a new set of variables into the model, 
	which specify whether a nonterminal generates
	a left or right complement.
Most verbs take one left complement,	
	which represents the subject.
So the verb \textit{bought} will generate one left complement,
	and the parser can assign the word \textit{IBM}
	to this role.
Of course, 
	all of these rules are expressed probabilistically:
	it is not impossible for a verb to have multiple
	left complements, just very unlikely.

Collins' third model
	attempts to handle the phenomena of 
	\textit{Wh}-movement and traces.
The importance of traces can be seen in the following examples:

\begin{quote}
1) The company (SBAR that TRACE bought Lotus). \\
2) The company (SBAR that IBM bought TRACE).
\end{quote}

\noindent
In the first sentence, 
	\textit{company} refers to IBM,
	which fills the trace position.
In the second sentence,
	\textit{company} refers to Lotus.
One way to handle \textit{Wh}-movement
	is to use the notion of a gap.
A gap starts with a trace,
	and propagates up through the parse tree
	until it finds something to resolve with.
In sentence \#2 above,
	the gap starts as the complement of \textit{bought}
	and propagates up the tree until
	it resolves with \textit{company}.
The gap element is added as another variable
	in the now quite complex parsing model.

After defining the third model,
	Collins makes a variety of additional 
	refinements to handle issues like 
	punctuation, coordination,
	and sentences with empty subjects.
The importance of coordination can be seen 
	by considering the phrase ``the man and his dog'',
	which has a head word $\textsc{NP(man)}$.
This component then expands 
	as $\textsc{NP(man)} \rightarrow \textsc{NP(man)} \textsc{CC(and)} \textsc{NP(dog)}$.
The issue is that,
	after the coordinator \textit{and} appears,
	it becomes very likely that another element will be produced.
A \naive~scheme, however,
	that does not take into account the presence of the coordinator,
	will put high probability on the outcome where
	no additional elements are produced.
The empty subject issue relates to 
	sentences like ``Mountain climbing is dangerous''.
The Penn Treebank tags this sentence
	as having no subject.
This linguistic analysis of the structure is problematic,
	because it causes the model to assign high probability
	to sentences with no subject.
An alternative analysis would conclude 
	that it is actually very rare for English sentences
	to have no subject.
To deal with this issue,
	Collins uses a preprocessing step	
	to transform trees with these kind of empty subjects
	into a simpler form.
	
\mpagebreak

\subsection{Evaluation of Parsing Systems}
	
As noted above,
	one of the most important developments
	in the history of statistical parsing research 
	was the appearance of parsed corpora
	such as the Penn Treebank (Marcus~\etal~\cite{Marcus:1993})
	and the Penn Chinese Treebank (Xue~\etal~\cite{Xue:2005}).
A treebank is a corpus of sentences
	with attached parse tree information,
	which has been produced by human annotators.
These parsed corpora,
	which are analogous to labeled datasets in machine learning research,
	allowed researchers to apply statistical learning
	techniques to the problem of parsing.

A key issue in the development of a treebank
	is that there is no single, objectively correct
	method or ruleset for parsing.
To parse a sentence,
	one must implicitly employ a theory of grammar
	which describes the rules of parsing
	and the elements included in a parse tree.
Though there is widespread agreement regarding basic elements like
	``verb'' and ``noun'',
	the precise content of grammatical theories
	are the subject of continuing research and 
	debate in the field of linguistics.
In order to construct a treebank,
	the developers must make choices about which 
	theory of grammar to use.
This issue is acknowledged by the developers
	of the Penn Chinese Treebank:
	
\begin{quote}
When we design the treebank,
	we consider (user) preferences and try to accomodate them when possible.
For instance, people who work on dependency parsers
	would prefer a treebank that contains dependency structures,
	while others might prefer a phrase structure treebank...
It is common for people to disagree on the underlying linguistic theories
	and the particular analyses of certain linguistic phenomena 
	in a treebank~\cite{Xue:2005}.
\end{quote}

\noindent
And later: 

\begin{quote}
Another desired goal is theoretical neutrality.
Clearly we prefer that this corpus survives ever changing linguistic theories.
While absolute theoretical neutrality is an unattainable goal,
	we approach this by building the corpus on the ``safe'' assumptions
	of theoretical frameworks ...
	the influence of Government and Binding theory
	and X-bar theory is obvious in our corpus,
	we do not adopt the whole package~\cite{Xue:2005}.
\end{quote}

A concrete example of this kind of issue
	relates to the choice of POS tags in the Penn Treebank~\cite{Marcus:1993}.
Marcus \etal~note that in the Brown Corpus,
	upon which the Penn Treebank is based,
	the contraction \textit{I'm} is tagged as PPSS+BEM.
PPSS indicates a ``non-third person nominative personal pronoun'',
	while BEM is a special tag reserved
	for \textit{am} or its contracted form \textit{'m}.
In contrast to the Brown Corpus,
	the Penn Treebank uses a much smaller number of tags.
This raises the issue of whether the treebank
	used the ``right'' or ``optimal'' set of tags,
	or if it even makes sense to discuss such an issue.
Marcus~\etal~also not that in some cases, 
	the correct tag of a word simply cannot be conclusively determined
	from the sentence.
In that case,
	the annotator marks the word with a double tag.
	
In addition to the conceptual issue 
	of how to choose a linguistic theory 
	to guide the annotation process,
	treebank developers must face the very practical
	issue of how to minimize the amount of human labor required.
Manually constructing a large number of parse trees
	is a highly time-intensive process.
This fact puts strong constraints on the ultimate form
	the database takes.
As Marcus~\etal~note,

\begin{quote}
Our approach to developing the syntactic tagset was highly pragmatic
	and strongly influenced by the need to create a large
	body of annotated material given limited human resources~\cite{Marcus:1993}.
\end{quote}
	
\noindent
One way to make the annotation work easier
	is to use an automatic parsing tool as
	a preprocessing step.
The human annotators then correct
	any errors the automatic tool may have made.
While this scheme saves human labor,
	it also puts subtle constraints on the 
	ultimate form of the annotation output.
The developers of the Penn Treebank
	used a program called Fidditch to 
	perform the initial parsing~\cite{Hindle:1989}.
Fidditch makes certain grammatical assumptions and 
	produces parse trees that reflect those assumptions.
The human annotators can correct small errors made by Fidditch,
	but do not have time to make comprehensive revisions.
Thus, the grammatical assumptions made by Fidditch are
	built into the structure of the annotations of the treebank.
	
Treebank developers rely on human annotators,
	who sometimes make mistakes.
This is to be expected,
	since parsing is a cognitively demanding task,
	and the annotators are encouraged to perform as efficiently as possible,
	to maximize the total number of words in the corpus.
Marcus \etal~note that the median error rate
	for the human annotators was 3.4\%~\cite{Marcus:1993}.
This implies that there are a substantial number of errors
	in the treebank.

Given the human-annotated parse tree,
	it is fairly straightforward to define
	a score for the machine-generated tree.
The basic quantities are $N^{*}$,
	the number of correctly labeled constituents,
	$N_{p}$, the number of constituents in the machine parse,
	and $N_{t}$, the number of constituents in the human parse.
For a constituent to be correct,
	it must span the same range of words as the human parse,
	and have the same label.
Research papers report performance in terms 
	of precision ($N^{*}/N_{p}$),
	recall ($N^{*}/N_{t}$),
	and $F$-score, which is the harmonic mean of precision and recall.
	
One notable aspect of research in statistical parsing
	is that the range of scores reported in the literature
	is quite narrow.
A paper published in 1995,
	two years after the publication of the treebank,
	achieved an $F$-score of 85\%~\cite{Magerman:1995}.
Eleven years later,
	a paper published in 2006 achieved an $F$-score of 92\%~\cite{Mcclosky:2006};
	this score appears to be comparable to the state of the art
	at the time of this writing.
It is not clear if additional research
	will produce additional improvements,
	or if there is some natural limit
	to the performance that can be achieved by statistical parsers
	on this problem.

% This is just a convention, not a real source of confusion.
%An interesting issue arises in Chinese that does not exist in English:
%	word segmentation.
%Chinese text consists of a stream of \textit{hanzi} characters,
%	with no explicit separating marks.
%In order to do POS tagging and syntactic bracketing,
%	the characters must first be segmented into words.
%The problem is that there is no obvious definition of what constitutes a word.
%Different criteria can lead to different word boundaries.

\mpagebreak

\subsection{Critical Analysis}

The major shortcoming of research in statistical parsing
	is that it is totally dependent on the existence
	of a parsed treebank.
Researchers rely on the treebank both to \textit{train}
	their systems and to \textit{evaluate} their systems.
This has several negative implications.

The first implication is that
	because the treebanks drive the development of the parsers,
	the assumptions made by the treebank authors
	are ``baked in'' to the parsers.
If the treebank developers use X-bar theory 
	to guide the annotation process,
	then systems will learn to parse sentences using X-bar theory~\cite{Jackendoff:1977}.
If it turns out that X-bar theory is incorrect,
	there is no way for researchers in statistical parsing to discover that fact.
This point is exactly analogous to the Elemental Recognizer thought experiment
	of Chapter~\ref{chapt:comprvsion},
	except that researchers are building systems based on 
	Chomsky's X-bar theory instead of Aristotle's theory of the 
	four elements.
	
In comparison to the process of building building systems 
	that learn to regurgitate the abstractions used by linguists,
	a far superior strategy would be for researchers to 
	use learning systems to test and validate those abstractions.
For example,
	an important issue in the design of the Penn Chinese Treebank
	is the question of how to characterize the special \textit{ba}-construction of Chinese.
Different linguists have argued that \textit{ba} is a verb,
	a preposition, a topic marker, and various other things.
After discussing the issue with various linguists,
	Xue~\etal~finally decided to categorize \textit{ba} as a verb~\cite{Xue:2005}.
As a result of this choice,
	all the statistical parsers trained on the Penn Chinese Treebank
	will learn to recognize \textit{ba} as a verb.
If that categorization choice was an error,
	the learning systems will be learning to duplicate an error.
It would be infinitely preferable
	if the parsing systems could be used as tools to answer linguistic questions,
	such as how to categorize the \textit{ba}-construction.

This general principle holds for many of the issues 
	related to parsing.
What is the optimal set of POS tags?
What is the optimal syntactic tagset?
Is a dependency structure representation of grammar
	superior to a phrase structure representation?
Computational linguists can answer these questions on their own,
	using grammatical introspection
	and other tools of traditional linguistic analysis,
	and then train the systems to produce the same answers.
Or they can attempt to use learning systems
	to automatically discover the optimal answer
	using some alternative principle,
	such as the compression rate.
	
A more concrete criticism of statistical parsing research
	relates to the amount of progress achieved
	compared to the amount of human effort required.
As noted,
	treebanks require a substantial investment of human time to develop.
Unfortunately,
	it is not clear how much of an improvement
	in the state of the art a treebank produces.
11 years of research relating to the Penn Treebank
	seems to have produced about a 7\% absolute 
	increase in $F$-score.
Does this improvement indicate a significant advance
	in the power of statistical parsers?
Given that parsers now achieve $F$-scores of around 92\%,
	does that mean parsing is a solved problem?
If not,
	will the research community need to construct a new
	and more difficult treebank in order to make further progress?
If each small advance in parsing technology requires
	a massive expenditure of time and money,
	it would seem that this approach is hopeless.
	
\mpagebreak

\subsection{\Compist~Formulation}

The fundamental principle of \compist~philosophy 
	is that in order to compress a dataset,
	one must understand its structure.
Since grammar and part of speech information
	constitute a crucial component of the structure of text,
	the compression goal leads naturally to a study of those phenomena.
Consider the following sentence:

\begin{quote}
John spoke \_\_\_\_\_.
\end{quote}

\noindent
To compress this sentence well,
	the compressor must predict what word will fill in the blank.
Assuming it is known that the sentence contains only three words,
	it is clear that the word in the blank is going to be an adverb
	such as ``quickly'', ``thoughtfully'', or ``angrily''.
The compressor can save bits by exploiting this fact.
This shows how an analysis of parsing and grammar 
	can be justified by the compression principle.
	
There is no reason to believe this idea
	cannot be scaled up to include more sophisticated techniques.
Indeed, it is easy to see how a PCFG can be used
	as a tool for text compression.
Consider the following highly simplified PCFG:
	
\begin{quote}
S \\
1: P=1 : $\rightarrow \mathrm{NP} \cdot \mathrm{VP}$ \\
VP \\
1: P=0.5 : $\rightarrow \mathrm{V} \cdot \mathrm{NP}$ \\
2: P=0.3 : $\rightarrow \mathrm{V} \cdot \mathrm{NP} \cdot \mathrm{NP}$ \\
3: P=0.2 : $\rightarrow \mathrm{V} \cdot \mathrm{NP} \cdot \mathrm{PP}$ \\
NP \\
1: P=0.4 : $\rightarrow \mathrm{N}$ \\
2: P=0.4 : $\rightarrow \mathrm{A} \cdot \mathrm{N}$ \\
3: P=0.2 : $\rightarrow \mathrm{D} \cdot \mathrm{A} \cdot \mathrm{N}$
\end{quote}

\noindent
So a sentence (S) always transforms into a noun phrase (NP) plus a verb phrase (VP).
With probability $P=.5$ a verb phrase 
	transforms into a verb (V) and a single noun phrase (NP),
	with probability $P=.3$ it transforms into a verb
	and two noun phrases,
	and with probability $P=.2$ it transforms into a verb,
	noun phrase, and prepositional phrase (PP).
A noun phrase can transform into a single noun ($P=.4$),
	an adjective and a noun ($P=.4$)
	or a determiner, adjective, and noun ($P=.2$).
The verb, noun, adjective, and determiner categories 
	correspond to actual words.
Consider how this highly simplified
	grammar can be used to parse the following sentence:
	
\begin{quote}
The black mouse ate the green cheese.
\end{quote}

\noindent
This sentence actually only contains three derivational rules.
First, the sentence splits into a noun phrase (\textit{the black cat})
	and a verb phrase (\textit{ate the green cheese}).
The verb phrase then splits into a verb and a noun phrase.
Both of the noun phrases split 
	into the determiner, adjective, noun pattern.
A compressor can encode the sentence using its parse tree as follows.
The $\mathrm{S} \rightarrow \mathrm{NP} \cdot \mathrm{VP}$ derivation requires zero bits,
	since there are no other possibilities.
To encode the derivation 
	$\mathrm{VP} \rightarrow \mathrm{V} \cdot \mathrm{NP}$,
	the compressor sends rule \#1 in the VP list,
	at a cost of $-\log_{2} (0.5) = 1$ bit.
To encode the two derivations 
	$\mathrm{NP} \rightarrow \mathrm{D} \cdot \mathrm{A} \cdot \mathrm{N}$,
	it sends rule \#3 in the NP list,
	at a cost of $-\log_{2} (0.2) \approx 2.3$ bits each.
So the entire parse tree can be encoded at a cost
	of about 5.6 bits.
This is quite reasonable, 
	given that a \naive~encoding for a single letter
	requires $-\log_{2}(26) \approx 4.7$ bits.
Finally, 
	the compressor transmits the information necessary to transform
	a terminal category (verb, noun, etc)
	into an actual word.
Knowing the category allows the compressor to save bits,
	since $P(\mathrm{mouse}|\mathrm{noun})$ 
	is much greater than $P(\mathrm{mouse})$.

Notice that using parsing techniques not only saves bits,
	it also sets the stage for more 
	advanced analysis techniques,
	that can save even more bits.
Consider the sentence ``John kissed Mary''.
Some bits can be saved by encoding this sentence
	using its parse tree.
But more interestingly,
	finding the parse tree also allows a higher level
	system to apply a more sophisticated analysis.
Such a system might notice
	that when the verb is \textit{kissed},
	the subject is almost always a human,
	and the object is usually also human.
This information can then be used to save additional 
	bits when encoding the names \textit{John} and \textit{Mary}.

\mpagebreak 

\section{Statistical Machine Translation}
\label{sec:statmachtrn}

Machine translation is one of the oldest subfields
	of artificial intelligence,
	dating back to a paper written by Warren Weaver in 1949~\cite{Weaver:1955}.
Initial expectations were high:
	various researchers predicted that success would be achieved
	within a couple of years.
That prediction, of course,
	turned out to be wildly overoptimistic:
	machine translation is still not a solved problem,
	though some systems achieve acceptable performance
	in some cases.
Originally, a major impetus for the field came
	from the American defense and intelligence establishment,
	which desired the ability to translate Russian scientific documents.
Even today,
	US government agencies provide a major share
	of the funding for translation research,
	and also help to organize competitions and evaluations.

A major issue in machine translation is the problem
	of evaluation.
For a long time,
	the only reliable method for evaluating a candidate solution
	to the translation problem
	was to assemble a team of human judges,
	and have the judges assign scores
	to the computer translations.
More recently,
	a variety of automatic schemes for evaluating
	machine translations have appeared~\cite{Papineni:2002,Snover:2006,Banerjee:2005}.
These schemes seem to spurred a new burst of research
	in machine translation,
	though their use is still controversial.
	
A major paper that began the transition away from rules-based
	and towards statistical language processing is 
	``Mathematics of Statistical Machine Translation: Parameter Estimation''
	by Brown~\etal~\cite{Brown:1994}.
The authors approach the problem by using what they call the Fundamental
	Equation of Machine Translation:
	
\begin{equation}
\label{eq:machntrans}
E^{*} = \arg \max_{E} P(F|E) P(E) 
\end{equation}

\noindent
Where $E^{*}$ is the produced English translation of a French (or Foreign)
	sentence $F$.
This approach decomposes the problem into two subproblems:
	modeling $P(F|E)$ and $P(E)$.
The authors focus most of their attention on the former,
	noting that other research exists that deals with the latter.
This problem of conditional modeling is still extremely hard.
One way to get a sense of the difficulty is to notice
	the immense size of the models.
% TODO: notation here.
If the average sentence has $W = 10$ words,
	and there are $T = 5\cdot 10^{4}$ words in the language,
	then the size of $P(E)$ is on the order of $T^{W} \approx 9\cdot 10^{46}$,
	while the size of $P(F|E)$ is $T^{2 W} \approx 9\cdot 10^{93}$.
	
The enormous sizes of the spaces involved
	cause one of the key problems 
	of complex statistical modeling: sparse data.
In principle, 
	if one had enough data,
	one could estimate the probability $P(F|E)$ by simply
	counting the number of times a French sentence $F$ is used 
	as a translation of an English sentence $E$.
Unfortunately, that strategy is completely impractical 
	due to the huge outcome spaces.
Indeed, it is quite difficult even to estimate
	the probability $P(f|e)$ of a French word $f$
	serving as the translation of an English word $e$;
	a~\naive~method for estimating this model
	would require $W^2 \approx 2.5 \cdot 10^{9}$.
In order to cut down on the size of the spaces,
	and thus the amount of data needed to estimate
	the model probabilities,
	a variety of simplifying assumptions are necessary.
	
The simplification strategy reflects
	one of the fundamental insights of statistical modeling.
To understand the idea,
	imagine one was trying to estimate a model of a distribution of heights.
Say the measurements range from 130 cm to 230 cm,
	and are accurate to a precision of 0.1 mm.
Then a \naive~method would be to simply count the number of observations
	in every 0.1 mm-width bin, 
	and use the observed frequency as the probability estimate.
For example,
	the probability of a height being between 150.05 cm and 150.06 cm
	would just be the number of observations in that range,
	divided by the total number of observations.
This strategy requires 10,000 statistics:
	the counts for each bin.
It will work in the limit of large data,
	but if there are a small number of observations,
	then the estimates obtained in this way will be extremely unreliable.
To solve this problem one computes a different set of statistics:
	the mean and variance of the data set.
This simplifying assumption requires far fewer parameters,
	and will produce a much more accurate model
	with a limited amount of data.
	 
Brown~\etal~follow an analogous strategy	
	to reduce the number of parameters required by their models.
They introduce a number of simplifying concepts,
	one of which is the idea of an \textit{alignment}.
An alignment $A$ is a matching between the French words and the English
	words of the two sentences.
If the alignment can be found,
	the probability of an English word in a given position can be modeled as:

\begin{equation*}
P(e_{i}|A, F) = P(e_{i}| f_{a_{i}})
\end{equation*}
	
\noindent
Where $e_{i}$ is the $i$th word in the English sentence,
	and $f_{a_{i}}$ is the French word aligned with it.
However, finding an alignment introduces its own challenges,
	because it is a hidden variable:
	it cannot be extracted directly from the corpus,
	but must be inferred from the sentence pair.
In order to find the most likely alignment,
	various pieces of knowledge about the translation parameters
	must be known.
Given a sentence-pair $(E,F)$ and a good model $P(E,A|F)$,
	then the most likely alignment can be found by:
	
\begin{equation*}
A^{*} = \arg \max_{A \in \cal{A}} P(E,A|F)
\end{equation*}

\noindent
Where $\cal{A}$ is the space of all possible alignments.
However, to use this optimization,
	it is necessary to have a good model $P(E,A|F)$.
If the corpus contained the alignment data,
	it would be possible to estimate the parameters for
	such a model from the data.
But since the alignment information is hidden,
	there is a chicken-and-egg problem:
	one requires alignment data to calculate a good model $P(E,A|F)$,
	but alignment data can only be obtained using the model.
	
There is a standard strategy 
	that can be used to solve this kind of chicken-and-egg problem
	called the Expectation Maximization algorithm~\cite{Dempster:1977}.
This algorithm works by first postulating a \naive~initial model.
This model is used to infer a set of alignments.
These alignments are then used to estimate new model parameters.
This process is repeated several times.

Brown~\etal~use a special modified version of the Expectation Maximization algorithm.
Instead of using the same model for each step,
	they actually swap in a new model after a couple of iterations.
The initial models contain a number of simplifying assumptions,
	which are relaxed in the later models.
One example of a simplifying assumption 
	is that each French word corresponds to exactly one English word.
Another such assumption is that
	the probability of a French word in a certain position
	depends only on the English word that it is connected
	to in the alignment.
Obviously these assumptions are not technically correct,
	but the simplified models can still serve as starting points
	in the EM algorithm.
The purpose of using the simpler models first
	is that they provide initial parameter estimates
	for the more complicated models.
For example,
	a good initial estimate of the word-to-word
	translation probability $P(f|e)$ can be obtained
	using the simple initial models.
This technique is useful because the more complex models
	are computationally expensive.
Also, the initial estimates help to ensure
	that the Expectation Maximization algorithm
	finds a good maximum,
	as opposed to a poor local maximum.
	
The idea of alignment, mentioned above,
	is related to another classic problem
	in large scale statistical modeling.
Consider the equation used to 
	calculate the conditional probability $P(F|E)$
	using the alignment-based model:
	
\begin{equation}
P(F|E) = \sum_{A} P(F|A,E)P(A|E) 
\end{equation}

\noindent
In other words, every possible alignment $A$ contributes to the net probability.
However, computing the probability in this way is totally infeasible
	from a computational perspective,
	because there are a vast number of possible alignments.
To grapple with this issue,
	it is necessary to approximate the sum by finding the 
	alignments with the highest possible values of $P(A|E)$
	and only summing over those.

\mpagebreak
	
A much simpler paper,
	but also influential, is ``Minimum Error Rate Training in Statistical Machine Translation'' 
	by Och~\cite{Och:2003}.
This paper describes a way of improving the output 
	of another system.
That is to say,
	it relies on an unspecified black box translation algorithm
	to produce a set of candidate translations.
The method then attempts to find the best of the set
	of candidate translations.
The point is that the initial translation algorithm
	doesn't have to be very good,
	because the reranking process
	will find the best translation from a list of candidates.

The method employs a MaxEnt model,
	but trains the model in a new way.
The standard way of choosing the optimal value
	for a parameter $\lambda_{i}$ in MaxEnt is given by the equation:

\begin{equation*}
\lambda_{i}^{*} = \arg \max_{\lambda_{i}} \sum_{s} \log P_{\lambda_{i}}(E_{s}|F_{s})
\end{equation*}

\noindent
Where $(E_{s}, F_{s})$ is the $s$th translation pair in the corpus
	and $P_{\lambda_{i}}$ is the model given the parameter value $\lambda_{i}$.
This scheme will produce
	a model that assigns high probability to the translation pairs in the corpus.
But it might not produce the translations with the lowest error rate.
To produce low-error translations,
	Och proposes to select parameters using the optimization:
	
\begin{equation*}
\lambda_{i}^{*} = \arg \min_{\lambda_{i}}  
	\big( \sum_{s} G(E_{s}, \hat{E}(F_{s}; \lambda_{i})) \big )
\end{equation*}

\noindent
%TODO - wtf is the difference between a reference translation and just another sentence pair 
% in the corpus
Where $G(\cdot, \cdot)$ is the particular error function being used, 
	and $E_{s}$ is the reference translation for the 
	$s$th sentence pair.
The translation guess $\hat{E}$ is given by:

\begin{equation*}
\hat{E}(F_{s}; \lambda_{i}) = 
	\arg \max_{E \in C_{s}} \big( \sum_{m} \lambda_{m} h_{m}(E|F_{s}) \big )
\end{equation*}

\noindent
Where $h_{m}$ are the context functions,
	and $C_{s}$ is a set of candidate translations of $F_{s}$
	produced by the initial system.
This alternate scheme of finding the $\lambda_{i}$ parameters
	will tend to minimize the error rate,
	instead of maximizing the probability.
	
There are a few issues that come up when using the new optimization scheme.
One issue is that, 
	because of the $\arg \max$ operation,
	the score is not continuous:
	a small change in $\lambda_{i}$ can cause
	a large change in the score.
The latter issue can be solved by using a
	``softened'' version of the $\arg \max$ operation
	which constructs a weighted combination
	of the sentences,
	with weights that depend on the model probabilities $P(E_{s}|F)$.	
	
\mpagebreak

\subsection{Evaluation of Machine Translation Systems}

In the most straightforward procedure for 
	evaluating machine translation results,
	a bilingual human judge assigns a score to an algorithm-generated translation
	by comparing it to the original source sentence.
Unfortunately, 
	this process is far too time-consuming to be carried
	out very frequently.
It also scales badly with the number of candidate systems:
	doubling the number of systems nearly doubles 
	the amount of human time required.
For this reason, the machine translation (MT) community has come to rely on a
	number of automatic scoring functions.
A scoring function 
	assigns a score to a machine-generated translation
	by comparing it to a set of reference translations
	contained in a bilingual corpus.
Once a bilingual corpus
	has been constructed,
	the scoring functions can be used as often as desired.
This provides crucial rapid feedback to MT developers.

The designers of translation scoring functions 
	must face several conceptual challenges.
One basic challenge is that 
	there exist a large number of valid translations
	of any given source sentence.
The scoring function only has access 
	to a small number of valid translations,
	that are contained in the bilingual corpus.
It is very possible for a translation to be good
	but bear little resemblance
	to any of the human translations.
Consider for example the following sentences:

\begin{quote}
H:  The fish tasted delicious.\\
T1: The salmon was excellent. \\
T2: The fish tasted rotten.
\end{quote}

\noindent
In this case,
	the first machine translation T1 should receive a high score,
	even though it shares only one word with the human translation H.
Conversely,
	the T2 translation has three words out of four in common with
	the human translation,
	but it should still receive a low score.	

One of the most widely used translation scoring functions
	is called \textsc{Bleu},
	which is an acronym for Bilingual Evaluation Understudy~\cite{Papineni:2002}.
\textsc{Bleu} uses a scoring metric called \textit{modified n-gram precision}.
The idea is to count how many times a given word sequence from
	the machine translation occurs in the human translations.
So if the human translation is ``John flew to America on a plane'',
	and the machine translation
	is ``John went to America on plane'',
	then the translation will get two unigram hits (John, plane),
	and one trigram hit (to America on).
The designers of \textsc{Bleu} chose to take into account only precision:
	the number of sequences in the machine translation
	that also appear in the human translations.
It does not explicitly penalize a machine translation
	for failing to include words from the human translations
	(i.e. it does not take recall into account).
This is somewhat awkward,
	since the sentence ``John did not commit the crime'' 
	means something very different if the word ``not'' is omitted.
To partially repair this deficiency,
	\textsc{Bleu} penalizes a machine translation 
	if it is substantially shorter than the human version.

The widespread use of \textsc{Bleu} scores
	has strongly influenced the field,
	but has also ignited some controversy.
Various authors have pointed out shortcomings in \textsc{Bleu}~\cite{Turian:2003,Callison:2006}.
One problem with \textsc{Bleu} is that an $n$-gram is scored if it appears in the human translation,
	even if it appears in a very strange position.
Callison-Burch~\etal~give the following example:

\begin{quote}
H: Orejuela appeared calm as he was being led to the American plane 
	that was to carry him to Miami, Florida.
	
T1: Appeared calm | when | he was | taken | to the American plane | ,
	| which will | to Miami, Florida | .
	
T2: which will | he was | , | when | taken | Appeared calm | to the American plane | 
	| to Miami, Florida |. 
\end{quote}
	
\noindent
The first sentence is the human translation,
	and the second two are machine translations,
	with the $n$-gram boundaries shown.
Though the second sentence T1 is not perfect,
	it is clearly better than T2.
But the \textsc{Bleu} score assigns both of these sentences
	the same value,
	because it does not take into account the order
	in which the $n$-grams appear~\cite{Callison:2006}.
Callison-Burch \etal~also noted
	that \textsc{Bleu} sometimes fails to agree with human judgments,
	and cite one case where the system that produced
	the best human-scored translations came in sixth out
	of seven in terms of \textsc{Bleu} score.

Several other translation scoring metrics have appeared
	in the last couple of years.
Snover \etal~developed a metric called \textsc{Ter},
	which is based on a concept called Translation Edit Rate:
	the number of changes an editor would need to make
	to convert the machine translation into one of the human sentences~\cite{Snover:2006}.
This system was designed to be used in conjunction with a human editor,
	who would find the correct translation that is 
	closest to the machine tranlation in terms of \textsc{Ter}.
Impressively,
	the authors show that the human-enhanced \textsc{Ter} score
	correlates better with human assessment of translation quality
	than the assessment of another human.

Another recent automatic scoring function is called \textsc{Meteor},
	developed by Banerjee and Lavie~\cite{Banerjee:2005}.
The \textsc{Meteor} metric uses a more powerful word-matching technique
	than \textsc{Bleu},
	which relies on explicit letter-by-letter equivalence.
\textsc{Meteor} matches words in a translation pair if they have the same root,
	or if they appear to be synonyms
	(a database called WordNet is used for checking synonymy).
\textsc{Meteor} also
	employs a more sophisticated word grouping method 
	than the $n$-gram scheme used in \textsc{Bleu}.
In principle,
	this allows it to repair the problems related
	to $n$-gram rearrangement noted above.

\mpagebreak
	
\subsection{Critical Analysis}

One of the most obvious shortcomings of
	research in machine translation 
	is that most methods do not use any actual linguistic knowledge.
This fact is actually considered to be beneficial in some cases:
	Brown~\etal~note in their abstract that because
	their method uses ``minimal linguistic content'',
	it is easy to port it over to a new language-pair~\cite{Brown:1994}.
This conceptual shortcoming is related 
	to one of the central themes of this book:
	the failure on the part of machine intelligence
	researchers to formulate their work as empirical science.
This failure prevents 
	them from exploiting structure that may be present in a problem,
	and from building effectively on previous work.

This conceptual limitation
	has important implications for machine translation.
As mentioned above,
	one of the key issues in machine translation is alignment:
	connecting words in the source sentence to words
	in the target sentence.
Brown~\etal~show how their alignment scheme works for
	the following pair of sentences:

\begin{quote}
What is the anticipated cost of administering and collecting fees 
	under the new proposal?\\
En vertu de les nouvelles propositions, 
	quel est le cout prevu de administration et de perception 
	de les droits?~\cite{Brown:1994}.
\end{quote}

\noindent
Impressively, 
	the algorithm is able to correctly align the English word ``proposal''
	with the French word ``propositions'',
	even though those words are located in very different positions
	in the two sentences.
This alignment technique is an essential component of the full
	translation system.
But the only reason the alignment problem
	is considered at all is that 
	the algorithms do not depend on any special linguistic knowledge
	or on any previously developed systems.
Researchers in computational linguistics have been working
	on parsers and part-of-speech taggers for a long time,
	but Brown~\etal~chose not to reuse any of that work.
It seems obvious that knowledge of part of speech
	would make the alignment problem 
	dramatically easier,
	if for no other reason than that it would 
	cut down the number of alignments that would need to be considered.
For example,
	in a 20-word sentence,
	there are in principle $20! \approx 2.4 \cdot 10^{18}$ possible alignments.
But if each sentence is known to consist of 5 nouns,
	3 verbs, 5 adjectives, 4 prepositions, and 3 particles,
	then only $5! \cdot 3! \cdot 5! \cdot 4! \cdot 3! \approx 5 \cdot 10^{7}$ 
	alignments need to be checked.
Furthermore, if it were possible to assign semantic roles
	to words such as ``subject'', ``action'', or ``topic'',
	it would be possible to cut down on the number of alignments
	by even more.

The more abstract argument here is that
	it seems unreasonable to attempt to do machine translation
	without extensive background knowledge of 
	both languages in the pair.
To see this point,
	imagine giving a person with no knowledge of 
	English or French a huge corpus of parallel text,
	and asking her to learn,
	on the basis of that corpus,
	how to translate between the two languages.
This seems absurd.
The reasonable approach is to first \textit{train} the person
	in English and French,
	and only then to ask her to learn to translate.
	
It is not hard to criticize the various scores
	such as \textsc{Bleu} and \textsc{Meteor}
	used for evaluation.
These methods seem somewhat arbitrary and \textit{ad hoc},
	and there is extensive controversy in the 
	community regarding their use~\cite{Turian:2003,Callison:2006}.
To justify the automatic metrics
	their developers needed to conduct meta-evaluations
	in which the automatic scores are correlated
	against human scores.
But the correlation evidence is hardly overwhelming proof of 
	the quality of an evaluator.
In some cases the correlation was achieved on the basis 
	of a small number of sentences;
	the \textsc{Meteor} metric was evaluated
	on the basis of 920 Chinese sentences and 664 Arabic sentences~\cite{Banerjee:2005}.
It could be that the scores 
	will fail to correspond with human judgment 
	when tested on larger databases, 
	or on text that is written in a different style.
Furthermore,
	the mere fact that an automatic scoring metric correlates
	with human judgment is not necessarily proof of its quality,
	since the correlation of human scores with one another
	is actually quite low~\cite{Turian:2003}.
	
Even if an initial evaluation shows that an automatic scoring metric
	correlates well with human scores,
	that correlation may evaporate over time.
This point was mentioned in Chapter~\ref{chapt:comprvsion},
	in connection with the idea of Goodhart's Law,
	which says that if a statistical regularity
	is used for control purposes, it will disappear.
Perhaps when they originally appeared,
	scores like \textsc{Bleu} and \textsc{Meteor}
	correlated well with human judgment.
But as researchers begin to use the metric	
	to design their systems,
	this pressure may destroy the correlation.
This problem is particularly relevant
	because many researchers now use
	techniques such as Minimum Error Rate Training
	to optimize their systems with respect
	to a particular scoring metric~\cite{Och:2003}.

In fairness to the developers of \textsc{Bleu} and other scoring metrics,
	these scores were not intended to fully replace
	human judgment for the machine translation task.
The community periodically holds contests or workshops
	in which machine translation results are evaluated
	by a panel of human judges.
This method cannot be criticized except for the fact
	that it is expensive and labor-intensive.
In particular,
	the human evaluation strategy scales badly with
	the number of candidate systems.
If there are $N$ sentences and $T$ systems,
	then the time requirement is proportional to $N \cdot T$.
This means the number of sentences used must be kept low,
	which may make the evaluations sensitive to random fluctuations:
	perhaps some systems work better on certain types of sentences,
	and by chance that type of sentence is over-represented in the small evaluation set.

\mpagebreak

\subsection{\Compist~Formulation of Machine Translation}

The \compist~philosophy suggests a clean and direct
	formulation of the machine translation problem:
	apply the Compression Rate Method to a large bilingual corpus.
A specialized compressor
	can exploit the relationship between the sentences
	in the translation pair to save bits.
In statistical terms,
	researchers will need to obtain a good model of $P(E|F)$ - 
	the probability of an English sentence given
	its French counterpart -
	in order to achieve good compression.
Of course, obtaining such models is not at all an easy problem,
	and involves a wide range of research questions. 

Once strong models of the form $P(E|F)$ or $P(F|E)$ are obtained,
	there are two ways to use them to generate actual translations.
The first method is simply to sample from the 
	$P(E|F)$ distribution.
The idea here is that if the model $P(E|F)$ is good,
	then samples of $P(E|F)$ will be veridical or realistic
	simulations of what a human translator would produce.
The second method of generating translations 
	is to apply the Fundamental Equation of Machine Translation:

\begin{equation*}
E^{*} = \arg \max_{E \in \cal{E}} P(F|E) P(E)
\end{equation*}
	
\noindent
This method is more difficult than sampling,
	because the optimizatiion over the huge space
	$\cal{E}$ may be hard to carry out.
The advantage of this method is that the use of the $P(E)$ model
	term can mitigate deficiencies in the $P(E|F)$ model~\cite{Brown:1994}.
	
The presence of the standalone model $P(E)$
	in the above equation can be seen as 
	simplistic evidence in favor of the Reusability Hypothesis.
\Compist~researchers
	studying a large database of raw English text
	will produce good models $P(E)$.
Improvements in the standalone models
	will produce immediate improvements in machine translation results.
While the importance of using a good model $P(E)$ 
	has been noticed in the MT literature,
	in practice this step seems to be overlooked.
Here is an example translation produced by Google Translate:

\begin{quote}
If locally heavy snow fell in a short time, 
	even life and logistics support under the direct control highway 
	was closed to traffic early on that country. 
To prevent congestion and stuck it in a hurry to snow removal vehicles. 
MLIT has blocked the national highway has been avoiding the highway, 
	unlike direct control, we changed course.
\end{quote}

\noindent
Evidently Google Translate does not use a very good model of $P(E)$.
Other automatic translation services routinely produce
	comparably unintelligible outputs.
Note that if $P(E)$ was highly sophisticated
	but $P(F|E)$ were substandard,
	in order to tell if a translation was good or not 
	a bilingual examiner would have to compare the meaning
	of the English output with the foreign original.
	
In addition to the simple reason given above,
	there is actually another reason why 
	it will be useful to learn standalone models for machine translation.
This is the fact that
	background knowledge of English and French
	will be highly valuable when attempting to do translation.
The idea here is similar to the point of the Japanese Quiz 
	thought experiment of Chapter~\ref{chapt:comprlearn}.
That example
	showed that it was impossible to construct good models
	for the category of a sentence 
	without extensive background knowledge of Japanese.
Analogously,
	in the context of machine translation,
	it may be impossible to construct good models
	of an English translation sentence
	without extensive background knowledge of French.
This idea is highly intuitive.
The current procedure of machine translation 
	is a bit like taking a large bilingual corpus of 
	English and French,
	giving it to a person with no knowledge of those languages,
	and asking her to learn to translate based solely on the corpus.	
A much more effective strategy would be for the person 
	to first learn both English and French,
	and then to polish her translation skills by 
	studying the bilingual corpus.

These arguments suggest an even better version of the problem formulation
	than the one mentioned above.
This is to package the bilingual corpus
	together with a raw English corpus 
	and a raw French corpus,
	and use the aggregate dataset as a target of a \textsc{CRM} investigation.
This will require both a model $P(E|F)$ 
	and standalone models $P(F)$ and $P(E)$,
	which can be reused to produce good translations
	with the optimization mentioned above.	
Computational tools and statistical models
	can be justified on the basis of the large raw corpora,
	and then immediately redeployed to attack
	the conditional modeling problem.
In concrete terms,
	the claim here is that knowledge of issues such as parsing,
	POS tagging, semantic role analysis, \textit{etc}, 
	will be useful not only for raw text compression
	but for conditional compression (and thus translation) as well.

As a metric for evaluating machine translation systems,
	the compression rate metric compares 
	very favorably to the other metrics
	such as \textsc{Bleu} and \textsc{Meteor}
	proposed in the literature.
As noted above,
	these metrics have serious shortcomings
	and have generated substantial controversy.
\textsc{Bleu} can be gamed,
	in the sense that very poor translations can be 
	found that will nonetheless receive high \textsc{Bleu} scores.
In contrast the compression metric is,
	in some sense, above criticism.
Certainly it cannot be gamed.
It is ultimately based on the same principles of
	prediction and falsification that drive traditional empirical science.
In order to criticize the use 
	of the conditional compression rate metric,
	a critic would need to show that it is possible for a system
	to achieve a strong compression score,
	but still produce poor translations.
Since a system that achieves a good compression rate must have 
	a good model of $P(E)$ and $P(F|E)$,
	this would effectively imply that 
	the Fundamental Equation of Machine Translation (Eq.~\ref{eq:machntrans})
	does not work.
Finally, the compression metric is more interesting:
	\compist~researchers can justify their work on the basis of pure intellectual curiousity,
	but no one would ever try to optimize \textsc{Bleu} scores
	if doing so did not supposedly produce good machine translations.
	
\mpagebreak	
	
\section{Statistical Language Modeling}
\label{sec:statlangmd}

Statistical language modeling is
	a subfield of computational linguistics 
	where the goal is to build statistical models
	of the probability of a sentence, $P(s)$
	(denoted $P(E)$ in the discussion of machine translation).
This is almost always done by breaking the problem up into a series of word probabilities:

\begin{equation*}
P(s) = \prod_{i} P(w_{i} | w_{i-1} \ldots w_{0})
\end{equation*}

\noindent
Where $w_{i}$ is the $i$th word in the sentence.
These models are trained by analyzing the
	statistics of a large text corpus.
Language modelers typically attempt to minimize the cross-entropy:

\begin{equation*}
\sum_{s} P^{*}(s) \log P(s) \approx \sum_{k} \log P(s_{k})
\end{equation*}

\noindent
Where $P^{*}(s)$ is the \textit{real} probability of a sentence.
The expression on the left is the actual cross entropy,
	and the sum is over all possible sentences.
This quantity can never actually be computed,
	not only because performing the sum is computationally infeasible,
	but also because the real distribution $P^{*}(s)$ is unknown.
The right hand side is an empirical estimate of the cross entropy,
	and it involves a sum over all the sentences $s_{k}$ in the corpus.
This empirical cross-entropy is exactly equivalent to the compression rate
	except that it does not involve a model complexity term.
As discussed below, this non-incorporation of a model complexity penalty has important ramifications;
	many language models use a large number of parameters 
	and are therefore probably overfitting the data.
Overall, though, language modeling is the area of current research
	that is most similar to the \compist~proposal.
	
As noted in Section~\ref{sec:statmachtrn},
	a good language model is an important component 
	of many approaches to machine translation,
	due to the role $P(E)$ 
	plays in the Fundamental Equation of Machine Translation.
It turns out that 
	an entirely analogous formulation works for speech recognition:
	
\begin{equation*}
E^{*} = \arg \max_{E \in \cal{E}} P(A|E) P(E)
\end{equation*}

\noindent
Where $E^{*}$ is the guess of the spoken English words,
	and $A$ is the audio data measured by the system.
The only difference between this equation and the one
	used for machine translation is that here
	the audio data $A$ takes the place of the French sentence.
In fact,
	the connection to speech recognition 
	the primary motivation for language modeling research.
Most papers on the topic describe a new language modeling technique,	
	show how it can reduce the cross-entropy,
	and in the final section demonstrate the reductions in word error rate
	that can be achieved by connecting the model
	to a speech recognizer.
	
By far the simplest,
	and the most widely used,
	approach to language modeling
	is the $n$-gram.
The conditional probabilities in a $n$-gram model are 
	obtained by simple counting; 
	for example, in a trigram model, the probabilities are:

\begin{equation}
\label{eq:ngrammodel}
P(w_{i}|w_{i-1}, w_{i-2}) = \frac{Count(w_{i}, w_{i-1}, w_{i-2})}{Count(w_{i-1}, w_{i-2})}
\end{equation}

\noindent
Where $Count(\cdot)$ denotes the count of a particular event in the training data.
The most common values of $n$ are 2 (bigram) and 3 (trigram).
Language modelers are often frustrated by the effectiveness of $n$-grams,
	as they find it difficult to do better by using
	more sophisticated techniques.
	
An early paper on statistical language modeling,
	and important influence on this book,
	is ``A Maximum Entropy Approach to Adaptive Statistical Language Modeling''
	by Ronald Rosenfeld~\keycite{Rosenfeld:1996}.
Rosenfeld begins by identifying a number
	of potentially useful information sources (or predictors)
	in the language stream.
The most basic predictors are based on $n$-gram models.
A more interesting type of predictor is a \textit{trigger pair}.
A pair of words $C$ and $D$ form a trigger pair $C \rightarrow D$
	when observing $C$ makes it more likely to observe $D$.
For example, if the word history contains the word ``stock'',
	it is much more likely than normal
	that the subsequent text will include the word ``bond''.
Trigger pairs can be automatically extracted from the corpus.
Another interesting technique is called long distance $n$-grams.
These are like $n$-grams except that they match against
	words farther back in the history.
Rosenfeld also experiments with a method for using word 
	stems to augment the trigger pair technique.
The idea here is that if the word ``film'' triggers
	the word ``movie'', 
	then the word ``filming'' probably does as well.
	
Rosenfeld's basic strategy is to write down many predictors,
	and use algorithmic techniques to combine them together
	to achieve the best possible model.
The issue is how exactly to do the combination.	
Most \naive~strategies for information integration,
	such as using a linear combination of the predictions,
	lack theoretical justification and 
	tend to underperform in practice.
Rosenfeld adopts the Maximum Entropy approach
	to solve this problem.
In this context,
	the MaxEnt model is of the following form:

\begin{equation*}
P(w| H) = \frac{1}{Z(H)} \exp (\sum_{i} \lambda_{i} f_{i}(w,H))
\end{equation*}

\noindent
Where the $f_{i}(w, H)$ are a set of context functions, 
	that operate on the history $H$ and the new word $w$,
	and $Z(H)$ is a quantity that ensures normalization.
As noted above,
	the power of the MaxEnt idea is that it allows 
	the user great flexibility in defining whatever
	kinds of context functions he may consider useful.
For example,
	one useful predictor might be a function that returns 1
	if $w=\mathrm{then}$ and the word \textit{if} appears in the history,
	and 0 otherwise.
The training algorithm for MaxEnt, 
	called Generalized Iterative Scaling,
	automatically finds the optimal parameters $\lambda_{i}$
	corresponding to each context function
	that maximizes the likelihood of the data.
An important feature is that there is no significant penalty 
	for trying out new predictors:
	if a predictor has no value, 
	the training algorithm will simply assign it a near-0 $\lambda$ value
	so that it has no effect on the actual probability.
	
Rosenfeld's paper influenced the development of the ideas in this book
	because his approach seemed like a kind of empirical science.	
His procedure 
	was to propose a new class of context functions,
	incorporate them into the MaxEnt model,
	and measure the resulting cross-entropy score.
If the new functions produced a reduction in cross-entropy,
	it meant that they captured some real empirical 
	regularity in the structure of the text.
In other words,
	by using MaxEnt along with the cross-entropy metric,
	Rosenfeld was able to \textit{probe the structure} of text
	by experimenting with different kinds of context functions.
	
Unfortunately, Rosenfeld did not carry this procedure
	as far as he could have.
Two reasons may have limited his motivation to do so.
First, new techniques tend to ``cannibalize'' one another:
	sometimes adding a new predictor is effective,
	but reduces the effectiveness of some other predictor,
	because the they both exploit the same underlying regularity.
Second, like many other language modelers,
	Rosenfeld's ultimate goal was to improve the performance
	of a speech recognition application.
At the end of the paper, 
	he notes that improvements in the cross-entropy score 
	tend to yield diminishing returns in terms of reductions in 
	word error rates for speech recognition.
These two effects, combined together,
	probably sapped his motivation to carry out a more systematic
	search for the best possible language model.

\mpagebreak

A more recent paper in this area is 
	``A Bit of Progress in Language Modeling'' by Joshua Goodman~\cite{Goodman:2001}.
Most language modeling papers present new techniques for language modeling.
Goodman's paper is different in that it describes a 
	systematic comparison and evaluation of an array of competing techniques.
It also examines how different modeling tools can be used in conjunction
	with one another.
	
Goodman starts off by showing that one of the most important components
	of an $n$-gram language model is a good 
	smoothing method.
Smoothing is important because of sparse data problems,
	which become increasingly severe as the value of $n$ becomes large,
	making the counts in Equation~\ref{eq:ngrammodel} unreliable.
To understand the sparse data issue,
	notice that if the number of words in the dictionary is $T=5\cdot10^{4}$, 
	then there are $T^{3} \approx 1.5 \cdot 10^{13}$ possible three-word sequences.
This implies that most three word sequences
	will never be observed in the corpus,
	so $C(w_{i}, w_{i-1}, w_{i-2})=0$.
This can be true even if a sequence has relatively high probability
	(say, $p \approx T^{-3}$).
Equation~\ref{eq:ngrammodel} would assign such a sequence zero probability,
	but clearly this behavior must be avoided.
To deal with this situation,
	language modelers use smoothing techniques.
One smoothing technique is to ``back-off'' in cases where a sequence
	has never been observed,
	and use a lower-$n$ model instead.
Goodman describes several other methods,
	such as Katz smoothing and 
	Kneser-Ney smoothing~\cite{Katz:1987,Kneser:1995}.

Goodman also describes
	a variety of language modeling techniques
	that work because of specific properties of text.
One such technique is called a skipping model.
A skipping model is like an $n$-gram,
	except that it skips a word.
So for example the model might be based on $P(w_{i} | w_{i-2}, w_{i-3})$,
	ignoring the influence of the immediately
	preceding word $w_{i-1}$.
This method works because of 
	phrases like ``went to John's party''.
If the pattern ``went to X party'' has been observed
	several times in the corpus,
	then a skipping model
	will assign a high probability to the word ``party''
	in the phrase ``went to Sarah's party'',
	even if that exact phrase has never been observed.

A second technique that works because it 
	reflects the real structure of text
	is called clustering.
The idea here is to model	

\begin{equation*}
P(w_{i} | w_{i-1} w_{i-2}) = P(w_{i} | W_{i}) P(W_{i} | W_{i-1} W_{i-2})
\end{equation*}

\noindent
Here $W_{i},W_{i-1},W_{i-2}$ are word \textit{categories}
	instead of specific words.
So the probabilistic effect of a word in the history
	is mediated through its category.
This also means that $w_{i}$ is conditionally independent
	of $w_{i-1}$ given $W_{i-1}$.
In theory,
	this technique can assign high probability
	to the phrase ``Jim flew to Canada'',
	even if only very approximately similar phrases
	such as ``Bob drove to Spain''
	have been observed in the corpus.
Furthermore, this technique can provide a decisive
	defense against sparse data problems,
	because there are far fewer word categories than words.
If there are $10^{2}$ categories,
	then a trigram-style model using categories
	will be only on the order of $10^{6}$.
The main challenge in using this technique is in obtaining
	the word categories;
	a variety of research deals with this problem~\cite{Brown:1992}.
%TODO check this claim.
Goodman's evaluation reports that the best-performing system
	is a model of this type. 

The extended version of Goodman's paper ends
	on a very pessimistic note:
	one section is entitled ``Abandon hope, ye who enter here''.
Goodman's frustration 
	is due to the difficulty of
	finding techniques that offer better practical 
	performance than $n$-gram models on the speech recognition task.
Techniques that achieve lower cross-entropy
	usually require far more computational and memory resources.
Furthermore,
	even if a technique achieves an improved cross-entropy,
	it often yields a relatively smaller reduction
	in word error rate.
% TODO: what about graph of word-error vs cross-entropy??
Goodman also complains that there is very little progress in the area,
	because people only compare their models to the trigram,
	not to other published research.
This makes it difficult to determine which 
	techniques actually represent the state of the art.

\mpagebreak
	
\subsection{Comparison of Approaches to Language Modeling}

Reserch in statistical language modeling
	is very similar to a \compist~inquiry
	targeted at a large text corpus.
The similarities between the two types of research
	offer ammunition to both critics and proponents
	of the \compist~philosophy.
Proponents can note that
	language modeling research has already demonstrated
	the basic validity of the Reusability Hypothesis:
	improvements in language models
	lead directly to reductions in the error rate of speech recognition systems.
On the other hand, 
	critics of the \compist~philosophy
	can cite the fact that language modeling
	has not produced spectacular results or dramatic progress.
Indeed, 
	language modelers often express frustration,
	noting that error rate reductions 
	are very difficult to achieve in practice.
If language modeling and similar lines of research
	are really going to produce important breakthroughs,
	why haven't any of those breakthroughs already been achieved?
	
While traditional and \compist~language modelers
	may appear to be working on the same kind of problem,
	they are guided by a different \textit{philosophy}.
To begin the comparison of the two mindsets,
	consider the following comment by Rosenfeld:	
	
\begin{quote}
Ultimately, the quality of a language model must be measured by its 
	effect on the specific application for which it was designed, 
	namely by its effect on the error rate of 
	that application~\cite{Rosenfeld:2000}.
\end{quote}

\noindent
Rosenfeld's comment illustrates what can be called the 
	toolbox mindset of language modeling.
The goal here is to create tools
	that produce performance improvements when
	connected to applications such as speech recognizers
	or machine translations systems.
This mindset has an important implied belief:
	there is no such thing as a uniquely optimal language model.
Instead, 
	there are a wide variety of tools 
	that are useful in different situations.
Perhaps tool A works best for speech recognition,
	while tool B is well suited to machine translation,
	and tool C can be used for information retrieval.
The goal, therefore, 
	is to create a toolbox that is as large and diverse as possible.
A researcher creates a new tool starting more or less from scratch,
	and validates his result by comparing it to
	the standard $n$-gram model.
If a tool is novel and interesting,
	and it performs well compared to the $n$-gram,
	then it must be a worthwhile contribution to the toolbox.
Few researchers attempt 
	to systematically compare the various tools, or 
	to improve an already existing tool.
No one wants to spend years of effort
	trying to refine someone else's tool
	to produce a minor performance improvement.
In the light of this diagnosis,
	it is worth noting that one of the most-cited papers
	in the subfield is entitled 
	``SRILM - an Extensible Language Modeling Toolkit''~\cite{Stolcke:2002}.

Empirical scientists completely reject all the views
	associated with the toolbox mentality.
Physicists believe in the existence 
	of a singular optimal theory of empirical reality.
They view their work as a systematic search for this theory,
	and because they can make decisive theory-comparisons,
	the search can proceed rapidly.
Physicists are not intrinsically interested in applications,
	and evaluate theories without regard for practical utility.
The same basic theory of physics is used by all engineers,
	regardless of whether they are constructing airplanes,
	bridges, skyscrapers, or automobiles.
A physicist who attempted to justify an alternative theory of mechanics
	by arguing that it would be more useful for bridge-building
	would be ridiculed and ignored, even by bridge-builders.
In physics it is extremely important and valuable
	to provide a modest improvement or refinement
	to the champion theory - 
	for example, by showing how quantum mechanics can be used
	to explain superconductivity.

\Compist~researchers adopt the mindset of physics
	in their approach to language modeling.
They reject the toolbox mentality,
	and believe in the existence of a single optimal 
	theory for describing a given text corpus.
They view their work
	as a systematic search for that optimal theory,
	and carry out the search
	by conducting decisive theory-comparisons.
They recognize the value of minor extensions or refinements
	of the champion theory,
	if those refinements produce codelength reductions.
They adopt the Circularity Commitment,
	and believe the Reusability Hypothesis
	promises that their highly refined theories
	will have practical applications.
	
A good example of how this difference in mindset
	plays out in practice relates to the 
	empirical regularities mentioned by Goodman.
These regularities allow models based 
	on techniques such as caching, skipping, and word categorization
	to achieve reductions in the cross-entropy.
Traditional language modelers notice and sometimes exploit these regularities,
	but do not attempt to systematically document and characterize them.
This is at least partially because,
	if the regularity is uncommon or difficult to describe,
	the amount of work required to exploit it
	is not justified by the minor potential improvement in speech recognition.
In contrast,	
	\compist~language modelers view these regularities
	as the focus of their research.
How can the regularities be categorized, organized, and modeled?
Which regularities represent independent phenomena,
	and which are manifestations of the same underlying effect?
Are the regularities present in English basically the same
	as those present in Japanese,
	or do they differ in some fundamental way?
To the \compist~language researcher,
	all of these questions are deeply interesting.

In addition
	to the change of mindset described above,
	there is actually an important technical difference
	between the two formulations.
The cross-entropy is actually not 
	equivalent to the compression rate,
	because it does not include a model complexity term.
This fact goes far in explaining the 
	frustrating apparent superiority of the $n$-gram methods.
These methods,
	though conceptually simple,
	are in fact inductively (or parametrically) complex.
If a corpus contains $T$ words,
	then the bigram model requires $T^2$ parameters to specify
	while the trigram model requires $T^3$.
If $T=10^5$, then the bigram model
	uses ten billion ($10^{10}$) parameters and 
	the trigram uses a million billion ($10^{15}$) parameters.
This indicates that while the $n$-gram methods may
	provide good performance using the cross-entropy score,
	they will not do well according to the complexity-penalized
	compression rate score.
The parametric complexity of $n$-gram methods
	also contributes to the chronic sparse data problems
	faced by language modelers.
Because of the large number of parameters used by $n$-gram models,
	the estimates of the parameter values
	will be unreliable unless there is a huge quantity of data.
This leads language modelers
	to focus a substantial amount of attention
	on smoothing techniques~\cite{Chen:1999}.
Smoothing may be interesting from a mathematical perspective,
	but if the goal is to describe the structure of text,
	it is mostly a distraction.
In contrast,
	\compist~researchers will need to find models with fewer parameters
	in order to achieve compression.
The use of more parsimonious models
	will naturally prevent sparse data problems,
	and also reveal interesting facets
	of the structure of language.

\mpagebreak
	
\section{Additional Remarks}
\label{sec:addtremark}

The discussion above treated three important areas 
	of computational linguistics,
	and showed how those areas can be reformulated
	as large scale text compression problems.
In this view,
	there is no meaningful distinction between
	the three areas.
Parsing is an integral part of language modeling,
	and both are necessary for machine translation.
Similar reformulations can also be established for 
	other areas of computational linguistics.

For example,
	document classification can be justified by
	adding document-level components to the language model.
Consider a newspaper article with the headline
	``Red Sox Win Game Three of Playoffs''.
A smart compressor should be able to guess
	from the headline that the article is about baseball.
When it comes to the text of the article,
	the compressor should assign higher probability
	to words like ``games'', ``Yankees'', and ``home run''.
Thus,
	document classification research can be justified
	by the compression principle.
Some reflection will reveal that several
	other standard tasks of computational linguistics,
	such as word sense disambiguation~\cite{Navigli:2009} 
	and semantic role analysis~\cite{Gildea:2002},
	can be reformulated as specialized text compression techniques.

This chapter has briefly criticized some aspects
	of the philosophical foundations of computational linguistics.
In general,
	many of the same conceptual failures
	described in relation to computer vision
	also exist in computational linguistics.
Research is not formulated as empirical science;
	the computer science mindset has an unhealthy influence.
Researchers attempt to build systems 
	that replicate the perceptual process of humans,
	but those perceptual processes and abstractions
	are too loosely defined.
The problem of evaluation is not well thought out;
	evaluation schemes are risky and time-consuming to develop,
	and may actually not work very well.
There is no parsimonious justification for research topics,
	allowing researchers too much liberty to 
	define their own problems.
	
A good example of the problem of esoteric 
	justification is the newly-popular task
	of word sense disambiguation (WSD).
One aspect of English and other languages
	which is vexing for language processing systems
	is that sometimes the same word
	can have different meanings in different contexts.
An example is the word ``bank'',
	which may signify a financial institution
	or the edge of a river.
WSD is the problem of determining which meaning
	is actually being used in a given sentence.
In a recent survey of the WSD subfield,
	Roberto Navigli mentioned a variety of problems 
	with the task~\cite{Navigli:2009}.
First, authors do not agree on what the fundamental definition of WSD,
	and this leads to different and incompatible formalizations.
Second, the WSD task relies heavily on background knowledge,
	but constructing knowledge databases
	is expensive and time-consuming.
According to Navigli,
	the difficulty of the task is evidenced by the fact
	that it has not been applied to any real-world tasks.
Given these conceptual difficulties,
	it is hard to understand why anyone would bother studying the problem.
The reason is probably sociological: WSD is \textit{new}.
Research on topics like parsing and machine translation
	has started to yield diminishing returns
	(because of conceptual flaws in the formulation of those tasks), 
	so the field's attention has wandered onto a new topic.

\mpagebreak

\subsection{Chomskyan Formulation of Linguistics}

In his famous early book ``Syntactic Structures'',
	Chomsky formalized the problem of linguistics in the following way:

\begin{quote}
The fundamental aim in the linguistic analysis of a language L
	is to separate to \textit{grammatical} sequences
	which are the sentences of L
	from the \textit{ungrammatical} sequences 
	which are not sentences of L
	and to study the structure of the grammatical sequences.
The grammar of L will thus be a device that generates 
	all of the grammatical sequences of L and 
	none of the ungrammatical ones.
One way to test the adequacy of a grammar 
	proposed for L is to determine whether or not the
	sequences that it generates are actually grammatical,
	i.e., acceptable to a native speaker, etc.~\cite{Chomsky:1957}
\end{quote}
	
\noindent
This passage describes the \textit{generative} approach 
	to the study of grammar,
	because it aims at constructing a ``device that generates'' grammatical sentences. 
This sentence-generating device must be very sophisticated,
	because the grammatical rules of natural language are complex.
The construction of such a device will therefore
	require a wide range of linguistic investigations.
Chomsky's definition is elegant because it simultaneously satisfies
	two important criteria:
	it is relatively concrete, 
	and it provides a parsimonious justification for many research
	topics in linguistics.
	
\Compist~language researchers aim at a goal
	that is only slightly different from the one given by Chomsky. 
To achieve good compression,
	researchers must construct models $P(s)$ of the probability of a sentence,
	that assigns low probability to rare sequences
	and high probability to common sequences.
If a text corpus is constructed by compiling
	a number of legitimate publications 
	such as books and newspaper articles,
	then ungrammatical sequences should be quite rare.
So by identifying the rules of grammar,
	it should be possible to save bits by reserving
	short codes for the grammatical sentences.

Furthermore, Chomsky's method of testing a proposed grammar,
	by generating sentences and showing them to a native speaker
	bears a striking resemblance to the veridical 
	simulation principle of Chapter~\ref{chapt:compmethod}.	
A \compist~researcher can generate sentences from the model $P(s)$
	by feeding random bit strings into the decoder.
The simulation principle asserts that,
	if the compressor is very good, 
	then the sentences generated in this way will be 
	legitimate English sentences.
If the sentences fail to observe some grammatical rule,
	then this indicates a deficiency in the model;
	if the researcher can correct this deficiency,
	the new model should achieve a better codelength.
The decoder component of a very good compressor
	is therefore very nearly equivalent to a 
	``device that generates'' the grammatical sentences 
	of a language.

Despite the close similarity between Chomsky's stated goal 
	and the \compist~formulation of linguistics,
	Chomsky rejects the applicability of statistical methods
	to syntactic research.
To justify this position,
	Chomsky posed the famous sentence
	``Colourless green ideas sleep furiously''.
This sentence is exceedingly strange,
	and one would expect that it should never occur in any normal
	corpus of English text.
However, the sentence is perfectly acceptable
	from a grammatical point of view.
To Chomsky,
	this implies that statistical methods,
	which rely on estimating the frequency of events,
	cannot be used to discover a theory of syntax.
Statistical methods,
	by equating ``probable'' with ``grammatical'' and vice versa,
	will fail to correctly assess the grammaticality of the green ideas sentence.

The flaw in Chomsky's argument can be seen by transplanting it
	into another domain.
Consider the event evoked by the sentence:
	``A sword fell out of the sky''.
This event is exceptionally rare,
	to the extent that most humans have never observed it,
	except perhaps in movies.
However, 
	the event is perfectly allowable
	according to the laws of physics.
It could certainly occur,
	if for example someone threw a sword off the roof
	of a tall building.	
Transplanting Chomsky's argument to deal with this event,
	one would conclude that statistical analysis of observed events
	cannot be used to infer the laws of physics.
% TODO
But this is obviously incorrect;
	many crucial physical insights were obtained through observation.

To analyze this point further,
	consider the following procedure.
A \compist~researcher constructs a model $P(s)$ of the probability
	of a sentence, 
	using some basic scheme such as $n$-grams.
The researcher then constructs a new model $P'(s)$ by revising $P(s)$
	in the following way.
In $P'(s)$, all of the sentences that do not contain a verb
	are assigned very low probability.
The probability freed up in this way is reassigned to
	the other sentences.
Because almost all English sentences have verbs,
	the model $P'(s)$ will achieve a substantially lower
	codelength than the model $P(s)$
	when used to compress a text database.
There is no reason to believe this principle cannot
	be scaled up to include new and more complex rules of grammar.
This argument shows that \compist~language research \textit{includes}
	syntactic investigations,
	though it also includes other kinds of inquiry as well.
The full extent of the overlap between
	\compist~language research and traditional linguistics
	cannot be known until such research is far advanced.
One possibility that might cause a schism between the two areas
	is if there are many grammatical structures 
	that humans will accept as legitimate,
	but will never actually use in writing (or speech).
\Compist~methods will fail to identify such structures as legitimate.

This concern is not very discouraging.
It is not obvious why the definition of grammaticality 
	as ``acceptable to a native speaker''
	should be superior to the definition as ``used in practice''.
The latter definition
	is probably far more relevant for practical purposes
	such as machine translation and information retrieval.
Overall,
	the fact that \compist~linguistics aligns strongly 
	with traditional linguistics,
	while also bringing a host of conceptual and methodological advantages,
	provides more than adequate motivation for interest in it.	
	
As a final point,
	it seems very probable that in a real compressor,
	making a distinction between syntactic and semantic
	modeling will be very useful from a 
	software engineering point of view.
Probably it will make sense to decompose the compressor
	into a series of modules.
One module will save bits by reserving short 
	codelengths for grammatical sentences.
Then a secondary module saves more bits
	by exploiting observations such as the fact 
	that ``ideas'' rarely have colors.

\mpagebreak

\subsection{Prediction of Progress}

Some critics may dismiss the philosophy of this book
	as intangible and unverifiable.
The following concrete prediction
	should counter this criticism
	and help to clarify the beliefs implicit
	in the \compist~philosophy.

Consider some small community that exists
	on the margins of mainstream linguistics research.
This community dedicates all of its efforts to the 
	problem of large scale compression of English text.
The researchers develop an extensive set of tools and methodologies to attack this problem.
They make repeated decisive comparisons between
	candidate theories.
The champion theory grows increasingly refined and complex,
	but this complexity is justified by improved descriptive accuracy.
This theory incorporates a number of abstractions
	related to the structure of text,
	including but not limited to grammatical rules.
The prediction, then, is as follows.
Once this strong champion theory is obtained,
	the \compist~researchers will reuse it
	to solve the standard problems of traditional linguistics
	such as parsing, machine translation,
	and word sense disambiguation.
At that time,
	the results achieved by the \compist~researchers
	will be dramatically superior to those achieved by more traditional approaches.
	
This prediction is just a restatement of the Reusability Hypothesis.
The history of science indicates
	that the Reusability Hypothesis holds for fields 
	like physics and chemistry.
There are reasons to believe the hypothesis will hold for \compist science as well,
	but as yet no concrete evidence.
If the prediction turns out to be true,
	it will vindicate the Reusability Hypothesis,
	and by extension the \compist~philosophy.
If, on the other hand,
	the champion theory proves to be useless for any other purpose,
	then the \compist~idea should be abandoned.

\chapter{Compression as Paradigm}
\label{chapt:crmparadgm}

\section{Scientific Paradigms}

Thomas Kuhn, in his famous book ``The Structure of Scientific Revolutions'',
	identified several abstractions that are very useful
	in describing the patterns of scientific history~\keycite{Kuhn:1970}.
Kuhn began by analyzing what most scientists do 
	in their day-to-day activities.
Most of the time,
	a typical scientist is not attempting
	to overthrow the cornerstone theory of his field.
Instead, he is attempting to articulate or enlarge the theory,
	by showing how it can be applied to a new situation,
	or how it can be used to explain a previously mysterious phenomenon.
Kuhn called this kind of incremental research ``normal science''.
A key property of normal science is that it is cumulative:
	while individuals may fail,
	once someone solves a problem,
	it is solved for all time and the field can move on.
Because of this cumulative effect,
	and because there are a large number of scientists,
	normal science can produce rapid progress.
However, several requirements must be met before
	normal science can take place.	
It is necessary for the scientific community to agree
	on the importance of various questions,
	and on the legitimacy of the solution methods used to answer them.
In other words, the community must be bound together 
	by a shared set of philosophical commitments and technical knowledge.

Kuhn calls this unifying body of conceptual apparatus a scientific paradigm.
In spite of their crucial importance,
	it is rare for a paradigm to be explicitly articulated.
This is because a paradigm is in many ways a set of 
	abstract philosophical commitments, 
	and scientists want to do science, not philosophy.
Instead of explicit instruction, 
	young scientists learn a paradigm by 
	studying exemplary research within a field, 
	and attempting to assimilate the rules and assumptions guiding the research.
Because the necessity of a paradigm is not always acknowledged,
	people often attempt to do science without one.
This leads to incoherence, 
	even if the individuals involved are disciplined and intelligent scientists.

Paradigms are never perfect, and sometimes collapse.
This event, called a scientific revolution,
	occurs when the inconsistencies and ambiguities 
	previously tolerated by the paradigm 
	start to become so glaring and burdensome as to prevent further progress.
As the paradigm breaks down, more and more researchers abandon it,
	and go in search of a replacement, or leave science altogether.
Scientific revolutions tend to be traumatic,
	because normally reasonable academic discourse
	becomes acrimonious and intractable.
The scientific community will splinter into opposing camps
	defined by new and mutually incompatible philosophical commitments.
Since the discarded paradigm defined the rules of scientific debate,
	there is no reason why a proponent of one camp should 
	accept the arguments made by another camp.
Members of communities defined by opposing paradigms
	might not even agree on what research questions
	are legitimate topics within a given field.
Thus, paradigm conflicts must be settled by methods that are essentially extrascientific.
Perhaps one paradigm leads to more practical applications,
	or perhaps it is simply more appealing to the younger generation of researchers.
In any event,
	further progress is impossible 
	until a new scientific community coalesces around a replacement paradigm.

\mpagebreak 

\subsection{Requirements of Scientific Paradigms}

In order to fulfill their purpose of providing coherence 
	to scientific community and enabling normal science,
	scientific paradigms must possess certain properties.
The most basic function of a paradigm 
	is that it serves as a body of shared background knowledge.
This knowledge is set forth in the textbooks used by students,
	and any practitioner in the field can be expected
	to be familiar with it.
This is crucial,
	because it means that researchers within a paradigm do not 
	have to waste time explaining basic ideas to one another.
	
A paradigm ensures that researchers share a core of technical knowledge,
	but perhaps more importantly,
	it ensures that they share a set of philosophical commitments.
A paradigm must give clear answers to the \textit{meta-questions}:
	what are the legitimate topics of inquiry within a field,
	and what are the important open problems?
In addition, a good paradigm should answer the question of evaluation:
	how can the quality of new results be determined?
If scientists needed to spend large amounts of time pondering
	these abstract questions, 
	they would have no time for concrete research.
The paradigm also provides an important concentration-enhancing role.
Practitioners become highly focused on the relevant problems
	defined by the paradigm.
This concentrating effect helps to insulate practitioners from distractions,
	such as social pressures from the outside world.
Scientists operating within a paradigm commit themselves to a certain intellectual path,
	upon which they can stride forward confidently,
	without going in circles or getting lost in the woods.

In addition to solving philosophical problems,
	a paradigm also helps to solve sociological issues.
Scientists tend to be intelligent, individualistic,
	and distrustful of authority.
There is no central committee or chief executive
	telling scientists what to do.
The paradigm is what allows this decentralized group 
	of often eccentric persons to make systematic progress.
If it were not for a paradigm,
	a scientific community would act more like a disorganized mob
	and less like a disciplined army.
Science is also, in many ways, a competitive sport,
	where the prizes include things like grant money,
	tenure, and fame.
The paradigm defines the rules of the game.
Without a paradigm,
	science would be like a baseball game
	in which the players do not agree about the location of the bases,
	the number of outs in an inning, or whether it is legal to bunt.
It is not a coincidence that the person who determines 
	whether to accept a paper submitted to a journal is called a ``referee''.	
		
Another, less obvious requirement of a good paradigm
	is that it must define a large number of technical questions, or ``puzzles''.
Ideally, these puzzles are not be exorbitantly difficult:
	an intelligent and hard-working practitioner
	should have a very good chance of solving 
	the puzzle he or she chooses to work on.
If this were not the case, 
	morale in the field would evaporate and it would cease
	to attract talented young people.
Since a scientific community commits to a paradigm,
	and the paradigm defines a set of puzzles,
	members of the community will recognize a new puzzle solution
	as a valuable research contribution.
Also, the puzzles must be cumulative in nature.
The puzzle solutions should be like bricks in a wall;
	each new brick both expands the wall and provides a 
	space for another brick to be placed on top of it.

If a scientific community is able to both generate good puzzles and solve them,
	it will naturally become increasingly esoteric.
The concepts, techniques, and phenomena it considers
	will become completely incomprehensible to a layperson.
This esotericity is a characteristic property of a mature scientific field.
A well-educated person can be expected to understand 
	Newton's laws and perhaps even special relativity.
But no layperson can be expected to understand 
	an abstract such as the following:
	
\begin{quote}
New torsional oscillator experiments with plastically deformed helium show 
	that what was thought 
	to be defect-controlled supersolidity at low 
	temperature may in fact be high-temperature 
	softening from nonsuperfluid defect motion in the crystalline structure~\cite{Beamish:2010}.
\end{quote}

\mpagebreak

\subsection{The Casimir Effect}

The following anecdote may help to illustrate
	the meaning of normal science.
In 2003 the author was a graduate student in physics
	at the University of Connecticut.
The physics department would often invite researchers from other universities
	to come and present their work.
At one point, a visiting researcher came to talk about
	his experiments involving a phenomenon known as the Casimir effect.

The Casimir effect is a fascinating implication
	of the theory of quantum electrodynamics.
This theory states that	
	there exist minute fluctuations in the electromagnetic field
	at every point in space, 
	even if there is no matter or forces at the point.
If space is thought of as a string,
	then these fluctuations correspond to very low frequency excitations
	that arise spontaneously.
Ordinarily, these fluctuations occur everywhere at more or less
	the same rate.
But if two mirrors are placed very close to one another 
	(on the order of one micron)
	then they dampen out certain frequencies of the spontaneous excitations.
Since the excitations contain energy,
	this implies that the region outside the mirrors has a slightly
	higher energy density than the region between them.
This causes a force, analogous to pressure,
	that pushes the mirrors together.
This phenomenon is called the Casimir effect.

The visiting researcher had performed a very sophisticated
	experiment to measure the Casimir effect.
The effect is hard to measure because
	it is quite weak except at small scales,
	where several other forces can interfere.
To measure the effect,
	one must make corrections relating to thermal fluctuations,
	and to the fact that real mirrors are neither perfectly
	smooth nor perfectly reflective.
Furthermore, 
	because of the weakness of the effect,
	a sophisticated and highly sensitive device called 
	an atomic force microscope must be used to measure it.
Finally, since air particles can also interfere,
	the experimental apparatus must be placed within
	a nearly perfect vacuum.

The researcher was able to obtain measurements
	for the Casimir effect that agreed with theoretical
	predictions to an accuracy of 1\%~\cite{Mohideen:1998},
	so the experiment was considered a great success.
It is worth asking, however,
	what would have happened if such an agreement 
	had not been achieved.
The most likely outcome is that the experiment would 
	have been interpreted as a failure, and not published.
If the negative results were published,
	they would undoubtedly have very little effect on the field,
	except perhaps to inspire other experimentalists to 
	try to tackle the problem.
Almost certainly,
	the negative result would not cause anyone to 
	conclude that quantum electrodynamics was wrong.
		
This anecdote illustrates the philosophical commitments
	that guide research in physics.
An intellectual with no understanding of the physics paradigm
	might reasonably conclude that the measurement was a colossal
	waste of time and money.
The Casimir effect is vanishingly weak and hardly ever arises in practical situations,
	the visiting researcher was not the first to measure it,
	the experiment was enormously difficult and expensive to carry out,
	and if the measurements had not agreed with predictions,
	the results would have been ignored.
In spite of all this,
	the experiment was widely recognized as a success,
	and in a congratulatory spirit the scientist 
	was invited to go on a lecture tour to present his research
	to aspiring young physicists.
The researcher's measurement was a small step forward,
	but it was decisive one,
	that clearly advanced the state of human knowledge.

\mpagebreak
	
\subsection{The Microprocessor Paradigm}

One of the most analyzed and commented-upon trends in recent technological history
	goes by the name Moore's Law.
There are various versions of Moore's Law,
	but the basic statement is that the number of transistors
	on a minimum cost chip doubles every year. 
Because the number of transistors included in a microprocessor
	is an important factor in its performance,
	this law roughly states that the performance
	of digital computers increases exponentially.

A remarkable aspect of Moore's Law
	is that it depends on the reliable appearance
	of ever more sophisticated technological insights.
The microprocessor depends on a complex web of technologies,
	that enable not only the thing itself,
	but also the processes that allow it to be manufactured efficiently.
Consider photolithography,
	which is but a single strand of the web.
Photolithography is a technique in which a geometric pattern
	is etched into a substrate using ultraviolet light.
The pattern is selected by shining light through a special mask.
Before exposure,
	a thin layer of a special chemical, called a photoresist,
	is deposited onto the chip.
After the deposition,	
	the chip is spun to ensure that the photoresist
	spreads out uniformly across the surface.
The areas of the photoresist that are exposed to the UV light
	become soluble in another special chemical.
This chemical is then used to remove the exposed
	areas of the photoresist,
	leaving behind the unexposed areas which match the pattern of the mask.
It is also necessary to use a special excimer laser
	that can produce deep ultraviolet light.
The full photolithographic process relies upon the use of specialized chemicals,
	advanced lasers, fluid mechanical modeling techniques,
	and optical engineering methods.
All of these moving parts need to work together
	seamlessly and with high reliability.
The brief description given above mentions only a couple of 
	innovative techniques that played a role in Moore's Law;
	there are many more.
In spite of the almost magical nature of these technologies,
	they all appeared right on time to satisfy the 
	next mandated doubling of processor speeds.
	
Another remarkable aspect of Moore's Law is that 
	it depends on a chaotic process of commercial competition
	to produce the predicted improvements.
Dozens of companies,
	such as IBM, Intel, Motorola, and Texas Instruments,
	participated in and contributed to the microprocessor revolution.
The history of the field records a large number of 
	new design offerings,
	many of which failed commercially or led nowhere.
For example,
	the Texas Instruments TMS 1000 was one of the first
	general purpose computers on a chip.
TI used the chip in its calculators
	and some other products,
	but it did not play an important role in later computers.
At the same time 
	as the TMS 1000,
	Intel released an all-purpose, 4-bit chip called the 4004.
The 4004 was not especially successful,
	but paved the way for later Intel chips 
	called the 8008 and 8080.
The 8080 was one of the first truly useful microprocessors.
Other general purpose chips of that era include
	the RCA 1802, the IBM 801, the Motorola 6800,
	and the MOS 6502.	
It is remarkable that,
	in spite of the fact that many individual products flopped
	and many companies went bankrupt,
	the field as a whole achieved rapid progress.

A major factor contributing to Moore's Law
	was economics. 
In the early decades of research in computers and electrical engineering,
	it was unclear whether there was a market 
	for ever-faster computer chips.
But at a certain point,
	it became clear that demand did exist.
Once computers came into widespread use,
	their basically generic nature 
	meant that improvements in processor speed 
	yielded improvements in a wide variety of applications
	(a faster CPU will speed up not only the spreadsheet application,
	but also the word processor, database, web browser, and so on).
These gave companies an incentive to spend 
	huge amounts of money on new research and development efforts.
	
While economics played an important role in microprocessor performance,
	philosophical factors played an equally important role.
In many ways, 
	microprocessor research served as a scientific paradigm.
The paradigm provided a clear answer to the meta-question:
	a new result was valuable if it improved the speed of a chip,
	or if it could make chip fabrication less expensive,
	or if it helped with a related issue such as power consumption.
Evaluation was straightforward and objective:
	did the chip compute the correct result?
Can it be manufactured cheaply and reliably?
Also, the paradigm posed a stream of puzzles
	for new researchers to attack.
These puzzles were certainly not easy,
	but the existence of the paradigm gave people the determination
	necessary to attempt to solve them.
Carver Mead,
	who did several early studies related to Moore's Law, remarks that:
	
\begin{quote}
[Moore's Law] is really about people's belief system, 
	it's not a law of physics, it's about human belief, 
	and when people believe in something, 
	they'll put energy behind it to make it come to pass~\cite{Kelly:2010}.
\end{quote}

\noindent
In other words, 
	people knew what they were looking for, and they
	knew that it was \textit{possible} to make rapid improvements,
	so the prospect of achieving yet another doubling
	of clock speed seemed like a routine challenge and not an impossible dream.
This expectation of success produced the institutional determination
	necessary to mobilize large groups of people.
Even if a each individual researcher achieved only a modest improvement,
	this was sufficient for rapid aggregate progress.
The economic and philosophical factors,
	combined together,
	implied that a group of people could dedicate themselves to a 
	highly specialized and esoteric research project
	with a strong degree of certainty that the fruit of their efforts 
	would constitute a substantial contribution to human knowledge and civilization.

\mpagebreak	

\subsection{The Chess Paradigm}

The problem of computer chess is nearly
	as old as the field of computer science itself.
Even before the first computers were built,
	Alan Turing wrote a chess algorithm
	and used his own brain to ``run'' it.
Claude Shannon suggested programming
	a computer to play chess, arguing that:
	
\begin{quote}
Although of no practical importance,
	the question is of theoretical interest,
	and it is hoped that...
	this problem will act as a wedge in attacking other problems...
	of greater significance~\cite{Shannon:1950}.
\end{quote}

The history of computer chess shows a record
	of gradual but systematic progress.
Dietrich Prinz, a student of Einstein and colleague of Turing,
	wrote the first limited chess program in England in 1951.
Alex Bernstein, a researcher at IBM,
	wrote the first program that could play a full game of chess in 1958.
The Association for Computing Machinery
	started to hold biannual computer chess competitions in 1970.
These competitions spurred a feeling of camaraderie and competitive spirit,
	and by the end of the 1970s the 
	the top programs were as good as upper-level human players.
Interestingly, 
	even as the field was making rapid progress,
	there was uncertainty about how far it could ultimately go.
In 1978, David Levy, an International Master,
	won a bet he made ten years earlier that no computer would beat him.
At an ACM conference in 1984,
	a panel of experts could not agree about
	whether a computer would ever defeat the top-ranked human player.
Of course,
	this event did eventually occur,
	when IBM's Deep Blue defeated Gary Kasparov in 1997.

The inexorable progress achieved in the field of computer chess
	can be explained by the fact that chess defines a paradigm.
The chess paradigm provides a clean answer to the meta-question:
	a new idea is a valid contribution to the field
	if it can be used to enhance the performance of a chess program.
The paradigm
	also supports clean, inexpensive, and objective evaluations of rival solutions.
Furthermore, 
	the paradigm defines a nice set of puzzles.
Computer chess researchers innovated a variety of techniques
	to attack the problem.
One such technique is alpha-beta pruning,
	which is a way of searching a tree 
	representing game states~\cite{Knuth:1975}.
The idea of alpha-beta
	is to prune subtrees that are known to be suboptimal,
	given that every other branching decision in the tree
	is made by the opposing player.
The chess problem even inspired 
	two Bell Labs researchers to develop
	a chess computer with special hardware 
	for evaluating board positions and generating lists of possible moves.

Most people would agree that computer chess
	was a successful research program
	that made an important contribution to human knowledge.
This fact is remarkable,
	given that chess is an abstract symbolic game
	with no connection to the real world.
An analogy to the game of tennis may be useful here.
A tennis game does not directly evaluate	
	a player's fitness level.
Nevertheless, 
	in order to win at tennis one must be highly fit.
So the discipline imposed by the rules of the game
	forces players to become fit,
	if they want to win.
The chess paradigm has an analogous effect.
The discipline imposed by the objective evaluation procedure
	forced researchers to innovate in order to win.
In comparison to the chess paradigm,
	the compression principle provides a similar level of discipline,
	which should force researchers to innovate.
Even if it turns out that \compist~research has no practical importance,
	it may act, as Shannon put it,
	as a wedge in attacking other problems of greater significane.

\mpagebreak

\subsection{Artificial Intelligence as Pre-paradigm Field}

This book has mostly avoided explicit discussion
	of the field of artificial intelligence,
	because it is too broad to be reasoned about as a whole.
But in the light of the discussion of paradigms,
	a simple diagnosis can be made that explains
	the gap between the enormous ambitions of the field 
	and the actual success it has achieved.
AI is, quite simply,
	a field without a paradigm.
Though many authors have attempted to define intelligence
	and thereby provide a direction to research,
	none of those definitions are widely accepted.
This lack of shared commitments 
	prevents the field from making systematic progress,
	except for short spurts of activity within narrowly defined domains.
The field consists of a large number of individuals
	carrying out idiosyncratic and mutually incompatible research programs.

Evidence in favor of this diagnosis
	can be obtained simply by perusing at the 
	standard AI textbook by Russell and Norvig~\cite{Russell:2009}.
In a textbook describing a mature field, 
	one would find a clear relationship between the early sections
	of the book and the later sections.
The early chapters would discuss fundamental ideas and theories,
	while the later ones would involve applications or specialized areas.
But the topics related in the first half of the AI textbook
	have nothing to do with the topics in the second half.
An early section of the book is entitled ``Problem Solving'',
	which includes topics such as heuristic search,
	adversarial search, and constraint satisfaction.
A later section is called ``Communicating, Perceiving, and Acting'',
	and discusses areas such as natural language processing,
	perception, and robotics.
A researcher in natural language processing or robotics
	has no special need to know anything about adversarial search or constraint satisfaction,
	and may very well be completely ignorant of these topics.
The incoherence of the textbook is not to be blamed on the authors,
	who have provided clear and cogent descriptions of the standard
	topics of the field.

This is not to say that AI researchers do not propose paradigms;
	in fact, they do quite often.
The problem is simply that the paradigms do not work very well.
One example of such a paradigm relates to the STRIPS planning formalism
	(STRIPS stands for Stanford Research Institute Problem Solver)~\cite{Fikes:1971}.
This formalism represents 
	the state of a system as a collection of propositions.
So \textsc{at(trucka, miami)} would assert that a certain truck is in Miami.
The agent has the option
	of modifying the state by executing an action,
	which may have preconditions, add effects, and delete effects.
An example action
	is \textsc{move(trucka, miami, houston)}.
This action would require \textsc{at(trucka, miami)} as a precondition,
	and also delete it, while adding \textsc{at(trucka, houston)}.
This formalism is expressive and general
	enough to allow users to encode a wide variety of planning problems.
This expressiveness implies
	that if a general purpose STRIPS solver could be found,
	it would be an extremely useful tool.
Motivated by the goal of obtaining such a tool,
	members of the automatic planning community have carried
	out several decades of research based on STRIPS.
To evaluate progress,
	the community has defined a number of benchmark problems
	that involve things like logistics
	(how to deliver a set of packages using the smallest amount of time and fuel)
	and airport traffic control~\cite{McDermott:2000}.
Unfortunately, 
	the mere act of defining shared benchmarks 
	has not allowed the community to make much progress.
Research in this area is hobbled by a variety of intractability theorems.
One such theorem shows that the basic STRIPS problem
	is NP-hard~\cite{Bylander:1994}.
Another shows that if the formalism is enhanced to include numeric quantities,
	the problem becomes undecidable~\cite{Helmert:2002}.
If new planning systems appear to show improved performance,
	this is probably because they been designed to 
	exploit special properties of the benchmark domains,
	not because they have achieved truly general applicability.
	
One of the purposes of a paradigm
	is to produce a set of puzzles
	that all members of a community will 
	recognize as important research topics.	
Because there is no clear paradigm 
	for artificial intelligence,
	researchers are generally at liberty to define their own research questions.
This situation might not be so deplorable,
	if researchers were honest about their reasons
	for considering a certain question important.
In practice,
	people have prefered solution methods,
	and define their research questions to be exactly
	those that can be solved, or at least attacked,
	by their methods.
	
\mpagebreak

\subsection{The Brooksian Paradigm Candidate}

In the early 1990s,
	an MIT researcher named Rodney Brooks
	wrote a series of papers
	articulating a critique of the philosophical foundations 
	of artificial intelligence research 
	and outlining a new paradigm for the field~\cite{Brooks:1990,Brooks:1991}.
Brooks believed that AI researchers suffered from a disease called ``puzzlitis'',
	which was caused by two factors.
One the one hand,
	everyone saw and felt the necessity of using
	objective measures to evaluate new research results.
But at the same time,
	the connection between AI research and the real world
	was tenuous at best.
This meant that the typical strategy for producing new research 
	was to invent some obscure logical puzzle that no one
	else had ever considered,
	and then show that a new system could solve the puzzle.
Brooks also deplored the fact
	that much research in AI was \textit{ungrounded}.
Researchers spent much time and effort on
	defining new abstractions,
	figuring out how to organize them,
	and deciding how they should relate to one another.
But these abstractions were not, for the most part,
	grounded in empirical reality.

To repair these problems,
	Brooks proposed a paradigm candidate for AI research
	based on the idea of real world agents.
Brooks defined intelligence to be the set
	of computational abilities
	that an agent would required to survive in the real world.
This definition provided an answer to the meta-question for the paradigm:
	a new result was a valuable contribution 
	if it helped an agent to navigate or operate in the real world.
Since real world agents encounter a wide
	variety of perceptual and computational challenges,
	this definition provided a parsimonious justification
	for a broad range of AI research.
Brooks noted that many old problems disappeared and many new problems arose
	when an agent was placed in the real world.
For example, the problem of building and manipulating 
	a complex internal world model largely disappeared.
Instead, agents could simply rely on extensive and constant
	sensing in order to keep track of the world state.

Brooks' research plan was also partly motivated by 
	an analogy to the path taken by evolution 
	in producing intelligence.
Evolution started with very simple creatures,
	that didn't do much other than wander around 
	and try to find food.
As evolution advanced, more complex sensory and cognitive abilities appeared.
Crucially, evolution spent huge amounts of time developing
	low-level skills like perception, motor control, and basic navigation.
In contrast, advanced human skills like logic, mathematics, and language
	appeared only recently in historical terms.
This implied that the real challenge was to develop the low-level skills.
Brooks argued that AI research should follow a similar path.
	
At the time of Brooks' original work,
	it seemed very plausible that the embodiment paradigm
	would satisfy many of the requirements for a good paradigm.
It produced a large stream of puzzles,
	that competent practitioners would have a good chance of solving.
It provided a set of shared philosophical commitments
	that would mandate all participants
	to recognize new puzzle-solutions as legitimate contributions.	
An example of such a puzzle-solution might be a 
	robot that could navigate around an office environment
	and clean up all the discarded soda cans.
At least in principle, 
	it seemed like the research in this area could be cumulative in nature.
Given a soda can cleaning robot, 
	it might be a tractable problem to improve it so that it 
	also has the ability to fetch coffee.
	
The debate between Brooks and other researchers pursuing more traditional
	approaches to AI
	provides a good illustration of how paradigms can be incommensurate.
To Brooks,
	topics such as theorem proving, automated planning,
	and computer chess were simply irrelevant distractions.
An artificial agent operating in the real world
	has no need to prove mathematical theorems or play chess.
Conversely,
	critics of Brooks' idea did not accept
	that computational tasks
	like wall-following or obstacle avoidance
	were relevant to intelligence. 
Neither group had any special reason to accept
	the other's assumptions and commitments.
The debate cannot be solved by normal scientific methods
	such as experimental verification or proof-checking.	
	
Twenty years after its initiation, 
	the Brooksian paradigm has not yet achieved its goals.
Perhaps the key failure was the assumption that this kind of research
	could achieve cumulative progress.
In reality, it is very difficult for a new researcher to pick up
	a robotics project begun by an earlier worker and 
	successfully enhance its capabilities.
One reason for this is that problems in robotics are not often  
	conclusively solved like mathematics problems are solved.
Rather, a typical robot will succeed in some situations,
	but fail badly in others.
This means that the foundation laid by earlier researchers is not 
	solid enough for their successors to build upon.
	
Another problem, somewhat foreign to the purist intellectual concerns
	of academic researchers, involves simple economics.
Robots are very expensive and time-consuming to build.
If a researcher devotes a million dollars and three years to the 
	development of a soda can cleaning robot,
	and then the robot doesn't work for technical reasons
	(perhaps its can-pincing mechanism is not precise enough),
	this is a disaster for the researcher's career.
This means that researchers must avoid risky projects,
	since so much effort is invested.
But as any entrepreneur will agree,
	risk aversion is poisonous to innovation.
	
While the Brooksian paradigm did not succeed in completely revolutionizing AI,
	it would be incorrect to label it a complete failure.
One of the few truly useful commercial robotics products on the market today
	is the Roomba.
This robot is not ``intelligent'' in the traditional sense,
	but is instead resilient and resourceful at navigating the real world.
Brooks' research on navigation and obstacle avoidance techniques 
	played a key role in developing the Roomba.
Many other intriguing devices emerge from the laboratories
	dedicated to the embodiment paradigm,
	and it is possible that they will inspire other, 
	more practical applications in the future~\cite{Pfeifer:2007}.

\mpagebreak

\section{\Compist~Paradigm} 
\label{sec:advantages}

The previous discussion mentioned some of the aspects
	of previous paradigms that enabled researchers
	operating within them to make progress.
Several issues arose in each case.
A paradigm requires an answer to the meta-question 
	and to the problem of evaluation.
In some cases,
	these two issues were combined together.
So in the chess paradigm,
	any contribution was important if it allowed a computer to play better chess,
	and new solutions were evaluated using the victory rate they achieved.
In physics, 
	the answer to the meta-question is less explicit,
	but researchers can use the long history of success in the field
	to hone their intuitions about what kind of research is valuable.
The embodiment idea is interesting because it came very close
	to meeting the criteria necessary for a good paradigm,
	but fell short for subtle reasons.

This section argues that
	the compression principle can provide a solid
	paradigm for research in a variety of topics.
The \compist~philosophy meets all of the necessary criteria.
It provides a definitive answer to the meta-question,
	as well as an objective evaluation process.
The compression principle provides a parsimonious justification
	for many research topics.
The following sections discuss these topics in greater detail.	
The text refers specifically to computer vision,
	but the remarks apply equally well to various related fields.
A summary of the comparison between
	the \compist~approach to vision and the traditional approach
	is given in Table~\ref{tab:cvtradcomp}.
	
\mpagebreak

\begin{center}
\begin{table}[h!]
\label{tab:cvtradcomp}
\caption{Comparison of Methodological Approaches}
\begin{tabular}{|c|c|c|}
\hline
 & Traditional & \Compist  \\
\hline
Meta-Question & No Standard Formulation & Clear, clean formulation  \\
\hline
Comparisons & Ambiguous & Stark, decisive \\
\hline
Bugs \& Fraud & Hard to Detect & Cannot Occur \\
\hline
Manual Overfitting & Hard to Prevent & Cannot Occur \\
\hline
Type of Science & Math/Engineering & Empirical Science \\
\hline
Product of Research & Suite of Tools & Single best theory \\
\hline
Motivation & Practical Utility & Open-Ended Curiousity \\
\hline
Data Source & Small Labeled & Vast Unlabeled \\
\hline
Evaluator Development & Expensive, risky & Cheap  \\
\hline
Evaluator Scalability & One-to-one & Many-to-one \\
\hline
Justification of Topic & Esoteric & Parsimonious \\
\hline
Cooperation & Constant Replication of Effort & Standing on Shoulders of Giants \\
\hline
\end{tabular}
\end{table}	
\end{center}

\mpagebreak

\subsection{Conceptual Clarity and Parsimonious Justification}

Any paradigm candidate for a scientific field must 
	provide clear answer to at least two questions.
The first is the meta-question:
	what topics or research questions are contained within the field?
Without a clear answer to this question,
	the field becomes incoherent.
In some fields,
	the meta-question is answered only implicitly.
Perhaps there is no standard definition of what 
	exactly constitutes a legitimate question in physics research.
But everyone can see the ridiculousness of trying 
	to publish a historical analysis of presidential elections
	in a physics journal.
The journal editors would reject such a paper out of hand,
	even though presidential elections are ultimately governed
	by physical law.
However, the rejection would not automatically imply that the paper
	was not a legitimate piece of scholarship,
	only that it does not belong in a physics journal and,
	equivalently, the editors cannot evaluate it.
This equivalence immediately suggest the second function of a paradigm:
	it defines the legitimate methods of evaluating
	new research within a field.
If members of a community do not at least approximately agree on this point,
	academic civility will break down due to disputes
	regarding the value of new contributions.

The modern field of computer vision provides only very weak
	answers to these fundamental philosophical questions.
The field has a set of topics of interest,
	such as image segmentation, stereo matching, and object recognition.
But these problems appear to lack deep justification.
Topics such as edge detection are oftened justified by neuroscientific findings,
	such as the discovery by Hubel and Wiesel
	of cells in the visual cortex whose receptive field
	suggests that their purpose is to detect edges~\cite{Hubel:1962}.
Ideas of Gestalt psychology provide another commonly cited motivation~\cite{Kanizsa:1979}.
While the basic idea of imitating the action of the human brain seems sound enough, 
	in practice this strategy produces a large number of incompatible
	and disconnected research results.
An orthogonal justification is practicality:
	certainly a solution to the face detection task is useful.
Unfortunately,
	this justification fails in almost all cases,
	because computer vision research is only very rarely useful.
Certainly computer vision solutions are not widely
	deployed in real-world commerical systems
	(an important exception here is factory settings, 
		where the visual environment can be tightly controlled).

So modern computer vision offers
	a diverse range of research based on a diverse set of justifications.
In stark contrast,
	the \compist~approach leads to an extensive and varied
	set of research problems justified by a \textit{single} principle.
Any concept that can be used to compress a database of natural images
	is part of \compist~vision science.
This includes a wide variety of contributions,
	including new mathematical theorems,
	algorithms, knowledge representation tools, 
	and methods of statistical inference.	
This many-to-one relationship between contributions and justification
	also exists in physics, 
	where all research is ultimately justified by a single problem: 
	predict the future.

To see how these philosophical issues 
	intersect with practical considerations,
	consider the problem of how a graduate student
	should choose a thesis topic.
Finding a good topic
	is a nonobvious research problem itself,
	and requires familiarity with the field and the state of the art.
But the student,
	by virtue of being a student,
	does not yet possess a rich understanding of the field,
	and so is at a loss to decide what kinds of questions are important.
Sometimes the student's supervisor steps in and 
	makes the choice for the student,
	but this solution is not ideal.
The \compist~philosophy provides a very simple method 
	for finding new research topics.
One simply downloads 
	a benchmark database and the corresponding champion theory,
	and looks for aspects of the data
	that the theory does not adequately capture.
This can be done by looking for segments of the database
	that require an abnormally high amount of codelength to encode.
The student can also find shortcomings of the champion theory
	by sampling from the model and examining 
	the veridicality of the generated samples.
	
\mpagebreak
	
\subsection{Methodological Efficiency}

A major benefit of the \compist~approach 
	is the surprising degree of methodological efficiency it permits.
To see this, 
	consider a hypothetical Journal of \Compist~Vision Research,
	which accepts two kinds of submisssions.
The first type includes reports of new shared image databases for use by the community.
A paper of this type describes the content of a new database,
	the mechanism by which it was constructed,
	the extent of the variation that it contains,
	and any other details that might be relevant.
These submissions are accompanied with the actual database.
The journal editors briefly inspect the paper and the database.
Unless the database contains some glaring deficiency,
	these kinds of contributions should in most cases
	simply be accepted.
The editors then publish the paper and post a link to the 
	database on the journal's web site
	where it can be downloaded by other 
	interested members of the community.

The second type of submission includes reports of new compression rates
	achieved on one of the shared databases.
A submission of this type must be accompanied by the actual compressor
	used to achieve the reported result.
As part of the review process, 
	the journal editors run the compressor, 
	verify that the real net codelength agrees with the reported result,
	and then check that the decoded version matches the original.
In principle, 
	the editors should accept any contribution
	that legitimately reports an improved compression rate.
In practice, it may be necessary for the editors to exercise some degree of 
	qualitative judgment: 
	a paper that merely tweaks some settings of a previous result
	and thereby achieves a small reduction in codelength is probably not worth publishing,
	while an innovative new approach that doesn't quite manage to
	surpass the current champion probably is. 
In spite of the above caveat,
	the journal editors have a remarkably easy job.
The simplified nature of the \compist~review process 
	contrasts strongly with the situation in most modern scientific fields,
	where peer review is a crucial and time-consuming activity.

The \compist~philosophy also considerably simplifies
	the process of database development.
A major limiting factor in traditional computer vision research is
	the effort required to build ground truth databases.
The \compist~approach mandates the use of unlabeled databases,
	which are much easier to construct.
For example,
	a researcher wishing to pursue the roadside video proposal of 
	Chapter~\ref{chapt:compmethod} needs only to set up a video camera
	next to a highway and start recording.
Even this relatively easy step can be bypassed,
	if an organization with video cameras already in place
	can be convinced to share their data.
Of course, not all databases are equally useful for research purposes.
It will be necessary to exercise some ingenuity and foresight
	when building target databases, especially in the early stages of research.
It is probably impossible to make a lot of progress by using
	an arbitrary collection of images downloaded from the internet.
Such an image collection would probably contain 
	an unapproachably large amount of variation.
	
The ease of data collection in \compist~research 
	is a consequence of its fundamentally new mindset.
Most researchers, 
	in computer vision and other fields,
	adopt a directed, utilitarian mindset.
They begin with the question:
	``what would it be useful to know in this area?''
This leads to specific questions like
	how to recognize faces, how to detect speech,
	how to perform part-of-speech tagging,
	whether a new drug will reduce the risk of heart disease,
	and whether a government stimulus will reduce unemployment.
To answer these questions, 
	the scientist must conduct a time-consuming empirical study.
The difficulty of the study limits 
	the amount of data that can be collected,
	which in turn limits the complexity of the model that can be used.
In contrast to the mindset of targeted research, 
	\compist~researchers adopt a stance of open-minded curiousity.
They begin with the question:
	``given this easily obtainable data source, what can be learned from it?''
This mindset is motivated by the belief that such databases
	\textit{do} contain valuable secrets.	
	
As noted above, 
	the \compist~philosophy makes it easy in principle
	to verify the quality of a new contribution.
The issue of verifiability is one of the main reasons
	why this book advocates the compression rate
	as a way of evaluating predictive power,
	as opposed to the log-likelihood or some other metric.
Consider a Maximum Likelihood Contest
	where the goal is to find the model that assigns
	the highest possible likelihood to a data set $T$.
The current champion theory has achieved a log-likelihood of $-3.4\cdot 10^{8}$ for $T$.
Now a challenger submits a new model,
	implemented as a software program.
The referee tests the program by invoking it on the data set.
After running for a while, the program prints out: 

\begin{quote}
Modeling complete: log-likelihood is $-2.2\cdot 10^{8}$.
\end{quote}

\noindent
If this claim is legitimate, 
	then the new model is superior and should be declared 
	the new champion theory.
But it is very difficult for the judge to verify 
	that the model follows the rules of probability theory and 
	that the software does not contain any bugs.
If a data compressor contains any significant bugs,
	these will be discovered when the decoded database
	fails to match the original one exactly.
	
The issue of software bugs and other kinds of mistakes
	is actually very relevant to machine intelligence research.
Software produced in vision research is far too complex
	to be thoroughly examined by reviewers.
Often researchers will not even release their source code for examination.
Typically,
	reviewers and the community must rely on the results
	published by the authors in their papers.
But if the authors produce those results using buggy software,
	it is very difficult for these errors to be discovered and repaired.
Consider the task of trying to determine
	if a segmentation algorithm is correctly implemented 
	according to the procedure set forth in a paper.
In some places,
	incorrect lines of code will cause the program to crash
	or produce an error.
But in other places,
	software bugs will simply cause the program to produce
	a different segmentation.
It is likely that many published results in computer vision
	are affected by bugs of this kind.
It is also possible that some errors are not entirely honest.
Given how difficult it is for a third party to check a result,
	and given the intense competitive pressure scientists operate under,
	it seems likely that some published results
	involve academic fraud.
It is very easy for a researcher, 
	when comparing his new algorithm for task X to some standard algorithm for X,
	to make an ``accidental'' mistake in implementing the latter.
The \compist~philosophy provides a strong defense against both honest mistakes
	and actual academic fraud.
	
\mpagebreak
	
\subsection{Scalable Evaluation} 

The word \textit{scalability} refers to the rate at which 
	the cost of a system increases as increasing demands are placed on it.
For example, 
	a process in which widgets are built by hand is not very scalable, 
	since doubling the number of widgets will approximately double 
	the amount of labor required.
A widget factory, in contrast,
	might be able to double its output using only a ten percent increase in labor, 
	and is therefore highly scalable.
A brief analysis shows that the 
	current evaluation mindset of computer vision 
	has very bad scaling properties,
	while the \compist~approach to evaluation is highly scalable.
	
The current evaluation strategy in computer vision
	can be characterized as a \textit{one-to-one} approach.
Since each specific vision task,
	such as object recognition, image segmentation, or stereo matching, 
	requires its own evaluator method,
	this approach scales badly.
If it turns out that a comprehensive articulation of computer vision 
	requires solutions for 1000 tasks,
	then the field will need to develop 1000 evaluators. 
This inefficiency is exacerbated
	by the fact that the process of developing an evaluator is difficult work,
	requiring both intellectual, and often manual, labor.
Furthermore, this arduous labor may very well go unrewarded.
Sometimes, after substantial effort has been invested in developing an evaluator,
	it turns out the scheme suffers from some flaw
	(a possible example of this is the difficulty related
	to the ROC curve evaluation scheme mentioned in 
	Section~\ref{sec:edgedteval}).
%~\cite{Bowyer:1999,Forbes:2000}).
Even if the evaluator is technically sound,
	it is not always obvious that it
	assigns consistently higher scores to higher-quality solutions.
It may be necessary 
	to conduct a meta-evaluation process
	in order to rate the quality of the evaluators.
Combined,
	these considerations imply that the current one-to-one
	approach to evaluation is painfully inefficient.

The importance of scalability becomes even more obvious 
	when one realizes that many vision tasks of current interest,
	such as image segmentation, edge detection, and optical flow estimation,
	are low-level tasks that are not directly useful.
Instead, the idea is that once good solutions are obtained for the low-level tasks, 
	they can be incorporated into higher-level systems.
These more advanced systems, then, constitute the ultimate goal of vision research. 
Under the current paradigm,
	each high-level task will require its own evaluator method
	to compare candidate solutions.	
Furthermore, since the high-level tasks are by definition more abstract,	
	performance on such tasks will be much harder to evaluate.
So the pile of work required to complete the project of 
	finding evaluators for \textit{current} tasks
	is tiny compared to the mountain of work 
	that will be required to develop evaluators for \textit{future} tasks.
	
In contrast to the current evaluation methodology,
	the Compression Rate Method provides the ability to evaluate a large
	number of disparate techniques using a single principle.
Chapter~\ref{chapt:comprvsion} showed that many tasks in computer vision
	can be reformulated as specialized image compression methods.
This implies that the compression principle provides
	a \textit{many-to-one} evaluation strategy.
As discussed in the proposal of Section~\ref{sec:cncrtprpsl},
	a single database of roadside video can be used to 
	evaluate the performance of a large number of components,
	such as motion detectors, wheel detectors, specialized segmentation algorithms,
	and learning algorithms that infer car categories.
The Visual Manhattan Project database of taxi cab video streams
	can be used to develop algorithms for detecting and analyzing the appearance
	of pedestrians, buildings, and other cars.

In addition to greatly reducing the number of man-hours required for evaluator development, 
	the compression principle provides a clean answer 
	to the question of what the high-level tasks \textit{are}.
In the compression approach,	
	a low-level system is one that achieves compression by exploiting
	simple and relatively obvious regularities,
	such as the fact that cars are rigid bodies obeying Newtonian laws of motion.
A higher-level system is built on top of lower-level systems
	and achieves an \textit{improved} compression rate 
	by exploiting more sophisticated abstractions,
	such as the fact that cars can be categorized by make and model.
Other techniques, 
	not specifically related to image modeling,
	can be justified by showing that they help to solve
	the problem of transforming from the raw pixel representation
	to the abstract description language.
	
\mpagebreak
	
\subsection{Systematic Progress}

Kuhn's book about scientific paradigms and revolutions contains	
	a number of profound aphorisms,
	but one of the most relevant is following rhetorical question:
	
\begin{quote}
To a very great extent, the term science is reserved for fields that do 
	progress in obvious ways. 
But does a field make progress because it is a science, 
	or is it a science because it makes progress?~\keycite{Kuhn:1970}.
\end{quote}

\noindent
As noted in Chapter~\ref{chapt:comprvsion},
	the field of computer vision 
	has a terrible problem involving constant replication of effort.
For any given task,
 	such as segmentation, edge detection, object recognition, or image reconstruction,
	there are hundreds of candidate solutions 
	to be found in the literature.
This endless rehashing and redevelopment of standard tasks,
	and the resulting lack of clear progress,
	is a clear sign of the immaturity of the field
	and the weakness of its philosophical foundations.
Consider, in contrast, how rare it is for physicists to submit
	new theories purporting to provide better descriptions of standard phenomena.
In and of itself, the proliferation of papers proposing rival solutions
	to the same task might not be so bad;
	any unexpected new experimental observation in physics might 
	merit a large number of proposed theoretical explanations.
The difference is that the physicists are able to ultimately 
	find the \textit{right} explanation,
	at which point people can move on to new areas of research.
	
The \compist~approach to vision not only allows the field to be formulated
	as an empirical science,
	but also effectively requires it to make progress.
Systematic progress is built into the very definition of the 
	Compression Rate Method.
In effect, an investigation guided by the compression principle	
	\textit{must} make clear, quantifiable progress, or grind to a complete halt.
The contrast between the potential for progress resulting from
	the \compist~approach and the lack of progress
	exhibited by the traditional approach seems, almost by itself,
	to be a sufficient argument in favor of the former.
	
The \compist~principle demands progress,
	but does not guarantee that progress will be easy.
However, there is a fairly clear
	path that a newcomer to the field 
	can take to build upon her predecessors' work.
Given a compressor that achieves a certain level of performance,
	the new researcher can simply add a specialized module
	that operates for data samples exhibiting a certain type of property, 
	and lies dormant for others.
For example, a researcher could take a generic image compressor such as PNG,
	and add a special module for encoding faces (see Section~\ref{sec:facedetmod}).
The enhanced compressor would then achieve much better codelengths 
	for images that include faces.
Another researcher could then add a module for encoding arms and hands
	(the pixels in these regions would be narrowly distributed around some mean skin color).
The improvement achieved by any single contribution may not be huge,
	of course, but this mode of research is guaranteed to achieve cumulative progress.
This description has glossed over some issues of software engineering;
	it is not, in practice, always easy to attach a new module
	to an existing system.
But this is exactly the kind of interesting
	technical challenge that computer scientists
	are trained to handle.
	
The issue of software design brings up
	another subtle advantage of the \compist~approach:
	it provides a software engineering principle to
	guide the integration of a large number of separate computational modules.
Given a set of software modules providing implementations of 
	segmentation, object recognition, edge detection, optical flow estimation, and so on,
	the traditional mindset of computer vision
	provides no guidance about how to package these modules
	together into an integrated system.
An attempt to do so would likely result in little more than a package of disjoint libraries,
	rather than a real application with a clear purpose.
Establishing compression as the function of the application
	provides a principle for binding the modules together.
The input is the raw image, 
	the output is the encoded version,
	and the software modules are employed in various ways
	to facilitate the transformation.

\chapter{Meta-Theories and Unification}
\label{chapt:metatheory}

\section{Compression Formats and Meta-Formats}

The last sequence of ideas dealt with by this book
	relate to the process of adaptation.
All modern compressors
	work by adapting to the statistics of the data
	to which they are applied.
For example,
	a text compressor invoked on a corpus of Spanish text
	will adapt to the language, 
	so that for example the suffix \textit{-dad} 
	(e.g. \textit{verdad, ciudad, soledad})
	will be assigned a relatively short code.
Early \compist~research will aim at the development
	of highly specialized, and thus less adaptive, compressors.
A \compist~theory-compressor of English will achieve
	superior short codelengths for a corpus of English,
	but fail completely when invoked on a Spanish corpus.
But the ultimate goal of \compist~research is to achieve
	both adaptivity and superior performance.
The following development describes how this can,
	in principle, be achieved.
	
To begin, consider a team of scientists who are planning an experiment,
	and wish to develop a specialized compressor
	for the data produced by the experiment.
Suppose the experiment was known to produce
	data that would follow a exponential distribution:
	
\begin{equation}
\label{eq:expondistr}
P(x) = k e^{-\lambda x}
\end{equation}

\noindent
Where $x$ is a non-negative integer,
	$\lambda$ is parameter of the system,
	and $k$ is a constant required to ensure normalization.
A simple calculation shows that $k = (1-e^{-\lambda})$. 
Now consider three scenarios involving different schemes for 
	encoding the data generated by this distribution.
	
In the first scenario, the true value of $\lambda = 2$ is known exactly in advance.
This allows the scientists to find the optimal codelength function:

\begin{equation*}
L(x) = -\log P(x) = -\log k + \lambda x
\end{equation*}

\noindent
The expected codelength achieved by using this codelength function
	can easily be calculated 
	by plugging Equation~\ref{eq:expondistr} into Equation~\ref{eq:entropyeqn}.
The result is $0.458$ bits,
	and this is the shortest possible expected per-sample codelength
	that can be achieved;
	also known as the entropy of the distribution $P(x)$.
A codelength function that diverged from the one shown above
	could do better on some outcomes,
	but it would underperform on average.
The ability of the scientists to achieve the best possible codelengths
	is based on their foreknowledge of $\lambda$, and thus $P(x)$.

\begin{figure}
\begin{centering}
\includegraphics[width=.5\textwidth]{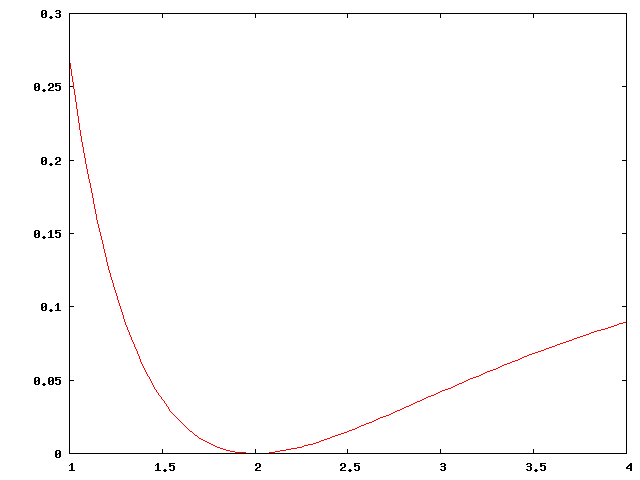} 
\caption{
Graph of codelength penalty per sample for various 
	choices of $\lambda$.
The real distribution has $\lambda=2$.
}
\label{fig:metalambda}
\end{centering}
\end{figure}
	
In the second scenario, 
	the scientists are sure the data will follow an exponential distribution,
	but do not know the value of $\lambda$.
% TODO: is this correct? Go over the math again.
They decide to make a guess,
	and use the value $\lambda_{g} = 3$.
This produces a model distribution $Q(x)$ 
	that is not the same as the real distribution $P(x)$.
Equation~\ref{eq:kldvrgence} can be used to calculate the average resulting codelength,
	which is $0.521$ bits per sample.
Due to the disagreement between $Q(x)$ and $P(x)$,
	this scheme underperforms the optimal one by $0.042$ bits per sample.
This can be interpreted
	as a penalty that must be paid for using an imperfect model $Q(x)$.
Figure~\ref{fig:metalambda} shows the relationship between
	the guess $\lambda_{g}$ and the codelength penalty.
The greater the difference between $\lambda$ and $\lambda_{g}$,
	the larger the penalty.
	
In the third scenario,
	the scientists decide not to try to guess the value of $\lambda$.
To get around their ignorance of $\lambda$,
	they instead decide to use a flexible meta-format,
	which works as follows.
When the compressor is invoked on the data,
	it reads all the samples,
	and performs a calculation to find
	the optimal value of $\lambda$.
It writes this value into the header of the format,
	and then encodes the rest of the data
	using it.
The decoder reads the $\lambda$ value in the header,
	and then uses it to decode the samples.
The price to be paid for using this scheme 
	is the extra bit cost of the header.
Assume $\lambda$ is encoded as a standard double precision floating point number,
	requiring 64 bits.
Then a simple calculation indicates that if $N > 1032$, 
	it is more efficient to use the adaptive encoding scheme
	than the ``dumb guess'' scheme.	
Of course, the superiority of the meta-format
	over the simple $\lambda_{g} = 3$ guess format can't be known in advance.
If the scientists get lucky and the guess turns out to be right,
	then the simple format will provide a better codelength.
	
The benefit of using a meta-format is flexibility:
	good performance can be achieved in spite of prior ignorance.
There are two drawbacks to the meta-format scheme:
	the additional overhead cost required to transmit the parameters (e.g. $\lambda$)
	and the computational cost involved in finding the optimal parameter values.
In the simple example discussed above,
	neither of these drawbacks seem very significant.
But when the models become more complex,
	so that instead of a single $\lambda$ value there
	are thousands or millions of parameters,
	the costs can increase dramatically.

This meta-format idea can be taken further.
Imagine that it was not known in advance that the data would follow
	an exponential distribution.
Instead, it was known that the data would be generated
	by either the exponential, Gaussian, Laplacian, Poisson, or Cauchy distribution.
To find the optimal specific format, 
	it is necessary to determine which distribution describes the data the best,
	and then to find the optimal parameter values
	(such as the mean and variance for the Gaussian).
This introduces no fundamental difficulties:
	one simply calculates the optimal parameters for each distribution,
	and then selects the distribution/parameter pair 
	that provides the shortest code.
This scheme requires a small additional codelength price to
	be paid in order to specify which distribution is being used,
	and a small additional amount of computation.
In return it provides a significant increase
	in the flexibility of the format.

The scientists can choose what meta-level to use
	based on the amount of data they expect to encode
	and their level of confidence about 
	what form the data will take.
The more data there is to encode,
	and the more ignorant they are of its structure,
	the higher up the meta-ladder they would want to go.
But there is a natural limit to this trend.
The highest meta-level choice is simply to specify a programming language
	as the format.
A programming language is the ultimate meta-format,
	because one can specify any possible specific format
	by transmitting the source code of a program that decodes the format.
For example, imagine that for a given dataset, 
	the bzip2 program achieves very short codes.
Then the programming language format can transform itself
	into the bzip2 format,
	simply by sending the source code for bzip2 as a prefix.
Furthermore, 
	this prefix is a one-time cost,
	that becomes insignificant as the volume of data being sent becomes large.
The programming language format provides the 
	largest possible degree of flexibility,
	and also the largest overhead cost. 
But in the limit of large data,
	the programming language format may very well be the best choice.

In the case of the simple exponential meta-format,
	it was a trivial task to find the optimal specific format,
	which was defined by a $\lambda$ value.
In contrast, 
	the problem of finding the optimal specific format
	when using the programming language meta-format is incomputable.
Finding the optimal specific format is equivalent to finding
	the Kolmogorov complexity of the data~\cite{Vitanyi:1997}.	
The impossibility of finding the optimal specific format
	when starting from the programming language meta-format
	does not, however, imply that the latter should not be used.
It hardly matters that the \textit{optimal} specific format cannot be found,
	when the programming language meta-format will provide codelengths
	only trivially worse, in the limit of large data,
	that \textit{any} other, less flexible meta-format.

%\begin{figure}
%\begin{centering}
%\includegraphics[width=.8\textwidth]{..//learnfctry.png} 
%\caption{
%The learning algorithm is a factory that 
%	transforms ``raw materials'' in the form 
%	of the meta-theory and the database,
%	and produces a specific local theory.
%% TODO: this schematic kind of sucks
%}
%\label{fig:learnfctry}
%\end{centering}
%\end{figure}

\mpagebreak

\begin{figure}[t]
\begin{centering}
\includegraphics[width=.5\textwidth]{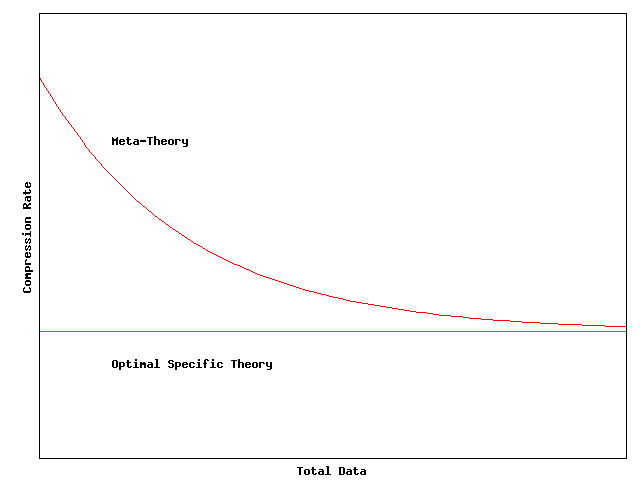} 
\caption{
Performance of the meta-theory vs. the optimal specific theory.
In this mode, the data is analyzed only once,
	and the meta-theory adapts as it is exposed to more and more data.
The $x$-axis is total data observed,
	and the $y$-axis is the compression rate per unit of data.
The area between the two curves is the amount by which the meta-theory
	underperforms the optimal specific theory.
}
\label{fig:metavsoptm}
\end{centering}
\end{figure}

\subsection{Encoding Formats and Empirical Theories}

Chapter~\ref{chapt:compmethod} discussed the No Free Lunch theorem,
	which states that no compressor
	can achieve lossless compression when averaged over all possible inputs.
The implication was that in order to achieve lossless compression in practice,
	a compressor needs to make an empirical assertion
	about the structure of the data set it targets.
If this assertion is correct, 
	compression is achieved;
	if not, the resulting encoded data will be larger than the original.
Thus a compression format can be thought of an empirical theory of 
	the phenomenon that generates the data.

The meta-format schemes mentioned above do not circumvent the No Free Lunch theorem,
	which holds for any valid compressor.
Since a meta-format can be implemented as a valid compressor,
	by using header data and prefixes,
	the theorem must hold for meta-formats as well.
Since a meta-format is still a format,
	it is also an empirical theory, 
	or more precisely an \textit{empirical meta-theory}.
	
To understand what this means, 
	consider again the discussion of the exponential formats.
A specific format makes the very precise assertion
	that the data will follow an exponential distribution with a particular $\lambda$ value.
If the real data distribution departs substantially from this assertion,
	the format will fail to achieve compression,
	and may end up inflating the data.
The meta-format, in contrast, makes the weaker assertion
	that the data will follow an exponential distribution 
	with \textit{some} $\lambda$ value.
Again, if this assertion turns out to be false,
	the meta-format will inflate the data.

At this point, 
	an important technical consideration must be mentioned,
	which is that there is a smarter way of using a meta-format
	than by using header data and prefix information.
This idea is called universal coding in the MDL literature~\cite{Grunwald:2007}.
The idea here is to calculate the optimal parameter values
	from the data, as it is being processed.
In the case of the exponential data,
	the scheme starts out identical to the dumb guess strategy,
	using some uninformed value of $\lambda_{g}$.
The difference is that as each $x$ data is processed and encoded,
	the value of $\lambda_{g}$ is updated.
After many data points are sent, 
	the updated value of $\lambda_{g}$ becomes nearly optimal.
This scheme achieves poor performance at the outset,
	but gets better and better as more data points are processed;
	this idea is shown in Figure~\ref{fig:metavsoptm}.
This strategy is technically superior to the header data approach,
	because it does not require any overhead costs for encoding the optimal parameters.
However, it has the same qualitative characteristics:
	it sacrifices performance in the low-$N$ regime for flexibility	
	and good performance in the high-$N$ regime.

\mpagebreak

\section{Local Scientific Theories}
\label{sec:locscithry}

\begin{table}[t]
\label{tab:birdsights}
\centering
\caption{Example Bird Sighting Database}
\begin{tabular}{|c|c|c|c|}
\hline
Date & County & Location & Species  \\
\hline
2/22/2011 & Coconino & Coconino & Rough-legged Hawk \\
\hline
2/19/2011 & Navajo & Heber & Pine Siskin\\
\hline
2/16/2011 & Pinal & Coolidge & Bald Eagle  \\
\hline
2/12/2011 & Navajo & Jacques Marsh & Tundra Swan \\
\hline
2/9/2011 & Apache & Greer & Wilson's Snipe \\
\hline
2/9/2011 & Maricopa & Peoria & Least Bittern \\
\hline
2/9/2011 & Apache & Eagar & Rough-legged Hawk \\
\hline
2/2/2011 & Pinal & Santa Cruz Flats & Rufous-backed Robin \\
\hline
2/2/2011 & Pinal & Santa Cruz Flats & Crested Caracara \\
\hline
2/1/2011 & Maricopa & Scottsdale & Zone-tailed Hawk \\
\hline
1/23/2011 & Pinal & Coolidge & McCown's Longspur \\
\hline
1/23/2011 & Maricopa & Arlington Valley & Red-breasted Merganser \\
\hline
1/19/2011 & Pinal & Kearney Lake & Ross's Goose \\
\hline
1/17/2011 & Navajo & Woodruff & Brewer's Sparrow \\
\hline
1/17/2011 & Navajo & Woodruff & Brewer's Blackbird \\
\hline
1/16/2011 & Maricopa & Arlington Valley & Harlan's Red-tail Hawk \\
\hline
1/16/2011  & Navajo & Woodruff & Sandhill Crane \\
\hline
1/15/2011  & Maricopa & Glendale Recharge Ponds & Western Sandpiper \\
\hline
\end{tabular}
\end{table}

For most people, 
	the word ``scientific theory'' generally refers
	to an idea like quantum mechanics or biological evolution.
These ideas seem to be universal, or at least highly general;
	physicists assume quantum mechanics holds everywhere in the universe, 
	and biologists believe that all 
	organisms are shaped by evolution.
This book employs a definition of a scientific theory
	that differs slightly from ordinary usage.
For a \compist~researcher, 
	a scientific theory is any kind of computational tool
	that can be used to predict and therefore compress an empirical dataset.
It is not even necessary for the predictions produced by the theory
	to be perfectly accurate,
	as long as they are better than random guessing.
Since the number of aspects of empirical reality 
	that can potentially be measured is enormous,
	the \compist~definition includes
	a very broad array of tools and techniques.
	
This more inclusive definition elevates to the status of science,
	and thereby identifies as worthy of examination,
	several simple tools that may otherwise have been overlooked.
One such example is the humble map.
A person using a map correctly
	can make reliable predictions about
	various observations such as street names and building locations.
Similarly, 
	a smart compressor could use a map to substantially compress
	a dataset of the form given in Table~\ref{tab:directions}.
Another example of a newly recognized scientific theory is a dictionary (word list),
	which can be used to predict word outcomes
	when scanning through a stream of text.
The dictionary will not, of course,
	provide \textit{exact} predictions,
	but it does cut down on the number of possibilities
	(there are far fewer real 10-letter words than there 
	are combinations of 10 letters).

\begin{figure}
\centering
\includegraphics[width=.4\textwidth]{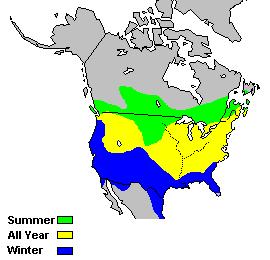}
\caption[Pixel difference histograms for a natural and a random image.]{
Range map for the American Goldfinch. 
This map is a computational tool that can be used to predict
	a database with the form shown in Table~\ref{tab:birdsights},
	and can also be extracted from the information contained
	in such a database.
% Image from Birding.com
}
\label{fig:birdranges}
\end{figure}

A less common example of this kind of computational tool
	is a bird range map, as shown in Figure~\ref{fig:birdranges}.
An observer using this map 
	can make predictions relating to the types 
	of bird species that might be sighted at different
	locations and times of the year.
Consider a data set of the form shown in Table~\ref{tab:birdsights},
	and imagine one is attempting to predict
	the fourth column (bird species) 
	from the first three columns (date and location information).
Because bird species have characteristic geographical distributions,
	information of the kind embodied in Figure~\ref{fig:birdranges}
	can be used to help with the prediction.
For example, if the location corresponds to an area in Alberta,
	and the date is in January,
	then the range map predicts that the American Goldfinch will not be observed.
The map also predicts that the goldfinch
	will not be sighted in Florida during the summer,
	and that it will never be sighted in Alaska.
One can easily imagine embellishing the range map with information
	relating to population statistics,
	allowing more accurate predictions that reflect the relative
	prevalence of various species.

The computational tools described above have several properties in common.
First, they are \textit{local}
	as opposed to universal theories:
	the map only works for one geographic area,
	while the dictionary only works for one language.
Second, while they are like physical theories in that they enable predictions,
	they employ an opposing set of inferential tools.
Physical theories combine a minimalist set of inductive steps 
	(the natural laws) with complex deductive reasoning
	such as calculus, linear algebra, and group theory.
Physical theories can therefore be characterized
	as inductively simple but deductively complex.
In contrast, 
	the local scientific theories mentioned above
	are logically simple to use,
	but contain a large number of parameters.
Therefore, they can be viewed as employing
	the inverse of the inferential strategy used in physics:
	complex induction combined with simple deduction.
	
A third, and crucial property of a local scientific theory 
	is that it can be \textit{extracted} from
	an appropriate dataset.
This is most obvious for the case of a word list,
	which can be extracted from a large text corpus.
To do so, 
	one simply programs an algorithm to find every
	unique combination of letters in the corpus.
More sophisticted techniques will, of course,
	lead to a more refined dictionary;
	a smart extraction algorithm might use knowledge of verb conjugations
	so that ``swim'', ``swims'', ``swimming'', and ``swam'' 
	are packaged together under the same heading.
It is also quite easy to imagine constructing a set of bird range maps
	from a large database of the form shown in Table~\ref{tab:birdsights}.
A street map can be constructed from a large corpus of observational data
	of the form shown in Table~\ref{tab:directions},
	though this is more complex algorithmically.
So not only can a local scientific theory be used to predict a particular database,
	it can also be extracted from the same database.

\mpagebreak
	
\begin{table}[h]
\label{tab:directions}
\centering
\caption{Table of driving observation data.
These were generated by an internet mapping service,
	but the same kind of data could be generated by exploration
	and observation.
}
\begin{tabular}{|c|c|c|c|}
\hline
Operation & Street Name & Distance & Time \\
\hline
\hline
Slight right & Garden St & 0.2 mi & - \\
\hline
Turn right & Mason St & 0.1 mi & - \\
\hline
Turn right & Brattle St & 344 ft & - \\
\hline
First left & Hawthorn St & 0.1 mi & 1 min \\
\hline
Continue onto & Memorial Dr & 390 ft & - \\
\hline
Turn left & Memorial Dr & 1.0 mi & 2 min \\
\hline
Turn right & Western Ave & 0.1 mi & - \\
\hline
Turn left & Soldiers Field Rd & 0.2 mi & 1 min \\
\hline
Take & I-90 Ramp & 0.3 mi & - \\
\hline
Keep left & I-90 East & 3.4 mi & 4 min \\
\hline
Take exit 24C & I-93 North & 0.7 mi & 1 min \\
\hline
Merge & I-93 South & 15.3 & 17 min \\
\hline
Take exit 1 & I-95 South & 80.0 mi & 84 min \\
\hline
Take exit 90 & CT-27 & 0.3 mi& - \\
\hline
Turn left & Whitehall Ave & 1.5 mi & 3 min \\
\hline
Turn right & Holmes St & 0.3 mi & 1 min \\
\hline
Turn right & East Main St & 0.2 mi & 1 min \\
\hline
Take 2nd left & Water St & 0.2 mi & - \\
\hline
Take 2nd right & Noank Rd & 459 ft & - \\
\hline
\end{tabular}
\end{table}

\mpagebreak

\section{Cartographic Meta-Theory}	

There is something very interesting about the concept of a map,
	or more precisely the concept of mappability,
	which the following thought experiment may help reveal.
Imagine an outbreak of polio has been observed in an 
	remote Indian town.
A team of health experts has been sent by the United Nations
	to administer the polio vaccine to everyone in the town.
The doctors plan to conduct a systematic door-to-door sweep of the town,
	administering the vaccine to everyone they can find.
However,
	they have encountered an obstacle in carrying out their plan:
	due to an odd prohibition promulgated by the local religious guru,
	there are no maps of the town
	(the guru believes that becoming lost is important for spiritual health).
The lack of a good map presents no special difficulties, however, 
	because the team can just write down a map on their own.

The interesting aspect of this thought experiment
	is the epistemic state the doctors are in before
	they construct the map.
They have never visited the town,
	and they possess only the most rudimentary information about it.
Nevertheless, they are completely confident 
	in their ability to construct a map of the town.
Furthermore,
	they are highly confident that once this map is written,
	it will help with the door-to-door vaccination effort.
The doctors therefore make a meta-prediction:
	the map, once created, will make good predictions.
They are highly confident in this meta-prediction,	
	in spite of the fact that the map has not yet been written
	and that they have no idea what it will contain.
They are, in effect, utilizing an empirical meta-theory,
	which can be called the cartographic meta-theory.

To see that the cartographic idea 
	must be an empirical theory,
	it is sufficient to show that it can be used to construct a compressor.
Consider a large database of observations 
	of the form of Table~\ref{tab:directions}.
Observe that a street map
	will be useful to predict and compress this data.
To see this,
	notice that a certain section of the observations turning
	right onto Western Ave. from Memorial Dr.
This specifies a precise location on the map.
The observations then invole turning left after a distance of 0.1 miles.
The map can now be used to predict that the name
	of this street will be ``Soldier's Field Rd'',
	because there is presumably only one road that intersects
	Western Ave. at that point.
Now,
	when the compressor begins, 
	it will not have a map, 
	and so it won't be able to make any confident predictions.
But if the corpus is large,
	then eventually the compressor will have enough data
	to infer a street map
	and use it to compress the subsequent observations.
So perhaps the compressor will achieve no compression
	for the first 10 Mb of data,
	but 60\% for the next 90 Mb.
Since it can be used to build a compressor,
	the cartographic principle must be empirical,
	and since it actually achieves compression,
	it must correspond to reality.
The success of the cartographic meta-theory
	reflects a certain aspect of reality: 
	the fact that streets and buildings
	occupy fixed geographic positions.
One can imagine a floating city made up of a shifting patchwork
	of boats and barges in which the cartographic principle will fail.
	
There is a strong similarity between the cartographic meta-theory
	and the adaptive-$\lambda$ scheme mentioned above.
In both cases, one starts with a flexible meta-format.
Then one performs some calculation or computation on the data.
This computation produces a set of parameters 
	that define a specific format.
The difference between the two schemes
	is only a matter of scale.
The cartographic meta-format
	requires many more parameters,
	and much more computation to find the optimal ones.
But the idea of using an algorithm to find a specific format
	based on a meta-format and a corpus of data
	is exactly the same.

\mpagebreak

\section{Search for New Empirical Meta-Theories}

It is rare for conceptual constructs to be unique;
	mathematics contains many distinct theorems
	and physics includes many natural laws.
Therefore it seems unlikely that the cartographic meta-theory
	mentioned above is the unique example of its class.
And indeed,
	a brief consideration will reveal several other examples.
For many languages,
	a lexicographic meta-theory holds,
	which asserts that dictionaries can be constructed.
Another meta-theory asserts that it is possible
	to build bird range maps.
Notice, again,
	that these other meta-theories
	make statements that are abstract,
	but still empirical.
The lexicographic meta-theory
	implies that languages are made up of words,
	and the set of legitimate words is 
	far smaller than the set of possible letter-combinations.
The avian meta-theory exploits
	the fact that bird species are not distributed randomly across the globe,
	but tend to inhabit characteristic regions
	that change from season to season,
	but not from year to year.

Because meta-theories correspond to meta-formats,
	and meta-formats are still formats,
	the assertions embodied by the meta-theories
	can be tested using the compression principle.
If bird species did not cluster in characteristic regions,
	the corresponding meta-theory would fail to achieve compression.
It is not, perhaps, 
	very interesting to prove the meta-theories mentioned above,
	because they are already intuitively obvious.
However, if there are \textit{other} valid meta-theories,
	some presently unknown,
	then the compression principle can be used to validate them.
	
Some reflection leads to new examples of other possible
	meta-theories.
One example, related to the roadside video dataset 
	proposed in Chapter~\ref{chapt:compmethod},
	is a list of car types such as the Toyota Corolla and the Honda Civic,
	that includes information about the dimensions of those types.
When a Toyota Corolla appears in the video data,
	the encoder could then specify the car type only and
	save bits by obviating the need to transmit shape information.
In the terminology of Chapter~\ref{chapt:comprvsion},
	this would correspond to defining a scene description
	language with a built-in function that draws a Toyota Corolla.
By using this specialized function when appropriate,
	shorter codelengths could be achieved in comparison
	with using an element that draws a generic automobile.
	
Another interesting possibility could be called a ``texture dictionary''.
Different kinds of materials have different visual textures:
	asphalt, brick, tree leaves, and clouds all have 
	very distinctive textures.
Such a dictionary might be very powerful in combination with an
	image segmentation algorithm.
The algorithm would attempt to identify the texture of a pixel region,
	and then package together pixels with similar textures.
Using the dictionary would save bits by using a prespecified
	statistical model for a pixel region.
Since textures are often associated with materials or objects,
	this technique could lead to further improvements
	relating to the understanding of characteristic object shapes.

\mpagebreak

\section{Unification}

A major theme in physics is the process of theoretical unification.
Unification occurs when two or more phenomena,
	which were previously described with completely unrelated theories,
	are shown to be describable by a single theory,
	which is more general and abstract.
Since a single unified theory is simpler than
	multiple specialized theories,
	unification indicates that the new theory is more correct and powerful.
	
% TODO: fix this description.
The history of science records 
	several examples of the process of unification;
	one of the most dramatic comes from the field of electromagnetics.
Until the time of Maxwell,
	the physical theories of electricity and magnetism
	were not packaged together.
Coulomb found that an electric field 
	was produced by the existence of a concentration of charge at some point in space.
The electric field points directly away from (or towards) the charged particles.
Ampere found that magnetic fields could be created by electrical current,
	and that the magnetic field lines circle around
	the direction of the current.
Faraday, in turn, 
	found that a change in the magnetic field creates an electric field.
These relationships were packaged together by Maxwell,
	who proposed the following compact set of equations:

\begin{eqnarray*}
\nabla \cdot \mathbf{E} &=& \frac{\rho}{\epsilon_{0}} \\ 
\nabla \cdot \mathbf{B} &=& 0 \\ 
\nabla \times \mathbf{E} &=& - \frac{\partial \mathbf{B}}{\partial t} \\
\nabla \times \mathbf{B} &=& \mu_{0} \mathbf{J} + \mu_{0} \epsilon_{0} \frac{\partial \mathbf{E}}{\partial t} 
\end{eqnarray*}

\noindent
Where $\mathbf{E}$ is the electric field, 
	$\mathbf{B}$ is the magnetic field,
	$\rho$ is a charge density, and $\mathbf{J}$ is the current.
These equations elegantly
	express the geometrical relationship of the various quantities 
	to one another.
The divergence operator, $\nabla \cdot$, 
	indicates straight lines radiating outward from a source,
	while the curl operator $\nabla \times$ 
	indicates that the field lines rotate around a central axis.
Maxwell's key contribution was the term 
	$\mu_{0} \epsilon_{0} \frac{\partial \mathbf{E}}{\partial t}$, 
	which expresses an effect that had not been measured at that time.
This term has crucial implications,
	because it shows that a magnetic field
	is produced not only by a current ($\mathbf{J}$), 
	but also by a change in the electric field.
This explains the phenomenon of light.	
Since a change in the electric field causes a magnetic field,
	and a change in the magnetic field causes an electric field,
	an initial disturbance in either field will propagate through space.
This propagation is a light wave.
Maxwell's equations therefore present a \textit{unified}
	description of three phenomena:
	electricity, magnetism, and light.
The following thought experiment shows how a similar
	process of unification can take place
	within \compist~science.

\mpagebreak
	
\subsection{The Form of Birds}

After his discussions with Sophie about the idea of virtual labels,
	Simon began to make rapid progress
	in the development of theories of birdsong.
He would use the synthesizer to create virtual labels
	for a large number of sound clips,
	then use the virtual labels to create a fast detector.
When the detector worked for the first time,
	and he ran outside to find a Blue Jay chirping in a tree next to the microphone,
	he experienced a feeling of vertiginous elation.
Unfortunately, the birdsong synthesizer program
	he found on the internet 
	was able to produce vocalizations for only a couple of bird species.
To go further, 
	it was necessary for Simon to develop 
	extensions to the synthesizer himself.
	
At the outset, Simon felt excited about this work.
Writing a synthesizer for a new species was an interesting challenge,
	requiring him to study the acoustic properties of each species' song,
	and figure out how to express those properties in computer code.
After the synthesizer was completed,
	he used it to create virtual labels and then train a recognizer
	for the bird.
Eventually, though, his old impatience 	
	got the better of him again.
The process of writing a synthesizer process began to feel like a tedious
	and painstaking chore.
It took Simon almost a month to write a synthesizer for a single species,
	and there are hundreds of bird species in North America.
Simon began to despair of ever completing his task.

During a pause in his efforts,
	Simon was assigned to read Plato's ``Republic''
	in his high school philosophy class.
Many of Plato's ideas seemed incomprehensible to Simon,
	but when he started to read about the theory of Froms,
	he felt that Plato's ideas connected somehow with his own studies.
Plato observed that all people agreed more or less on the meaning
	of the word ``dog'', 
	and the properties attached to the dog-concept.
A man from Sparta would mostly agree with a man from Athens about the concept,
	even if the two men had never observed the same dog.
Plato explained this phenomenon
	by postulating the existence of Forms -
	abstract versions of each particular entity.
Thus, all real dogs
	were simple imitations or manifestations of the Form of Dogs,
	which existed on an entirely higher plane.
By observing many dogs,
	the Spartan and the Athenian both come to perceive the Form of Dogs,
	and this explains why they can agree on what the word ``dog'' means.
The Forms also shape perception.
When one observes a dog,
	one perceives the dog in terms of its Form.
The Form specifies the range of variation allowed in each particular instance.
Thus dogs may be small or large, with floppy ears or pointed ears,
	and so on.
To understand a particular dog,
	one combines the relevant observed properties with the abstract invariants
	determined by the Form.
By doing so,
	one obtains a good understanding of any newly observed dog,
	without needing to conduct a comprehensive study of it.

Simon thought about the idea of Forms for a long time,
	while also trying to figure out how to expedite his 
	project of detecting and recognizing birdsong.
After several months,
	he came to believe that the idea of Forms held
	the secret to his problem.
Each particular bird species had a unique song.
But all of the songs shared general properties.
%TODO: syrinx is the name of bird... larynx
% TODO: this section does not make sense.
These general properties were dictated by the 
	biomechanical properties of the birds' vocal system,
	and the range of frequencies to which 
	their ears were sensitive.
Simon did not know exactly what the precise characteristics
	of these general properties were.
But that information was implicitly contained in the database;
	he just needed to extract it.
So he only needed to write a learning algorithm 
	that inferred the properties of each species'
	song from the data.
	
After several months of hard intellectual effort,
	Simon succeeded.
He obtained a program that,
	when invoked on the database of treetop recordings,
	would produce specific local theories
	describing the song of each species.
As more and more data was obtained,
	the performance of the automatically generated local theories
	became just as good as Simon's original manually-developed theories
	(Figure~\ref{fig:metavsoptm}).
The local theory 
	for a given species
	contained both a synthesizer (generative model) and a detector
	for the corresponding song.
For that reason, 
	Simon's research was of interest to bird-watchers and environmentalists.
By working with a group of ecologists studying the Amazon rain forest,
	Simon obtained a large database of recordings from that ecosystem.
By applying the meta-theory to that database,
	Simon was able to detect the existence of a previously unknown bird species,
	which he named after his aunt.
	
After the meta-theory of birdsong was complete,
	a natural next step was to study the appearance of birds.
The necessary data was easily obtained,
	by putting cameras in treetops and other promising locations.
Simon was initially stunned by the difficulty of working
	with image data.
Visual effects such as occlusion and lighting,
	in combination with the simple fact that 
	a small change in pose could cause a substantial change in appearance,
	made the problem devilishly hard.
But by narrowing his attention down to a single species,
	and by assiduously following the compression rate,
	Simon was able to build a local theory for the American Goldfinch.
Then he developed theories for the Blue Jay,
	Spotted Owl, Osprey, and the Raven.
Finally,
	after writing several local theories,
	Simon was able to construct a meta-theory of the appearance of birds.

\mpagebreak

\subsection{Unification in \Compist~Science}

The thought experiment given above illustrates the process
	of unification in \compist~science.
Simon was able to jump from a set of unrelated theories
	describing specific birdsongs to
	a unified and abstract theory describing
	birdsong in general.
In addition to its practical time-saving value,
	this abstract theory is obviously much 
	more satisfying from an intellectual perspective.
Of course, 
	the unification process is incomparably easier to describe
	as a thought experiment than to actually perform.

To understand the nature of this difficulty,
	compare the hypothesized birdsong meta-theory
	to the well-understood cartographic meta-theory (the Form of Maps).
When combined with a large quantity of observational data,
	the latter produces a specific local theory describing
	a geographical region.
The basic idea of the cartographic meta-theory
	is understood by everyone.
Nevertheless,
	it is still quite difficult to discover
	the precise algorithmic process by which to construct
	a specific theory out of a large corpus of data.
The birdsong meta-theory
	performs an analogous role:
	given a large corpus of treetop recordings,
	it produces several specific local theories 
	of the songs of individual bird species.
However, unlike the cartographic meta-theory,
	the nature of the birdsong meta-theory is currently unknown.
To complete it
	will require an extensive study of birdsong,
	and an effort to discover the invariants shared by all songs,
	as well as the abstract dimensions of variation between songs.
Then additional work will be required to 
	find a learning algorithm that can automatically extract
	the relevant parameters from an appropriate dataset.

If the ideas above are at all clear,
	it should be obvious that there are a wide range of categories 
	susceptible to the process of unification.
Consider the roadside video camera database
	proposed in Chapter~\ref{chapt:compmethod}.
One component of that research
	might be the use of templates describing
	various car designs,
	such as the Toyota Corrolla.
Such templates will allow the compressor to save 
	bits by encoding only the category of the car,
	instead of its precise geometry.
At the outset,
	researchers will probably need to design these templates by hand.
But after templates have been developed
	for several car types, 
	researchers should begin to discern the basic pattern of the process.
Once templates have been created for the Corrolla,
	the Honda Civic, the Ford Explorer, and the Mazda Miata,
	it will not be much harder to construct 
	the template for the Hyundai Sonata.
At this point it should become 
	possible to construct a meta-theory of the appearance of cars (the Form of Cars),
	and a corresponding learning algorithm.
The learning algorithm uses the meta-theory
	and a large corpus of video to obtain
	specific theories for each car type.
A similar process may work 
	for trees, insects, flowers, motorcycles, clothing, and houses.
The next section argues the process may also hold for language.
	
\mpagebreak
	
\subsection{Universal Grammar and the Form of Language}

A concept called Universal Grammar (UG) is a major theme and goal
	in the field of linguistics~\cite{Chomsky:1995}.
At least two key observations motivate the search for UG.
First, any normal human child growing up in any linguistic environment
	will be able to achieve fluency in its native tongue.
A child of Japanese ethnicity growing up in Peru will
	become perfectly fluent in Peruvian Spanish.
Second, children are able to infer facts about grammar and syntax
	from a limited amount of data.
In particular,
	children receive very little exposure
	to negative examples (ungrammatical utterances).
In spite of this,
	they reliably learn to distinguish grammatical sentences
	from ungrammatical ones.
These ideas imply that humans must have a powerful innate 
	ability to learn language.

Related to the idea of UG is the observation
	that human languages seem to obey a set of linguistic universals.
All languages have grammar and use words that 
	can be categorized into parts of speech
	such as ``noun'' and ``verb''.
The linguist Joseph Greenberg proposed a set of 45 linguistic universals,
	of which the following is an example:

\begin{quote}
In declarative sentences with nominal subject and object, 
	the dominant order is almost always one in which the subject precedes the object~\cite{Greenberg:1963}.
\end{quote}

\noindent
From a learning-theoretic point of view,
	the existence of linguistic universals is not surprising,
	given the fact mentioned above 
	that children can learn correct grammar from
	a small number of examples.
If there were no such structure or constraints on the form of language,
	children (or any other learning system)
	would require much more data to infer the correct grammatical rules.	
	
The idea of Universal Grammar,
	and the search for its properties, 
	can be formulated in terms of \compist~philosophy as follows.
Linguistic competence represents 
	a specific local theory of some language.
This local theory contains a variety of components,
	including but not limited to vocabulary lists,
	part of speech information,
	and grammatical rules.
However, humans are not born with this specific theory.
Instead, they are born with a meta-theory.
This meta-theory,
	when combined with a reasonably sized corpus of data, 
	produces a specific theory.
	
The above claim can be transformed into a concrete computational statement.
Consider a set of linguistic databases of several different languages
	(English, Chinese, and so on).
For each individual database,
	there is some  specific optimal theory-compressor
	that can be used to compress the database to the smallest possible size.
The \compist~translation of the UG claim
	is that there exists a single meta-theory that can adapt to 
	the structure of each language,
	and produce compression rates that are asympotically equivalent
	to the rates produced by the language-specific theories
	(Figure~\ref{fig:metavsoptm}).

The key point that bears reiteration is that the meta-theory
	must still be \textit{empirical}.
It must make assumptions about the structure of the data to which it is applied,
	or else it cannot achieve compression.
These assumptions will take the form of abstract statements,
	such as the one due to Greenberg mentioned above.
Proposal theories of UG can therefore be tested using the compression principle.
If the theories are too restrictive,
	they will fail to describe some languages well,
	and therefore fail to compress the corresponding databases.
If the theories are too loose,
	they will fail to exploit structure that exists,
	and underperform relative to more aggressive theories.
Therefore,
	the compression principle can be used to find the optimal meta-theory
	that describes linguistic universals while also
	acknowledging linguistic diversity.

\mpagebreak
	
\subsection{The Form of Forms}

After several years of studying birds,
	Simon's attention began to wander.
There were still some small challenges related to birds,
	such as attempting to identify individuals,
	but he felt that he had solved all the major problems.
One day he was studying an 
	image of a Spotted Owl
	that was partially obscured by tree leaves.
Because the bird was obscured,
	his compressor had difficulty with the image.
He found himself studying the shapes of the leaves
	and the branches of the tree.
He realized that trees had their own unique characteristics 
	that were also worthy of study.

Working with roboticists, 
	he designed a robot that drove through the forest
	and photographed trees.
The robot was programmed to take close-up images
	of the bark, roots, and leaves of each tree,
	in addition to taking full scale photographs
	of the entire tree.
Several copies of the robot were built,
	and deployed in forests around the world.
By studying the database of images sent back by the robots,
	Simon began an inquiry into the appearance of trees.
He learned about different textures of bark
	and different types of leaves,
	and how those two features were correlated.
He was able to deduce 
	a cladogram of tree species
	by analyzing the similarities in their outer appearance.
Finally, after building specific theories for oaks,
	pines, elms, birches, and palm trees,
	he was able to construct a meta-theory of trees.
The tools Simon developed in his study were useful to park rangers and logging companies,
	who were interested in monitoring the number of 
	trees of each type in a given region.

While trees were an interesting challenge,
	they only held Simon's attention for a couple of years.
He met another group of roboticists,
	who where working on an underwater robot,
	and he persuaded them to help him construct a 
	large database of images of fish.
By studying this database, 
	Simon learned theories of the appearance of 
	marlins, trout, catfish, sharks, and lampreys,
	and finally he obtained a meta-theory of fish.
These theories helped conservationists to monitor	
	the populations of different fish species,
	and helped fishermen locate their targets.
Simon then moved on to a study of insects.
He built specialized camera systems that were able
	to track insects as they crawled or flew,
	and take high frame rate image sequences of their motion.
Based on the resulting database,
	he began to construct and refine theories
	of the appearance of insect species.
He began with bees, 
	and then moved on to ants, and then to cockroaches.
The difficulty he had with each new species
	caused him to develop a grudging respect for the tremendous
	biodiversity of insects.
Finally, after years of study, 
	he was able to obtain a unified meta-theory of insects
	that successfully described the entire class,
	including everything from wasps and spiders to ladybugs and butterflies.
These tools helped farmers to protect their crops against insects,
	and so reduce their dependence on chemical pesticides.
They also helped scientists in Africa to monitor the activity of mosquitoes,
	thereby contributing to efforts in the fight against malaria.
	
After more than a decade of studying the biological world,
	Simon switched his attention to the social world,
	and he began a study of written language.
He assembled large corpora of text,
	and applied Sophie's method to these databases.
By doing so he learned about grammatical rules,
	parts of speech, 
	and linguistic topics such as \textit{Wh}-movement
	and the distinction between complements and adjuncts.
After studying written language for a while,
	he broadened his inquiry to include spoken language as well.
He learned that spoken language obeyed a different set 
	of grammatical rules than written language.
After long efforts spent developing local theories of English, Japanese,
	Arabic, and Gujarati,
	he began to see the abstract patterns common 
	to all human languages, 
	and he was able to discover a meta-theory of language.
The tools he developed became useful
	for a variety of purposes
	such as information retrieval, machine translation, 
	and speech recognition.
	
Though he much prefered images of trees and animals 
	to images of cars and people,
	he realized the next logical step was
	to study the human visual environment.
He persuaded a taxi cab company to mount cameras on top of their cars,
	and collect the resulting video streams.
From this data,
	he studied the appearance of pedestrians, buildings, cars,
	and many other aspects of the urban panorama.
He discovered meta-theories for each category of object.

In each area Simon studied,
	his procedure was the same.
First he would develop a series of local scientific theories
	that described a specific set of elements of a class very well.
In studying trees,
	he first built theories of the appearance of oaks, then elms,
	then birch, then pine trees.
As he constructed each new local theory,
	he would begin to see the underlying structure of the class.
A local theory of a particular tree species
	required an analysis of its bark,
	its leaf pattern, its branching rate, and a couple of other components.
Once he understood these parameters of variation,
	he would construct a meta-theory describing the entire class.
The meta-theory, combined with the large database and a learning algorithm,
	would automatically construct local theories for each particular member of the class.
	
At first the different categories,
	such as birds, trees, languages, and buildings,
	seemed to have almost nothing in common with one another.
When starting in each new area,
	Simon would have to begin again almost from scratch.
But as he studied more and more categories,
	he began to notice their similarities.
Just as computer scientists constantly reuse
	the list, the map, and the for-loop,
	and physicists constantly reuse calculus and linear algebra,
	Simon found himself using the same set of tools over and over again.
He began to sense the possibility of a still more abstract theory.
Instead of producing local theories directly,
	this new meta-meta-theory would,
	when supplied with a sufficient quantity of data,
	produce meta-theories, which would in turn produce local theories.
A meta-theory was like a factory,
	which instead of transforming raw material into cars,
	airplanes, and electronics,
	transformed data into local theories.
The meta-meta-theory would be like a factory for constructing factories.	
By combining all his previous datasets together,
	and liberally adding a wide variety of new ones,
	Simon constructed the largest database he had ever used.
Then he began his ultimate quest:
	the search for the Form of Forms.

%\minclude{vsionphysc}
%\minclude{crmvsphscs}
%\minclude{intellignc}
%\minclude{otherideas}

\appendix
%\minclude{maximappnd}
\chapter{Information Theory}
\label{chapt:infoappend}

\section{Basic Principle of Data Compression}

This appendix presents a very brief overview of the some of the core 
	concepts of information theory required to understand the book.
For a fuller treatment, see~\cite{Cover:1991}.

The motivating question of information theory is:
	how can a given data set $\{x_{1}, x_{2} \ldots x_{N}\}$
	be encoded as a bit string using the smallest
	possible number of bits?
In spite of the very technical nature of this question,	
	the answer turns out to be quite deep.

To motivate this problem,
	imagine trying to encode an immense database of DNA data.
The DNA information is represented as a long sequence of letters G,A,T, and C,
	as follows:
	
\begin{quote}
AAAAGGTCTAGTCAAAGTCGAGAAAGCTCGGGAAAAATGCAATC ...
\end{quote}
	
\noindent
Since there are four different outcomes,
	an obvious scheme would be to encode each outcome using 2 bits.
For example, the following assignment would work:
	$A \rightarrow 00$, $G \rightarrow 01$, $T \rightarrow 10$, 
	$C \rightarrow 11$.
Here, each outcome requires 2 bits to encode,
	so if the database includes $10^6$ outcomes,
	then this scheme would require $2 \cdot 10^6$ bits to encode it.
	
However,
	suppose that it is found that some DNA outcomes are much more
	common than others.
In particular, 50\% of the outcomes are A, 25\% are G, 
	while C and T both represent 12.5\% of the total.
In this case,
	it turns out that a different encoding scheme will achieve better results.
Let $A \rightarrow 1$, $G \rightarrow 01$, $T \rightarrow 001$, 
	$C \rightarrow 000$.
This scheme achieves the following codelength for the full database:

\begin{eqnarray*}
L &=& N (P(A) L(A) + P(G)L(G) + P(T)L(T) + P(C)L(C)) \\
&=& 10^6 (\frac{1}{2} \cdot 1 + \frac{1}{4}\cdot 2 + \frac{1}{8}\cdot 3 + \frac{1}{8}\cdot 3) \\
&=& 10^6 (1.75) \\
\end{eqnarray*}

\noindent
Where $P(x)$ is the frequency of an outcome $x$,
	and $L(x)$ is the number of bits used by the encoding scheme
	to represent $x$.
Evidently, 
	this scheme saves $2.5 \cdot 10^5$ bits compared to the previous scheme.
Since the $A$ outcomes are very common,
	it is possible to save bits overall by assigning them a very short code,
	while reserving longer codes for the other outcomes.
When rearranging codes like this,
	there is an important rule that must be obeyed,
	which can be illustrated by imagining that
	that the encoding scheme used 
	$A \rightarrow 1$, and $G \rightarrow 11$.
The problem is that when the decoder sees a 1,
	it will think the outcome is an A,
	even if the encoder wants to send a G.
In general,
	the rule is that no outcome can be assigned a code
	that is the prefix of some other outcome's code.
This is called the prefix-free rule.
This rule can be used to derive an important relationship,
	called Kraft's inequality, 
	governing the codelengths $L(x)$ that are used to encode a data set:
	
\begin{equation}
\sum_{x} 2^{-L(x)} \leq 1
\end{equation}
	
\noindent
This sum runs over all possible outcomes,
	so in the DNA example there would be four terms,
	corresponding to A, G, T, and C.
Kraft's inequality 
	is a mathematical expression of the intrinsic scarcity of short codes:
	if there are 16 possible data outcomes,
	they cannot all be assigned codelengths less than 4 bits,
	because the sum of the $2^{-L(x)}$ will be greater than 1.
Kraft's inequality holds for any viable prefix-free code,
	whether or not it is any good.
But if the sum is substantially less than one,
	it means that the code is actually wasting bits.	
For any codelength function $L(x)$ that is not trivially suboptimal,
	Kraft's inequality becomes an equality.
Because of this,	
	the function $P(x) = 2^{-L(x)}$
	satisfies the criteria for probability distribution:
	$0 \leq P(x) \leq 1$,
	and the normalization requirement $\sum_{x} P(x) = 1$.
This suggests a relationship 
	between the probability distribution of a set of data outcomes
	and the codelength function one should use to encode them.
	
The DNA example given above suggests the basic strategy of data compression.
In order to save bits on average
	in spite of the instrinsic scarcity of short codes,
	it is essential to assign long codes to the uncommon outcomes,
	while reserving short codes for the common outcomes.
This suggests a relationship between the probability $P(x)$ of 
	an outcome and the optimal codelength $L(x)$
	that should be used to encode it.
But what exactly is the relationship between the two functions?
In fact,
	the actual relationship was hinted at a moment ago.
Claude Shannon proved that the optimal codelength $L(x)$
	is related to the probability $P(x)$ 
	as follows:
	
\begin{equation}
\label{eq:codeprobeq}
L(x) = -\log P(x)
\end{equation}

\noindent
Here and in the following,
	all logarithms are assumed to be in base 2.
Take a moment to consider what this equation is saying.
It does not dictate the choice of codelength functions;
	rather, it says that any other function will perform worse on average.
The caveat ``on average'' is necessary
	because it is not impossible for the observed frequency
	of an outcome to be very different from its probability.
Consider a case where the probability $P(A)$ of A is 50\%,
	but by chance it is observed at a frequency of only 10\%;
	in such a case a codelength function that
	corresponds to the observed frequency will do better
	than one corresponding to the predicted probability.
Also, the codelength-probability relationship 
	only specifies the lengths of the codes;
	it gives no guidance about how to construct them.
	
The relationship between codelength and probability specified
	by Equation~\ref{eq:codeprobeq} indicates a deep relationship
	between statistical modeling and compression.
In order to achieve the shortest possible codes,
	a sender wishing to transmit a data set $\{x_{1}, x_{2} \ldots \}$
	must know the distribution $P(x)$
	that generated the data.
It also indicates that there is a one-to-one relationship
	between a discrete statistical model and a compression
	program for the type of data specified by the model.

Given the codelength-probability equivalence relationship, 
	it is natural to define the \textit{entropy}:

\begin{equation}
\label{eq:entropyeqn}
H(P) = \sum_{x} P(x) L(x) = -\sum_{x} P(x) \log P(x)
\end{equation}

\noindent
These sums run over all possible $x$ outcomes.
The entropy of a distribution $P(x)$ 
	is the expected codelength achieved by the optimal code for $P(x)$.
There are several important facts to notice about the entropy.
First, it is a function of the distribution $P(x)$.
Second, upper bounds for the entropy can easily be found.
For example, if there are 16 possible outcomes,
	then the entropy is as most 4 bits.
Generally speaking, the greater the number of outcomes in the distribution,
	the larger the entropy.

A question that arises often in statistical modeling is:
	what happens if an imperfect model $Q(x)$ 
	is used instead of the real distribution $P(x)$,
	which may be unknown?
Since $Q(x)$ is imperfect, the codelengths specified by $Q(x)$
	will be suboptimal.
The codelength penalty for using imperfect codes based on the model $Q(x)$
	instead of perfect codes based on the real distribution $P(x)$ is:

\begin{eqnarray}
\label{eq:kldvrgence}
D(P||Q) &=& \sum_{x} P(x) L(x) - \sum_{x} P(x) L_{q}(x) \\
&=& \sum_{x} P(x) \log \frac{P(x)}{Q(x)}
\end{eqnarray}

\noindent
The quantity $D(P||Q)$ is known as the Kullback-Liebler divergence (KLD).
It is easy to show that the Kullback-Liebler divergence is always non-negative,
	and becomes zero only if $Q(x) = P(x)$.
This implies that no scheme for assigning codelengths to data outcomes
	can achieve better codelengths on average than $L(x) = -\log P(x)$.
It is possible to get better codes for certain outcomes.
But by reducing the codelength of some outcomes,
	it is necessary to increase the codelength of other outcomes,
	and this will hurt the overall average.
	
Though the entropy and the KLD are fascinating concepts,
	they are not very useful in practice,
	because they both depend on $P(x)$,
	which is generally unknown.
A more useful quantity is the realized codelength:

\begin{equation}
\sum_{k=1}^{N} L(x_{k}) = -\sum_{k=1}^{N} \log Q(x_{k})
\end{equation}

\noindent
Here the sum goes over the $N$-sample data set $\{x_{k}\}$.
Since this quantity depends only on the model $Q(x)$
	and the data set,
	it can be computed in practice.
Also, 
	if the data set is large,
	it should be related to the entropy and the KLD, since:
	
\begin{eqnarray*}
- \frac{1}{N} \sum_{k=1}^{N} \log Q(x_{k}) &\approx& -\sum_{x} P(x) \log Q(x) \\
&=& -\sum_{x} P(x) \log P(x) + \sum_{x} P(x) \log \frac{P(x)}{Q(x)} \\ 
&=& H(P) + D(P||Q) 
\end{eqnarray*}

\noindent
In words,
	the average realized per-sample codelength for the model $Q(x)$
	is about equal to the entropy $H(P)$ plus the KLD $D(P||Q)$.
This provides one motivation for the goal of minimizing the codelength.
$H(P)$ is a constant of the distribution $P$,
	so it can never be reduced.
Thus, reducing the codelength is equivalent to reducing the KLD.
The only way to reduce the KLD is to 
	obtain better and better models $Q(x)$,
	since as $Q(x) \rightarrow P(x)$,
	$D(P||Q) \rightarrow 0$.
	
A second motivation for attempting to minimize the codelength
	is the relationship to the Maximum Likelihood Principle.
Given a model family $Q(x;\theta)$ that depends on parameters $\theta$,
	this principle suggests choosing $\theta$ to maximize 
	the likelihood of the data given the model:
	
\begin{eqnarray*}
\theta^{*} &=& \arg \max_{\theta} \prod_{k} Q(x_{k}; \theta) \\
&=& \arg \min_{\theta} \sum_{k} -\log Q(x_{k}; \theta) 
\end{eqnarray*}
	
\noindent
In words,	
	the codelength is minimized when the 
	likelihood is maximized.
	
\subsection{Encoding}

The codelength-probability relationship of Equation~\ref{eq:codeprobeq}
	specifies the \textit{lengths} of the optimal codes.
But how then can the code itself be constructed?
That is to say, given a model $Q(x)$,
	how do we transform a series of outcomes $\{x_{1}, x_{2} \ldots x_{N}\}$
	into a bit string $S$ that satisfies~\ref{eq:codeprobeq}?
This is called the encoding problem, and it is non-trivial.
Fortunately, however, 
	very general solutions to the encoding problem exist.
The first method proposed that was guaranteed to achieve near-optimal 
	codelengths is called Huffman encoding~\cite{Huffman:1952}.
Another method, 
	called arithmetic encoding~\cite{Rissanen:1976,Witten:1987},
	is also very useful.
	
The encoding techniques allow compression researchers to avoid 
	a variety of technical problems.
One such problem is:
	given that probabilities can take on arbitrary values between zero and one,
	but a bit string needs to have an integral length,
	how can $L(x) = -\log P(x)$ produce integral bits?
The solution is to chain together multiple outcomes,
	so that while $L(x)$ is not quite integral,
	not much is lost by simply rounding it up.

\chapter{Related Work}
\label{chapt:relappendx}

This appendix describes research that is closely related
	to the \compist~idea.
In several cases,
	the related research has had a strong influence in shaping the ideas of the present book.
This is particularly true for the Hutter Prize,
	Hinton's philosophy of generative models,
	and the research on statistical language modeling that 
	was described in Chapter~\ref{chapt:linguistic}.
	
\mpagebreak
	
\section{Hutter Prize}
	
Matt Mahoney, a researcher at the Florida Institute of Technology, 
	was the first to advocate the problem of large scale lossless data compression.
Mahoney saw the deep interest of the compression problem
	and its methodological advantages.
In particular, 
	he argued that achieving prediction accuracy on text comparable
	to what a human can achieve 
	is related to the Turing Test
	and therefore nearly equivalent to the problem of artificial intelligence.
Mahoney's justification is included in the following paper:

\begin{itemize}
\item Matt Mahoney, ``Rationale for a Large Text Compression Benchmark''~\keycite{Mahoney:2006}
\end{itemize} 

Based on this philosophy,
	Mahoney set up a text compression contest 
	targeting a 100 Mb chunk of the English Wikipedia.
The contest was funded by and named after Marcus Hutter,
	a researcher who became interested 
	in the compression problem for theoretical reasons~\cite{Hutter:2004}.
Mahoney developed a compressor called PAQ
	which achieved the first strong result on the Wikipedia data.
Afterwards, 
	a small group of researchers from around the world
	contributed improved techniques, 
	leading to a gradual improvement in the compression results.
Most of these contributions 
	involved either preprocessors or modifications to PAQ.

While Mahoney's ideas and the Hutter Prize are a strong influence on the present book,
	there is a substantial gap between the two philosophies.
Mahoney did not clearly articulate the essential component of 
	the \compist~philosophy,
	which is the equivalence between lossless data compression and empirical science.
Mahoney and Hutter both seem to suffer from the computer science mentality,
	which suggests the existence of one single, general, all-purpose data compressor.
There are two reasons why this belief 
	had a negative influence on the research related to the Hutter Prize.
	
The first result of the negative influence
	was that contributions to the contest were mostly all general-purpose compressors.
PAQ itself is a smart algorithm for integrating statistics (predictions)
	from many history-dependent context functions.
Because of the linear nature of text,
	it is possible to make relatively good predictions based on the immediate history,
	and PAQ works well for that reason.
PAQ would also work well for several other types of data,
	such as music and programming language source code.
Few participants in the contest seemed interested
	in developing special-purpose compressors targeting the Wikipedia data.
Most seemed to believe
	that a smart general-purpose compressor could be found that would 
	give compression rates nearly as good as any specialized 
	compressor could.
	
The second negative influence of the computer science mentality
	was that it led Mahoney to choose the Wikipedia data as the target.
This data has an abundance of XML tags,
	image links, URLs, and related cruft intermixed
	with the text data. 
Mahoney probably thought that a smart general-purpose compressor could
	be found that would automatically deal with the formatting information.
If Mahoney had selected a cleaner batch of text data as the compression target,
	the contest participants would probably have realized the link
	between language modeling and text compression,
	and started to apply sophisticated language modeling techniques to the problem.
Moreover,
	the cruft-laden nature of the target database probably
	deterred participants from attempting to develop specialized compressors.
If the data was pure text,
	a specialized compressor would be equivalent to a language model,
	and would therefore probably be reusable
	for reasons discussed in the book.
But a specialized compressor for the Wikipedia data
	would have to include thousands of lines of code
	to deal with the various forms of cruft.

\mpagebreak

\section{Generative Model Philosophy}

An important thread of research in machine learning deals with the
	construction of generative models.
These models attempt to describe the observational data
	instead of, or in addition to,
	the quantity of interest.
In other words,
	the generative approach attempts to build models $P(X)$ or $P(X,Y)$ 
	instead of just $P(Y|X)$.
These kind of models are called generative because
	they can produce data similar 
	to the original $X$ data by sampling
	(the degree of similarity depends on how good the model is).
The Collins parser described in Chapter~\ref{chapt:linguistic}
	was an example of a generative model,
	because it defined a joint distribution $P(T,S)$ 
	of the probability of a sentence $S$ and a parse tree $T$.
There is also a clear connection 
	between the idea of generative models and 
	Chomsky's idea of generative grammar.

The generative model approach can also be used to approach
	the standard task of supervised learning 
	when the $Y$ data comes in a small number of discrete labels.
Consider the much-studied problem of handwritten digit recognition,
	which involves ten labels.
To use the generative approach, 
	one builds ten models $P_{1}(X), P_{2}(X), \ldots P_{10}(X)$.
Then given new sample $x$,
	the system outputs a guess $y_{k^{*}}$, 
	where $k^{*}$ is chosen using the following rule:
	
\begin{equation}
k^{*} = \arg \max_{k} P_{k}(x)
\end{equation}
	
\noindent
While the generative approach can be used for the standard task,
	it usually provides worse performance 
	in terms of error rate
	than the normal approach of learning $P(Y|X)$ directly.

One major advocate of the generative approach is Geoff Hinton,
	a professor at the University of Toronto.
Hinton's work,
	and its underlying philosophy, 
	has had a significant influence on the present book.
Unfortunately that philosophy is never articulated clearly and directly,
	but important aspects of it can be gleaned from 
	the following three papers:

\begin{itemize}
\item Geoff Hinton, ``To recognize shapes, first learn to generate images''~\keycite{Hinton:2007}.
\item Geoff Hinton, Peter Dayan, Brendan Frey, Radford Neal,
	``The wake-sleep algorithm for unsupervised neural networks''~\keycite{Hinton:1995}.
\item Geoff Hinton, Simon Osindero, Yee-Whye Teh,
	``A Fast Learning Algorithm for Deep Belief Nets''~\keycite{Hinton:2006b}.
\end{itemize}

\noindent
The title of the first paper sums up the approach quite well.
Hinton advocates
	the indirect approach to learning,
	in which a generative model $P(X)$ is learned during the first step.
If a good model $P(X)$ can be found,
	it should not be much harder to find a joint model $P(X,Y)$,
	which can then be used to solve the supervised problem.
Hinton emphasizes that the same features (or patterns)
	that are important in modeling $P(X)$ are also probably important
	in modeling $P(Y|X)$.
Furthermore,
	the raw data models $P(X)$ can be made far more complex 
	than the conditional $P(Y|X)$ models,
	because the raw data contains much more information.
Section~\ref{sec:compistref} contains a key passage
	from reference~\keycite{Hinton:2006b} that articulates this idea.

The second paper,
	about the wake-sleep approach,
	describes an algorithm for using the generative model idea
	to train a neural network.
The key insight
	is that if the neural network model has adequately captured
	the statistics of the data set,
	the samples it generates should look like the real data.
The authors present a special type of neural network,
	that alternates between a ``wake'' phase,
	where it processes the input data using its current connection weights,
	and a ``sleep'' phase, where it samples from the current model,
	and compares the samples to the real data.
The authors then show how a simple learning rule,
	based on a statistical comparison of the sampled data to the real data,
	can be used to improve the network model.
	
The design of the wake-sleep algorithm highlights the 
	reciprocal roles of interpretation and generation,
	which are analogous to encoding and decoding in data compression.
These ideas can be understood easily in terms of 
	the abstract formulation of computer vision 
	described in Chapter~\ref{chapt:comprvsion}.
A generative model is a graphics program $G$ that
	transforms scene descriptions $D$ into an image $G(D)$.
It is easy to sample from the model:
	create a description $D$ at random,
	and feed it into the graphics program.
It is much more difficult,
	however, to obtain a good description $D$
	that reproduces an target image $I$
	such that $I \approx G(D)$;
	this is called the interpretation step.
The wake-sleep algorithm uses a smart statistical interpretation
	scheme to produce descriptions.
In terms of the neural network,
	a description is a vector describing the activity level
	of each node in the network.

The third paper,
	about deep belief nets,
	demonstrates the potential power of a two-phase (indirect) process
	in which the first phase learns highly complex structures
	by modeling the raw data.
Hinton~\etal~propose to learn models of the raw data 
	by using a stack of neural network layers.
Each layer learns about the patterns and features
	in the output of the previous layer;
	the base layer learns directly from the input images.
By learning about the raw data instead of the image-label relationship,
	this strategy repairs a standard problem with training a multi-layer network,
	which is that there is a large distance between
	the input and the output.
This distance makes it harder to determine
	what effect a small change in a low-level weight
	will have on the final output.
Another problem with the standard approach 
	is that multi-layer networks are by definition complex,
	so they will tend to cause overfitting when
	trained on limited data problems.
Training on the raw data overcomes this problem as well,
	since by modeling raw data it is possible
	to use complex models without overfitting.
	
Hinton~\etal~applied their stacked neural network
	algorithm to the MNIST database of handwritten digit images,
	a standard benchmark in machine learning.
The results they achieved, using the indirect approach,
	were slightly better than the previous state of the art results, 
	which were obtained using sophisticated direct
	modeling algorithms such as the Support Vector Machine.
However, 
	a more visually impressive feat
	was a demonstration of veridical simulation:
	the images produced by sampling from the learned model
	were almost indistinguishable from real handwritten digit images.
	
Hinton's philosophy has several elements in common with the \compist~approach.
Both philosophies advocate indirect learning
	as a way to circumvent the model complexity limitations
	inherent in the direct approach.
Both philosophies 
	emphasize the veridical simulation principle.
The reciprocal relationship between interpretation (encoding)
	and generation (decoding) 
	comes up in both approaches.

In spite of these similarities, 
	there are several crucial differences between the two philosophies.
For technical reasons,
	Hinton's models can be used to sample from the distribution $P(X)$,
	but cannot be used to explicitly calculate 
	the probability of a data sample.
This makes it impossible to compare one model to another,
	using the log-likelihood, compression rate, or an analogous quantity.
Perhaps this limitation can be repaired,
	but Hinton appears to have little motivation to do so.
This is because of the toolbox mentality criticized in Chapter~\ref{chapt:linguistic}.
Hinton and his coworkers conceive of their work 
	as the development of a suite of general-purpose tools.
Since different tools work well in different situations,
	it is difficult or meaningless to make rigorous comparisons between them.
Because the tools are thought of as general purpose,
	they can be tested using whatever databases happen to be available,
	such as the MNIST handwritten digit database or the problems
	in the UCI machine learning repository~\cite{Lecun:2006,Asuncion:2007}.
Thus, in this mindset,
	there is not much point in creating specialized databases,
	or in creating targeted models 
	that exploit the empirical properties of such databases.
This procedure is a core component of the \compist~philosophy.

\mpagebreak
	
\section{Unsupervised and Semi-Supervised Learning}

Unsupervised learning is the problem of learning without a set of labels
	or other guidance provided by a human supervisor.
Instead of attempting to learn the relationship between
	a set of images $\{x_{i}\}$ and a set of labels $\{y_{i}\}$,
	unsupervised methods attempt to learn something 
	about the images themselves.
The (huge) advantage of unsupervised methods is that
	they do not necessitate the construction of a labeled dataset,
	which is almost always a tedious and time-consuming project.
Furthermore,
	unsupervised methods can justify highly complex models
	without overfitting.	
These features of the problem formulation overlap with 
	the \compist~idea.
The goal of a related subfield,
	called semi-supervised learning,
	is to improve the performance of supervised learning
	by performing unsupervised learning in advance~\cite{Chapelle:2010}.
This notion is related to the thought experiments of Chapter~\ref{chapt:comprlearn}
	about indirect learning,
	and the Reusability Hypothesis that forms a core component of the \compist~philosophy.
	
Given a large set of raw data objects and 
	the general goal of learning \textit{something} about those objects,
	a critical question immediately arises:
	what quantity should the system optimize?
The subfield mandates no specific response to this question,
	and different answers will lead to widely varying
	lines of research.
Notice how in the case of supervised learning,
	the basic form of the answer is obvious:
	the goal is to learn the relationship between
	the data objects $\{x_{i}\}$ and the labels $\{y_{i}\}$.
Different researchers answer the optimization question
	by proposing quantities that reflect their expectations
	about what kind of structure the data set will exhibit.
For example, a common expectation is that the raw data samples
	will be distributed in clusters,
	and so the computational goal should be to identify those structures.
This reveals a general conceptual shortcoming of unsupervised learning:
	is that it is hard to know if the structure discovered by the algorithm
	is really present in the data,
	or if it is simply imposed on the data by the researcher's preconceived notions.

Many researchers have formulated the problem of unsupervised learning
	as one of \textit{dimensionality reduction}.
Most raw data objects 
	(images, sound, text, etc) 
	are very high dimensional.
For example, a typical image might have $9 \cdot 10^{5}$ pixels;
	so one image can be thought of as a point
	in a space with $9 \cdot 10^{5}$ dimensions.
The goal of dimensionality reduction techniques is to transform
	this kind of data sample into a point in a lower-dimensional space,
	while preserving most of the ``relevant'' information.
One common technique in this family is called 
	Principal Components Analysis (PCA)~\cite{Pearson:1901}.
Given a dataset of $P$-dimensional points,
	PCA first computes the $P \cdot P$ array of correlations
	between each pair of dimensions.
It then constructs a new basis by calculating the top $D$ eigenvectors of this matrix,
	where $D < P$.
This new basis expresses most of the variation in the data set,
	while requiring a smaller number of dimensions to do so.
Another popular dimensionality reduction technique is called Isomap,
	developed by Tenenbaum~\etal,
	which is a refinement of a procedure called Multi-Dimensional Scaling (MDS)~\cite{Tenenbaum:2000}.
Based on an $N \cdot N$ matrix $A_{ij}$ of point-point distances,
	MDS attempts to find a set of vectors in $D$-dimensional space
	that preserves the distance mappings: $||x_{i} - x_{j}|| \approx A_{ij}$.
Since $D$ can be smaller than the original dimensionality of the points,
	MDS provides a way of projecting the points into a lower-dimension manifold
	that preserves their similarity (distance) relationships.
Tenenbaum~\etal~define a special graph to produce the distance matrix.
In this graph, 
	only nearby points have edges between them,
	with small weights for very close together points.
The distance between two points is then defined as the 
	shortest path between the points in the graph.
Because this definition preserves the local structure of the relationships between the points,
	it can solve some complex manifold-discovery problems
	that other methods cannot.

Another important vein of unsupervised learning research is called automatic clustering.
The idea here is that data is often naturally distributed in clusters.
Data samples within a cluster often resemble one another much more
	than they resemble other samples.
Zoology provides a good example of naturally occuring clusters.
Categories like ``fish'' and ``mammal'' are in some sense real,
	and reflect legitimate order in the natural world.
The key point, again, is that clusters can in principle be learned on the basis
	of the raw data (e.g. images).
Many clustering algorithms have been proposed in the literature;
	one immensely popular algorithm is called $K$-means~\cite{MacQueen:1967}.
The goal of this algorithm is to partition the data points into $K$ sets
	in such a way as to minimize the distance between each point and
	the mean of its containing set.
The algorithm operates by choosing some initial guesses for the mean points,
	and then assigning each data sample to a cluster based on its distance
	to the means.
Then the means for each cluster are recomputed using the data points in the cluster.
Ng~\etal~proposed a clustering algorithm based on spectral graph theory~\cite{Ng:2001}.
The basic idea here is to find the eigenvectors of the graph's Laplacian matrix,
	which is the difference between the degree matrix and the adjacency matrix.
The second eigenvector is related to the solution of the minimum cut	
	problem for partitioning the graph into two subcomponents.

The variety of answers to the optimization question
	is a key conceptual weakness of research in this area.
Because there is no standard answer,
	there is no strong method for comparing candidate solutions.
Researchers operate according to the toolbox mindset,
	already criticized in various places in this book.
Perhaps the $K$-means algorithm will produce clusters
	that work well for a particular application; perhaps it won't.
If it doesn't,
	that is not the fault of the algorithm,
	it just means that the tool is not well-suited to the particular task.
It is the responsibility of the application developer
	to search through the toolbox to find 
	the one that works the best for her particular project.
To prove the potential usefulness of a new contribution,
	an unsupervised learning researcher 
	need only show that it is useful for \textit{some} task.
This results in a proliferation of tools.
	
Another flaw in the philosophy of unsupervised learning research
	is that methods are thought to be ``general'':
	they can be applied to any data set that has the correct form.
The implicit belief is that there exists some ideal or optimal algorithm
	that can discover the correct categories or hidden structure in any data set.
This belief is a holdover from traditional computer science,
	the discipline in which most machine learning researchers receive their early training.
In traditional computer science,
	it is perfectly reasonable to search for general algorithms.
Methods like QuickSort
	and Dijkstra's algorithm are fully general solutions to the problems
	to which they apply.

The \compist~approach to learning is thus related to unsupervised learning,
	but quite different in emphasis.
\Compist~researchers agree on the compression rate as the singular
	quantity of interest.
This enables them to rigorously compare candidate solutions.
At the same time, the No Free Lunch theorem
	shows that no truly general compression algorithm can exist.
To achieve improved compression rates, therefore,
	researchers must study the empirical properties of the data they are analyzing.
Furthermore,
	\compist~researchers study vast data sets
	and construct those data sets with some degree of care,
	so as to facilitate the study of particular aspects of the data
	(such as English grammar or the appearance of New York buildings).

\mpagebreak
	
\section{Traditional Data Compression}	
	
Data compression itself is not a new idea;
	it dates back at least to the work of Claude Shannon in the 1940s.
Researchers have been working 
	on various problems in this area for decades.
Indeed, many important applications of the ``dot com''
	era of the 1990s and 2000s became possible
	due to innovations in this field.
Perhaps the most important innovation was the development
	of the MP3 audio compression format,
	that rocked the music industry 
	by making it possible to distribute music files
	over a modestly fast internet connection.
Another important set of innovations relate to video compression,
	which allows people to watch movies online 
	or on their smartphones.

It should be clear that the research proposed in this book
	is only superficially related to traditional data compression research.
There are several crucial differences in emphasis and methodology
	that lead to wide divergences in practice.
	
The most basic difference is that traditional data compression aims
	for real practicality,
	while \compist~research aims for 
	specialized understanding of various phenomena.
This means, that, for example,
	a traditional compression researcher would 
	never bother to set up a roadside video camera
	and attempt to compress it.
No one else is interested in transmitting such a data set,
	so whatever he discovered, it would be of no interest to anyone else.
Also, the specialized insights that would allow superior compression rates
	for the roadside video camera would not be especially useful
	for other types of data.
	
Because of their primary interest in practical data transmission,
	traditional compression researchers consider
	a variety of other factors of additional factors
	when evaluating a new system.
One important consideration is speed;
	most users do not want to wait an hour for the decompressor to 
	produce an image.
Another consideration is simplicity of implementation.
If a compression format is to become a standard,
	it must be relatively easy for different groups to implement it.

Traditional compression researchers also devote a significant amount
	of effort the lossy version of the problem,
	where what goes into the encoder does not match exactly
	what comes out of the decoder.
This is worthwhile because the human perceptual system
	simply cannot detect some of the minute details of an image,
	so bits can be saved by leaving those details out.
The disadvantage of this approach is that
	it normally introduces an element of subjectivity.
It means that a human being must observe the lossy compressed image
	and decide if it provides a good quality for a low price in bits.
This removes a critical component of the philosophy articulated in this book,
	which is the methodological rigor associated
	with objective comparisons.
	
Another key difference between traditional compression research
	and \compist~research is that the former deals with 
	relatively small objects such as single images or text files,
	while the latter deals with databases that are 
	orders of magnitude larger.
This change in focus completely changes the character of the problem.
To see why, consider the problem of building a car.
If one wants to build only a single car,
	then one constructs each component of the car individually using machine tools,
	and fits the parts together by hand.
The challenge is to minimize the time- and dollar-cost 
	of the manual labor.
This is an interesting challenge,
	but the challenge of building ten million cars is entirely different.
If the goal is to build ten million cars,	
	then it makes sense to start by building a \textit{factory}.
This will be a big up-front cost,
	but it will pay for itself by reducing the marginal cost of each additional car.
Analogously, when attempting to compress huge databases,
	it becomes worthwhile to build sophisticated computational tools into the compressor.
For example, it is worthwhile to use a specialized hubcap detector
	for the roadside video data 
	(but not for a single car image).
The development of these sophisticated computational tools is the real goal
	of the research.

%\minclude{keyrefernc}

%\bibliographystyle{Classes/CUEDbiblio}
\bibliographystyle{plain}
\renewcommand{\bibname}{References} % changes default name Bibliography to References
\bibliography{../papers/latex/bibtex/allrefs} % References file
\addcontentsline{toc}{chapter}{References} %adds References to contents page

\end{document}